\documentclass{article}

\usepackage{arxiv}

\usepackage[utf8]{inputenc} 
\usepackage[T1]{fontenc}    
\usepackage{hyperref}       
\usepackage{booktabs}       
\usepackage{commath}
\usepackage{amsfonts}       
\usepackage{amsmath}
\usepackage{nicefrac}       
\usepackage{microtype}      
\usepackage{graphicx}
\usepackage{subfig}
\usepackage{float}
\usepackage{comment}
\usepackage[ruled,vlined]{algorithm2e}
\graphicspath{ {./images/} }
\usepackage{siunitx}
\usepackage{mathtools}
\usepackage{physics}
\usepackage{graphicx,pstricks}
\usepackage{moreverb}
\usepackage{epsfig}
\usepackage{microtype} 
\usepackage{url}            
\usepackage{tikz}

\mathchardef\mhyphen="2D

\title{Sapinet: A sparse event-based spatiotemporal oscillator for learning in the wild}

\author{
 Ayon Borthakur \\
  Field of Computational Biology\\
  Cornell University\\
  Ithaca, NY 14853 \\
  \texttt{ab2535@cornell.edu} \\
}

\newcommand\copyrighttext{%
\footnotesize
This work is a reproduction, with minor modification, of Chapter 5 from the author's Ph.D. dissertation, entitled "Mechanisms and Architectural Priors for Learning in the Wild" (Cornell University, Ithaca, NY, USA). ProQuest Publication Number: 28652661. Submission Date: 2021-07-28. Creative Commons License: Attribution-NonCommercial-NoDerivs (CC BY-NC-ND).  Copyright \textcopyright 2021, Ayon Borthakur.}
\newcommand\copyrightnotice{%
\begin{tikzpicture}[remember picture,overlay]
\node[anchor=south,yshift=10pt] at (current page.south) {\fbox{\parbox{\dimexpr\textwidth-\fboxsep-\fboxrule\relax}{\copyrighttext}}};
\end{tikzpicture}%
}

\begin{document}
\maketitle
\begin{abstract}
We introduce Sapinet -- a spike timing (event)-based multilayer neural network for \textit{learning in the wild} -- that is:  one-shot online learning of multiple inputs without catastrophic forgetting, and without the need for data-specific hyperparameter retuning.  Key features of Sapinet include data regularization, model scaling, data classification, and denoising. The model also supports stimulus similarity mapping. We propose a systematic method to tune the network for performance. We studied the model performance on different levels of odor similarity, gaussian and impulse noise. Sapinet achieved high classification accuracies on standard machine olfaction datasets without the requirement of fine tuning for a specific dataset. 
\end{abstract}

\copyrightnotice

\section{Introduction}

Understanding spike timing-based computation and learning is of common interest for both neuroscience and neuromorphic computing. Neuromorphic computing is primarily concerned with developing AI solutions exhibiting low latency and ultra-low power consumption when deployed on neuromorphic hardware. The use of spiking neural networks (SNNs) is a common requirement of almost all of the neuromorphic computing architectures presently in use. Owing both to this event-based computation model, and to the colocalization of memory and computation in neuromorphic hardware (sharply limiting the ability to trivially find global solutions), developing algorithms for such architectures require radically different ways of thinking about algorithms. For example, neuromorphic chips such as Loihi do not support multiply accumulation operations, which are essential for most current artificial neural networks. As of now, there is a paucity of task dependent SNN based algorithms for training as well as inference ~\cite{Davies2019, christensen20212021}. Based upon some coordinated sets of mechanisms, hypothesized to be part of the functioning of the mammalian olfactory bulb, we here present an SNN-based plastic multilayer algorithm that can \textit{learn in the wild} - that is, operate effectively under several of the conditions that normally impair real-world performance. Specifically, it can effectively learn multiple odor signatures in series, based on one-shot training without catastrophic forgetting, and incorporating solutions for mitigating sensor drift, varying classifier confidence, including a \textit{None of the above} class, and identifying odors despite competitive interference (see Methods for details). 

Both the mammalian olfactory system and artificial noses (which use metal oxide, polymer, or other sensor arrays) face the common problem of identifying odors within highly occluded environments, where multiple chemical species compete for the same set of sensors, hence disrupting the cross-sensor activation patterns on which odor identification depends. In addition, odors occur in nature at different and varying concentrations, and under different temperature and humidity conditions -- all of which can alter sensor responses. Unlike contemporary artificial systems, the mammalian olfactory system is adept at identifying odors despite these challenges. Hence, we take inspiration from biology to engineer an artificial system exhibiting these advantages. The mammalian olfactory system is based fundamentally on structured plasticity, employing multiple layers of processing involving heterogeneous cell types, network connectivity patterns, and localized excitatory and inhibitory synaptic learning algorithms -- all of which can be effectively implemented on neuromorphic hardware that incorporates on-chip plasticity (e.g., Intel Loihi). In our Sapinet network, sensor data are first conditioned and regularized by a network layer inspired by the glomerular layer (the first computational layer) of the mammalian olfactory bulb. These regularized data then are transformed into a sparse spike timing-based representation measured against an underlying network rhythm, inspired by the $\gamma$-band field potential oscillations of the olfactory bulb's external plexiform layer (the bulb's second computational layer). This spike timing representation is generated in mitral cells (MCs; the principal neurons of the olfactory bulb), Fig ~\ref{mod_arch}.  The relative spike times of groups of MCs are learnt by interneurons (granule cells, GCs), which in turn learn to shape the spike timing of MCs in a recurrent feedback loop. This two-layer architecture enables the network to perform signal conditioning, data regularization, classification and also signal denoising. As SNN parameters are hard to tune for use with data of different dimensions, we also here provide a model scaling technique for the systematic tuning of SNNs with respect to sensor array size -- an essential requirement of \textit{learning in the wild}. Moreover, as shown for the signal conditioning preprocessor layer ~\cite{BorthakurCleland2019, Borthakur2019B}, we demonstrate that the Sapinet architecture can utilize rapid online learning without catastrophic forgetting as a solution for mitigating sensor drift. 

In this work, we demonstrate the core requisites of \textit{learning in the wild} -- data regularization, model scaling, rapid one-shot learning, prediction without catastrophic forgetting, and signal denoising -- on multiple datasets, both synthetic and real world, each with unique features.

For data regularization and model scaling, we first demonstrate Sapinet's effectiveness on a $100$-dimensional synthetic dataset with samples drawn from a range of distributions, including uniform, gaussian, poisson, and rayleigh. The data dimension for studying model scaling was varied from as low as $10$ to a maximum of $640$. We then show that this these model scaling and data regularization techniques are effective on two publicly available chemosensor response datasets, the UCSD Gas Sensor Array Drift Dataset at Different Concentrations Data Set ("gas sensor drift") ~\cite{UCIMachineLearningRepositoryGasSen,vergara_chemical_2012}  and the UCSD Gas Sensor Arrays in Open Sampling Settings ("wind tunnel") dataset ~\cite{UCIMachineLearningRepositoryGassen_wind,VERGARA2013462},  which provide data challenges including substantial sensor drift, wide changes in source concentrations, and unpredictable plume-based concentration variance.  In some studies, we also inserted impulse noise in the test set as a model for the unpredictable interference of background odors on net sensor responses.  Specifically, in these cases, we randomly selected a number of sensors during testing and replaced their activity values with random numbers. 

Specifically, we first trained and tested the network on a $1$-shot, $1$-epoch sequential learning task using a set of synthetically generated, sequentially similar "odor" inputs and introducing gaussian and impulse noise in the test set. We varied the degree of odor similarity, the standard deviation of the gaussian noise, and the level of occlusion (proportion of affected sensors, for both gaussian and impulse noises) over wide ranges. The network was also trained and tested on all $10$ batches of the UCSD gas sensor drift dataset ~\cite{vergara_chemical_2012} and on the odor plume variances of the UCSD wind tunnel dataset ~\cite{VERGARA2013462}. We show that the network rapidly learns odors and performs robust classification and signal denoising on all the datasets.  In addition, these performances on multiple datasets, arising from various distributions and dimensionalities, were achieved using only data regularization and model scaling without any other hyperparameter retuning required across datasets.

\section{Results}

\subsection{Network architecture}

\begin{figure}
  \centering
  \includegraphics[width=0.75\linewidth]{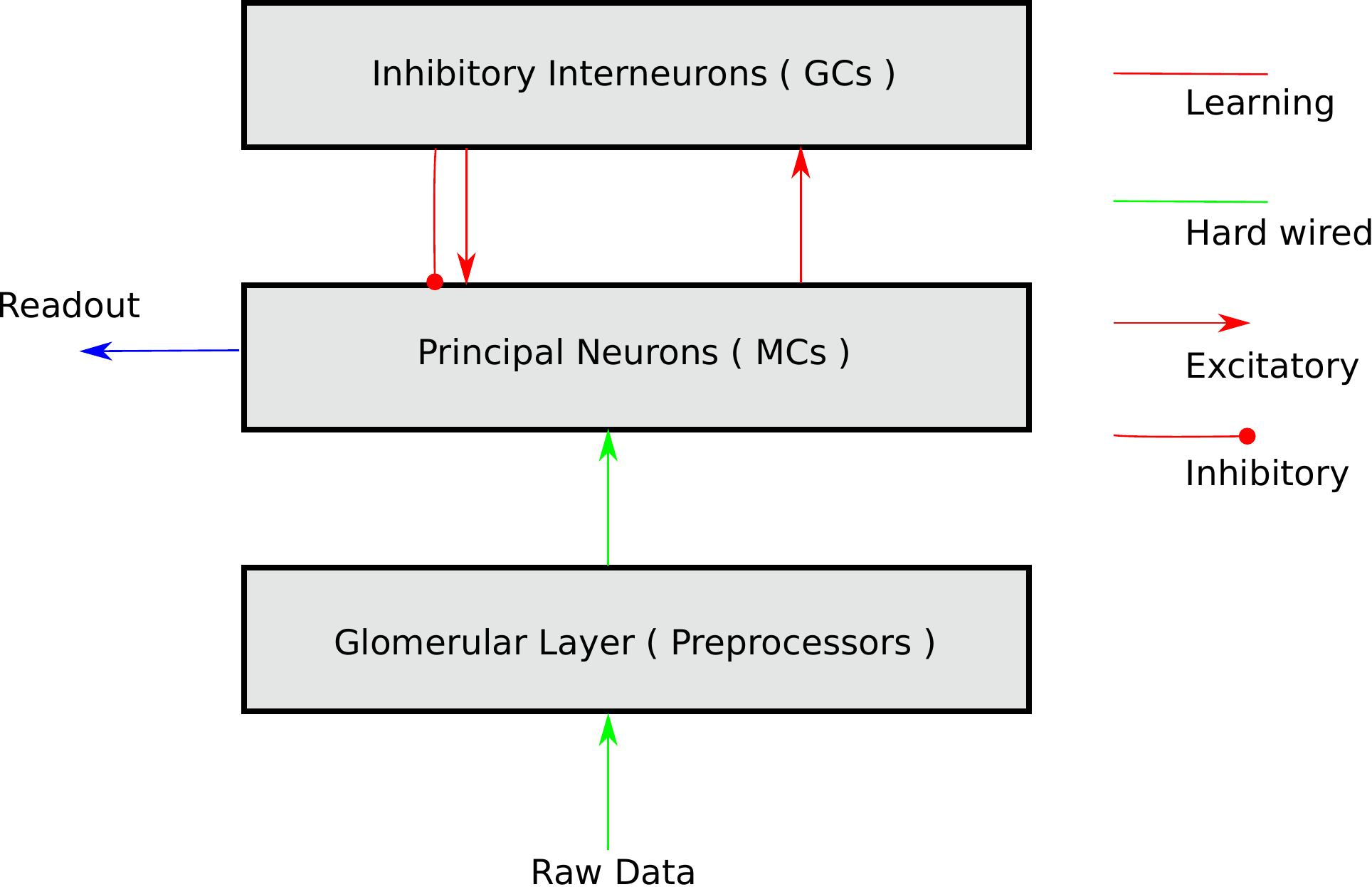}
  \caption{Sapinet model schematic}  
  \label{mod_arch}
\end{figure}

Sapinet is a multi layer recurrent neural network model inspired by the  microcircuitry of the rodent olfactory bulb ~\cite{cleland_construction_2014}. Key computational principles of rodent olfactory included in Sapinet include (also discussed in ~\cite{borthakur_neuromorphic_2017, BorthakurCleland2019, Borthakur2019B, imam_implementation_2012, imam_rapid_2019}): 
  \begin{itemize}
        \item The data preprocessing capacity of the excitatory and inhibitory neurons of the olfactory bulb glomerular layer. 
        \item The significance of sparse random projections and heterogeneous activity levels of sister mitral cells. 
        \item The importance of $\gamma$-discretized spike timing-based coding in the mitral cell somata, governed by external plexiform layer (EPL) interactions. 
        \item The ability of mitral cell dendrites to excite almost any interneurons (granule cells) thereby forming higher order receptive fields (HORFs) in granule cells using local excitatory learning -- specifically, asymmetric spike timing dependent plasticity (STDP). 
        \item Local (column-specific) inhibitory effects of granule cells on mitral cells. 
        \item Role of the recurrent external plexiform layer (EPL) in the denoising of signals.   
        \item The importance of adult neurogenesis in maintaining lifelong learning without catastrophic forgetting. 
        \item The importance of parameter heterogeneity in robust network performance.  
   \end{itemize}  

In the mammalian olfactory system, odors sensed by the genetically coded receptors are delivered to the glomerular layer -- the first computational layer of the olfactory bulb. Likewise, raw sensor data are delivered to the neurons of the Sapinet glomerular layer (Fig ~\ref{mod_arch}).  A glomerular circuit in Sapinet is comprised of multiple excitatory external tufted (ET) cells (neurons) and a single inhibitory periglomerular (PG) cell, both of which target the excitatory apical dendrites of olfactory bulb principal neurons, known as mitral cells (MCs).  MC apical dendrites in Sapinet are a functional unit and are abbreviated as ApiMCs.  Odor stimuli excite both ET cells and PG cells simultaneously. The ET cells can excite ApiMCs of only the respective glomerulus whereas the PG cells can mediate inter-glomerular inhibition by delivering inhibition to any of the ET cells in the network.  The ApiMC activation pattern then is presented to the network for multiple $\gamma$ cycles (in the present model, $8$ gamma cycles are presented, each representing $25 ms$ of biological time) for a total "sniff" duration of $200 ms$ in both the training and testing phases. The activation level of each ApiMC excites the corresponding MC soma (MCsoma) on each gamma cycle, which generates action potentials based on this afferent excitation.  Specifically, in the MCsoma, ApiMC excitation levels are transformed into a $\gamma$ precedence code, with higher input levels inducing the MCsoma to fire earlier on each recurring cycle of the $\gamma$ oscillation sequence. Different odors, consequently, generate different MCsoma population spike timing codes. These external input driven network activity levels (in ApiMC compartments) are reset at the end of every $\gamma$ cycle, inspired by the phasic inhibition of MCs by GCs. 

MCs and GCs reciprocally connect to one another in a particular configuration.  Specifically, as the MC dendrites extend broadly across the biological EPL, in Sapinet any GC in the network is equally likely to be excited by a spike emitted by any given MCsoma. We modeled the MCsoma to GC connection pattern as sparse random excitatory synapses with local learning (asymmetric STDP, see Methods). Similar to previous works ~\cite{BorthakurCleland2019, Borthakur2019B}, these synapses were modeled as double exponentials.  The effect of this plasticity model was that, during training, granule cells (GCs) formed higher order receptive fields, learning to respond only when particular sets of MCs fired together, hence becoming selective to odor-diagnostic patterns among specific groups of co-activated MCs.  

In contrast, GCs delivered their inhibitory effect only onto MCs within their column (i.e., the MC associated with the same glomerulus, and hence the same sensor.  GCs were associated with a specific column for this purpose only).  We modeled the granule cell (GC) to MCsoma connection as a mechanism for applying \textit{inhibitory drive} (see Methods for details). Briefly, the spiking GCs of a column learn spike timings of sister MCs of a column for an odor. During any training/testing $\gamma$ cycle, the spike timing of an MCsoma is determined by the level of afferent input received from its ApiMC as well by the \textit{inhibitory drive} applied by the GCs of that column. During testing, if a noisy odor -- with a signal impaired by any of multiple sources of error -- is presented to the network as input, the GCs work with MCs iteratively across multiple $\gamma$ cycles to reconstruct the relevant trained MC spike timing pattern. For classification, the spike timing of MCsoma is compared with all of the trained MCsoma timing patterns. 

\subsection{Spike timing encoding}

\begin{figure}
  \centering
  \includegraphics[width=1.0\linewidth]{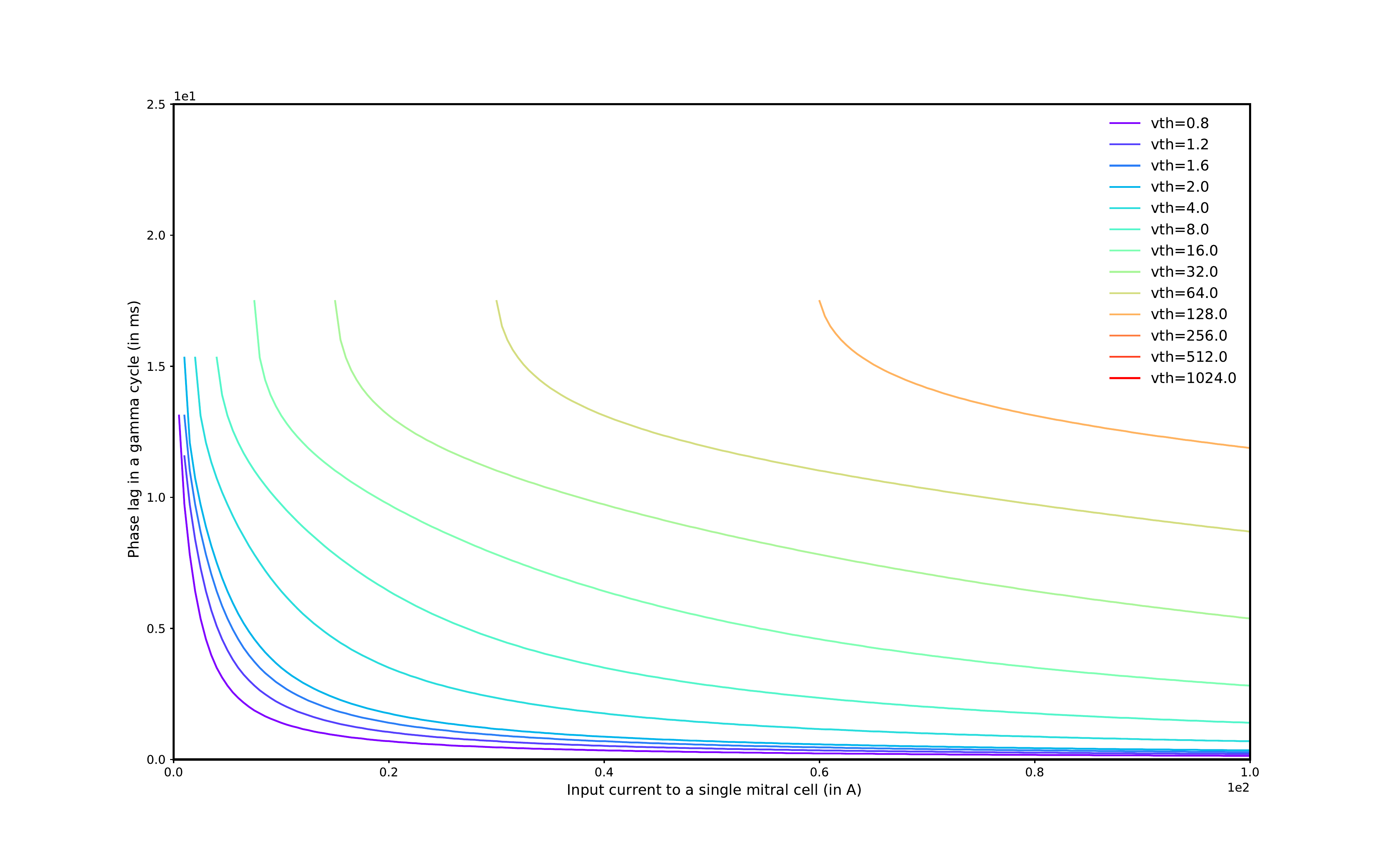}
  \caption{Variation of Mitral cell spike phase lag w.r.t input current for different spiking thresholds ($v_{th}$)}  
  \label{phase_mc}
\end{figure}

Using a single glomerular network model, we observed the variation of MC spike timing w.r.t input current level for different levels of MC spiking thresholds ($v_{th}$), keeping everything else constant. \\
The spike timing ( phase ) of spiking neurons is dependent on threshold and input current (Fig ~\ref{phase_mc}). In our $\gamma$ precedence coding scheme, a very high threshold and/or low input current prevents neurons from spiking. Similarly, very low threshold and/or very high input current causes neurons to spike too early. From a population coding perspective, both are undesirable. Hence, there is a need to regularize the \textbf{input current} and the \textbf{spiking threshold}. For our \textit{learning in the wild} scenario, we decided to keep the MC thresholds unchanged across datasets. Fig ~\ref{phase_mc} indicates that the mitral spike phase lags are unchanged ($~1.5 \, ms$) when the input current is greater than $~20A$. Hence, for Sapinet, we set the MC input current range to $0-20 A$ approximately. 

\subsection{Data regularization}

Spiking neural networks are sensitive to the absolute value of inputs received. If the inputs are too low, in a spike precedence coding regime, there might not be any spike at all. When communication in a network is designed to be through spike events, this will be a problem. For example, there might be a situation where no neurons spike for a sample which will deter the model's classification performance. The problem becomes much more severe while \textit{learning in the wild} as data statistics are uncertain. Resorting to hyperparameter tuning is infeasible as it compromises with algorithm's rapidity. We propose, \textbf{data regularization} as a better alternative solution - so that the network works just fine for data sets with different statistics. 

The goal of data regularization is to ensure - 
\begin{itemize}
    \item Neurons spike for all samples
    \item Neuron spike count fraction is never above a set threshold ( 1 indicates all neurons spiked ). 
    \item The fraction of neurons that spike across samples is consistent ( the neurons that actually spike can be different across samples )
\end{itemize}

We studied data regularization in terms of spike counts of MCs and GCs across samples. 

To assess the consistency of MC and GC spike counts, the goodness of preprocessing as described in Borthakur \& Cleland ~\cite{Borthakur2019B} is used with additional constraint - spike counts of MCs / GCs cannot be greater then a set $threshold$ ( set to  $0.9$ here. )

Mathematically,

\begin{equation}
\begin{split}
\centering
& Goodness \,of \,processing(g_{p}) = No \,spike\, penalty\, \times Sparsity\,factor\, \times Spike\, count \,similarity\\
& No \,spike\, penalty = min(v)\\
& Sparsity\,factor\, = \left\{
    \begin{array}{ll}
      1, & \mbox{if $max(v) < threshold$}.\\
      0, & \mbox{otherwise}.
    \end{array}
  \right.\\
& Spike\, count \,similarity = \frac{\sum{\frac{v_{i}}{max(v)}}}{n}\\
\end{split}
\end{equation}

where $v$ is the spike count of neurons across $n$ samples. \\
For MCs of dimensions $100$ and $5$ samples, length of $v$ will be $5$ and each entry will be between $0 - 100$. \\

In order to study the requirement of data regularization in spike timing based coding, we generated $40$ synthetic odor samples from various distributions together comprising a wild scenario as described below:
\begin{itemize}
    \item Samples $1-9$ were drawn from folded normal distributions with three different means and three standard deviations for each mean. 
    \item Samples $10-18$ were the same as above, scaled by a factor mimicking odors at different concentrations. 
    \item Samples $19-21$ and $22-24$ were uniform and power distribution respectively with three different means. 
    \item Samples $25-29$ were poisson distributions varying w.r.t mean. 
    \item $30-32$ were drawn from Rayleigh distributions varying w.r.t mean. 
    \item For samples $33-37$, the sensor responses varied linearly w.r.t sensor indices as $y=mx$, where $x$ is the sensor index and $y$ being the sensor response, $m$ being five different slope values. 
    \item For sample $38$ - all sensor responses were set to $200$, for sample $39$ - first $25\%$ of sensor responses were set to $5$, and for sample $40$ - first $75\%$ of samples were set to $5$. 
\end{itemize}
\subsubsection{Network output for raw data}

\begin{figure}
  \centering
  \subfloat[]{\includegraphics[width=0.49\linewidth]{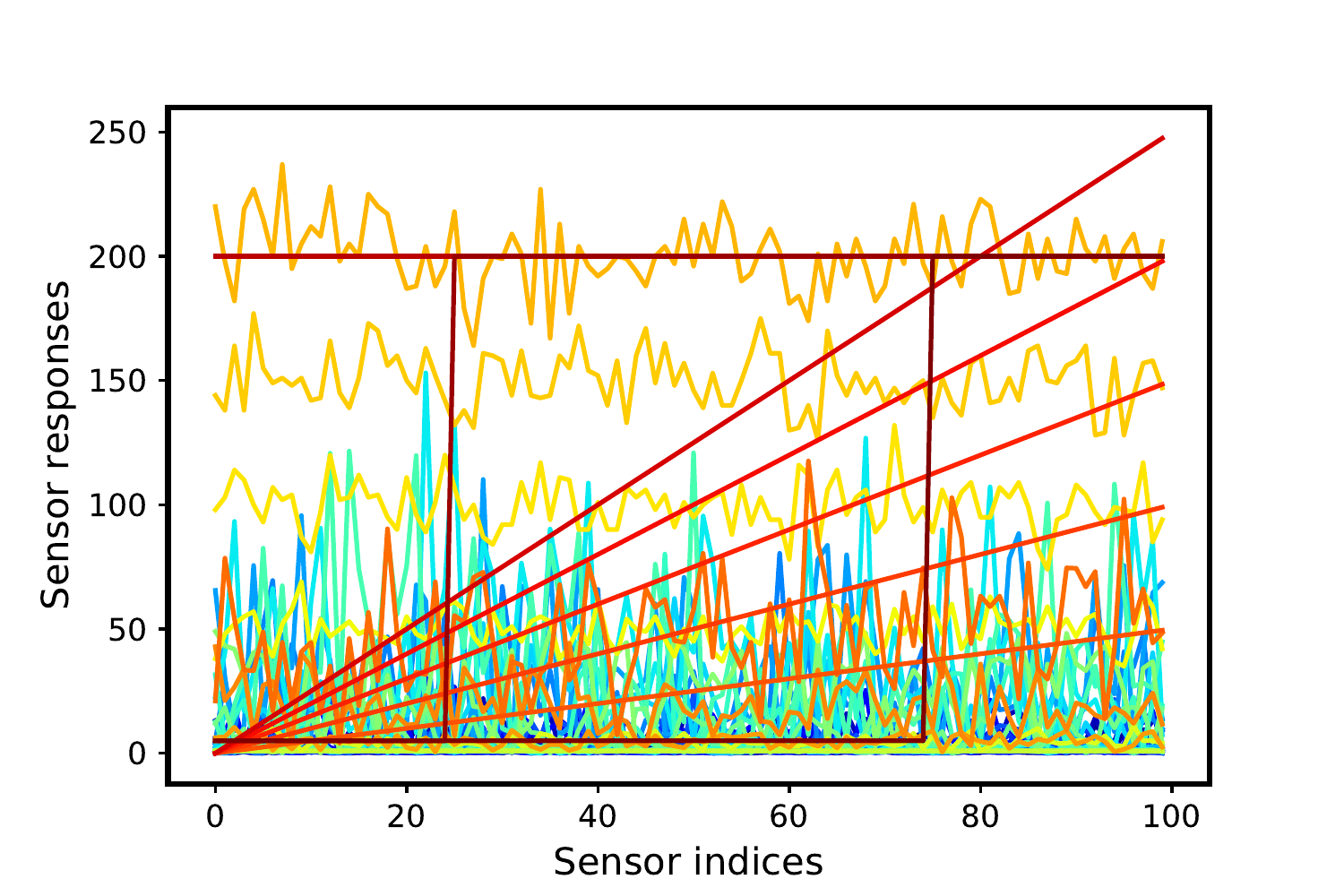}}
    \hspace{0.01in}
\subfloat[]{\includegraphics[width=0.49\linewidth]{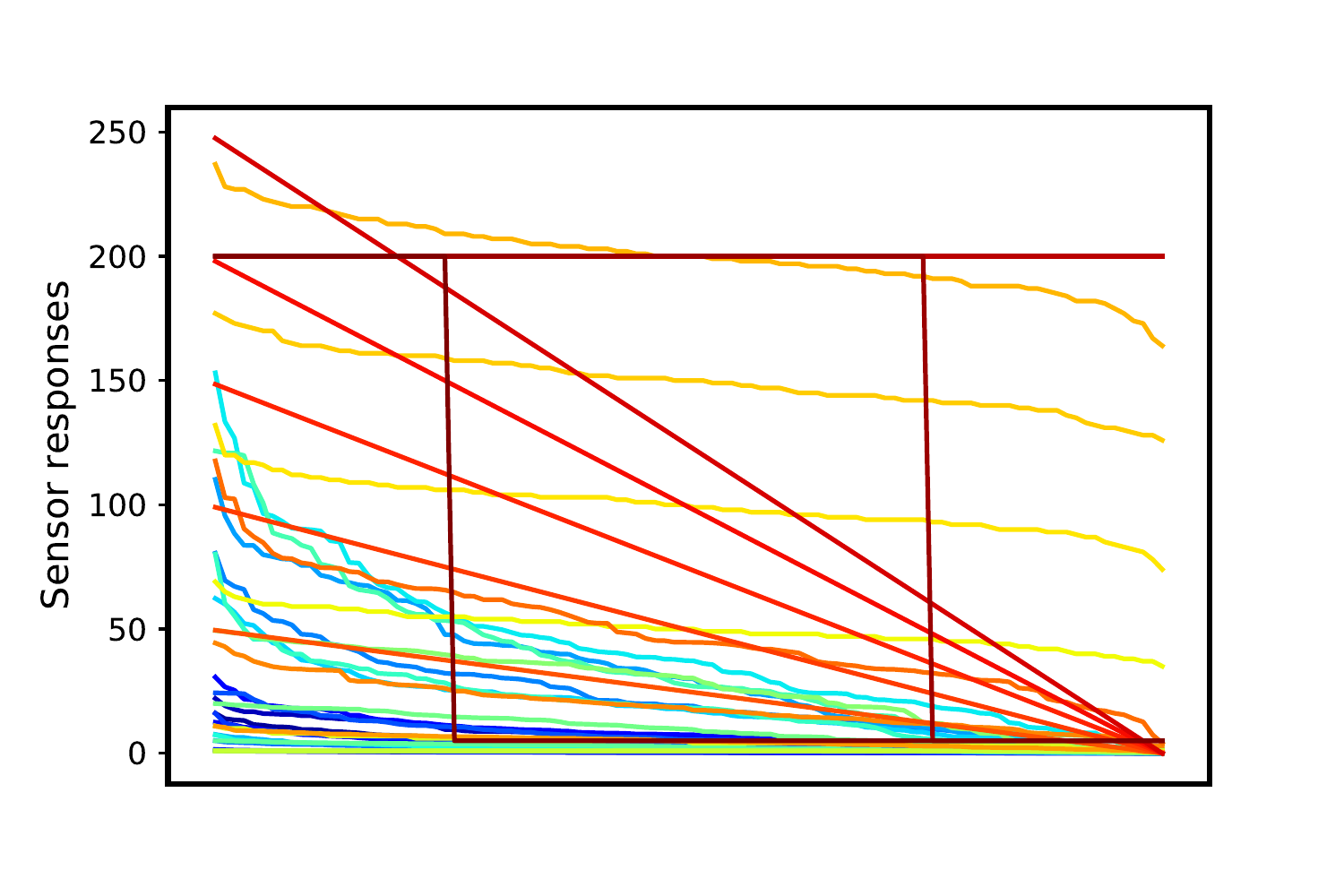}}
\hspace{0.01in}

  \caption{$40$ wild odor samples drawn from various distributions such as normal, poisson etc ( details in text). a) Raw synthetic sensor response data. b) Same as a but with responses sorted by amplitudes. }  
  \label{raw_synth}
\end{figure}

Fig ~\ref{raw_synth}describes the $40$ wild odor samples with hotter colors corresponding to samples with higher indices. Fig ~\ref{raw_synth}b describe the same set of odors where all sensor responses are sorted by amplitude - useful for discussing the importance of data regularization. 

\begin{figure}
  \centering
  \subfloat[]{\includegraphics[width=0.32\linewidth]{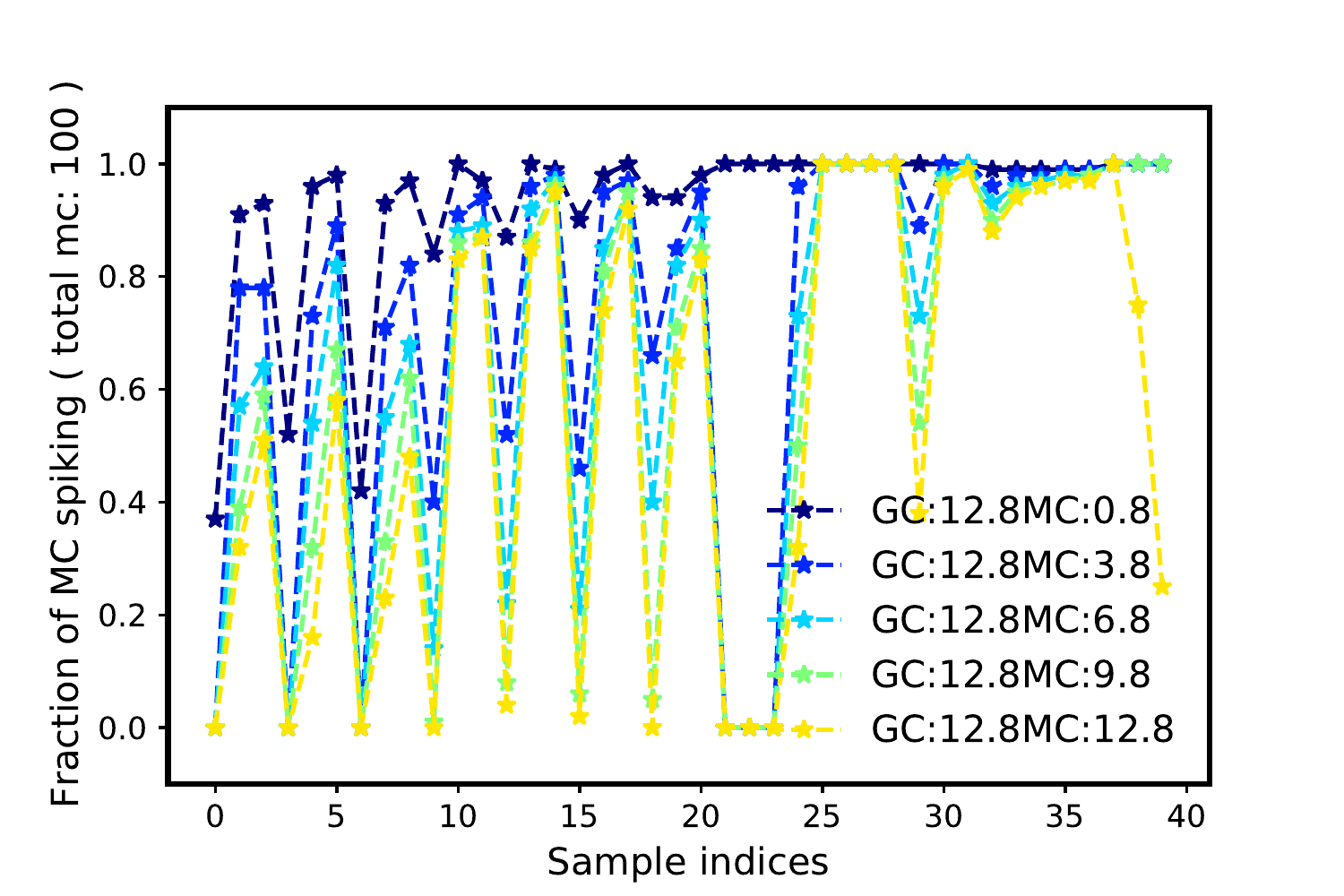}}
    \hspace{0.01in}
\subfloat[]{\includegraphics[width=0.32\linewidth]{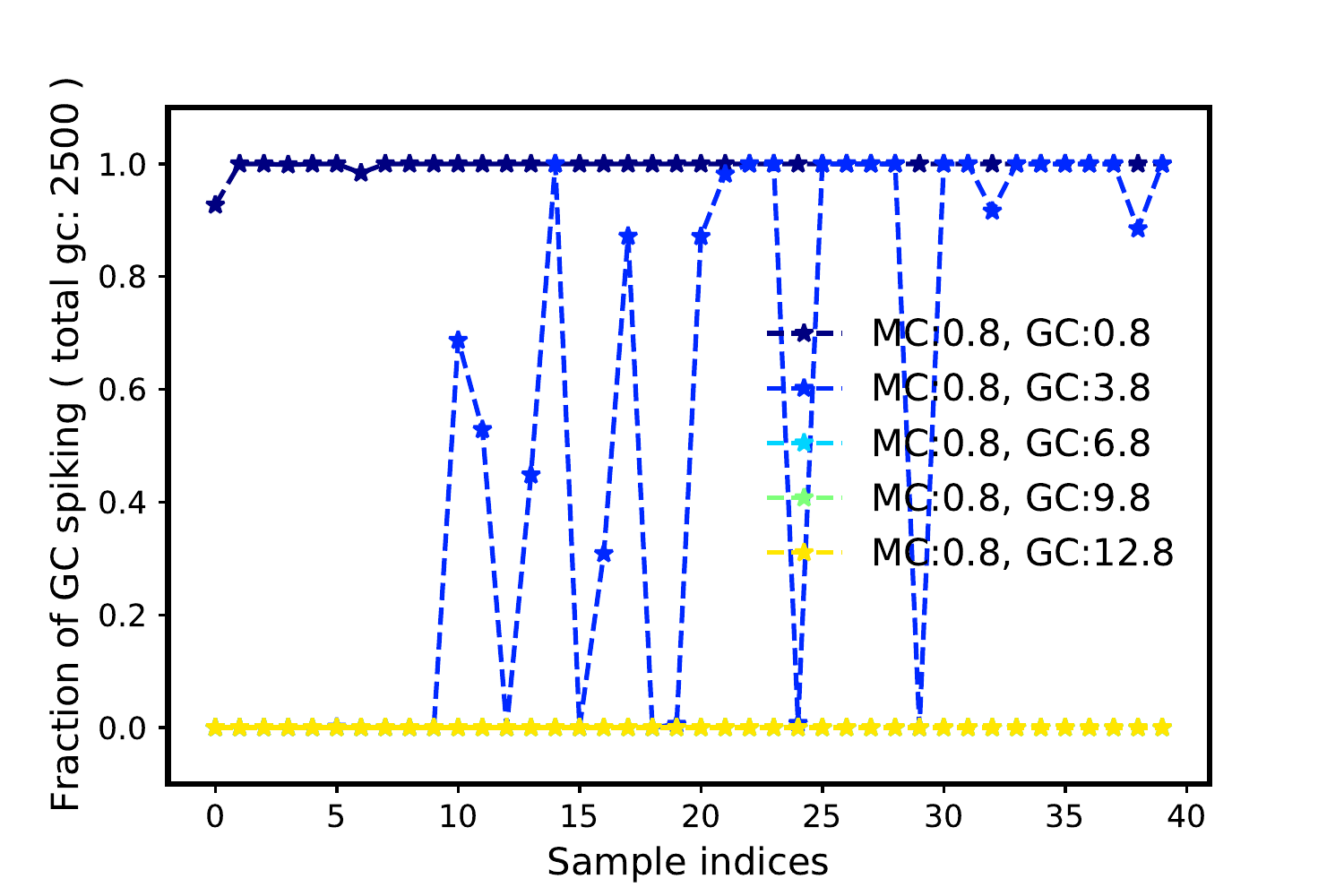}}
\hspace{0.01in}
\subfloat[]{\includegraphics[width=0.32\linewidth]{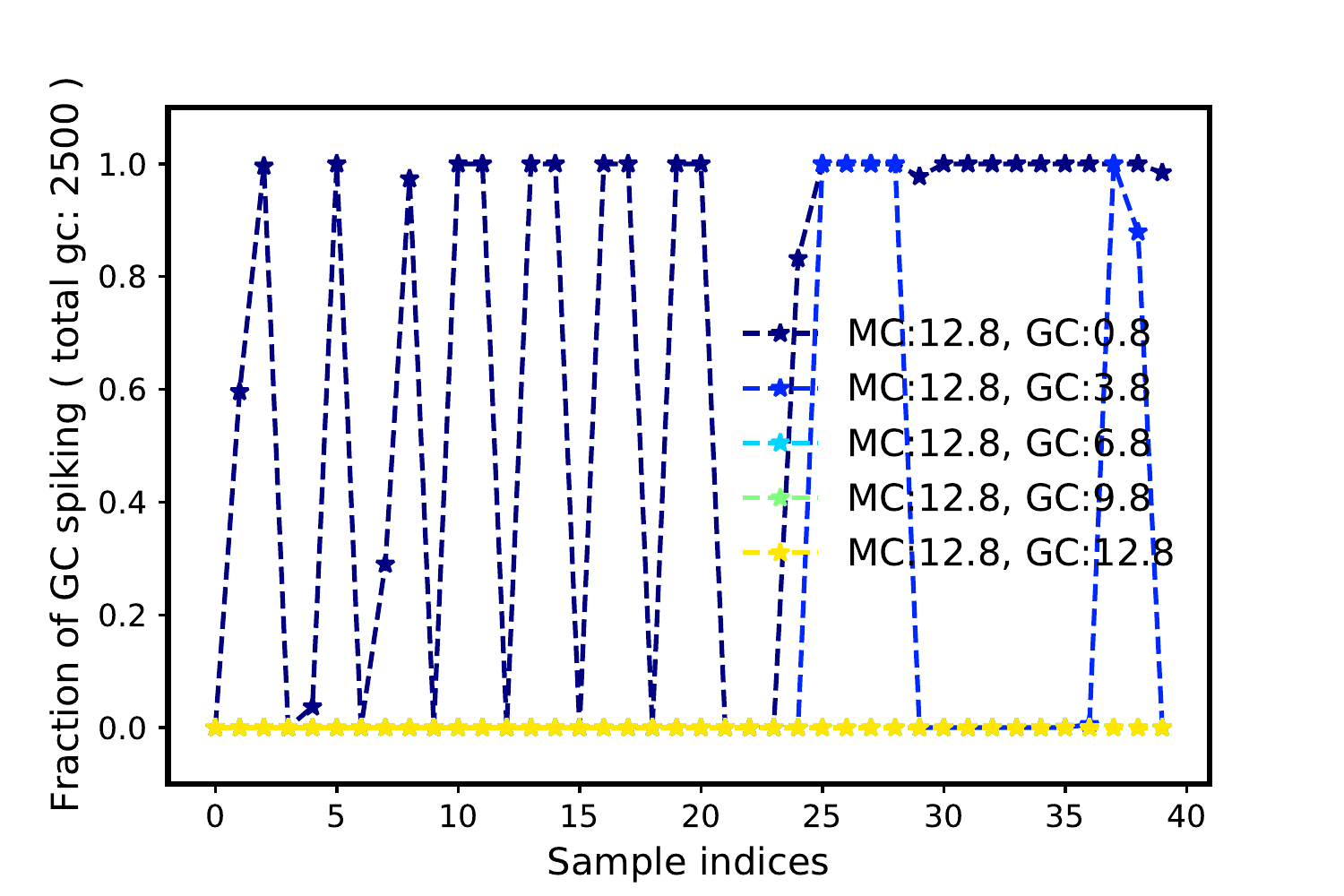}}
\hspace{0.01in}

  \caption{Raw data - Fraction of active MC/GCs w.r.t samples in a feedforward MC-GC network. Legends indicate MC/GC spiking thresholds. }  
  \label{raw_cnt}
\end{figure}

Fig ~\ref{raw_cnt}a, b,c shows that when raw data is fed directly as current as current inputs to MCs, the fraction of active MCs/GCs are highly inconsistent across samples for different values of MC/GC spiking thresholds. The goodness of preprocessing value, $g_{p}$ was zero, Table~\ref{tab:mc_gp}, ~\ref{tab:gc_gp}. 

This scenario is undesirable in a spike event based communication network. For example, if the network is implemented as an anomaly detector, no neuron spike scenario will fail to trigger an anomaly alarm. 

\subsubsection{Network output for preprocessed data}

\begin{figure}
  \centering
  \subfloat[]{\includegraphics[width=0.49\linewidth]{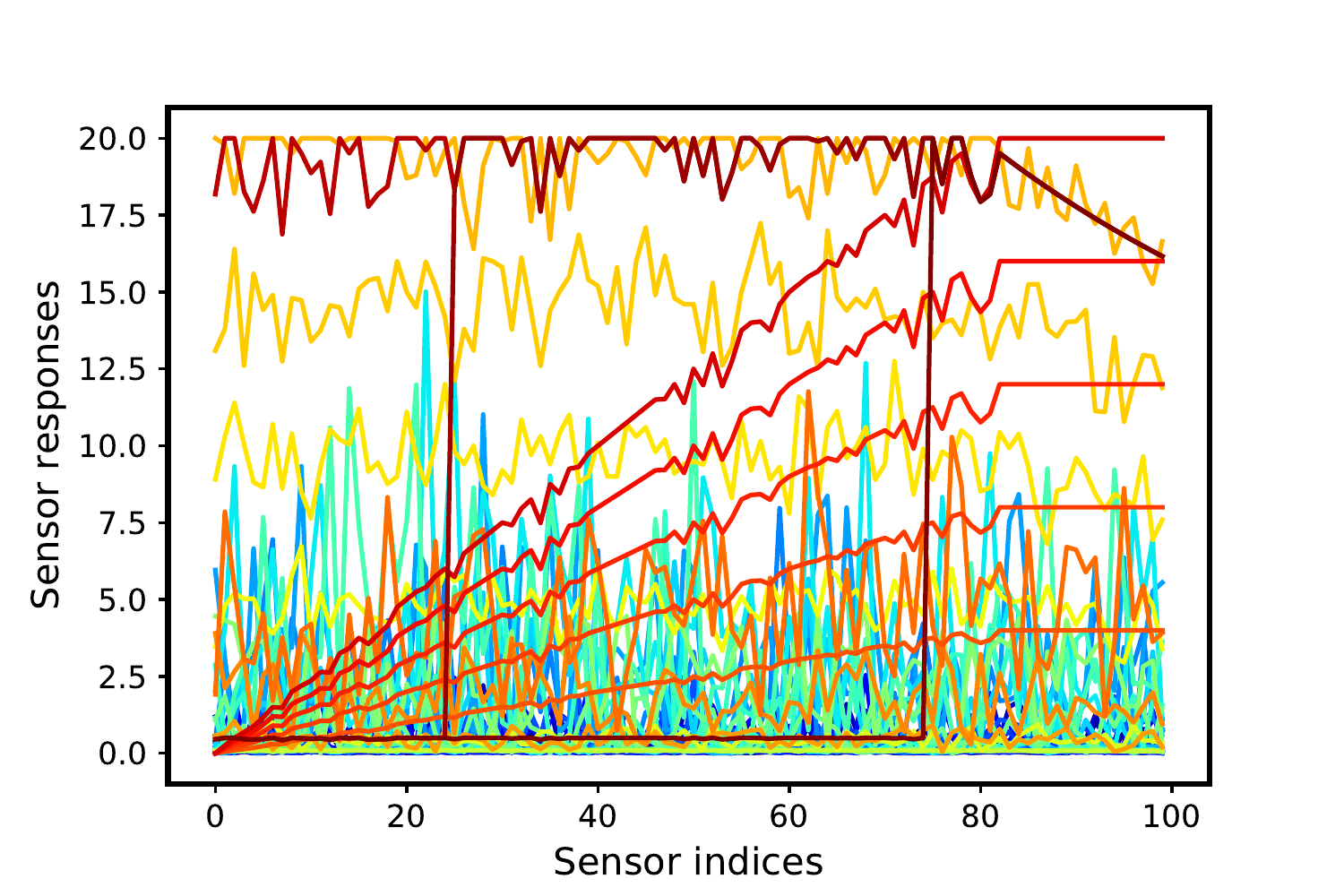}}
    \hspace{0.01in}
\subfloat[]{\includegraphics[width=0.49\linewidth]{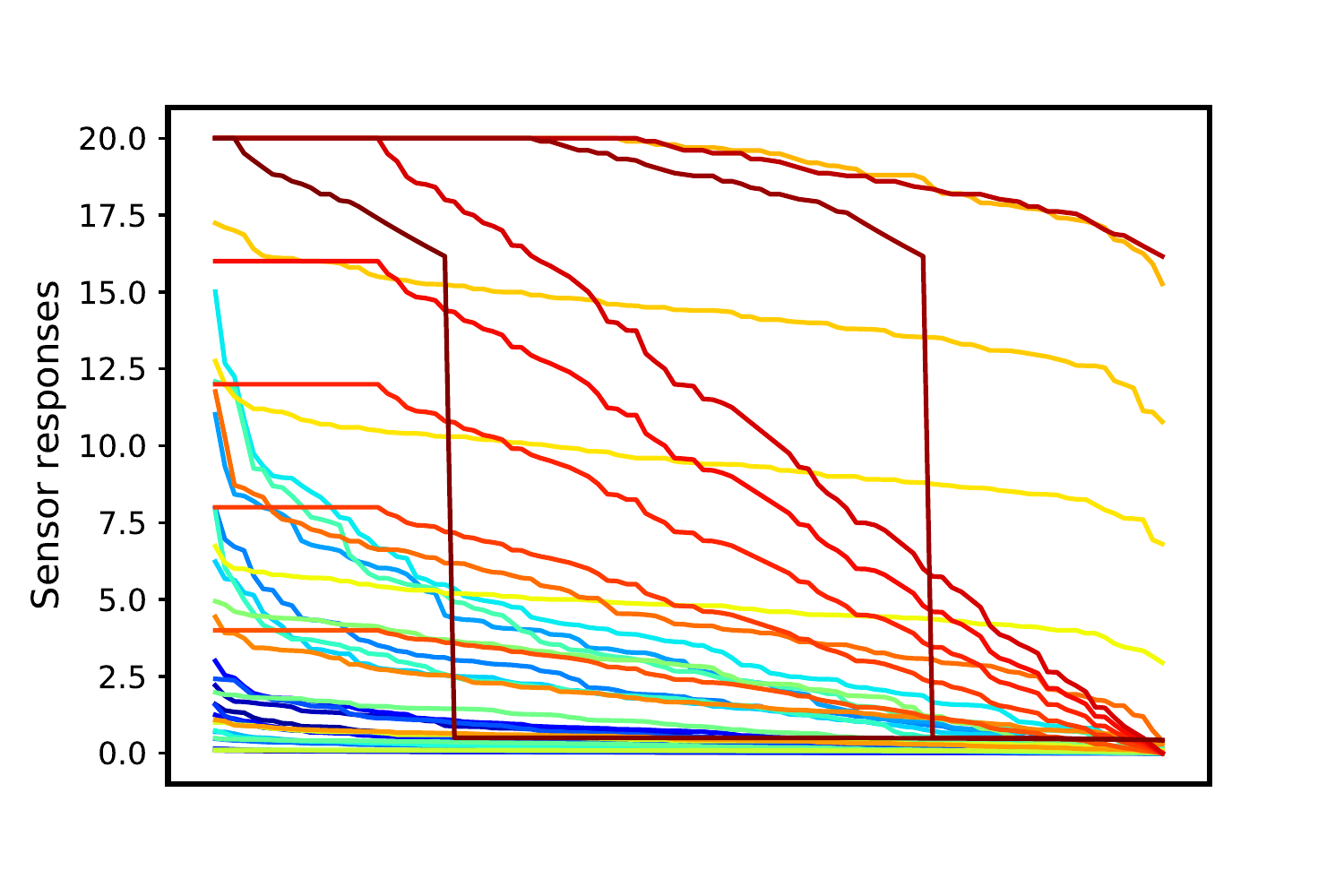}}
\hspace{0.01in}

  \caption{$40$ wild odor samples drawn from various distributions such as normal, poisson etc ( details in text) after application of sensor scaling. a) Scaled synthetic sensor response data. b) Same as a but with responses sorted by amplitudes.}  
  \label{scaled_synth}
\end{figure}

\begin{figure}
  \centering
  \subfloat[]{\includegraphics[width=0.49\linewidth]{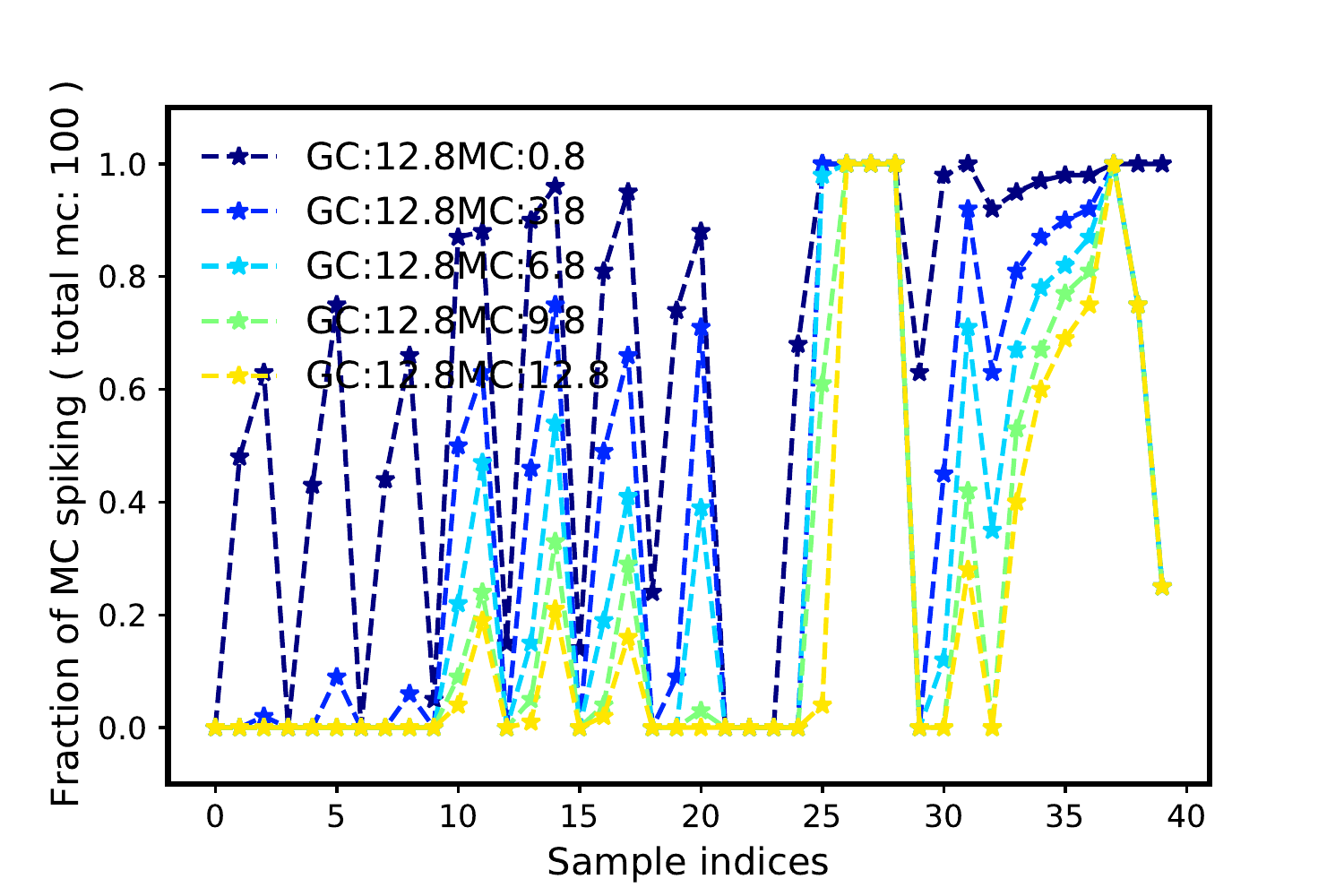}}
    \hspace{0.01in}
\subfloat[]{\includegraphics[width=0.49\linewidth]{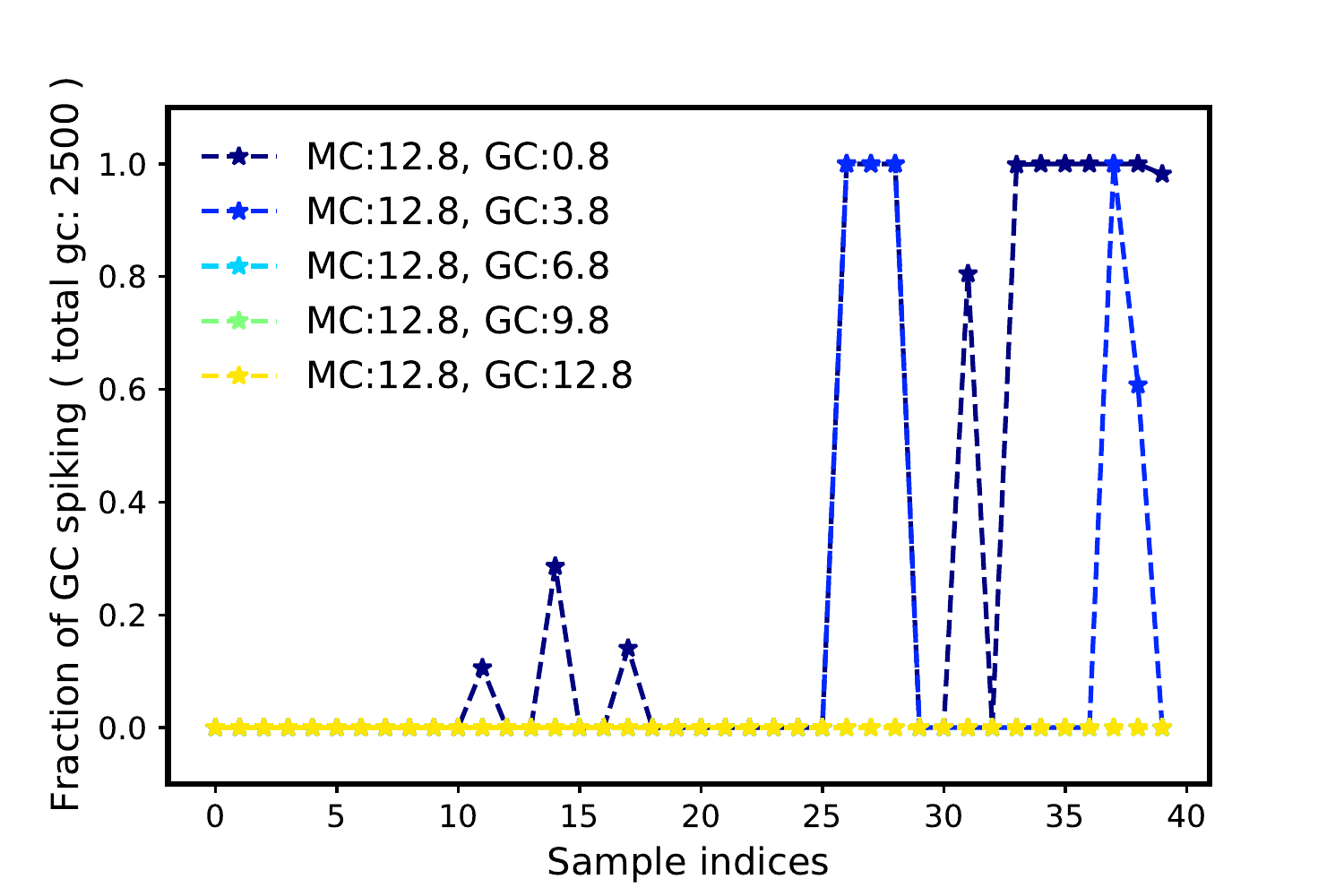}}
\hspace{0.01in}

  \caption{Scaled data - Fraction of active MC/GCs w.r.t samples in a feedforward MC-GC network. Legends indicate MC / GC spiking thresholds. }  
  \label{scaled_cnt}
\end{figure}

\paragraph{Scaled data}
In this well known step, the raw data is scaled to be within the desirable range of MC inputs $(0-20)$.
Similar to raw data, scaling didn't improve MC/GC spike count similarity, Fig ~\ref{scaled_synth}, ~\ref{scaled_cnt}a, b and hence the goodness of preprocessing value, $g_{p}$ was zero, Table~\ref{tab:mc_gp}, ~\ref{tab:gc_gp}. 

\paragraph{Unsupervised concentration tolerance}

Odors in the natural environment occur at different concentrations. It's impossible to train the model with all levels of concentration variants. Unsupervised concentration tolerance / intensity normalization serves to normalize the sensor responses across all concentrations. Details of these implementations have been discussed in our earlier work ~\cite{Borthakur2019B}. 

Mathematically,

\begin{equation}
    X^{d} = \frac{X^{d}}{\sum \, X^{d}}
\end{equation}
where $X^{d}$ is a $d$ dimensional odor sample.

\begin{figure}
  \centering
  \subfloat[]{\includegraphics[width=0.49\linewidth]{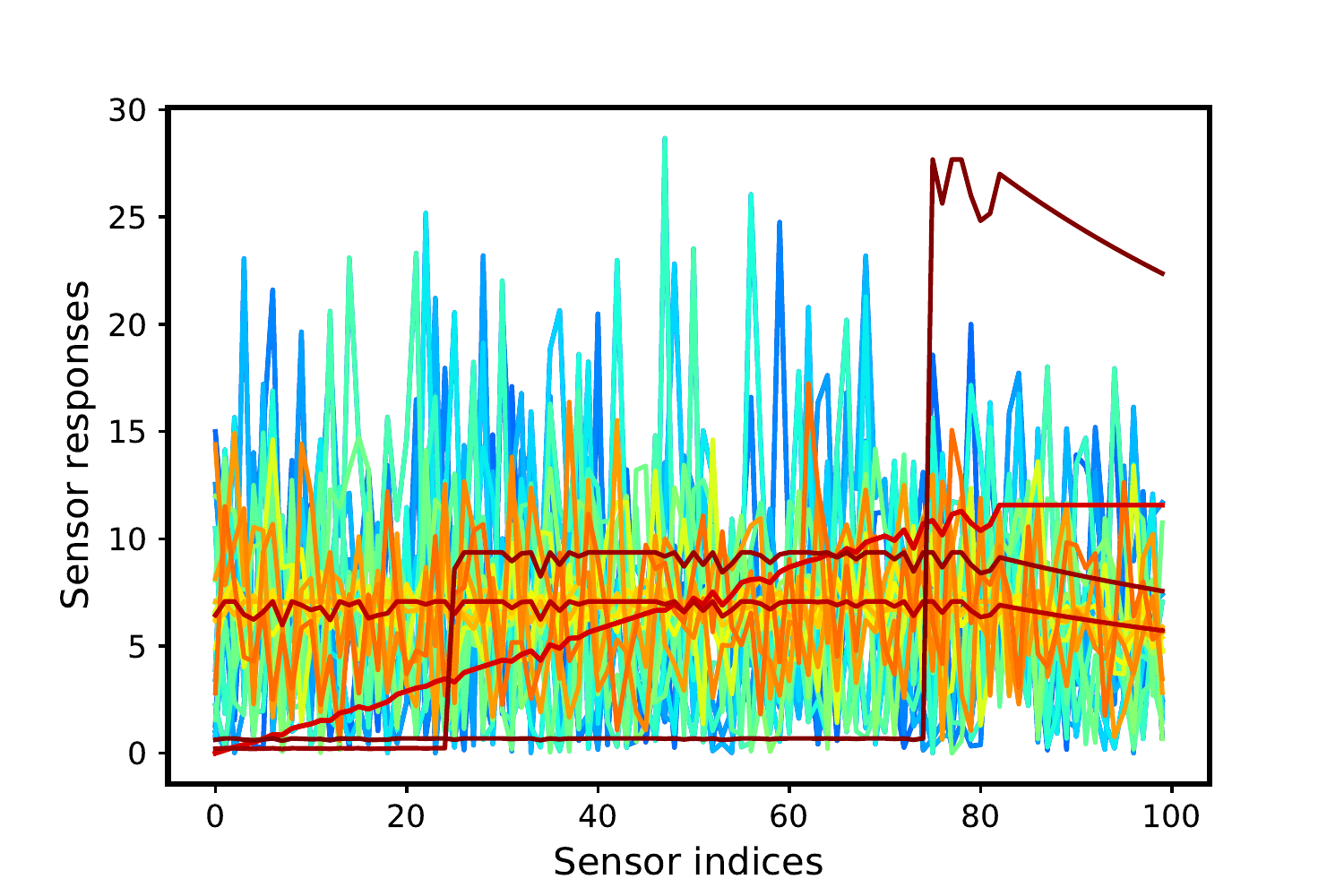}}
    \hspace{0.01in}
\subfloat[]{\includegraphics[width=0.49\linewidth]{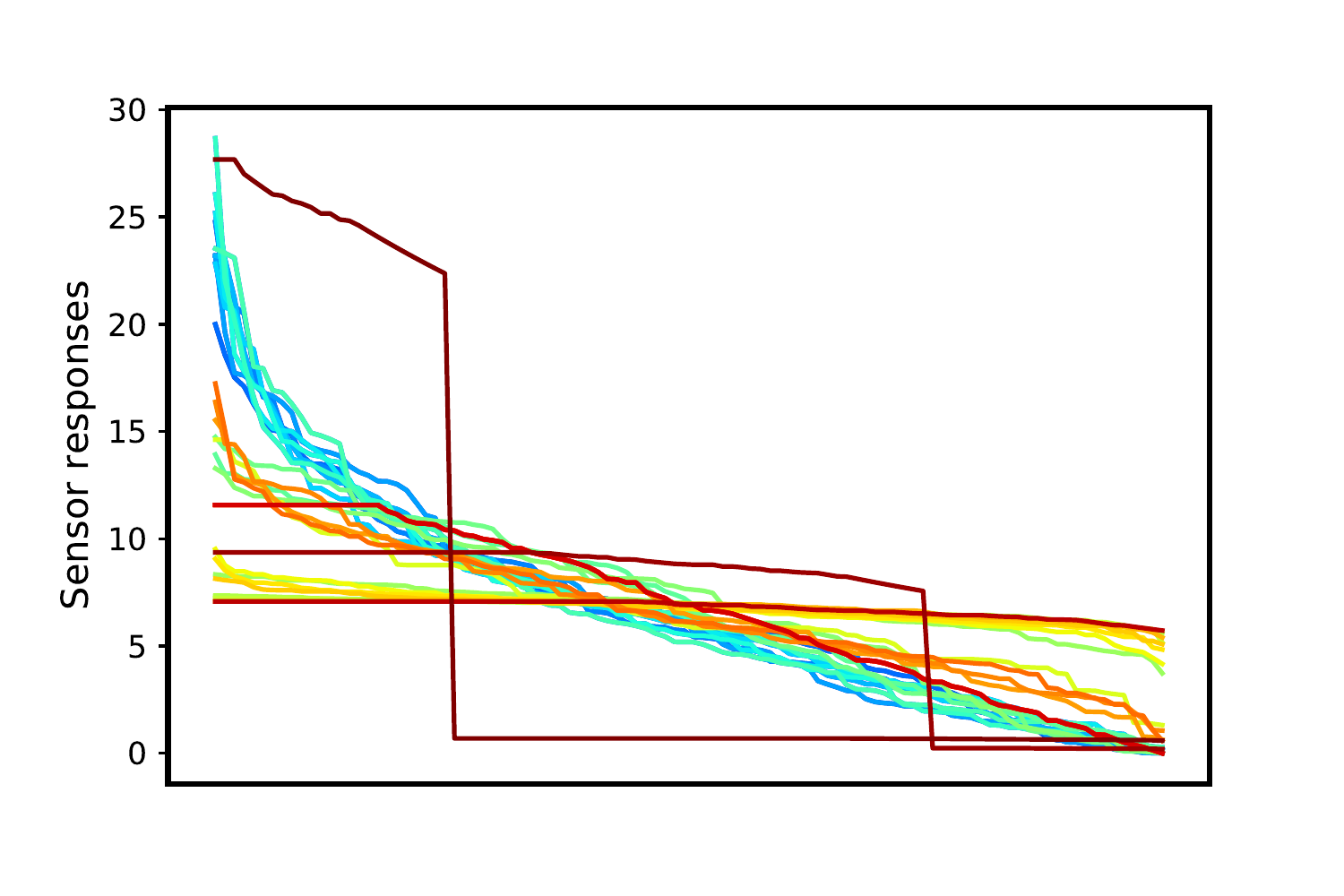}}
\hspace{0.01in}

  \caption{$40$ wild odor samples drawn from various distributions such as normal, poisson etc ( details in text) after application of unsupervised concentration tolerance. a) Concentration normalized synthetic sensor response data. b) Same as a but with responses sorted by amplitudes.}  
  \label{cn_synth}
\end{figure}

\begin{figure}
  \centering
  \subfloat[]{\includegraphics[width=0.49\linewidth]{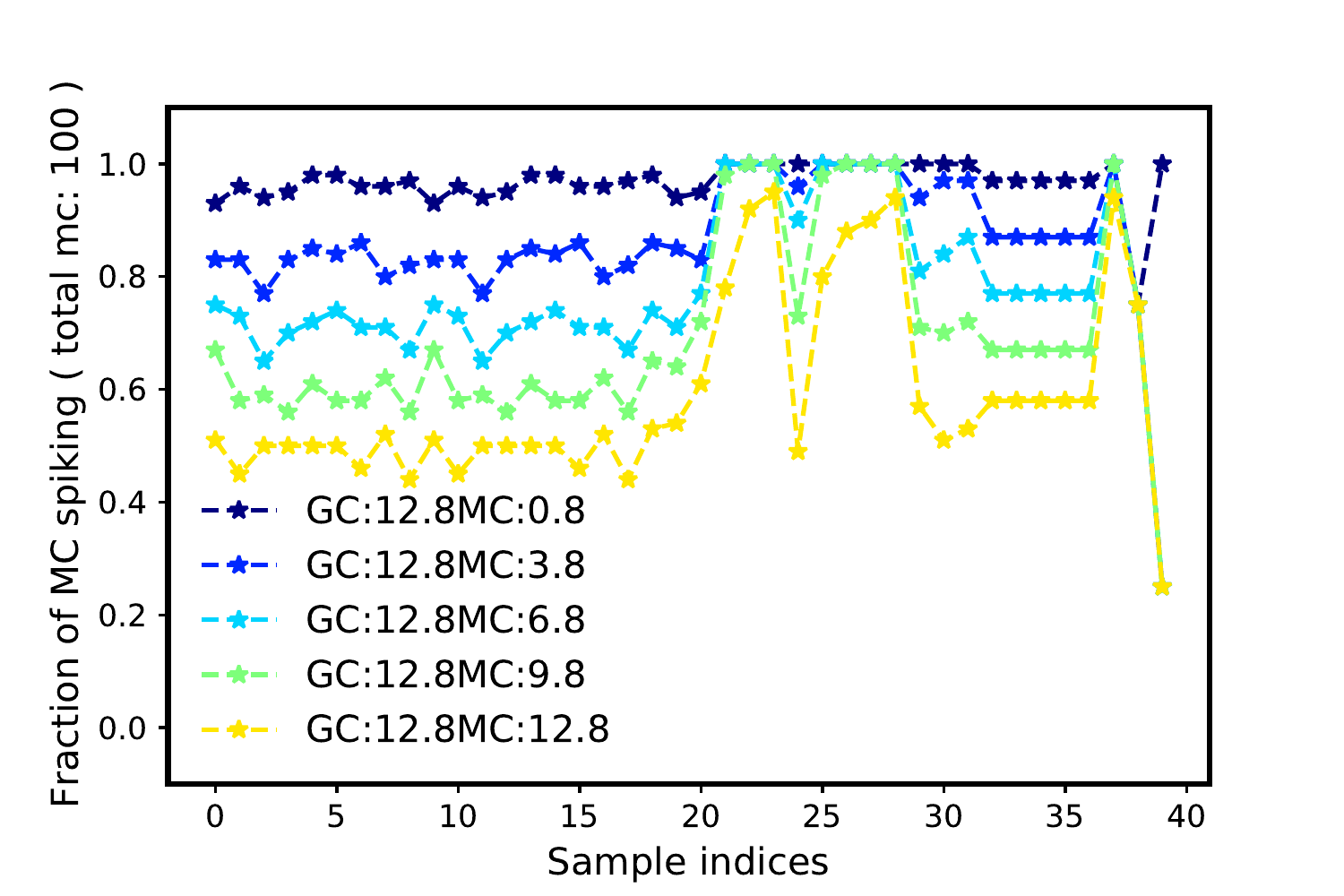}}
    \hspace{0.01in}
\subfloat[]{\includegraphics[width=0.49\linewidth]{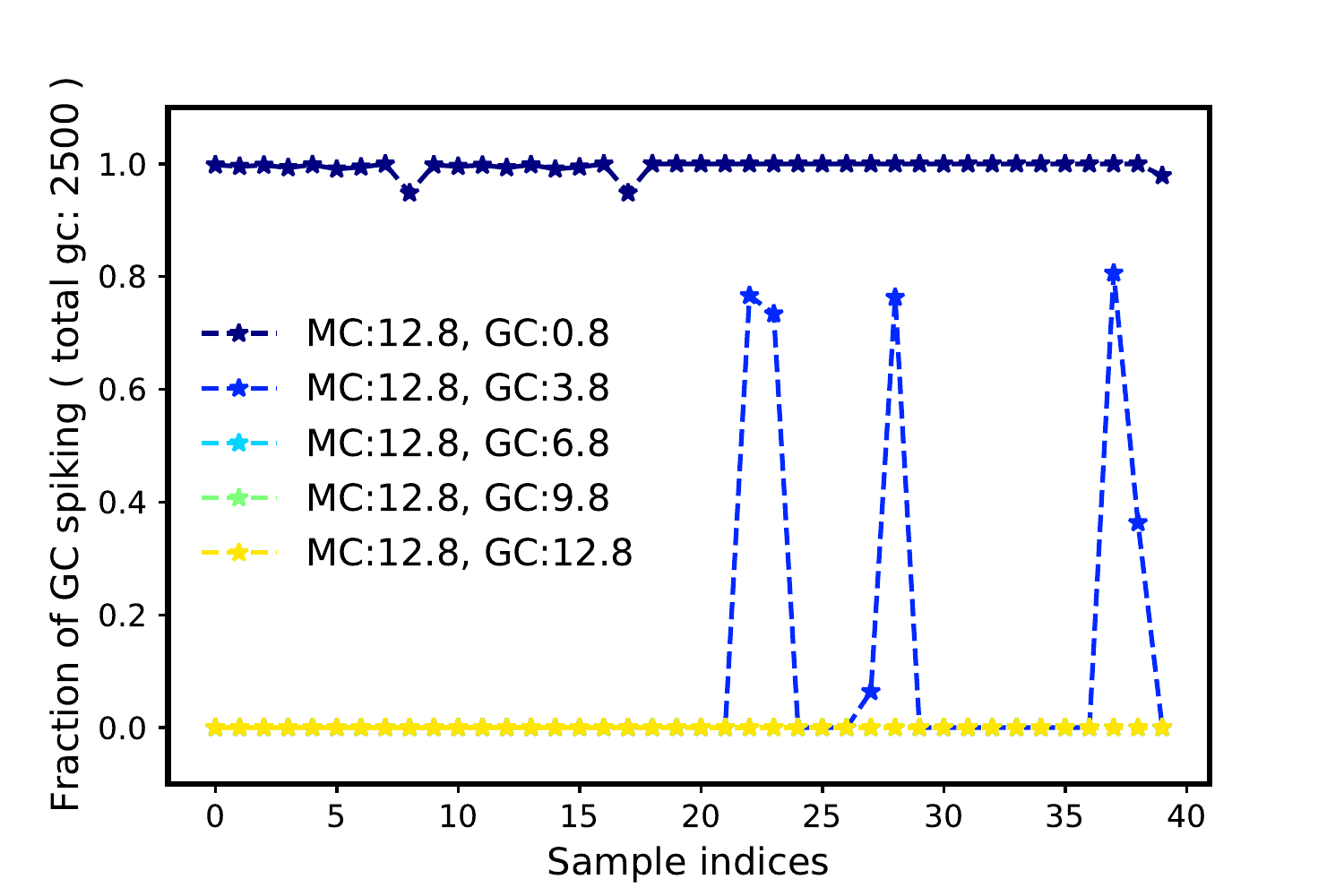}}
\hspace{0.01in}

  \caption{Normalized data - Fraction of active MC/GCs w.r.t samples in a feedforward MC-GC network. Legends indicate MC / GC spiking thresholds. }  
  \label{cn_cnt}
\end{figure}

Fig ~\ref{cn_synth} shows the data after application of unsupervised concentration tolerance. Since, samples $1-9$ and $10-18$ are the same but at different concentrations, the spike count distribution trajectory for MC/GC, Fig ~\ref{cn_cnt} from $1-9$ is similar to $10-18$. The spike count similarity did improve for MC compared to previous steps. The goodness of preprocessing value, $g_{p}$ was zero, Table~\ref{tab:mc_gp}, ~\ref{tab:gc_gp}.

\subsubsection{Network output after duplication and threshold heterogeneity}

\begin{figure}
  \centering
  \subfloat[]{\includegraphics[width=0.49\linewidth]{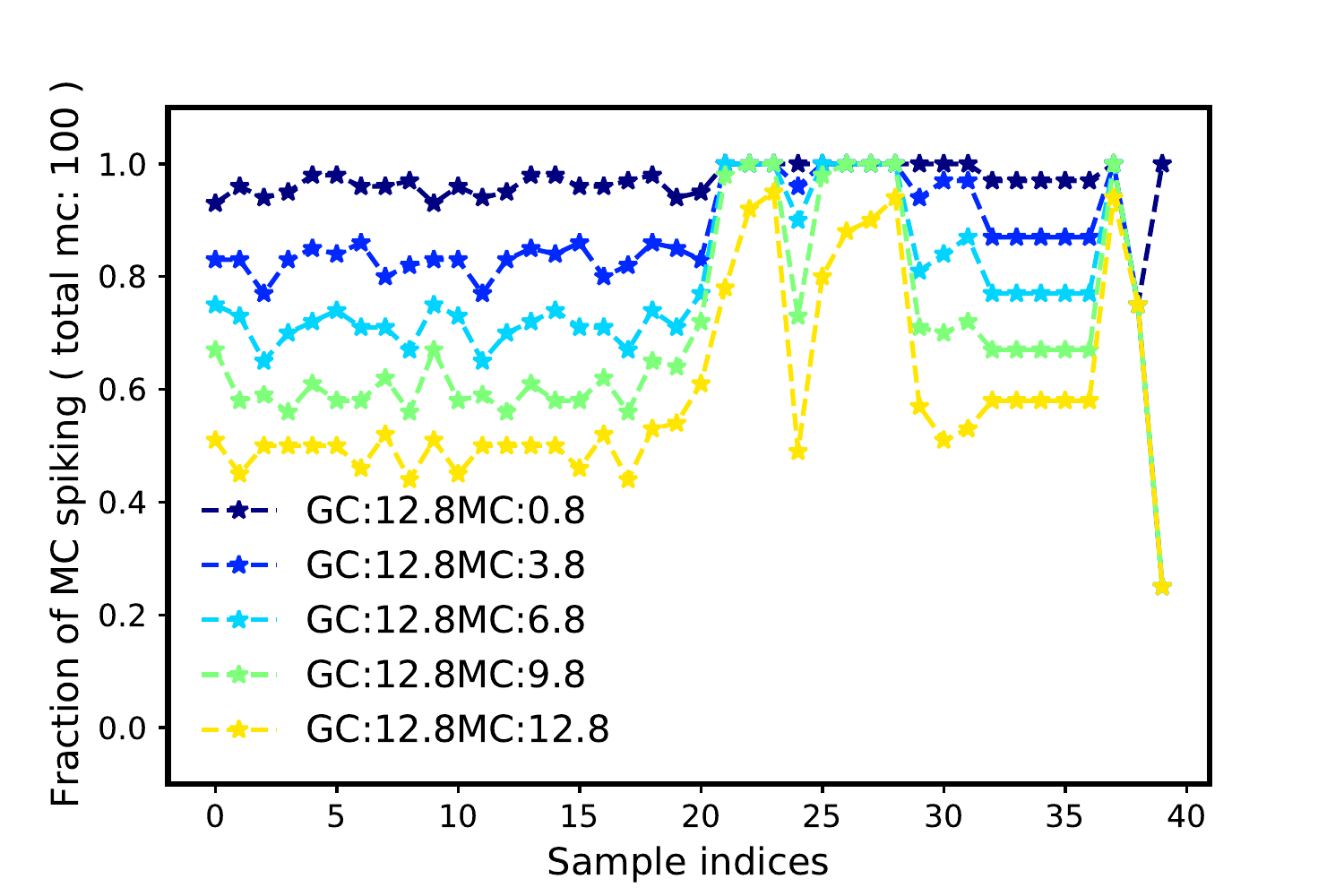}}
    \hspace{0.01in}
\subfloat[]{\includegraphics[width=0.49\linewidth]{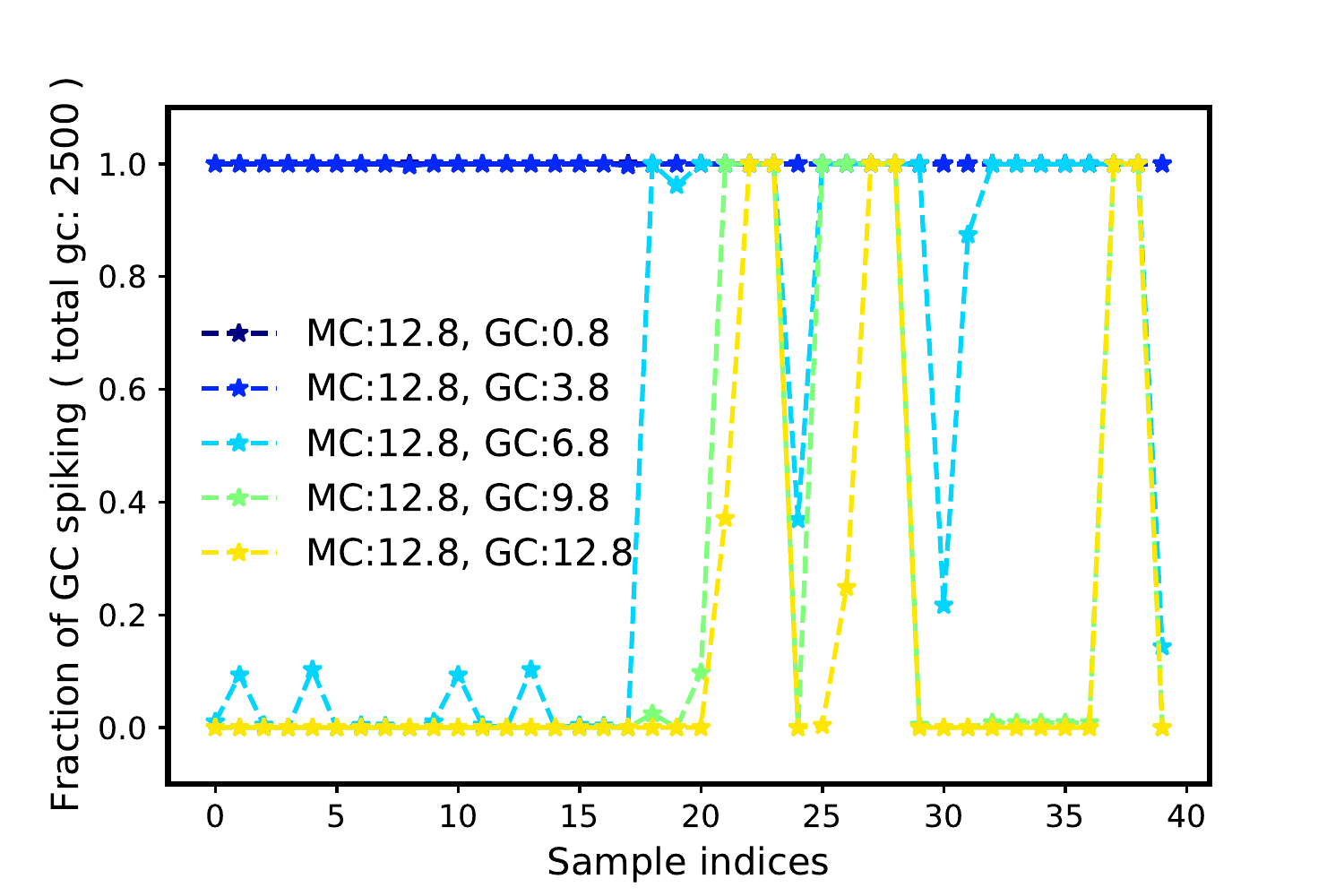}}
\hspace{0.01in}

  \caption{Duplicated data - Fraction of active MC/GCs w.r.t samples in a feedforward MC-GC network. Legends indicate MC/GC spiking thresholds. }  
  \label{dup_cnt}
\end{figure}

\paragraph{Duplication}
Each glomerulus is  composed of multiple mitral, external tufted cells. Inspired from this organization, we introduced column specific duplication of MCs, keeping everything else the same. 
Fig ~\ref{dup_cnt} describe the MC and GC spike count distributions after column specific duplication of the data from $100$ data to $100 \times 5 = 500$. The spike count similarities for MC and GC didn't change. Absolute GC spike counts are increased as the number of MCs increases five fold. The goodness of preprocessing value, $g_{p}$ was zero, Table~\ref{tab:mc_gp}, ~\ref{tab:gc_gp}. 

\paragraph{MC and GC threshold heterogeneity}

\begin{figure}
  \centering
  \subfloat[]{\includegraphics[width=0.49\linewidth]{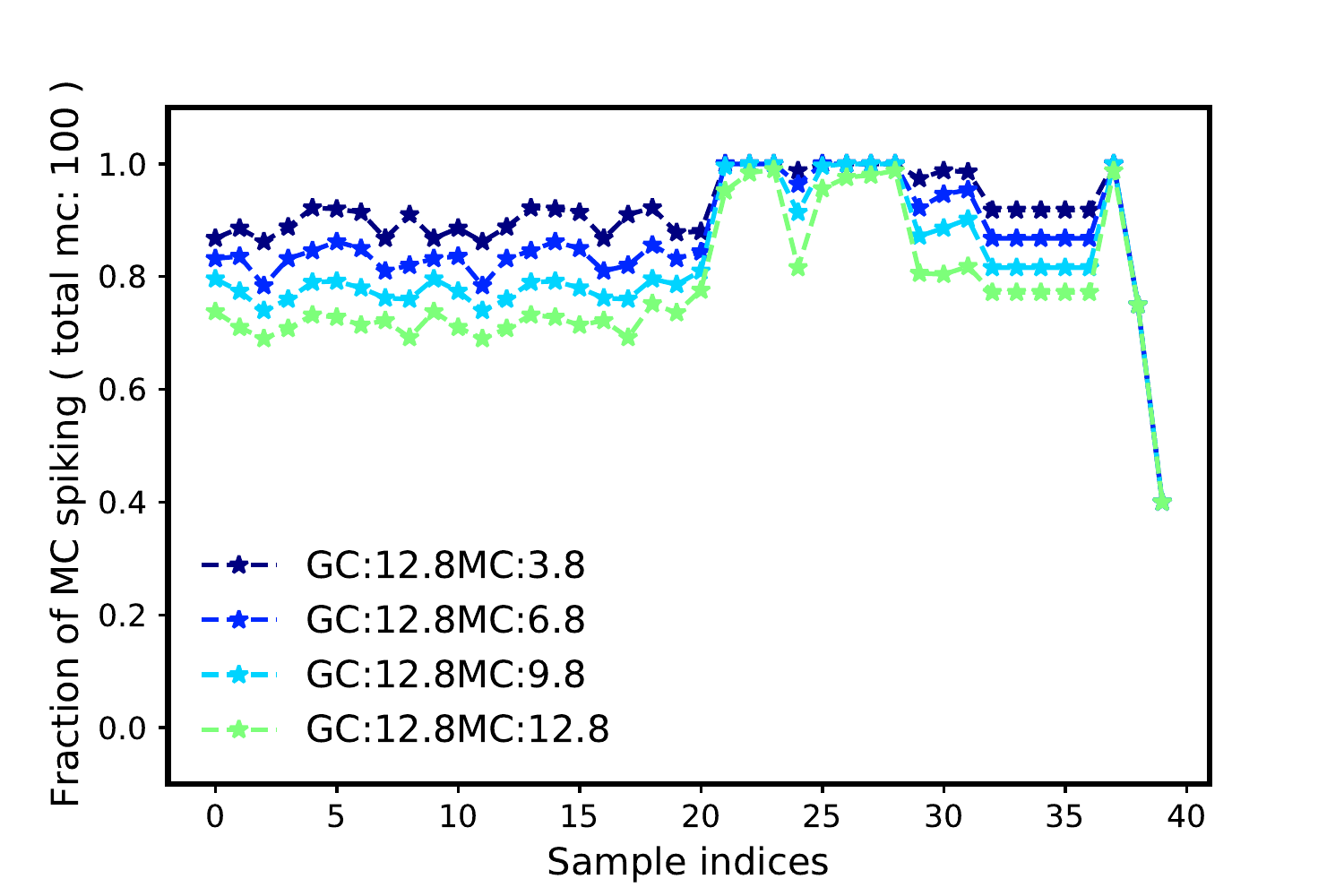}}
  \hspace{0.01in}
  \subfloat[]{\includegraphics[width=0.49\linewidth]{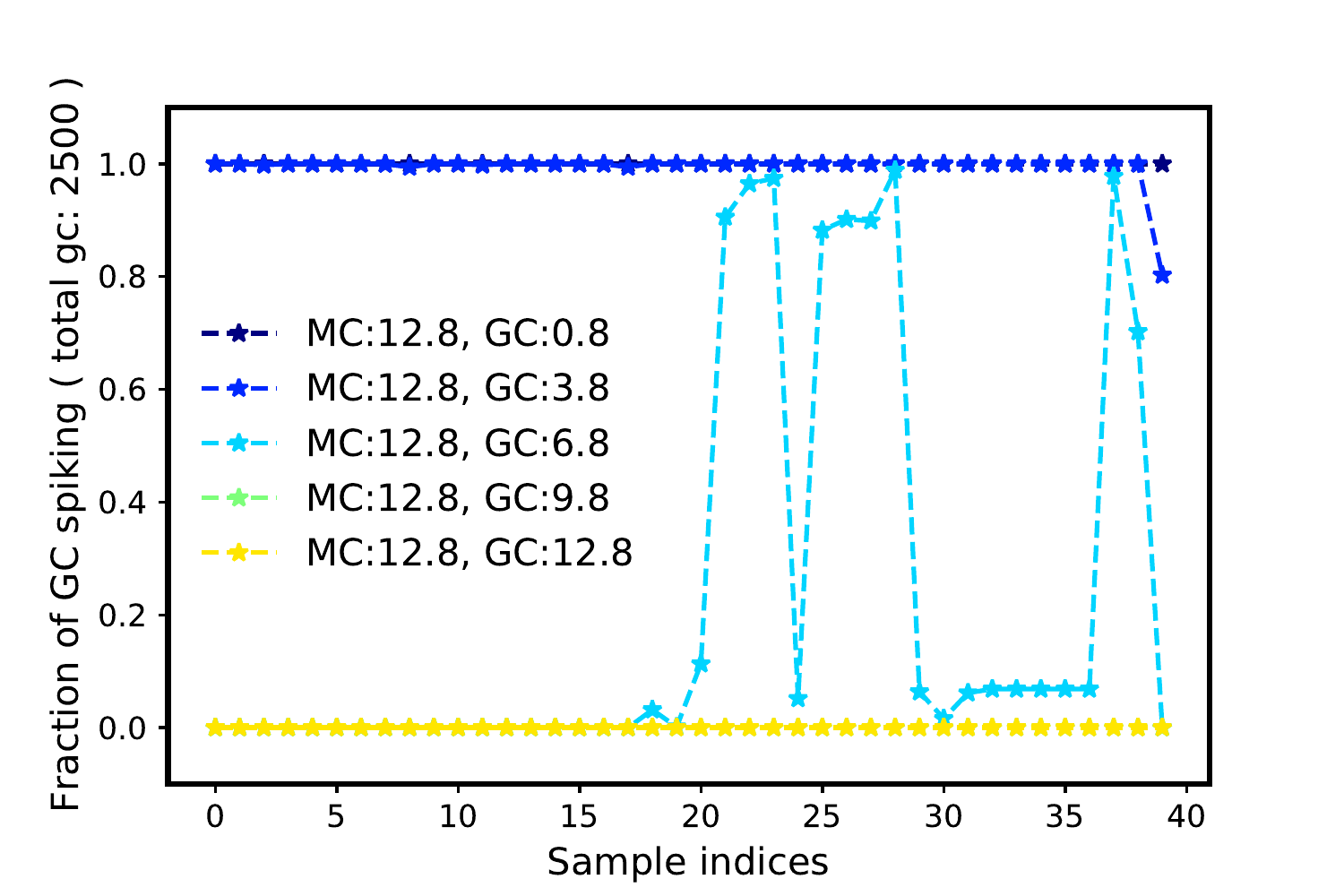}}
  \caption{MC threshold heterogeneity data - Fraction of active MC/GCs w.r.t samples in a feedforward MC-GC network. Legends indicate maximum MC/GC spiking thresholds. }  
  \label{mcvth_cnt}
\end{figure}

Duplication only was not enough. Sister Mitral cells are found to have correlated but different spike latency for an odor. Hence, after duplication, we applied threshold heterogeneity to the model. Fig ~\ref{mcvth_cnt} shows the fraction of active MCs and GCs spiking after the sister mitral cells in a column are assigned a range of spiking threshold. The minimum possible threshold was always $0.8v$ and the maximum can be $3.8, 6.8, 9.8, 12.8$. If the number of sister mitral cells is $5$, then a similar number of uniformly distributed thresholds from this range are assigned.  The goodness of preprocessing value, $g_{p}$ was zero, Table~\ref{tab:mc_gp}, ~\ref{tab:gc_gp}.

\begin{figure}
  \centering
  \subfloat[]{\includegraphics[width=0.49\linewidth]{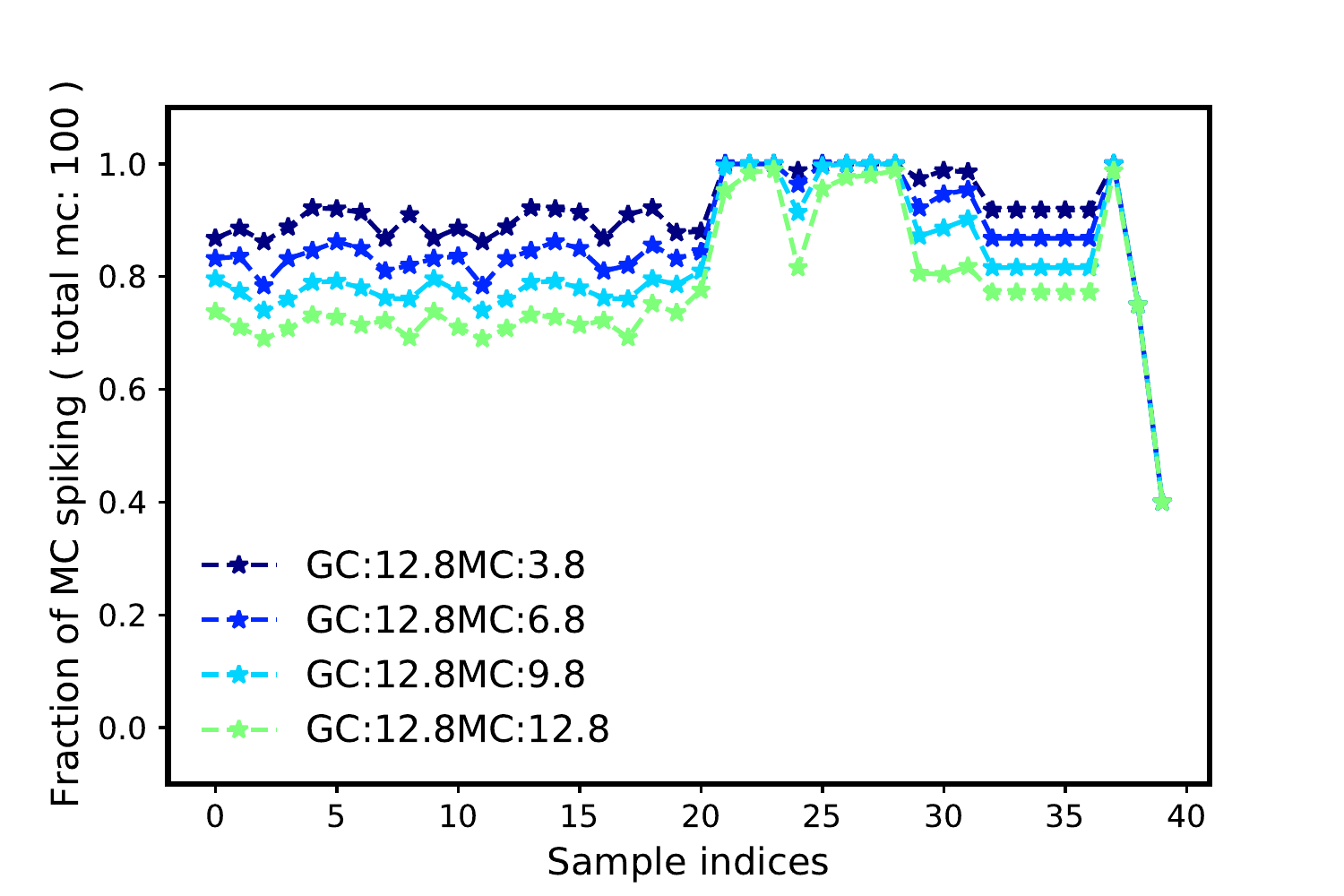}}
    \hspace{0.01in}
  \subfloat[]{\includegraphics[width=0.49\linewidth]{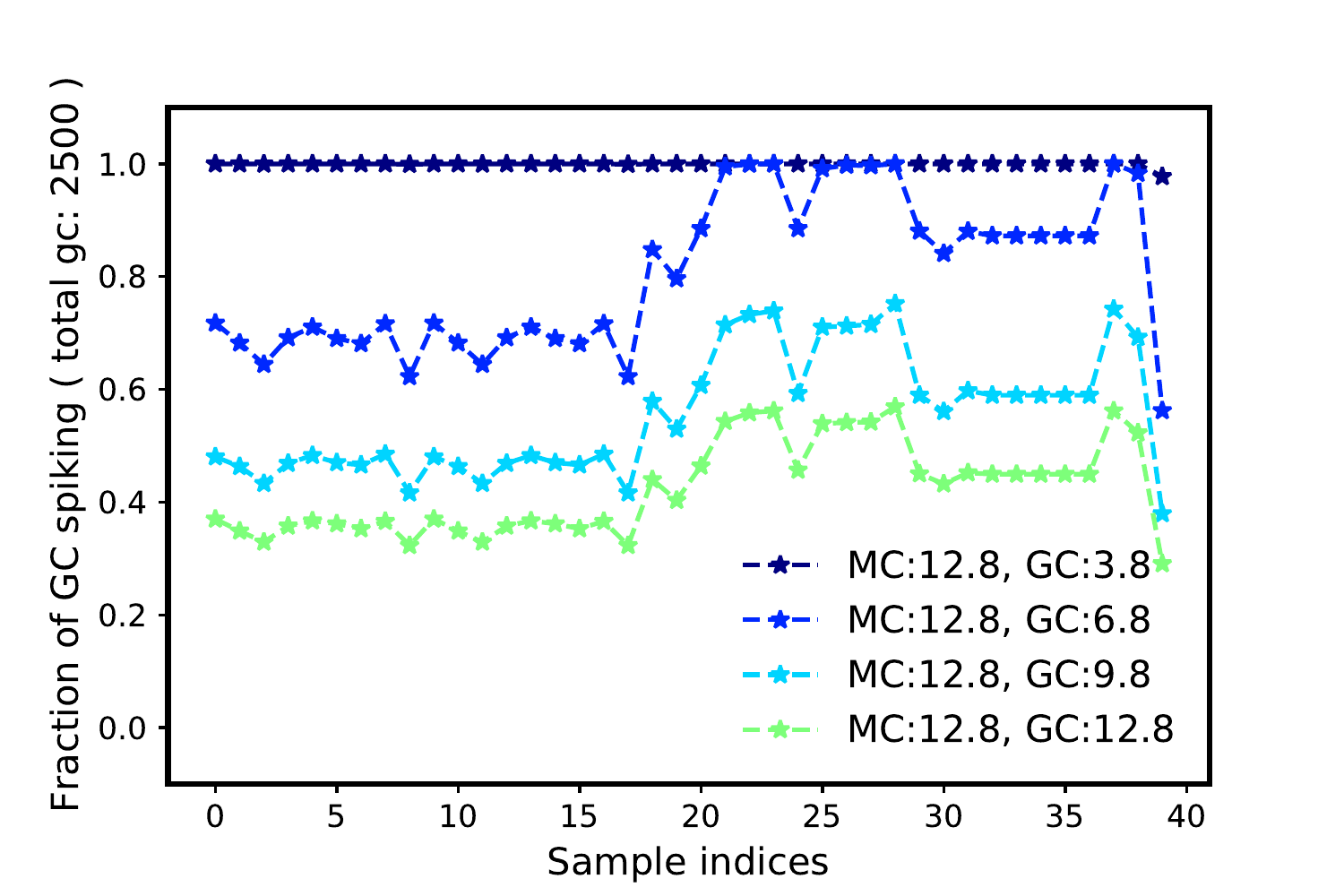}}
  \hspace{0.01in}
  \caption{GC threshold heterogeneity data - Fraction of active MC/GCs w.r.t samples in a feedforward MC-GC network. Legends indicate maximum MC/GC spiking thresholds. }  
  \label{mcgc_vth_cnt}
\end{figure}

Fig ~\ref{mcgc_vth_cnt} shows the spike count distribution after application of both MC and GC threshold heterogeneity. As this is a feedforward network, the MC spike count distribution is similar to Fig ~\ref{mcvth_cnt}. For GCs, the number of heterogeneous thresholds per column was equal to $number \, GCs \, per \, column \times number \, of \, sister \, mitral \, cells $. In these simulations, GCs had $25$ different spiking thresholds in the range $0.8-vthmax$. $vthmax$ can be $3.8, 6.8, 9.8, 12.8$. After application of GC spiking threshold, the spike count distribution across samples improved significantly, Fig. ~\ref{mcgc_vth_cnt}b. The goodness of preprocessing value, $g_{p}$ was zero for MC but was $0.74$ for GCs, Table~\ref{tab:mc_gp}, ~\ref{tab:gc_gp}. 

\subsubsection{Network output after heterogenous duplication}

This step, heterogeneous duplication is also inspired from the glomerular layer connectivity pattern of the olfactory system. In each glomerular column, there are multiple sister mitral cells. They receive inputs from the external tufted cells (ET). We observed that if a sparse random matrix is used as the connection matrix from the external tufted (ET) cells to the mitral cells (MC), with no intraglomerular connection - the input data to mitral cells get regularized significantly. 

Mathematically, 

\begin{equation}
    X_{MC}=w_{ET-MC}^T X_{ET}
\end{equation}

Where  $w_{ET-MC}$ is a sparse random projection matrix (with no learning) with dimension $np\ \times nq$, $q$ being the number of sister mitral cells (MCs) per column. $w_{ET-MC}$ no-zero values are always drawn from a uniform distribution of range $0-0.65$. The value $0.65$ is essential to constrain the maximum possible input to MC to around $20A$. $X_{ET}$ is the concentration normalized input scaled always by a factor of $10$. $X_{MC}$ is the current input to mitral cells (MCs).

\begin{figure}
  \centering
  \subfloat[]{\includegraphics[width=0.49\linewidth]{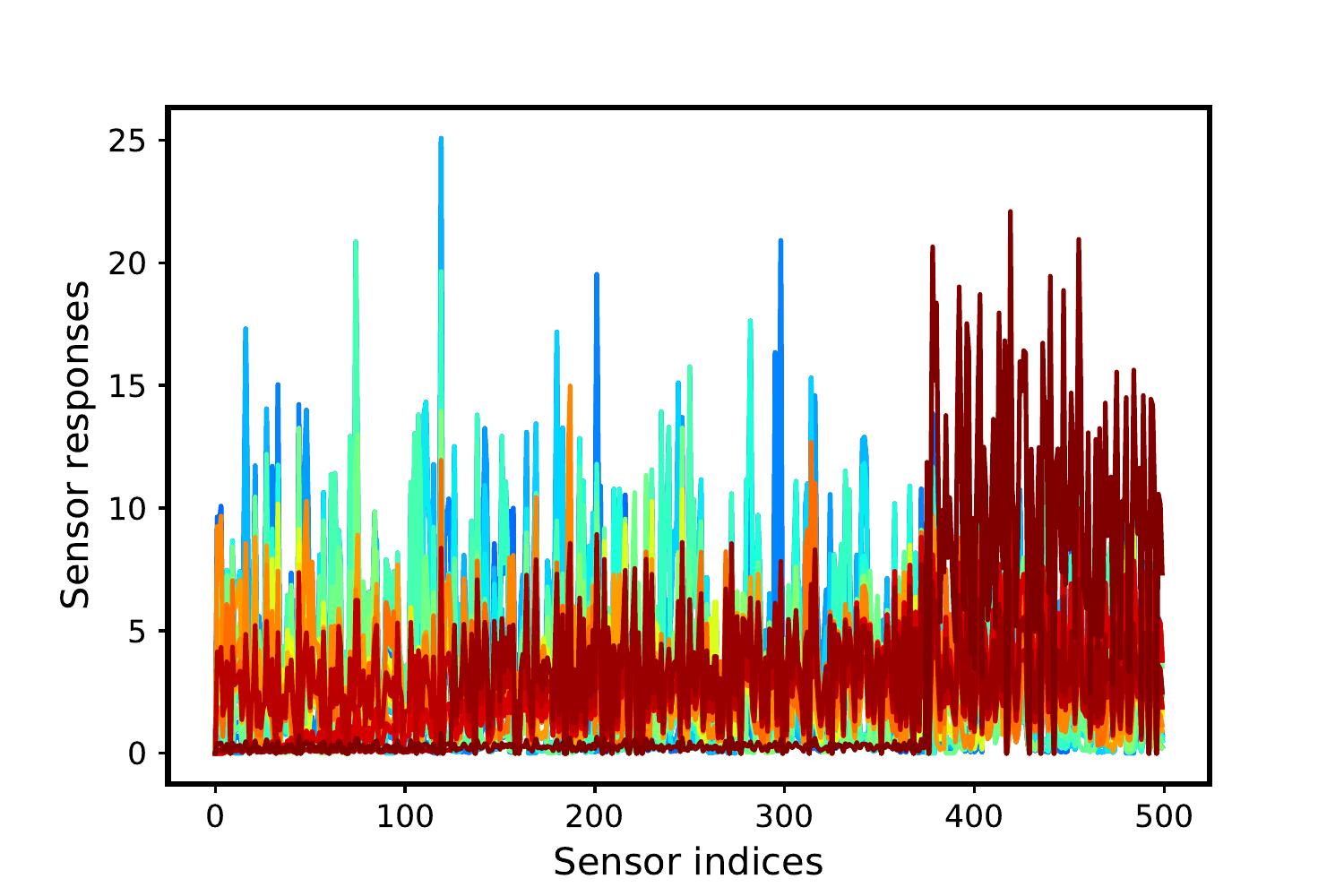}}
  \hspace{0.01in}
  \subfloat[]{\includegraphics[width=0.49\linewidth]{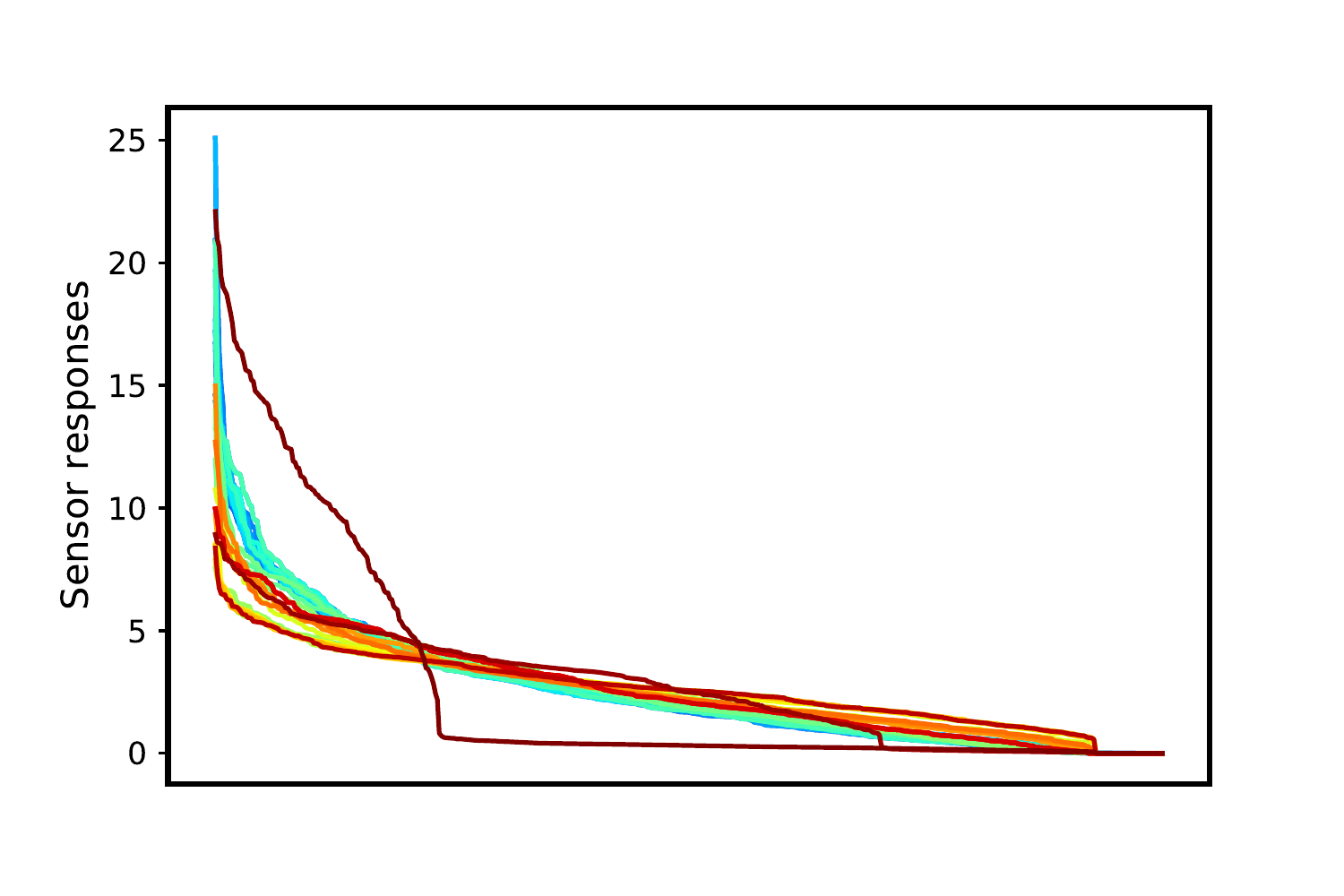}}
  \hspace{0.01in}
  \caption{Heterogeneous duplication: Sensor responses after application of heterogeneous duplication. a. Regularized data b. Same as a but sorted by amplitude. }  
  \label{hetdup_synth}
\end{figure}

\begin{figure}
  \centering
  \subfloat[MC spike count]{\includegraphics[width=0.48\linewidth]{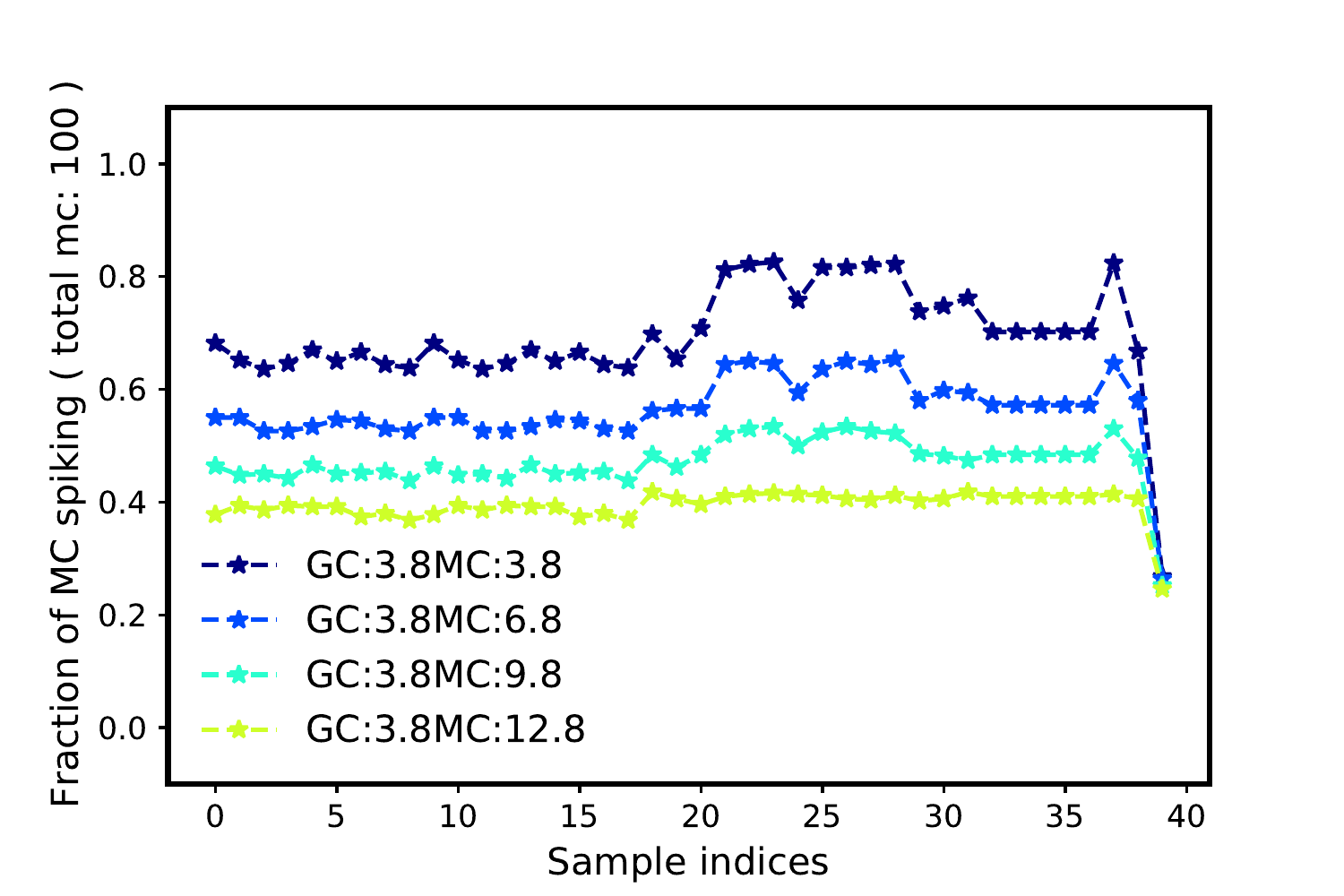}}
  \hspace{0.01in}
  \subfloat[MC spike count]{\includegraphics[width=0.48\linewidth]{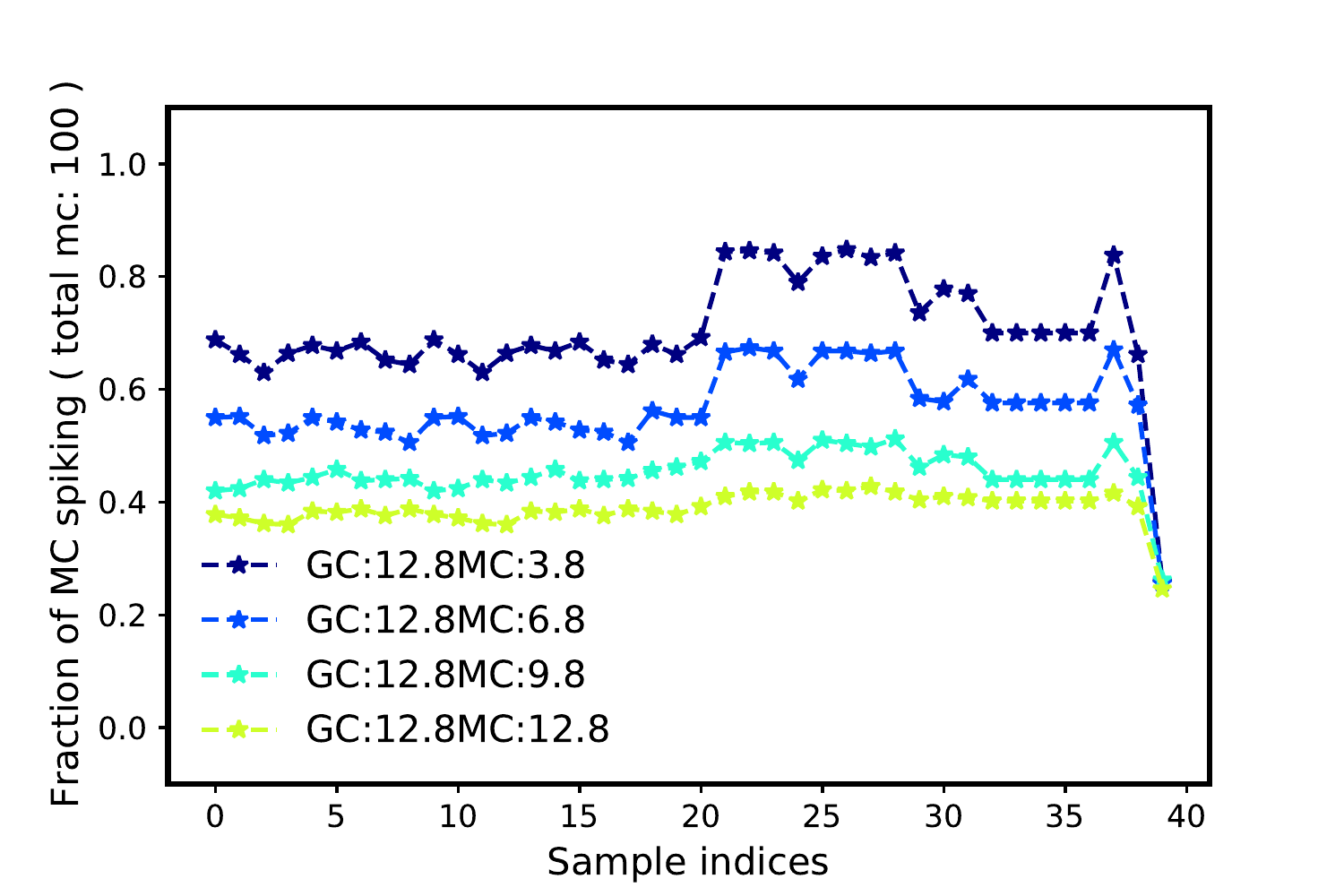}}
  \hspace{0.01in}
  \subfloat[GC spike count]{\includegraphics[width=0.48\linewidth]{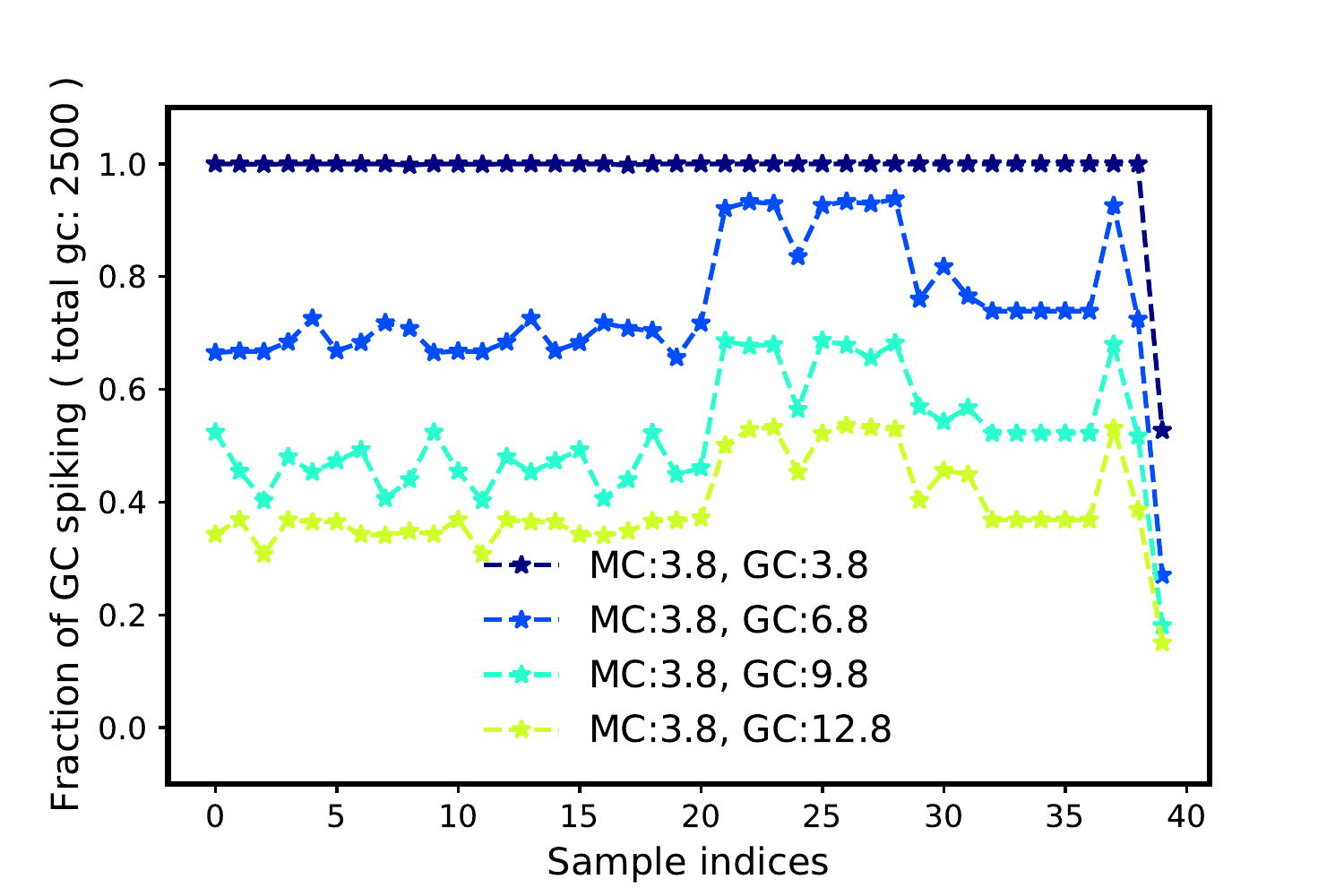}}
  \hspace{0.01in}
  \subfloat[GC spike count]{\includegraphics[width=0.48\linewidth]{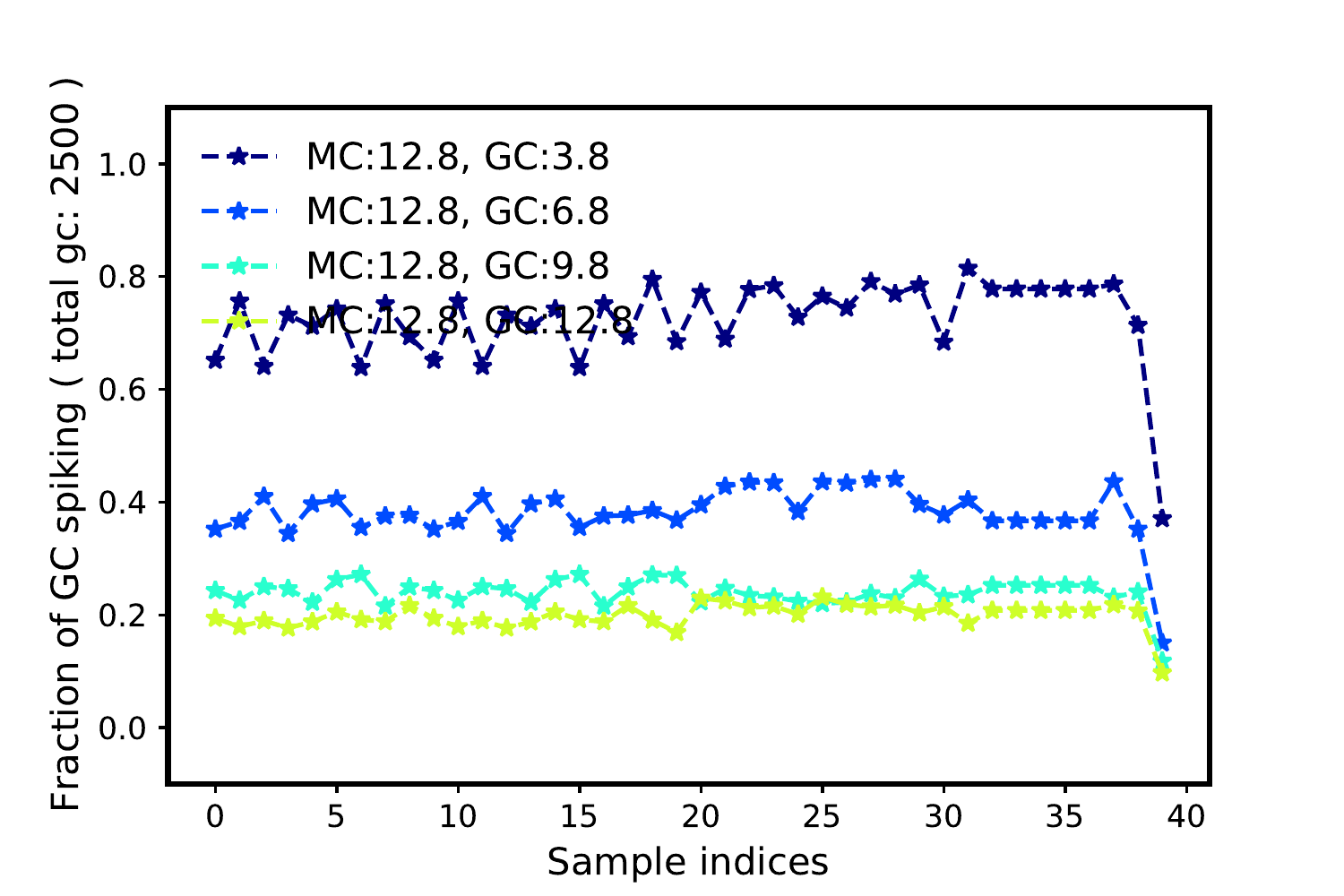}}
  \hspace{0.01in}
  \caption{Regularized data - Fraction of active MC/GCs w.r.t samples in a feedforward MC-GC network. Legends indicate MC/GC spiking thresholds. }  
  \label{hetdup_cnt}
\end{figure}

Fig ~\ref{hetdup_synth} depicts the regularized data after application of heterogenous duplication. From Fig ~\ref{hetdup_synth}b, it is evident that this step regularized the data. Hence, as shown in Fig ~\ref{hetdup_cnt}, the fraction of active MCs and GCs are within acceptable range. The similarity in spike counts improved as the spiking threshold heterogeneity range widened. Table ~\ref{tab:mc_gp} and Table ~\ref{tab:gc_gp} describe the good of preprocessing ($g_{p}$) for MCs and GCs respectively. The $g_{p}$ values of MCs and GCs were both high, $0.94$ and $0.88$ respectively.

\begin{table}
  \centering
  \caption{$g_{p}$ of MC spike count values assessed from each experimental dataset following the sequential application of each preprocessor.
  $+MS$ implies model scaling along with data regularization.}
  \label{tab:mc_gp}
  \vspace{0.2in}
  \resizebox{\textwidth}{!}{
  \begin{tabular}{||cccccccc||}
    \toprule
     & Raw & Scaled & Intensity norm. & Dup. & MC Het. & MC, GC Het. & Het. duplication \\
    \midrule
     Synthetic & 0 & 0 & 0 & 0 & 0 & 0 & \textbf{0.94} \\
    \midrule
     Drift (+MS) & 0 & 0 & - & - & - & - & \textbf{0.86}\\
     \midrule
     Wind Tunnel (+MS) & 0 & 0 & - & - & - & - & \textbf{0.92}\\
  \bottomrule
  \end{tabular}}
\end{table}

\begin{table}
\centering
  \caption{$g_{p}$ of GC spike count values assessed from each experimental dataset following the sequential application of each preprocessor. $+MS$ implies model scaling along with data regularization. }
  \label{tab:gc_gp}
  \vspace{0.2in}
  \resizebox{\textwidth}{!}{
  \begin{tabular}{||cccccccc||}
    \toprule
     & Raw & Scaled & Intensity norm. & Dup. & MC Het. & MC, GC Het. & Het. duplication \\
    \midrule
     Synthetic & 0 & 0 & 0 & 0 & 0 & 0.74 & \textbf{0.88} \\
    \midrule
    Drift (+MS) & 0 & 0 & - & - & - & - & \textbf{0.72}\\
    \midrule
    Wind Tunnel (+MS) & 0 & 0 & - & - & - & - & \textbf{0.76}\\
  \bottomrule
  \end{tabular}}
\end{table}

\subsection{Plasticity:  excitatory}

We observed that heterogeneous granule cell (GC) threshold is useful for data regularization. With a goal of understanding the relation between GC spike count and heterogeneity, we studied the variation of GC spike count w.r.t spiking threshold by using the same network as in Fig ~\ref{hetdup_cnt}: MCs having $5$ heterogeneous spiking thresholds uniformly distributed in the range $0.8-12.8 \, mV$ inclusive, GC spiking thresholds having a single value in the range $0.8-16.8$ with increment step size of $0.5$. 

Using the synthetic odor generation method ( see materials \& methods ), we generated $5$ odors of dimension $100$, first $4$ being sequentially similar with inter odor distance of $0.5$ and one non overlapping odor. 

\begin{figure}
  \centering
  \includegraphics[width=0.9\linewidth]{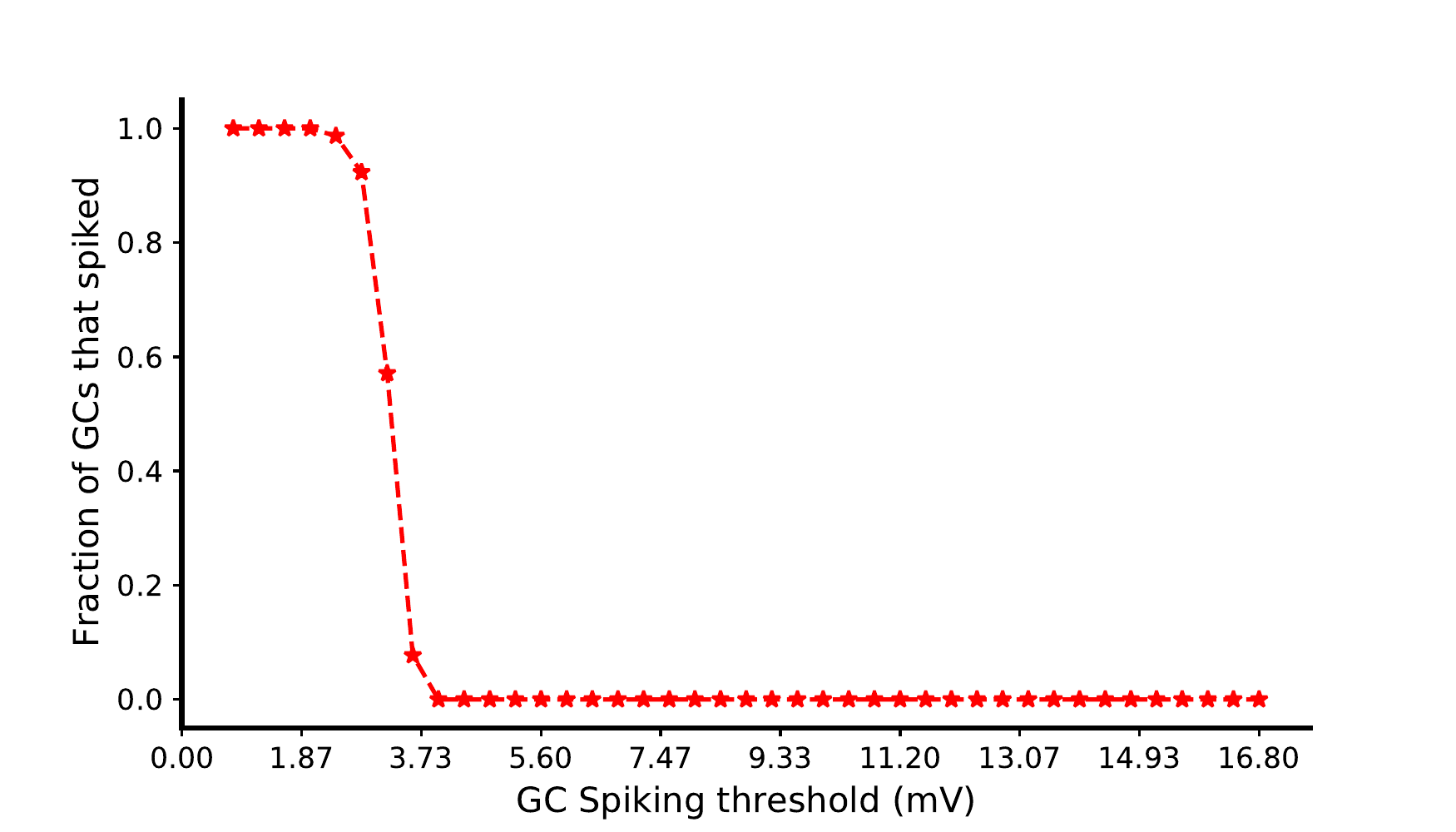}
  \caption{Variation of GC spike count fraction w.r.t threshold when a regularized data sample is fed to a network of $100$ MC columns and $100 \times 5 \times 5=2500$ GCs. All parameters same as Fig ~\ref{hetdup_cnt}}
  \label{gc_vth_cnt}
\end{figure}

Fig ~\ref{gc_vth_cnt} shows that GC spike count fraction is $1.$ for threshold of $0.8 \, mV$. But this value reduces as the spiking threshold is raised. For very high threshold GCs ( $ > \, 3.73 \, mV$), there was no spike at all. 

\begin{figure}
  \centering
  \subfloat[$v_{th}=0.8 \, mV$]{\includegraphics[width=0.32\linewidth]{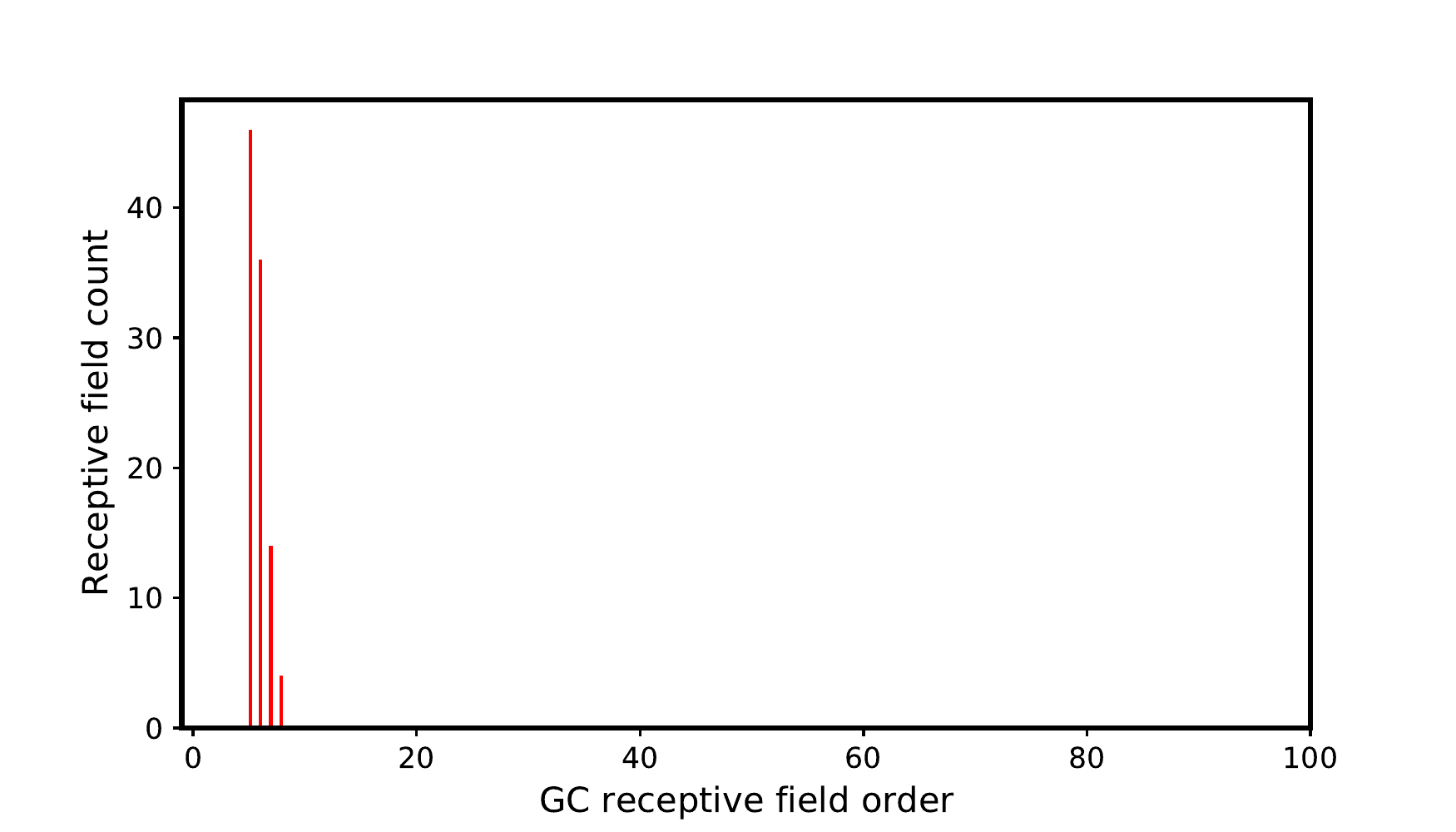}}
  \hspace{0.05in}
  \subfloat[$v_{th}=2.8 \, mV$]{\includegraphics[width=0.32\linewidth]{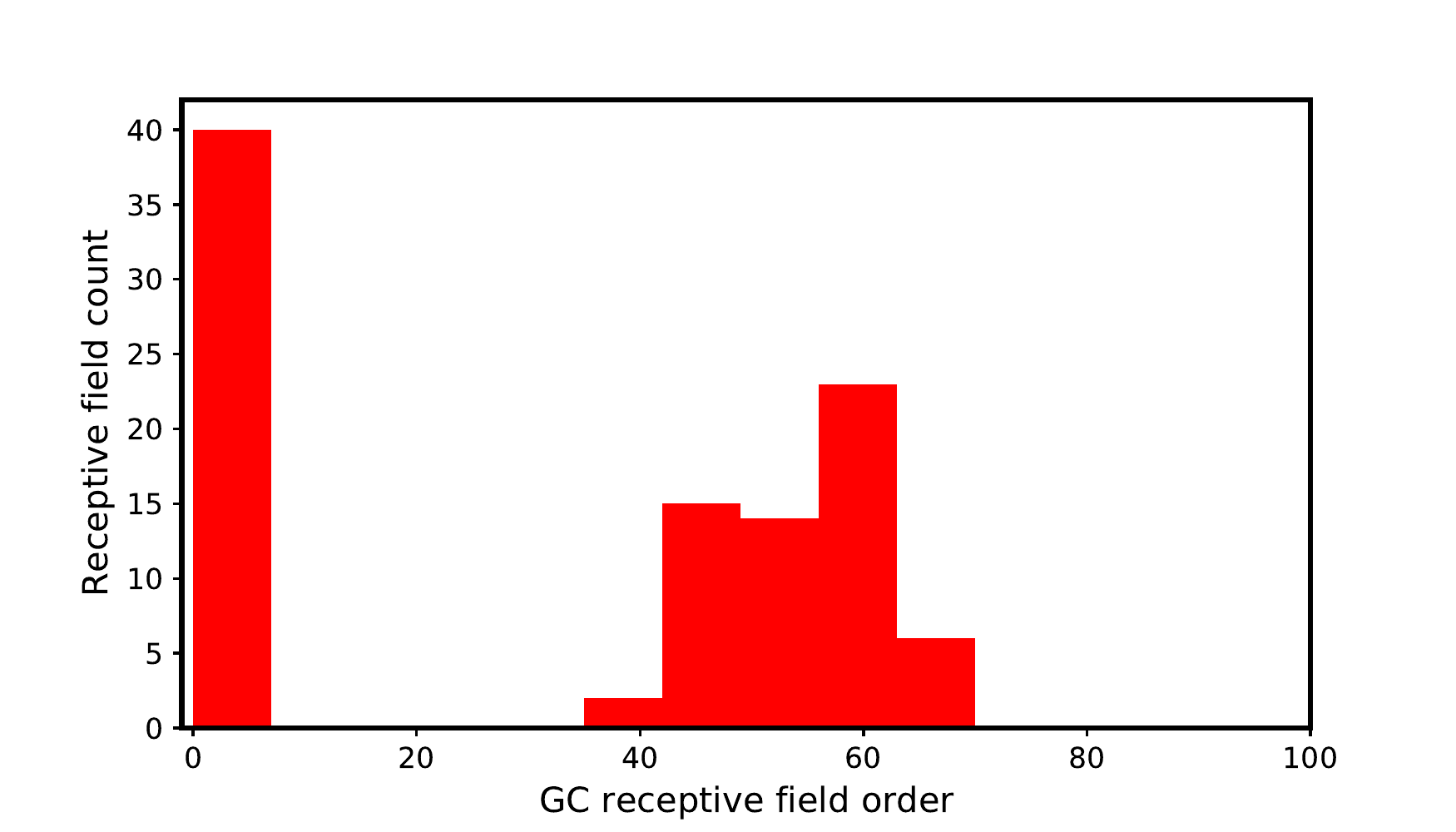}}
  \hspace{0.05in}
  \subfloat[$v_{th}=3.55 \, mV$]{\includegraphics[width=0.32\linewidth]{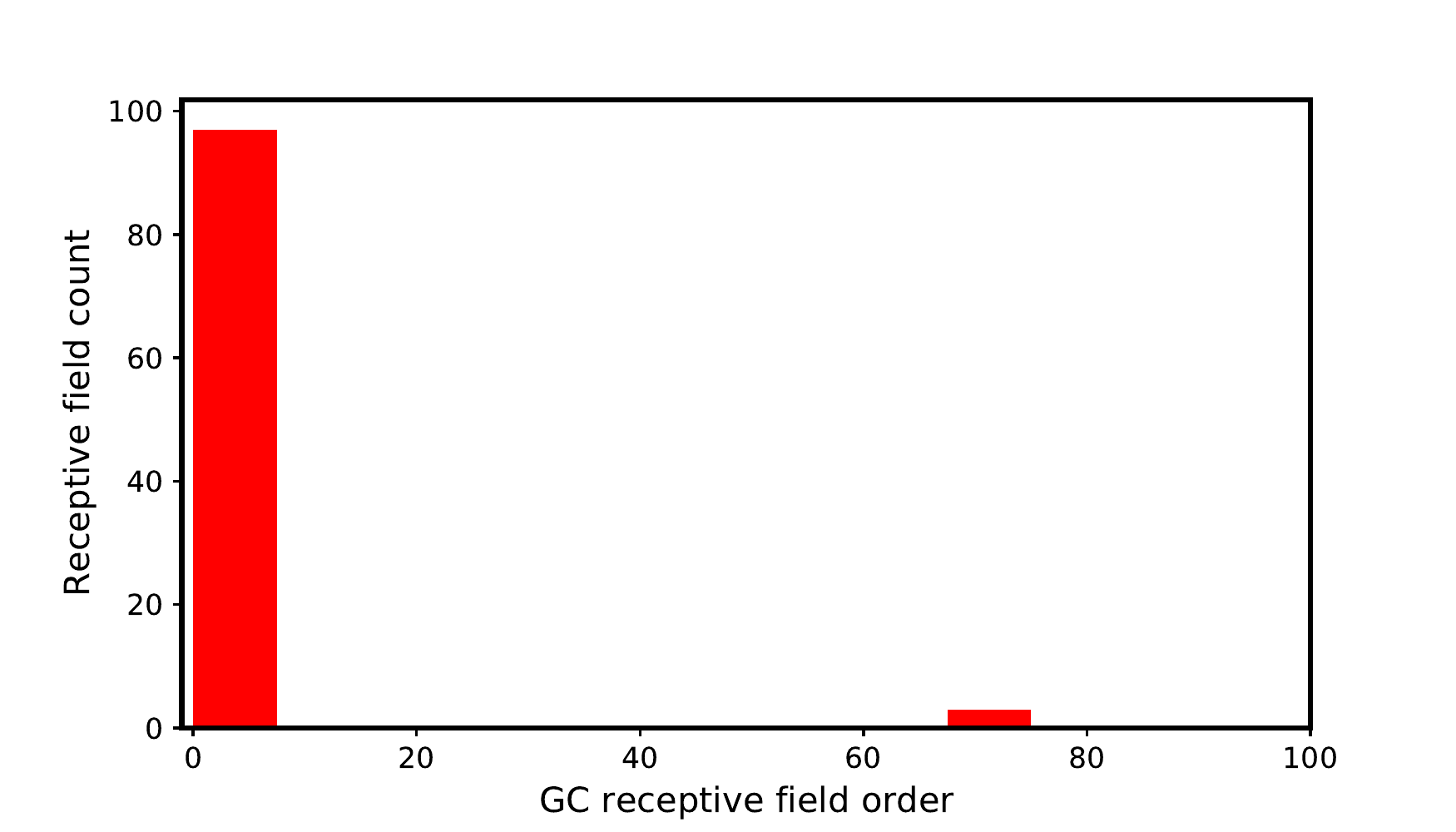}}
  \hspace{0.05in}
  \caption{Variation of GC receptive fields w.r.t GC spiking thresholds after learning an odor. a, b, c depicts receptive field order distributions for GC groups with spiking thresholds of $0.8, 2.8 \, and \, 3.55 \, mV$ respectively.}
  \label{HORF_vth}
\end{figure}

We next introduced threshold heterogeneity in GCs of the network and accordingly, $25$ threshold values per MC column drawn uniformly in the range $0.8-6.8 \, mV$ were assigned to the network. The odor is learned at the MC-GC synapses using an excitatory asymmetric spike timing dependent plasticity (STDP) rule ( see materials \& methods for details ). A close look at Fig ~\ref{HORF_vth} shows that the receptive field order increases with the increases with threshold - the distribution shifts towards right. This implies, higher the threshold, more the number of presynaptic neurons (MCs) required to make it spike. This also explains the absence of spike from very high thresholds - the required number of presynaptic neurons is higher than the available.  

\begin{figure}
  \centering
  \includegraphics[width=0.9\linewidth]{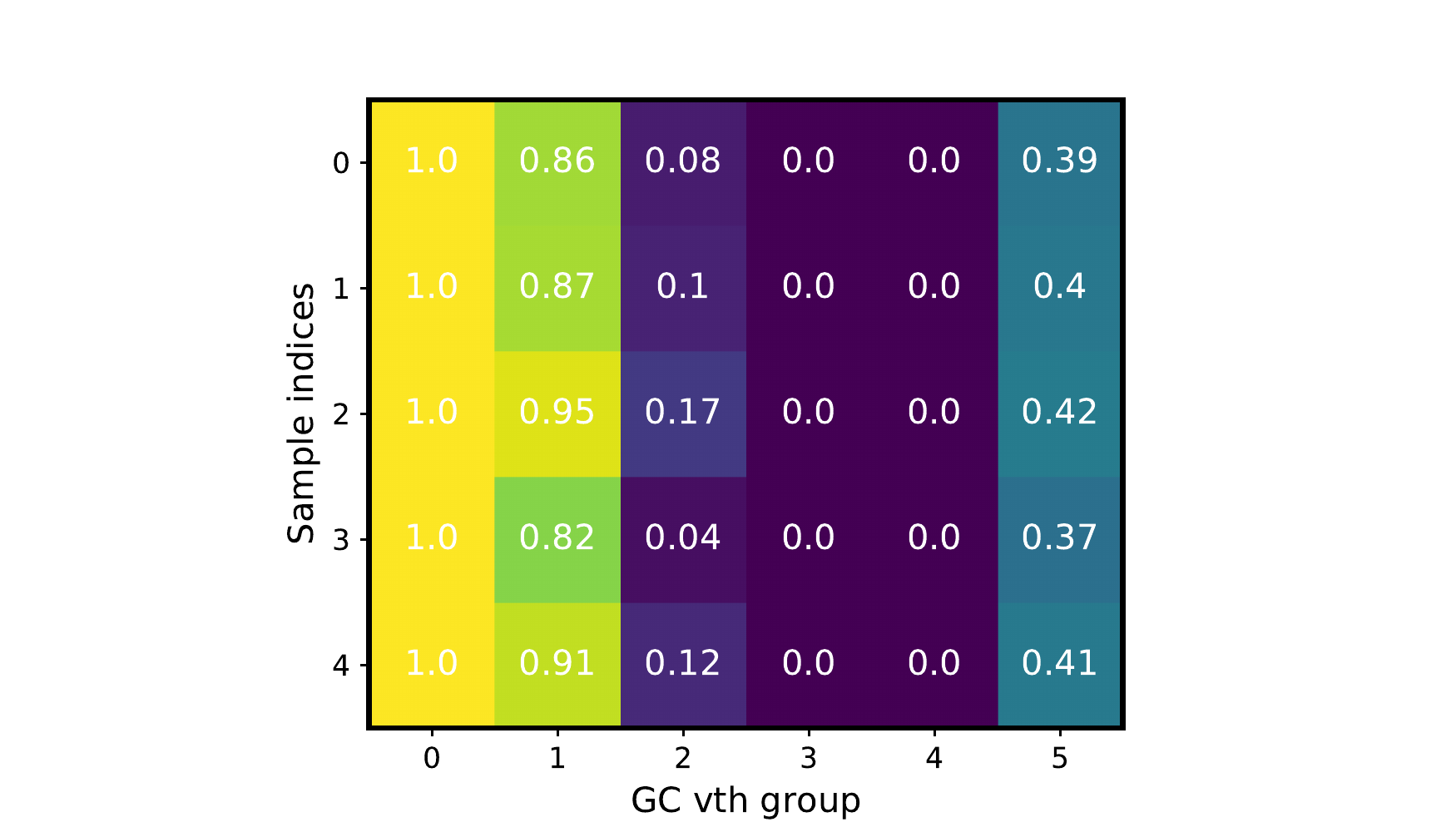}
  \caption{GC spike counts w.r.t different threshold groups for sequentially similar odors with MC-GC standard random connection probability of $0.4$.} \label{gc_cnt_seq_sim}
\end{figure}

For better visualization, we grouped the $25$ GC thresholds into $5$ groups. Fig ~\ref{gc_cnt_seq_sim} shows the variation in GC spike count with respect to threshold range when sequentially similar odors are input to the network. We observe that, for all odors, the spike count of GCs reduces with threshold. 

\paragraph{Heterogenous random connection}

\begin{figure}
  \centering
  \includegraphics[width=0.9\linewidth]{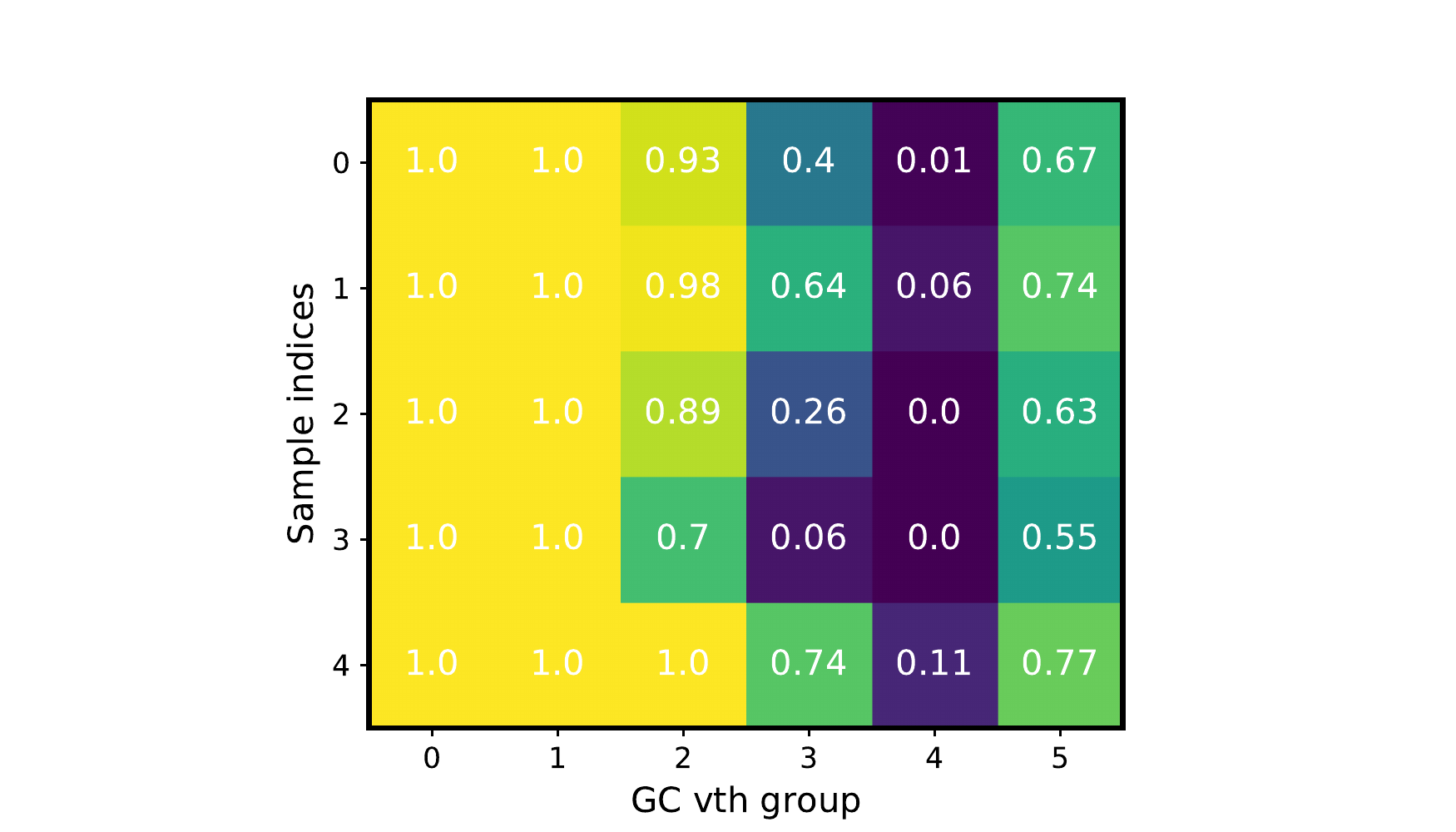}
  \caption{GC spike counts w.r.t different threshold groups for sequentially similar odors with MC-GC heterogenous random connection probabilities in the range $0.4-0.8$.}  
  \label{het_cp_vth}
\end{figure}

\begin{figure}
  \centering
  \subfloat[]{\includegraphics[width=0.48\linewidth]{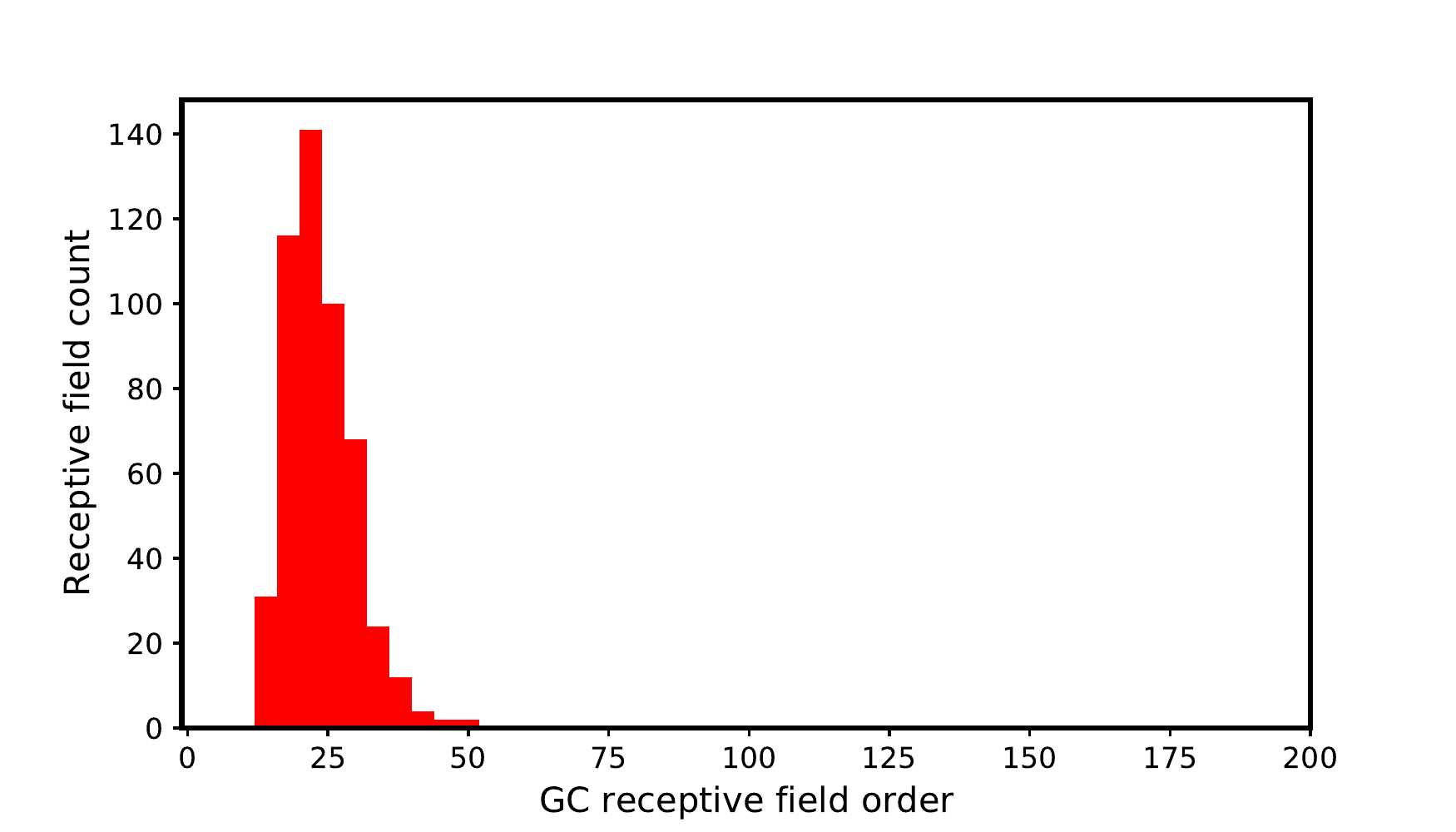}}
    \hspace{0.001in}
  \subfloat[]{\includegraphics[width=0.48\linewidth]{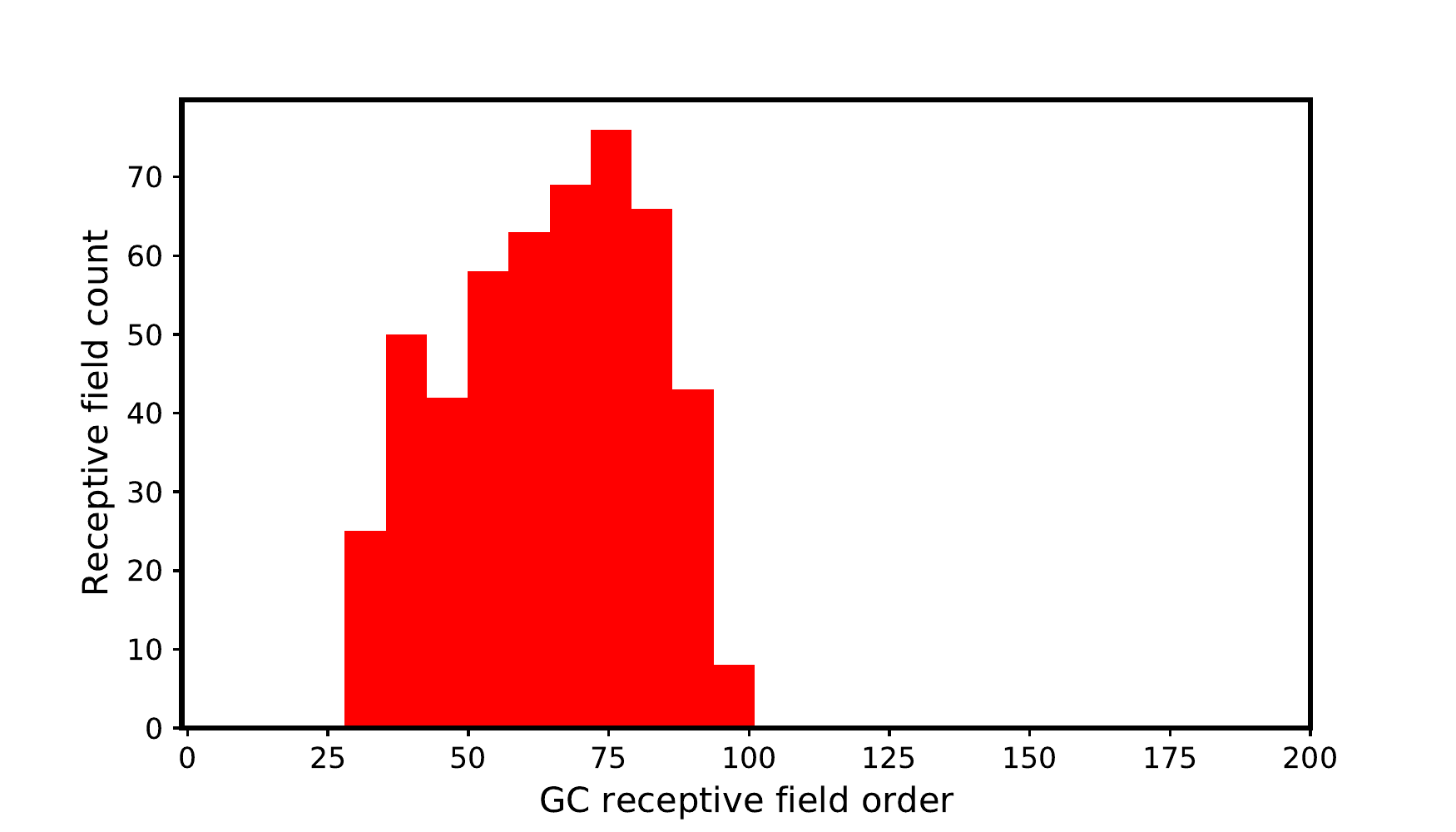}}
  \hspace{0.001in}
  \subfloat[]{\includegraphics[width=0.48\linewidth]{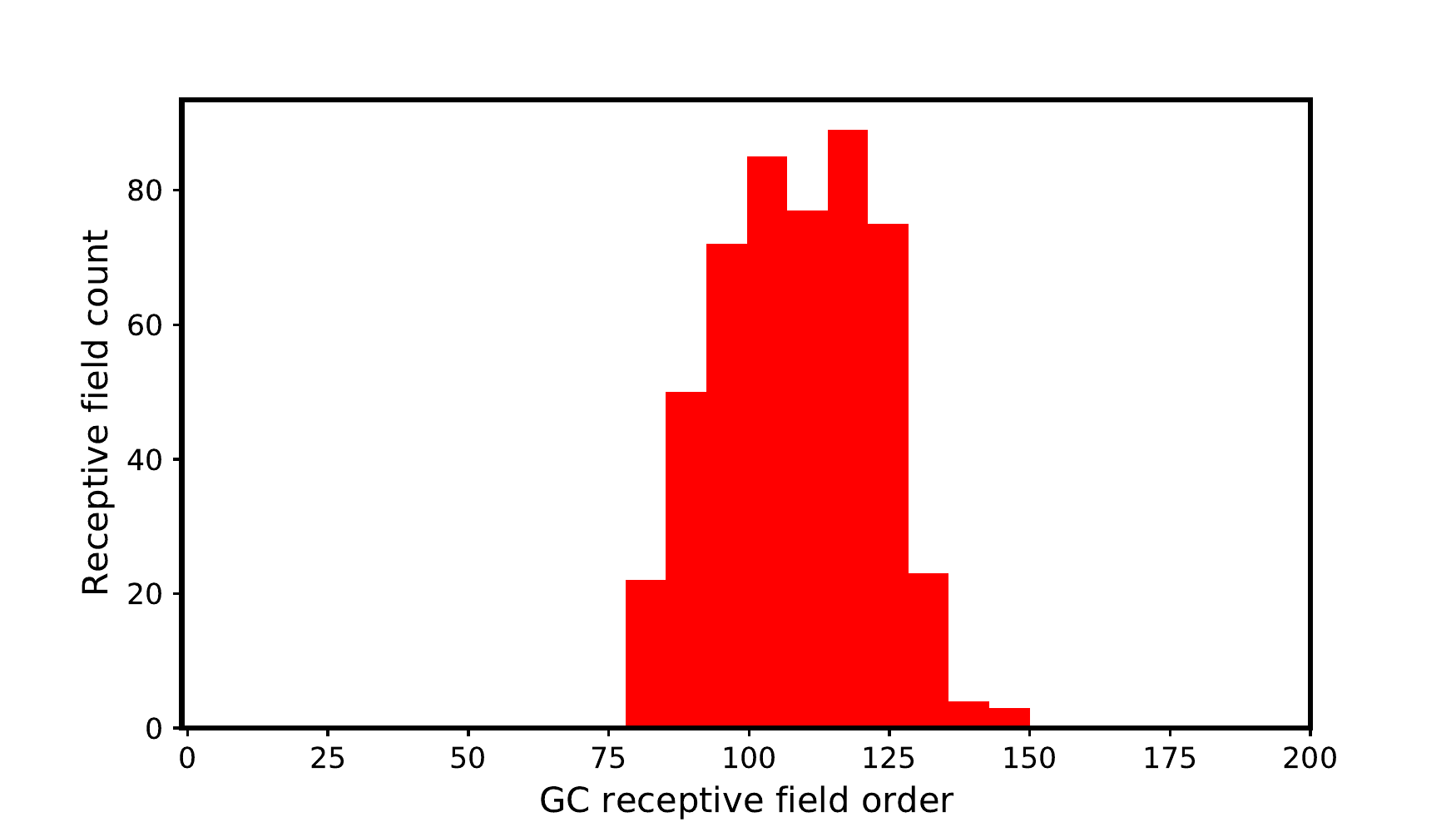}}
  \hspace{0.001in}
  \subfloat[]{\includegraphics[width=0.48\linewidth]{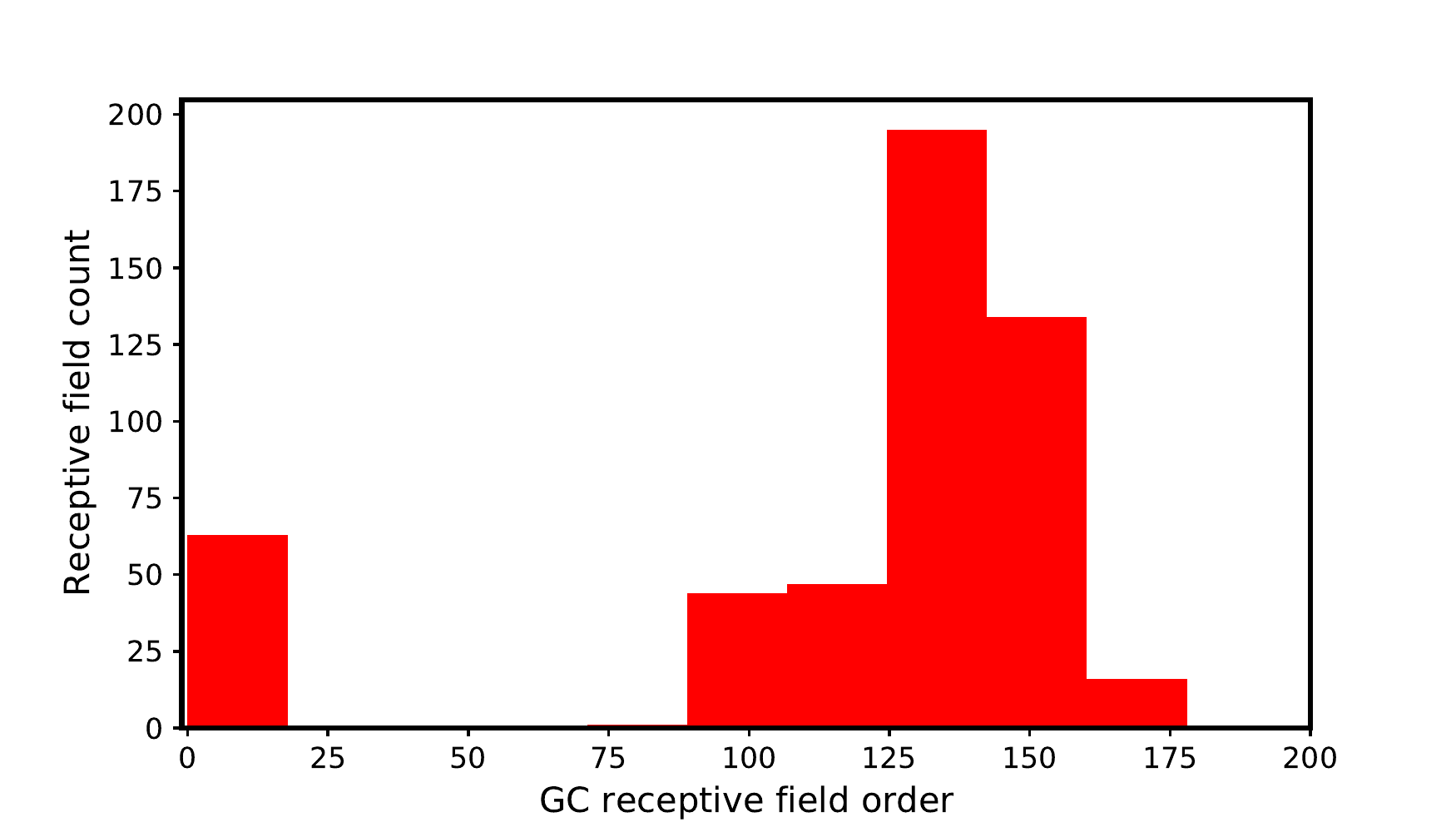}}
  \hspace{0.001in}
  \subfloat[]{\includegraphics[width=0.48\linewidth]{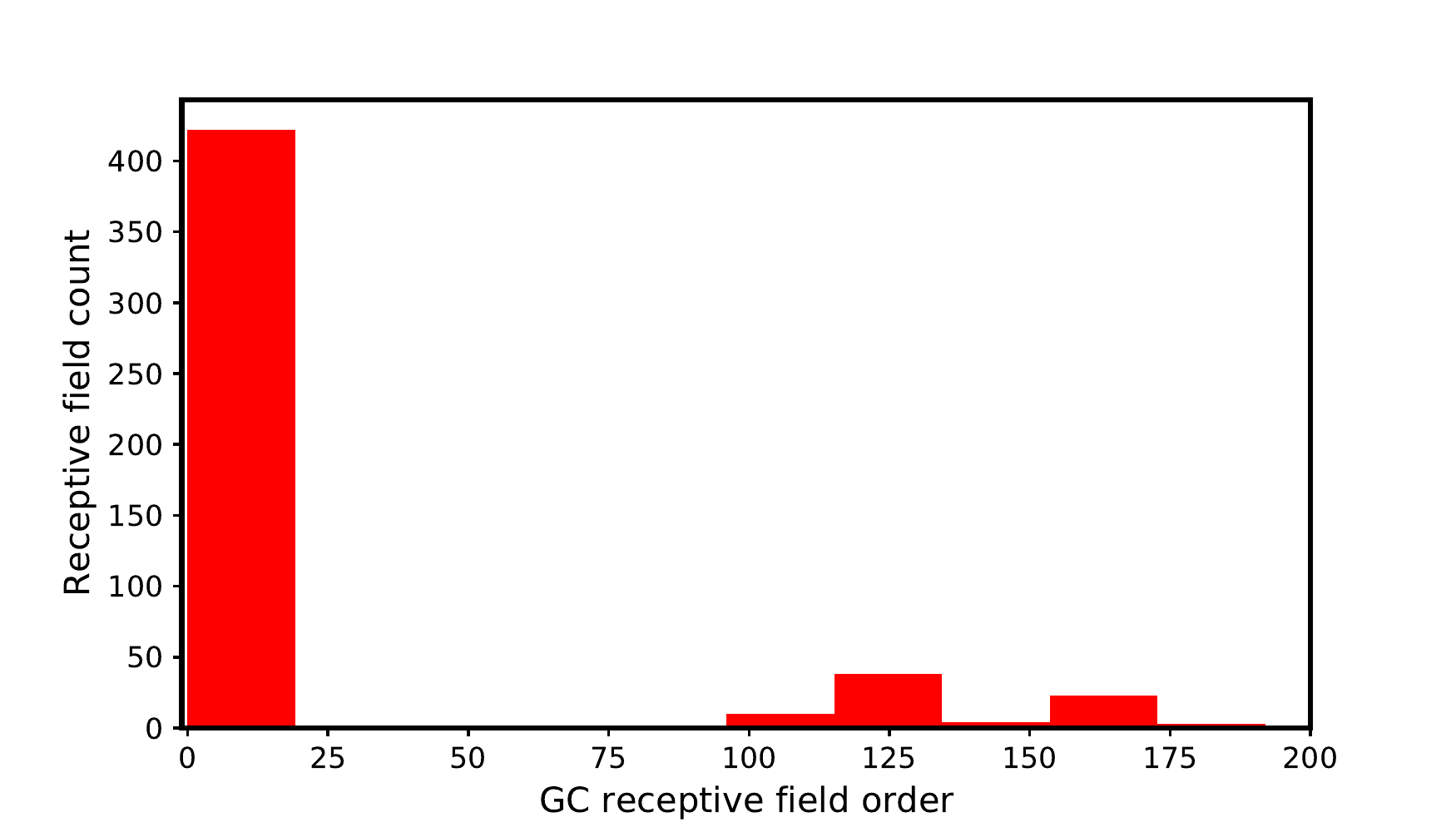}}
  \hspace{0.001in}
  \caption{Variation of GC receptive field distribution w.r.t GC spiking threshold. a, b, c, d \& e correspond to receptive fields of different threshold groups.}  
  \label{het_conn_horf}
\end{figure}

The higher threshold neurons need more presynaptic neurons to spike. One way to ensure that higher threshold neurons spike is by assigning higher initial connection probabilities. On the other hand, the lower threshold neurons need to have lower connection probabilities in order to ensure spike sparsity. Hence, we introduced a heterogeneous connection regime in which the number of MCs initially converging to a GC increases with the GC threshold. Fig ~\ref{het_cp_vth} shows the GC spike count w.r.t threshold when sequentially similar odors are input to the network. We varied the MC-GC convergence uniformly from $0.4-0.8$ which ensured spikes for even higher threshold GCs.  After learning, the non participating synapses are pruned. The use of heterogeneous connection probabilities also maximizes the receptive field formation with higher order for all threshold groups, Fig ~\ref{het_conn_horf}. 

\subsection{Sequential learning}

Sequential learning refers to training networks with only a single sample at a time ( $batch \, size \, = \, 1$), often combined with intermittent inference / testing periods. 

\subsubsection{Similarity}

One essential element of \textit{learning in the wild} is a property referred to as \textit{true generalization}~\cite{Shepard1317}. Briefly, this property encompasses the ability to identify inputs based on the explicit experience-dependent shaping of consequential regions within a physical similarity space~\cite{edelman_1998}. That is, stimuli are not simply to be learned discreetly; their relationships also must be learned~\cite{ClelandNarlaBoudadi}, so that, for example, physically similar stimuli predicting different consequences can be efficiently distinguished (\textit{perceptual learning}), whereas distinct but related stimuli can be embedded within an adaptive hierarchy of meaning.  Pursuant to this goal, we here report on our development of \textit{similarity-aware representation} of our model. 

\begin{figure}
  \centering
  \subfloat[Scaled]{\includegraphics[width=0.48\linewidth]{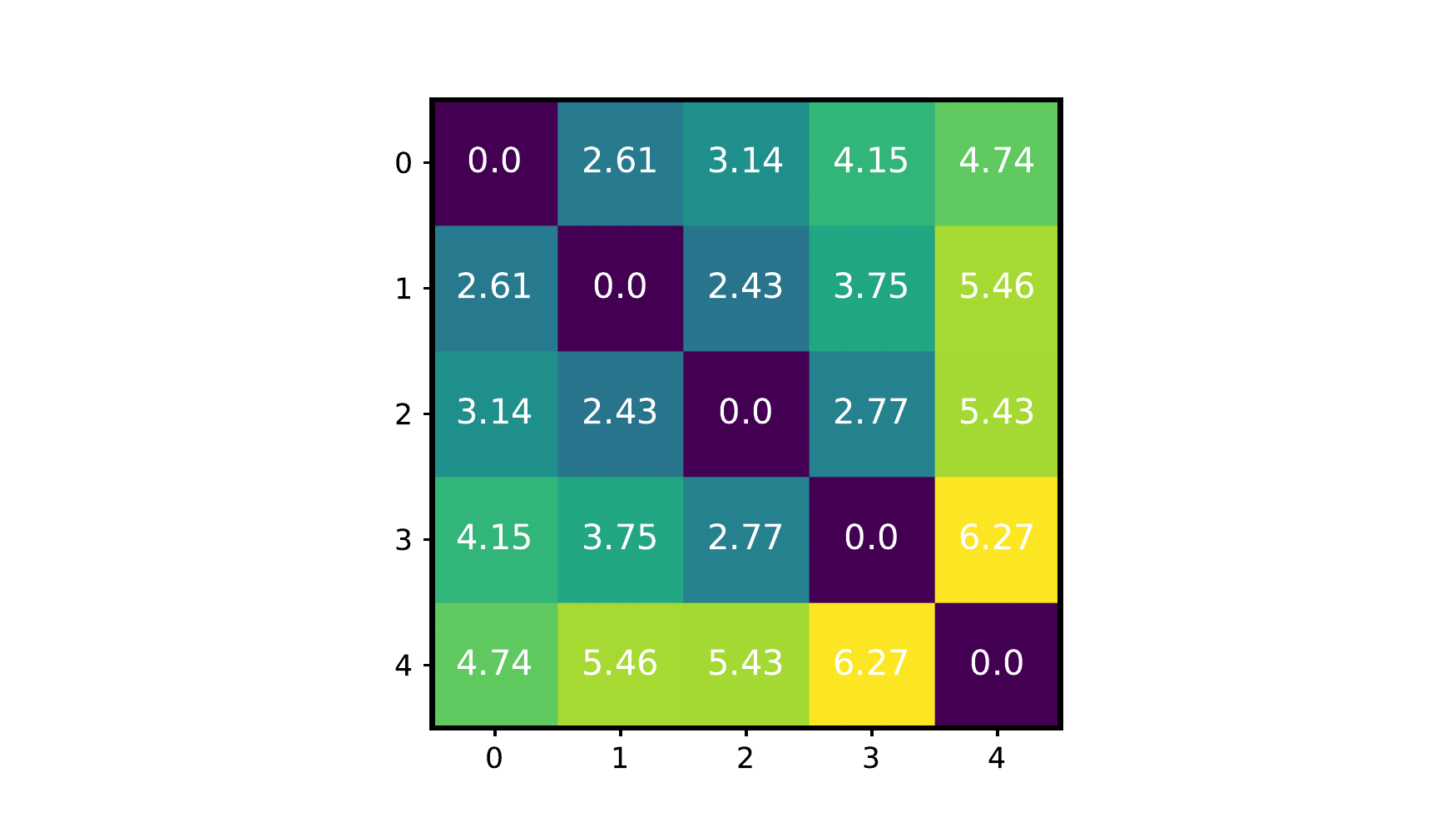}}
  \hspace{0.001in}
  \subfloat[Normalized]{\includegraphics[width=0.48\linewidth]{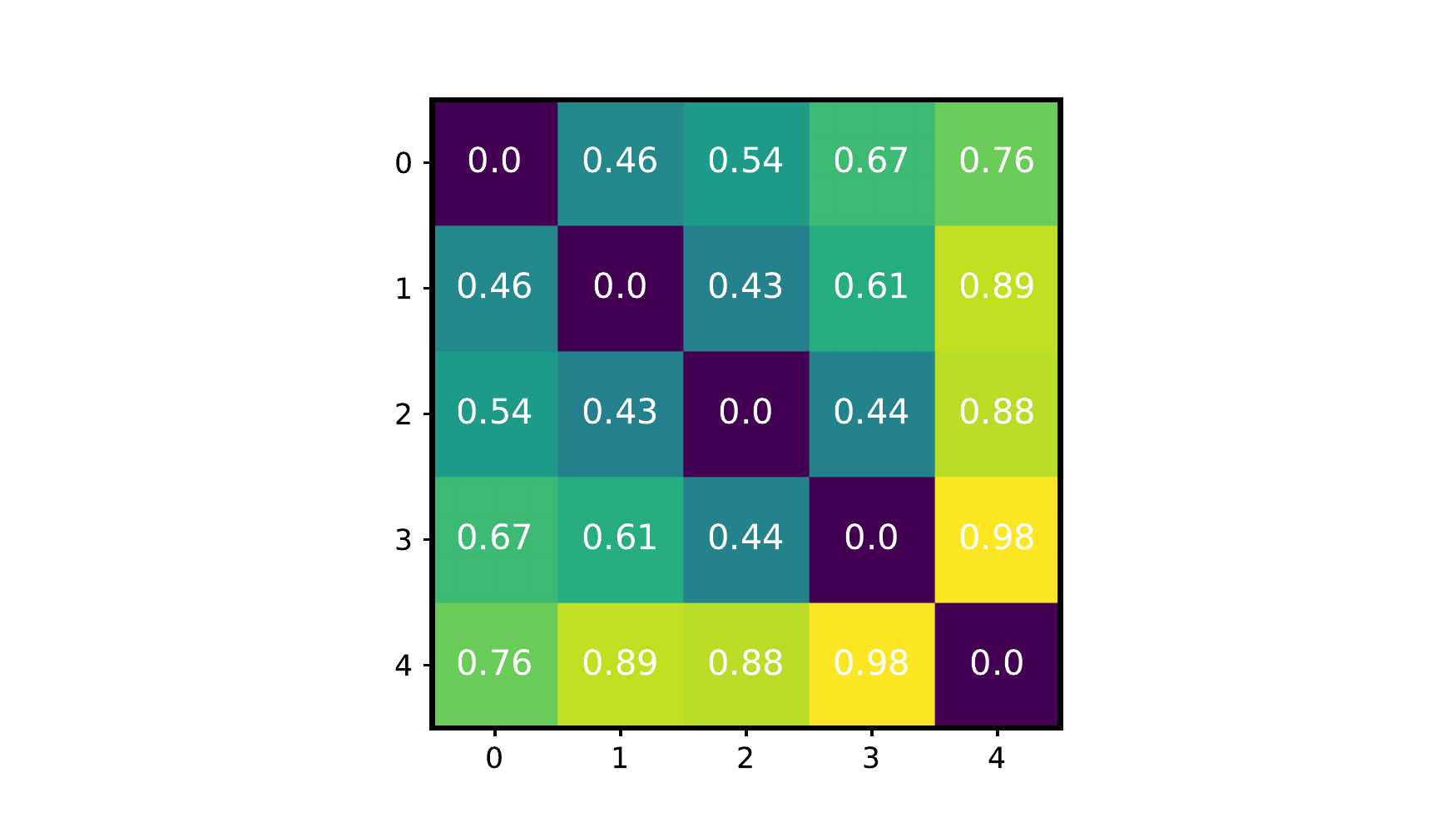}}
  \hspace{0.001in}
  \subfloat[Heterogenous Duplication]{\includegraphics[width=0.48\linewidth]{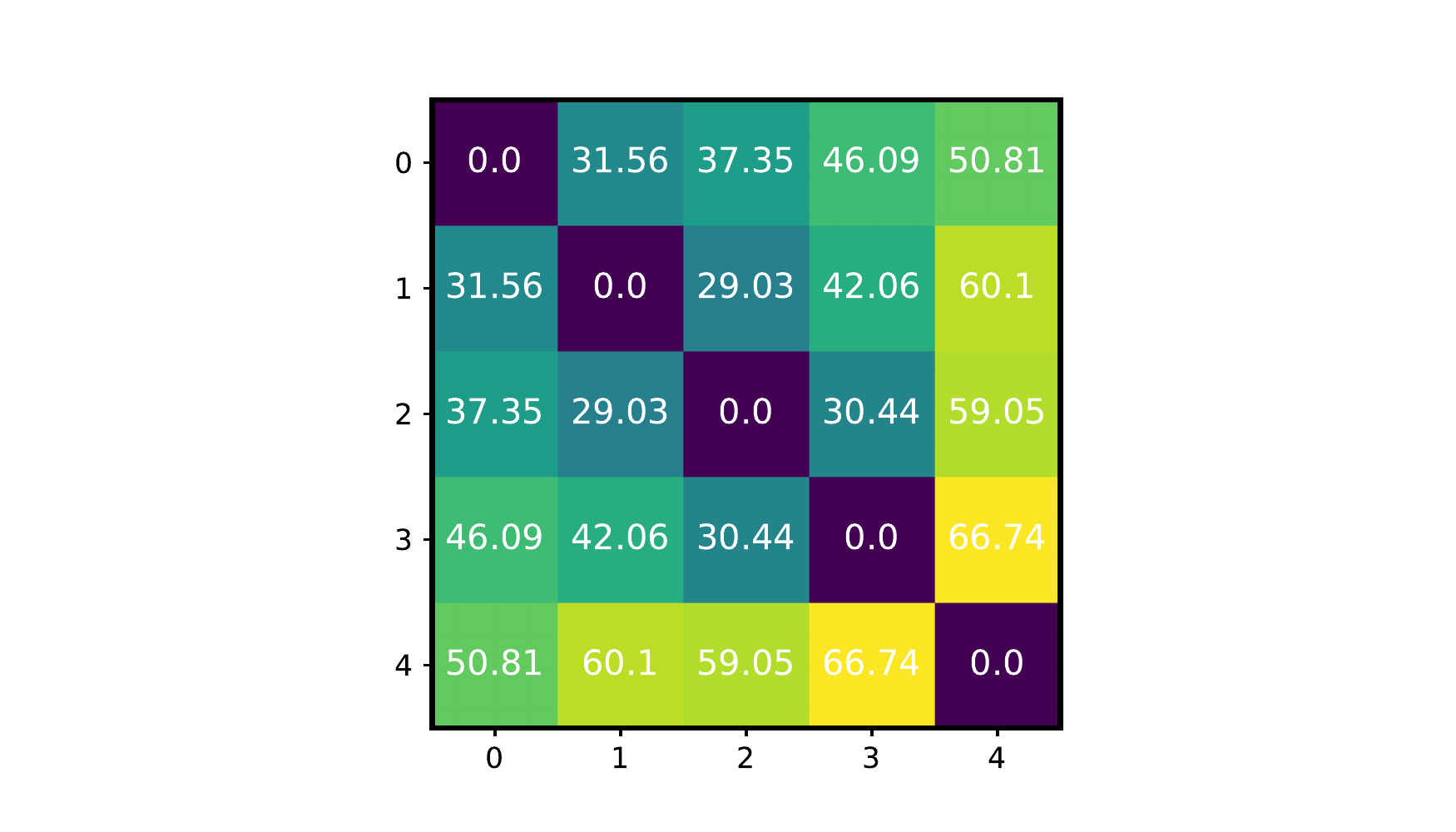}}
  \hspace{0.001in}
  \caption{Euclidean distance based similarity relationships of the sequentially similar odors at various stages of data regularization. a) After scaling the data. b) After application of concentration normalization. c) After application of heterogeneous duplication. }  
  \label{all_preprocessors}
\end{figure}

To study our model's similarity regulation capacity in a systematic manner, we again generated a set of synthetic high-dimensional (odorant-like) stimuli with systematic similarity relationships( $inter \, odor \, distance \, = \, 0.5$, see materials \& methods for details).  We first generated four sequentially similar stimuli (\#0-\#3) ~\cite{Cleland2002} and a non-overlapping stimulus (\#4).  These stimuli then were preprocessed using data regularization techniques to obtain activity distributions identical to described earlier. Fig ~\ref{all_preprocessors} shows at different stage of data regularization such as \textit{scaling}, \textit{concentration normalization} and \textit{data regularization}, similarity relationships between odors are retained, e.g the euclidean distance between odors $0$ and $1$ is always less than odor $0$ and $4$. 

To compute similarity, we used the interneuron spiking overlap metric described in ~\cite{BorthakurCleland2019, Linster2010}. Briefly, the similarity between two stimuli here is defined by the proportion of interneurons activated by both stimuli.  

\begin{figure}
  \centering
  \subfloat[]{\includegraphics[width=0.48\linewidth]{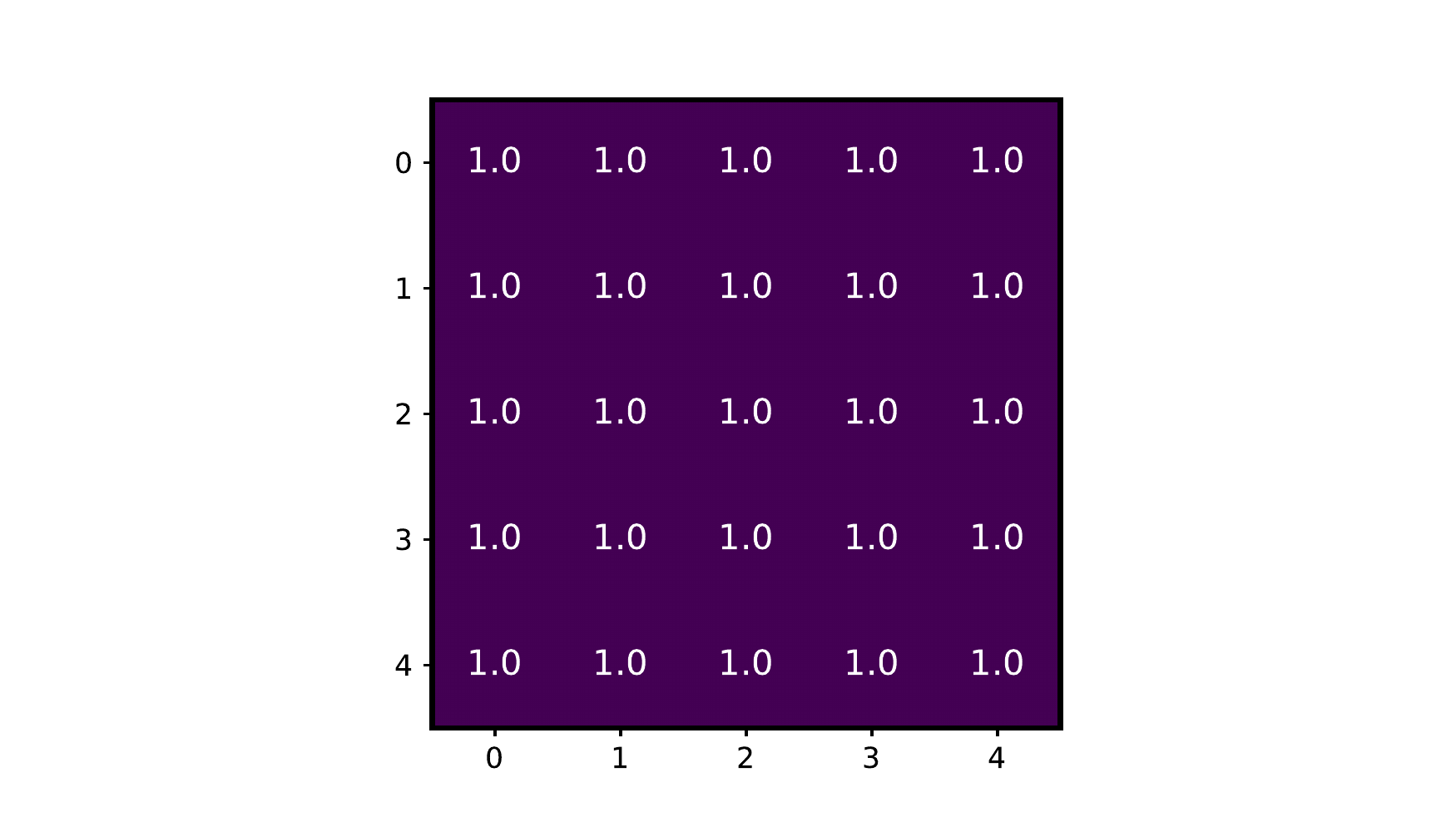}}
    \hspace{0.001in}
  \subfloat[]{\includegraphics[width=0.48\linewidth]{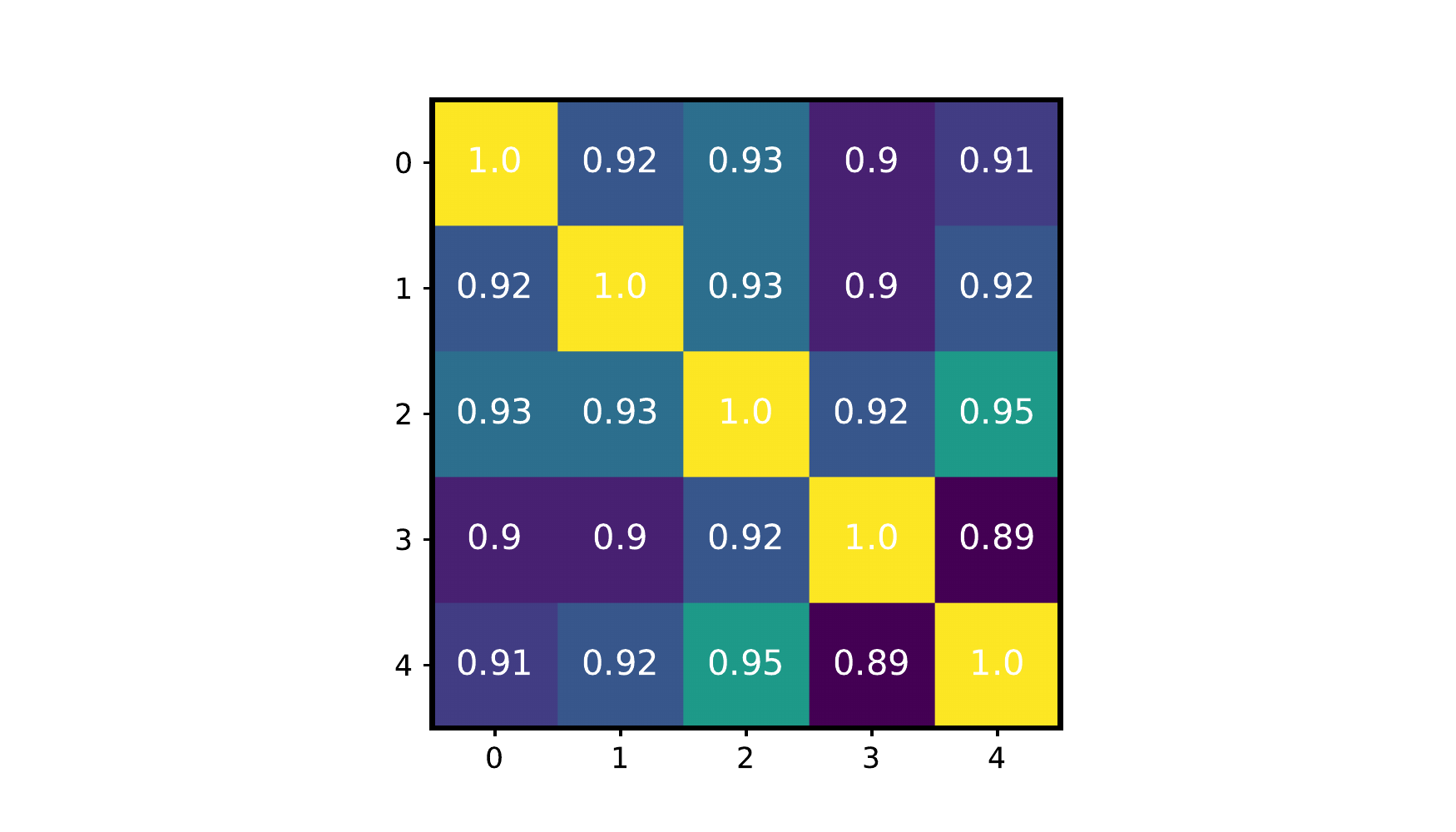}}
  \hspace{0.001in}
  \subfloat[]{\includegraphics[width=0.48\linewidth]{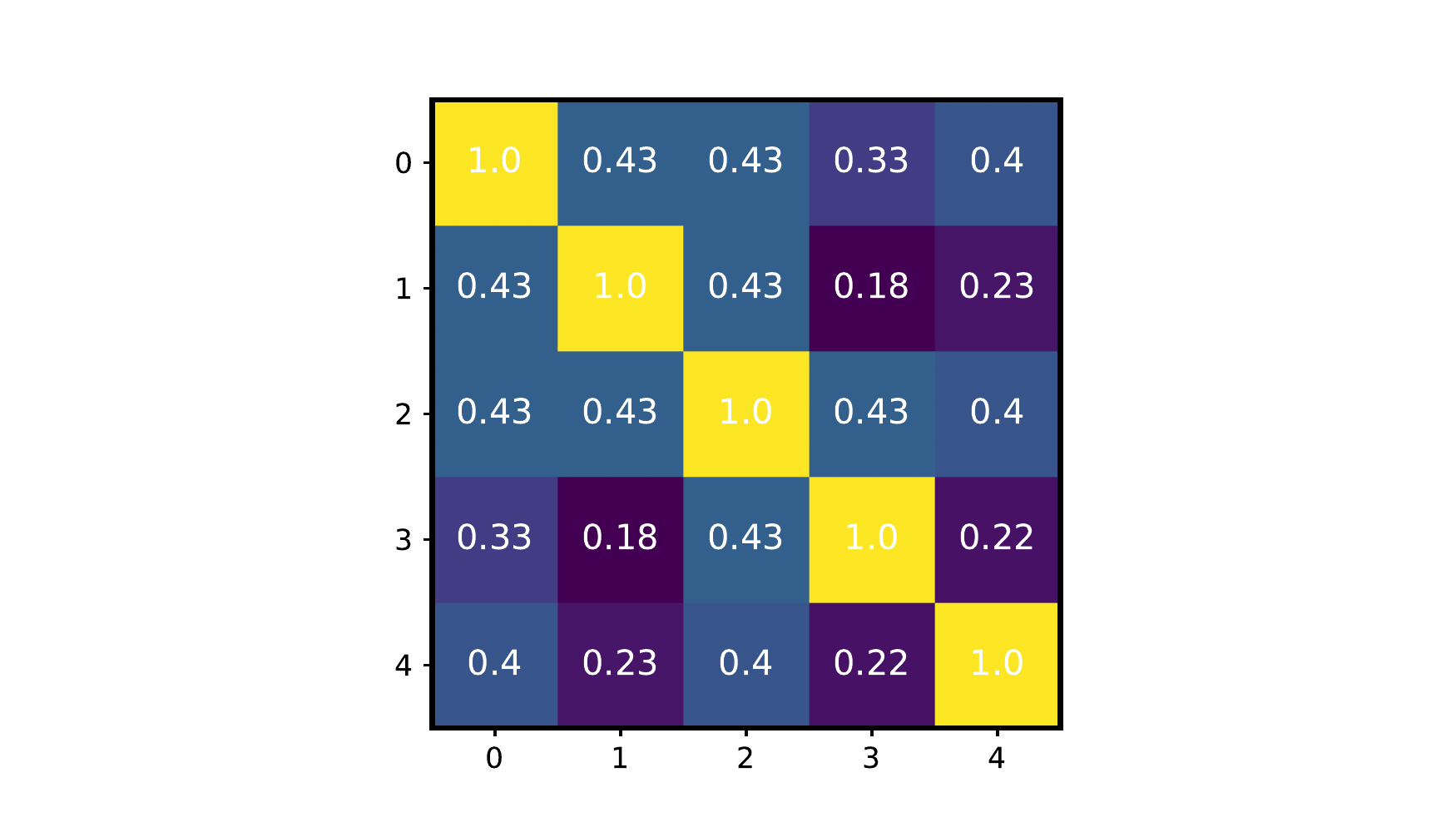}}
  \hspace{0.001in}
  \caption{GC / Interneuron spike count overlap as a metric of odor similarity mapping without sequential learning in the network. $a, b \, \& \, c$ correspond to the first three lower GC threshold group. }  
  \label{ovlp_nl}
\end{figure}

Prior to learning, with sparse random projection in the mitral to granule network, each granule cell is responsive ( can spike ) to multiple odors. This is because, though not fully connected, a significant number of mitral cells have excitatory projections to a granule cell. Consequently, the granule spike count overlap among all five stimuli was high and their representational differences were minimal (Figure~\ref{ovlp_nl}).

\begin{figure}
  \centering
  \subfloat[]{\includegraphics[width=0.48\linewidth]{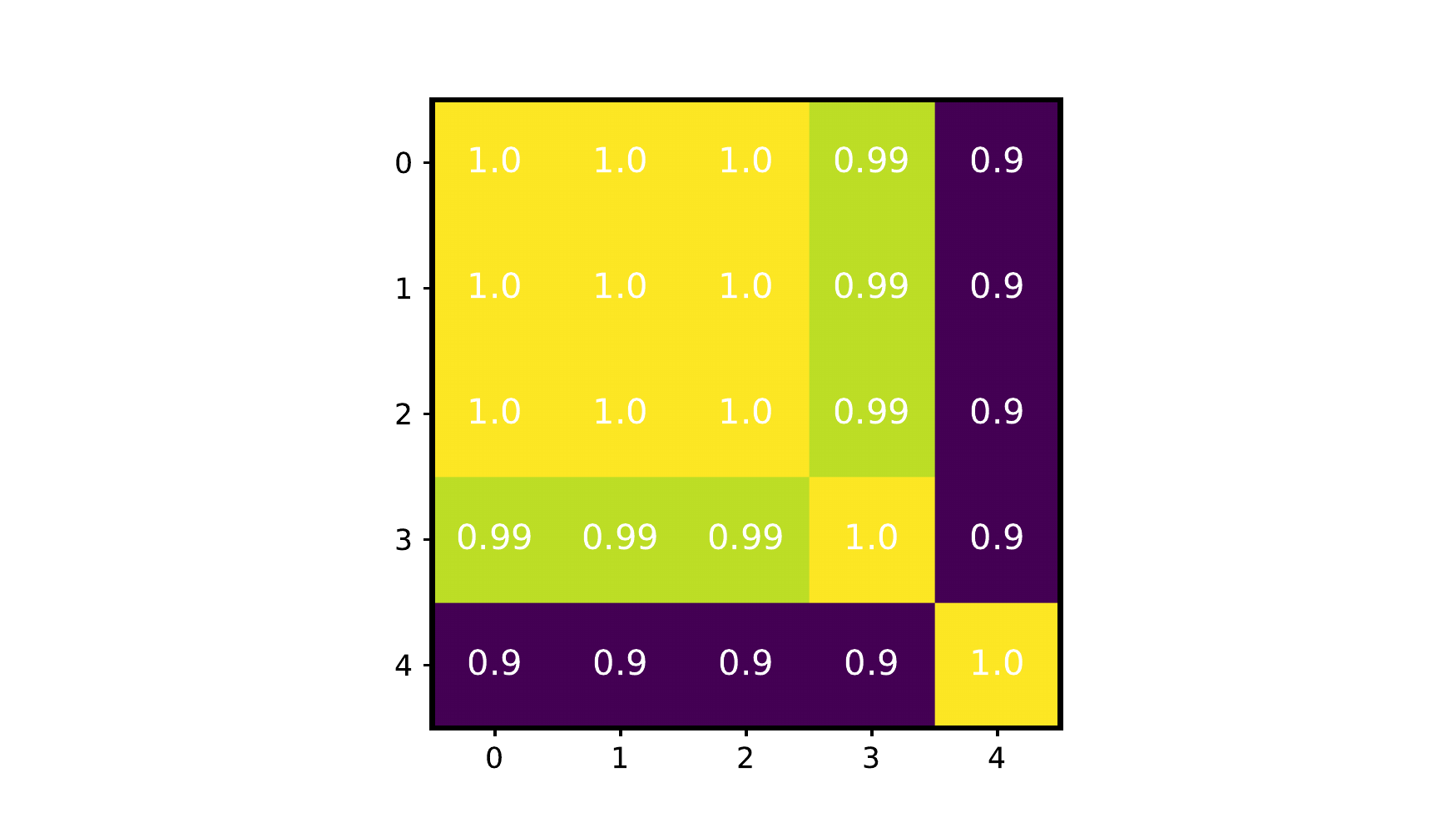}}
  \hspace{0.001in}
  \subfloat[]{\includegraphics[width=0.48\linewidth]{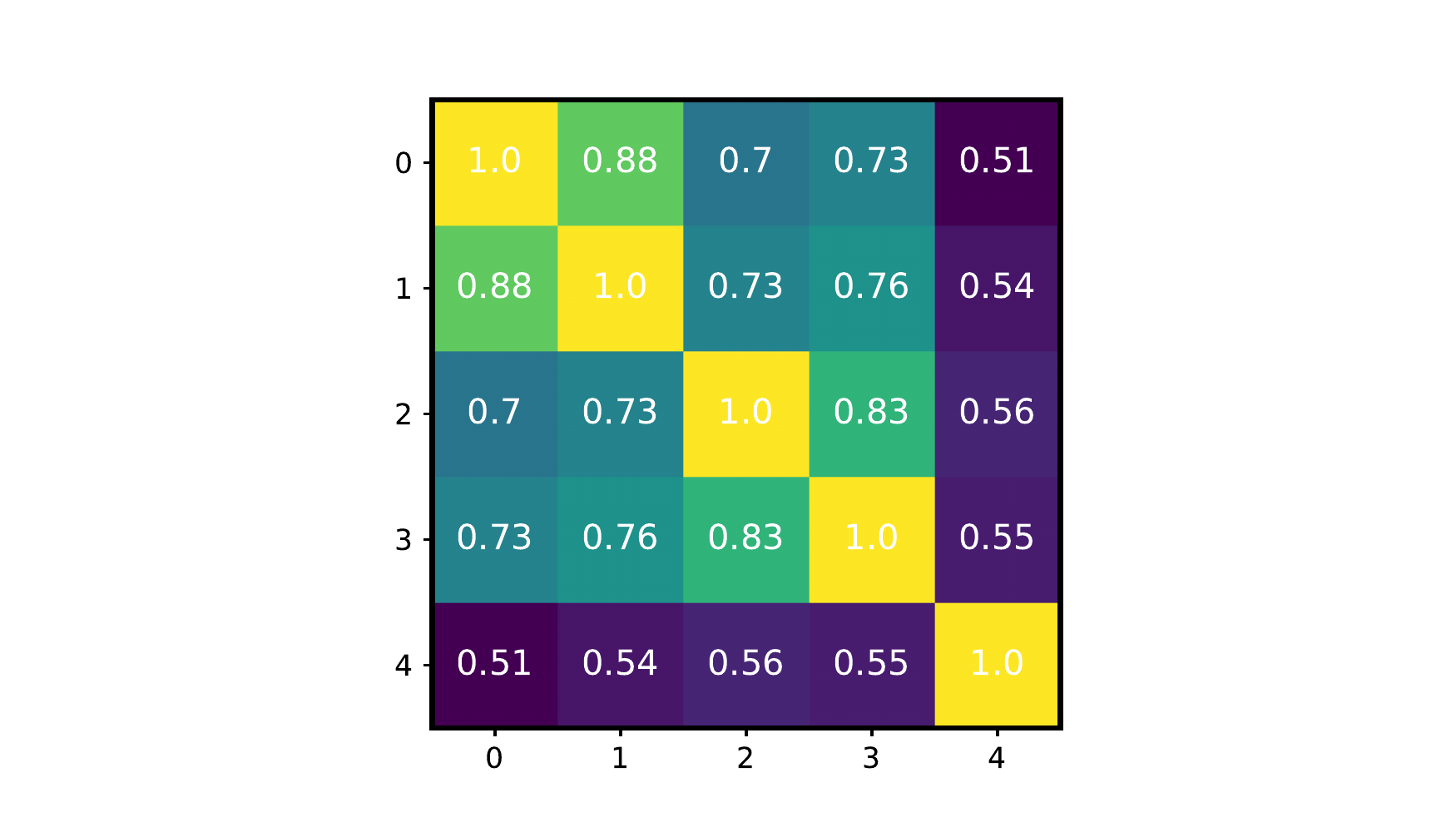}}
  \hspace{0.001in}
  \subfloat[]{\includegraphics[width=0.48\linewidth]{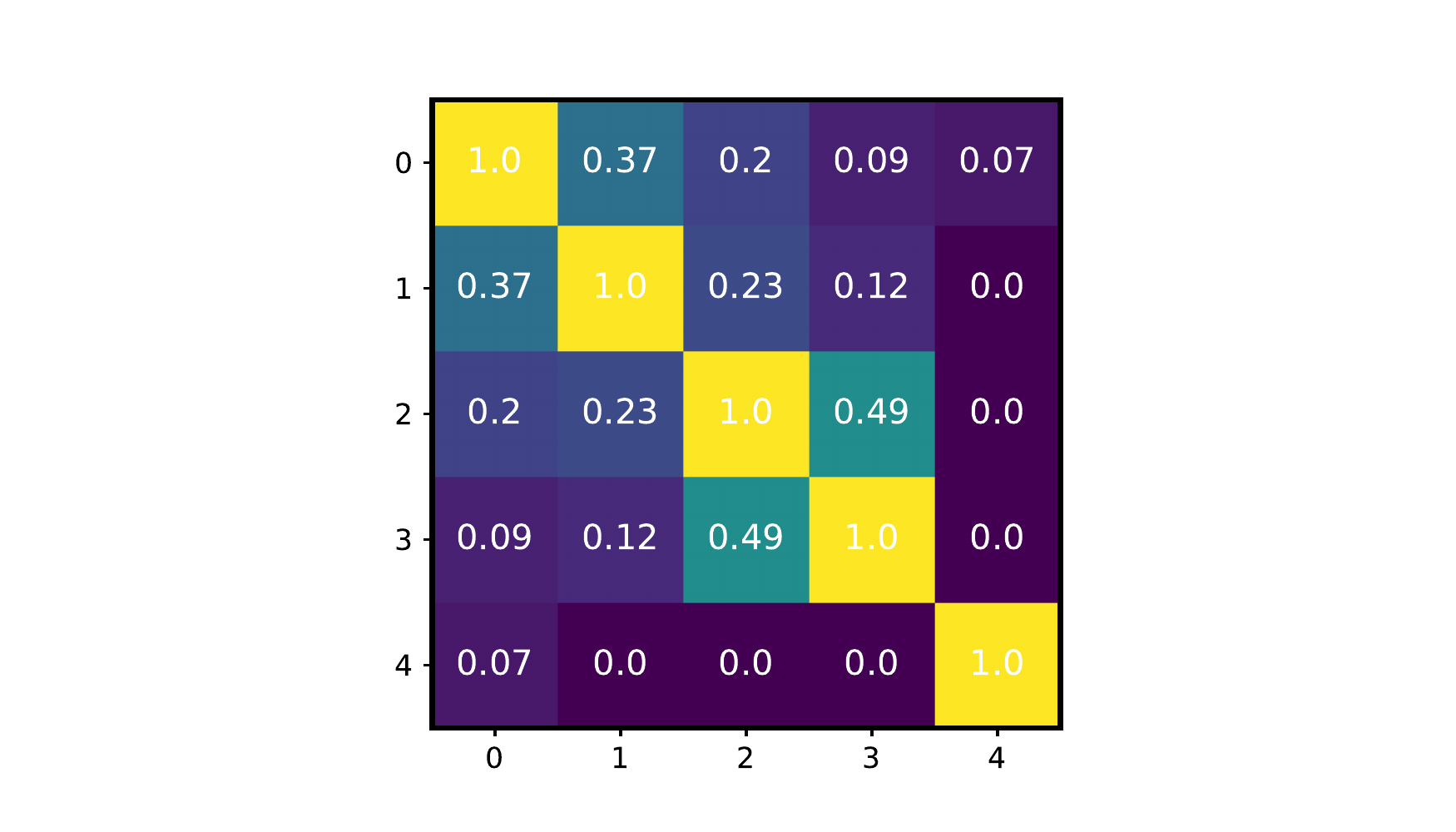}}
  \caption{GC / Interneuron spike count overlap as a metric of odor similarity mapping due to sequential learning in the network. $a, b \, \& \, c$ correspond to the first three lower GC threshold groups.}  
  \label{ovlp_no_neurogen}
\end{figure}

Whereas, sequential learning in the network allocates resources selectively to trained stimuli, which progressively reduces their similarities over the course of learning (Figure~\ref{ovlp_no_neurogen}b).  For example, the interneuron overlap between odor $0$ and $3$ is reduced after learning (Fig ~\ref{ovlp_no_neurogen}b vs Fig ~\ref{ovlp_nl}b). 

\paragraph{Importance of heterogeneous granule cell interneuron thresholds}
\begin{itemize}
    \item As discussed earlier, introduction of GC spiking thresholds is essential for regularizing GC spike counts. 
    \item The responsiveness of GCs to an odor is dependent upon spiking thresholds. Fig ~\ref{ovlp_nl}a, Fig ~\ref{ovlp_no_neurogen}shows the overlap of interneurons / GCs for the lowest group of GC thresholds. These neurons are responsive to multiple odors even after learning and hence the overlap is high. Whereas, the GCs are much more selective to odors as threshold is raised. Hence, as can be seen in Fig ~\ref{ovlp_nl}b,c; Fig ~\ref{ovlp_no_neurogen}b,c, the overlap between odors are reduced. In other words, as the threshold is raised, similarity between odors expressed in terms of GC overlap decreases. Also, the more the similarity of odors, higher is the requirement of spiking threshold for achieving orthogonality of odors. 
\end{itemize}

\subsubsection{Need for neurogenesis}

\begin{figure}
  \centering
  \includegraphics[width=0.9\linewidth]{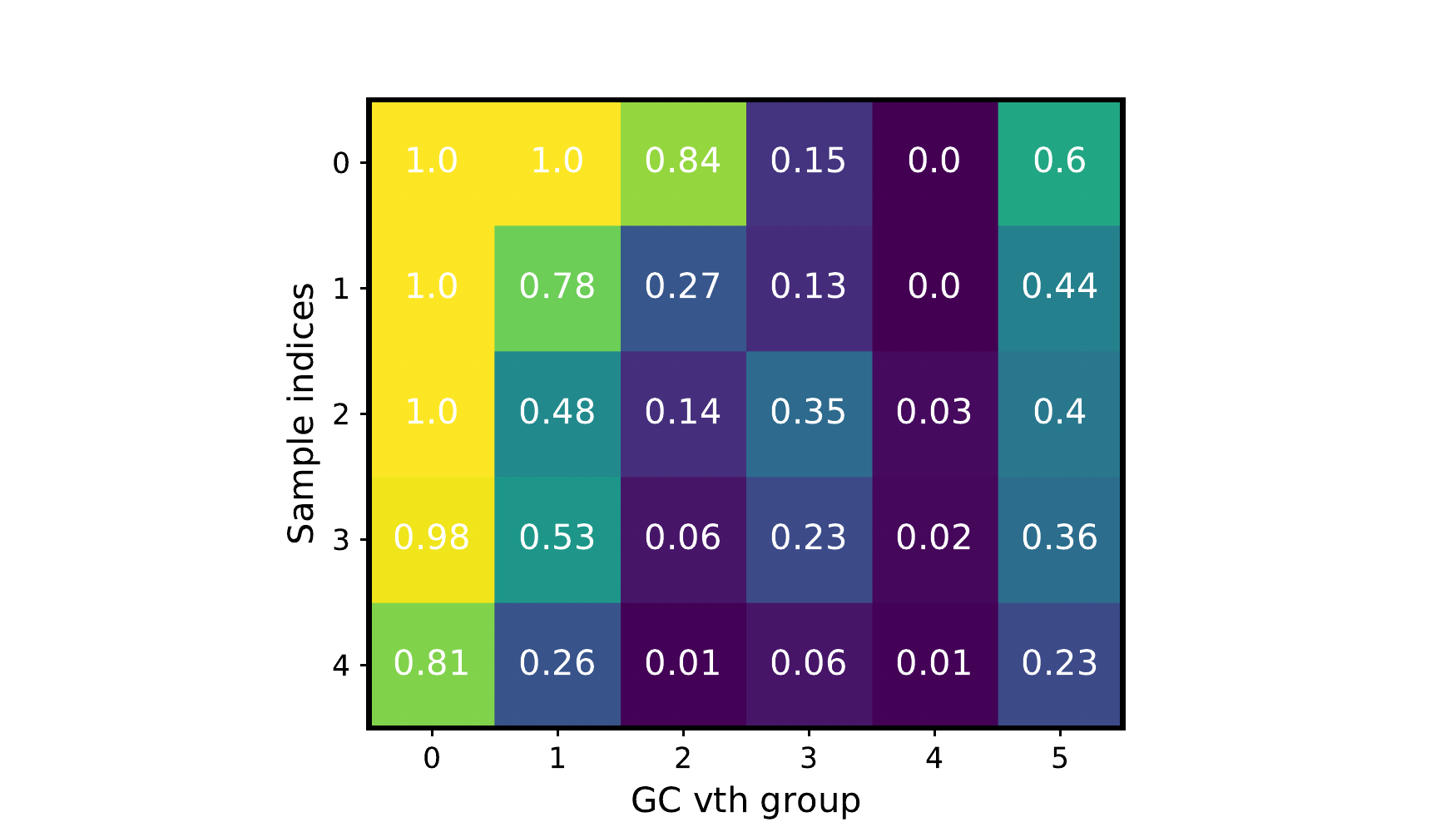}
  \caption{Variation of GC spike counts w.r.t threshold group when the network learns similar odors sequentially. }  
  \label{no_neurogen_cnt}
\end{figure}

Fig ~\ref{no_neurogen_cnt} shows the GC / interneuron spike counts w.r.t the different GC threshold groups while sequentially learning the similar odors keeping everything else constant. Fig ~\ref{het_cp_vth} shows that for the same scenario but without learning, the GC spike counts are almost the same for say threshold group 1. But after learning of odor 1, Fig ~\ref{no_neurogen_cnt} shows that the GC spike counts for odor 2 reduce significantly. This decreases even more for the subsequent learning. 

\begin{figure}
  \centering
  \includegraphics[width=0.9\linewidth]{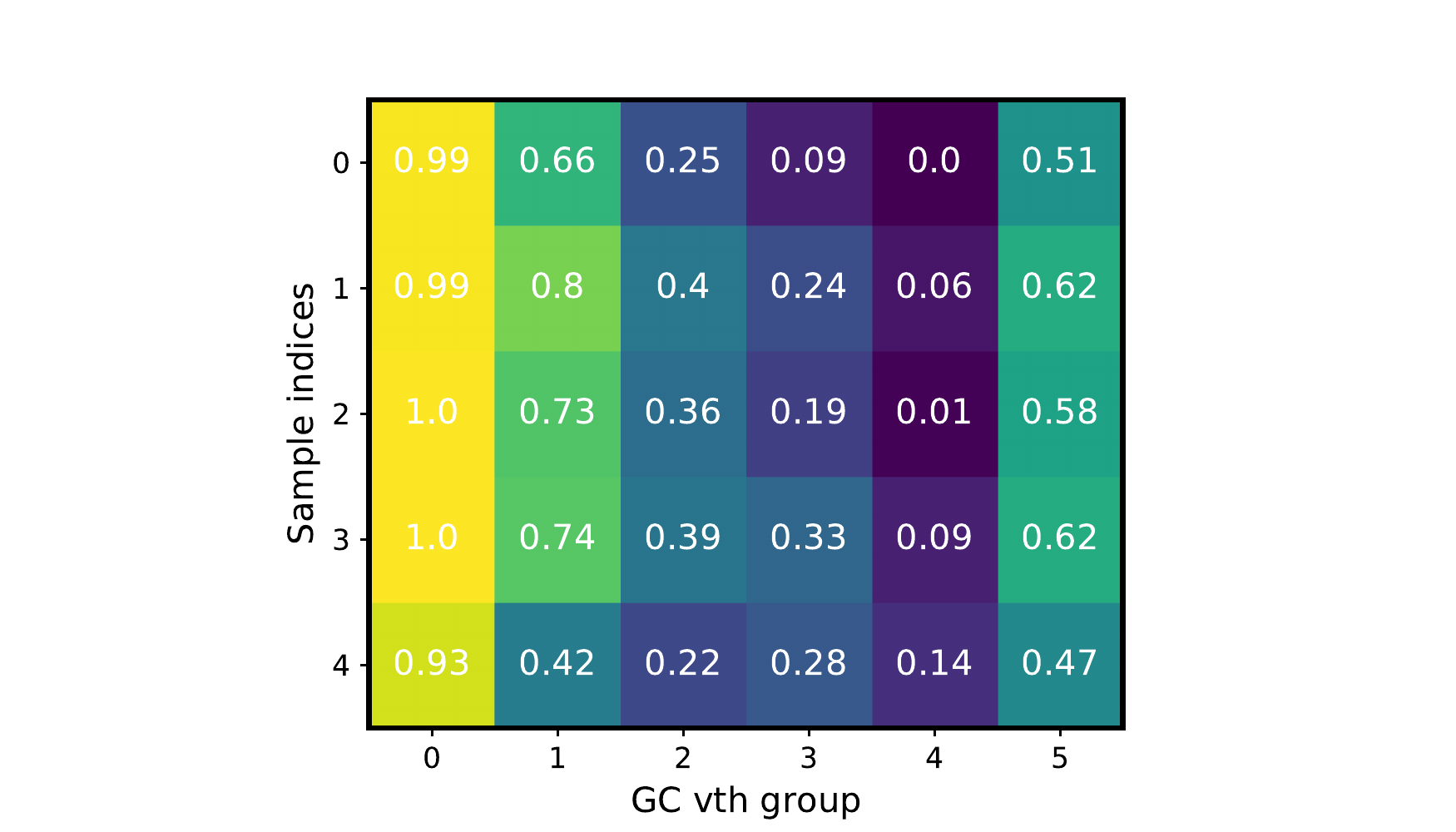}
  \caption{Variation of GC spike counts w.r.t threshold group when the network learns similar odors sequentially and neurogenesis is applied.}  
  \label{neurogen_cnt}
\end{figure}

In the mammalian olfactory bulb, the external plexiform layer is one of the sites of neurogenesis. By this method, new born GCs populations are added to the layer as time progresses. Multiple experimental studies have indicated its existence and its role in odor coding ~\cite{murthy, Shani-Narkiss2020}. But the exact mechanism / criteria for neurogenesis is not yet clear. Inspired by this phenomenon, Imam \& Cleland ~\cite{imam_rapid_2019} introduced a new group ( also of equal size ) of GCs for every new odor learning.  We here seek to implement neurogenesis in a more biologically plausible way which enables mapping of stimuli similarity, and ensures better scalability of models. Accordingly, we introduced a column specific differentiation dependent neurogenesis method. As per method, for any column, newborn neurons are added for first differentiated neurons. With learning, the number of GC per column increases. We applied this method and as observed in Fig ~\ref{neurogen_cnt}, the GC spike counts are not affected by sequential learning. 

\begin{figure}
  \centering
  \subfloat[]{\includegraphics[width=0.48\linewidth]{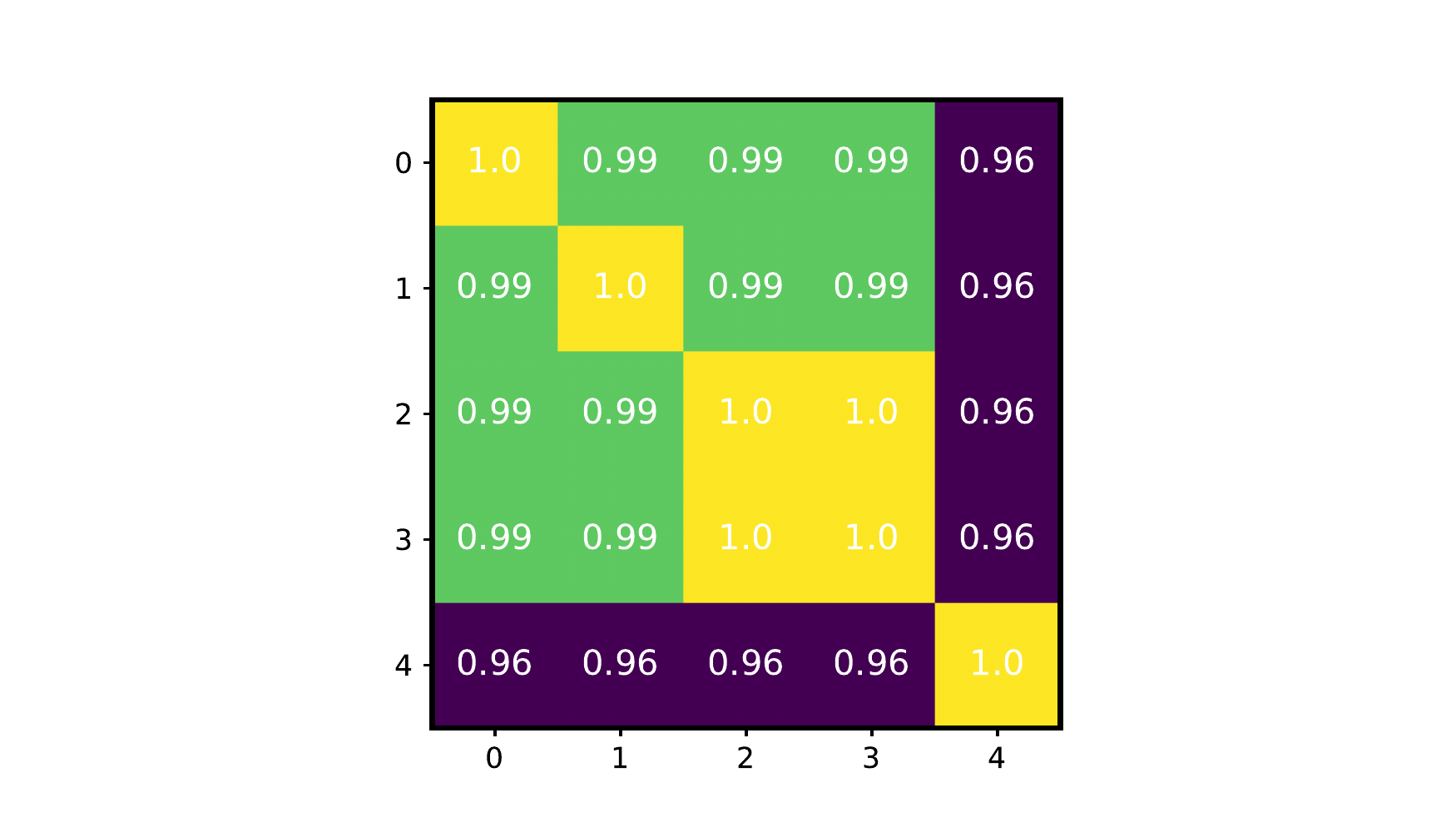}}
  \hspace{0.001in}
  \subfloat[]{\includegraphics[width=0.48\linewidth]{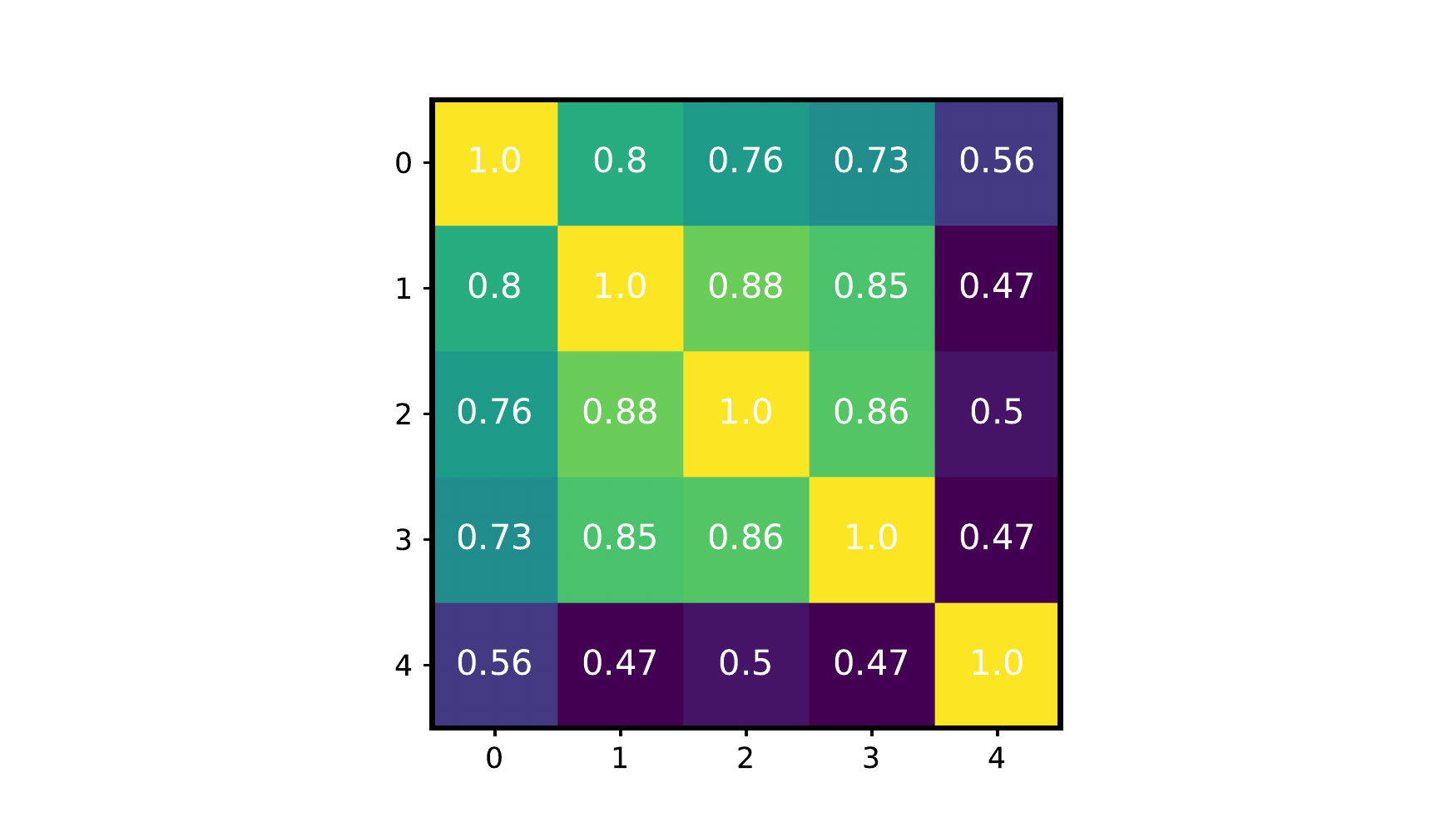}}
  \hspace{0.001in}
  \subfloat[]{\includegraphics[width=0.48\linewidth]{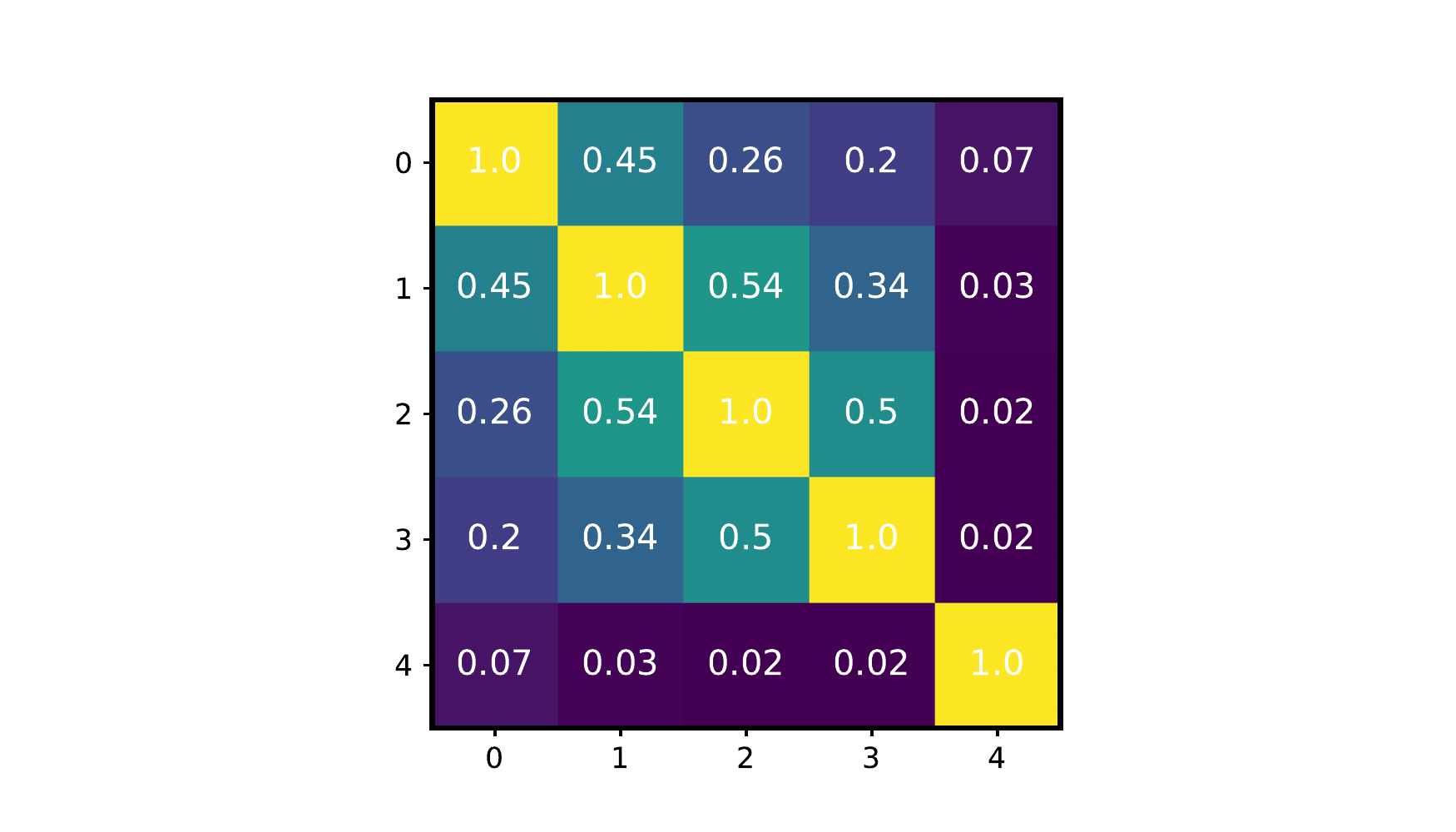}}
  \hspace{0.001in}
  \subfloat[]{\includegraphics[width=0.48\linewidth]{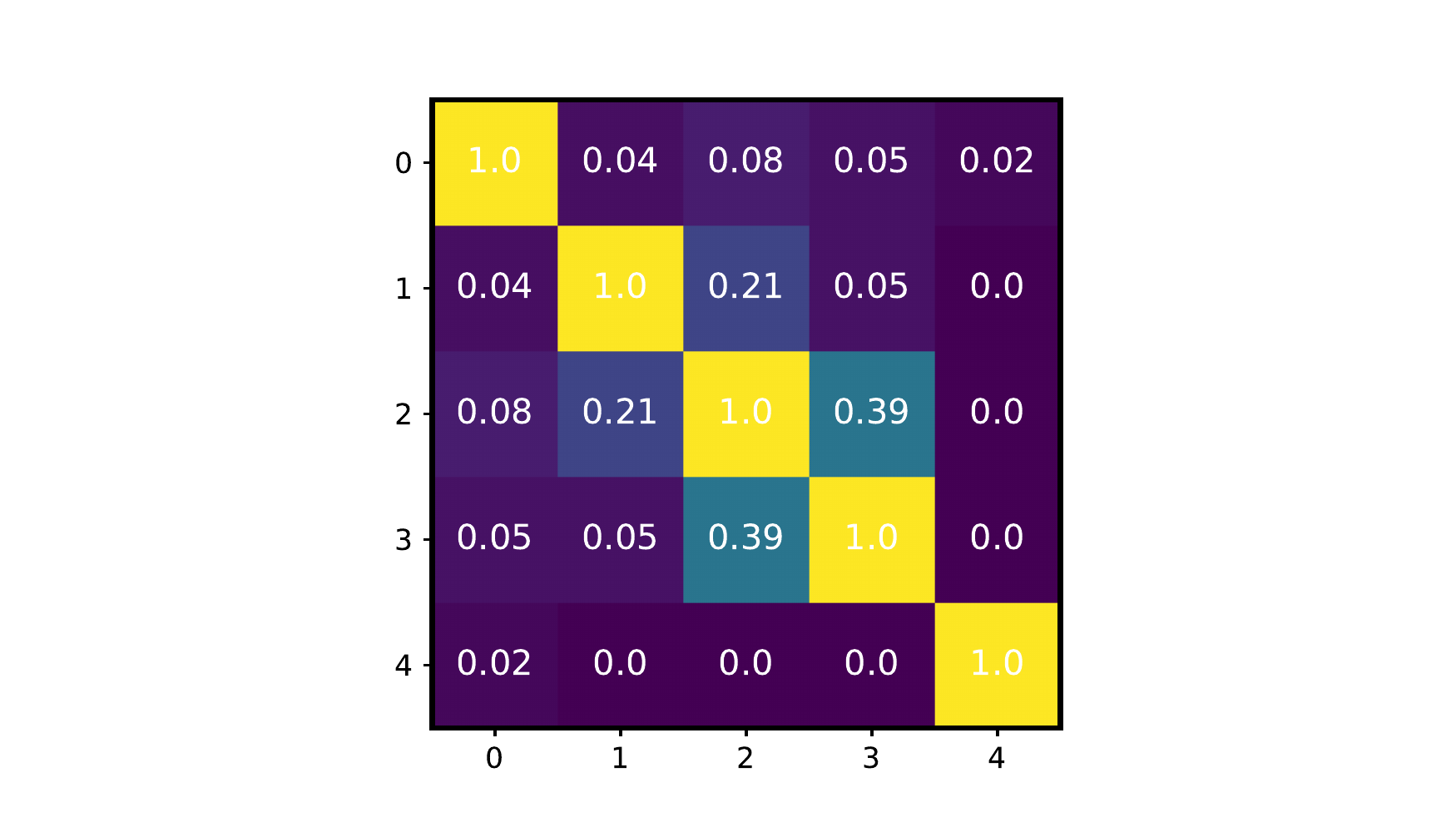}}
  \caption{GC / Interneuron overlap after sequential learning of similar odors with neurogenesis in the model. $a, b, c \, \& \, d$ correspond to figures for the first four threshold groups ( in increasing order of threshold value).  }  
  \label{ovlp_neurogen}
\end{figure}

Fig ~\ref{ovlp_neurogen} shows the overlap of interneurons / GCs for the first four GC threshold groups after introduction of neurogenesis. Neurogenesis doesn't destroy the similarity mapping capacity of the feed-forward network. 

\subsection{Model Scaling}

\begin{figure}
  \centering
  \subfloat[]{\includegraphics[width=0.32\linewidth]{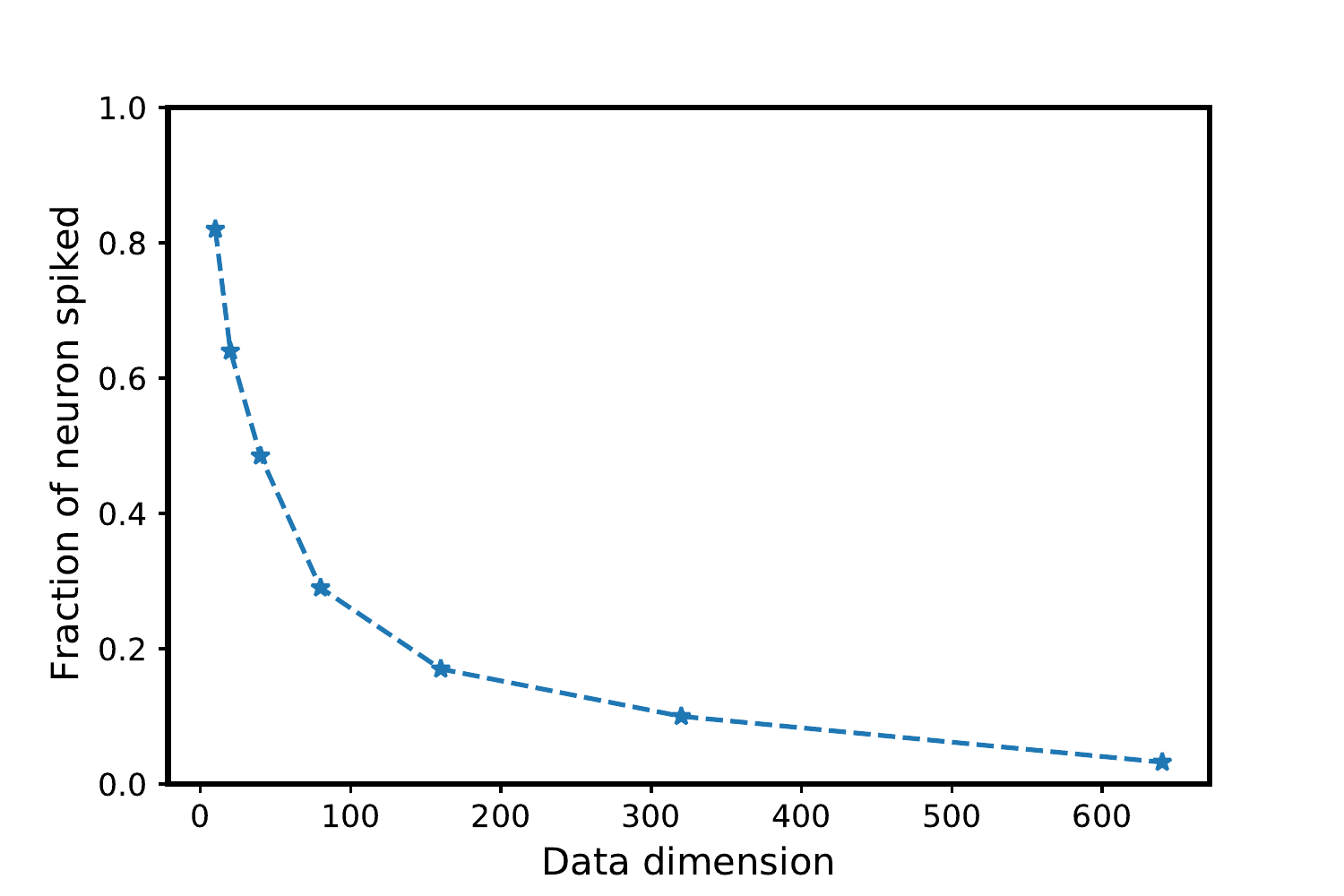}}
  \hspace{0.01in}
  \subfloat[]{\includegraphics[width=0.32\linewidth]{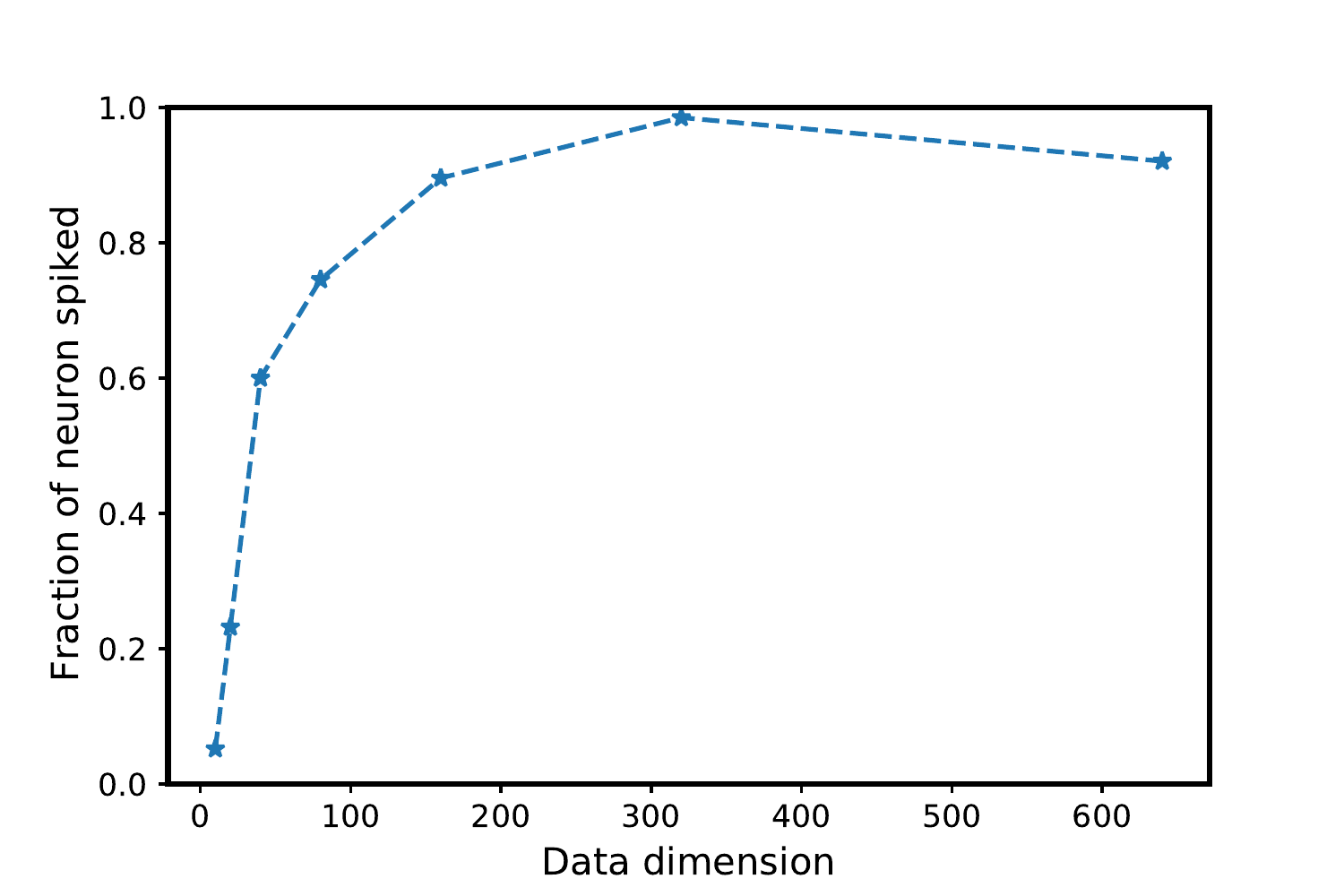}}
  \hspace{0.01in}
  \subfloat[]{\includegraphics[width=0.32\linewidth]{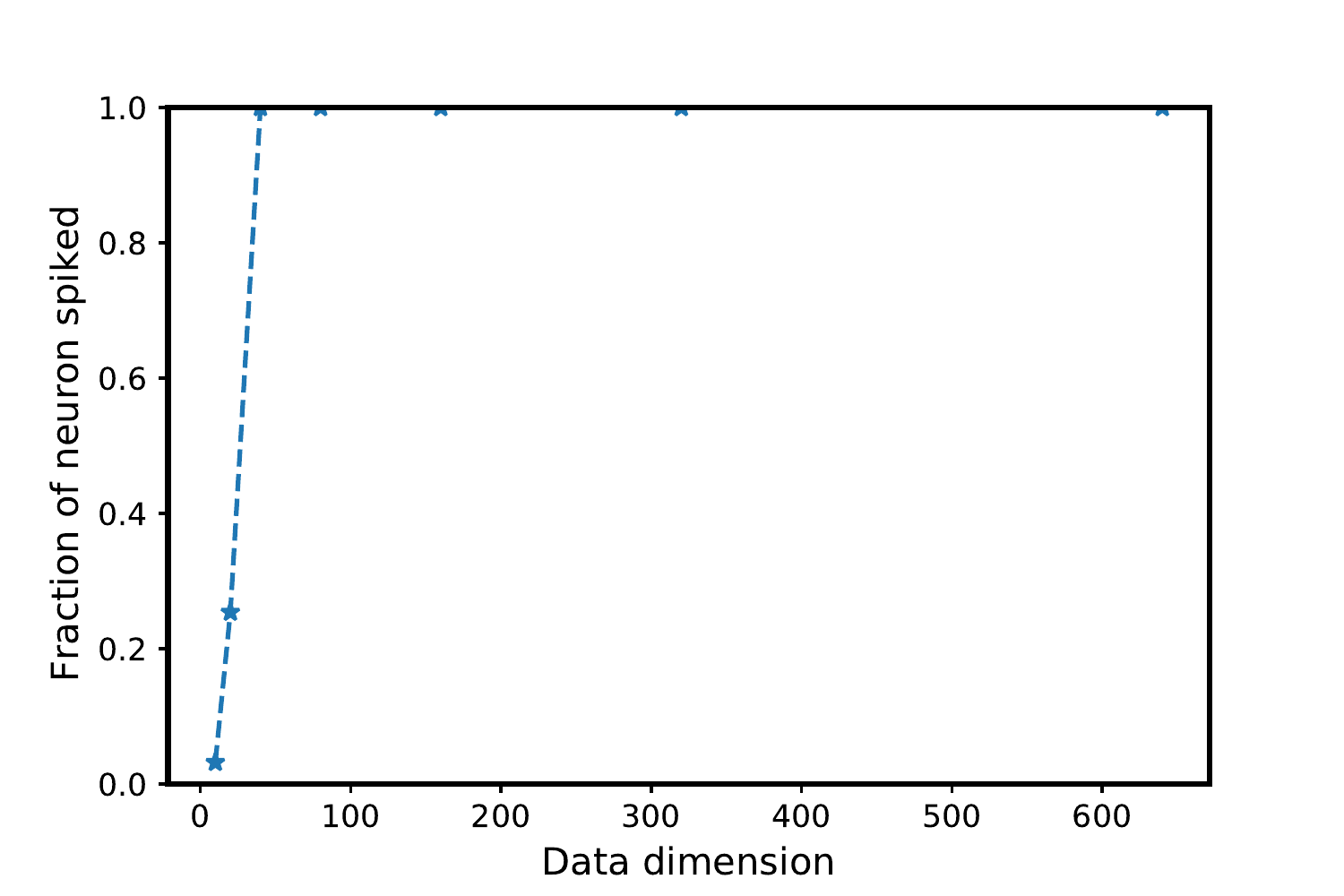}}
  \hspace{0.01in}
 \subfloat[]{\includegraphics[width=0.32\linewidth]{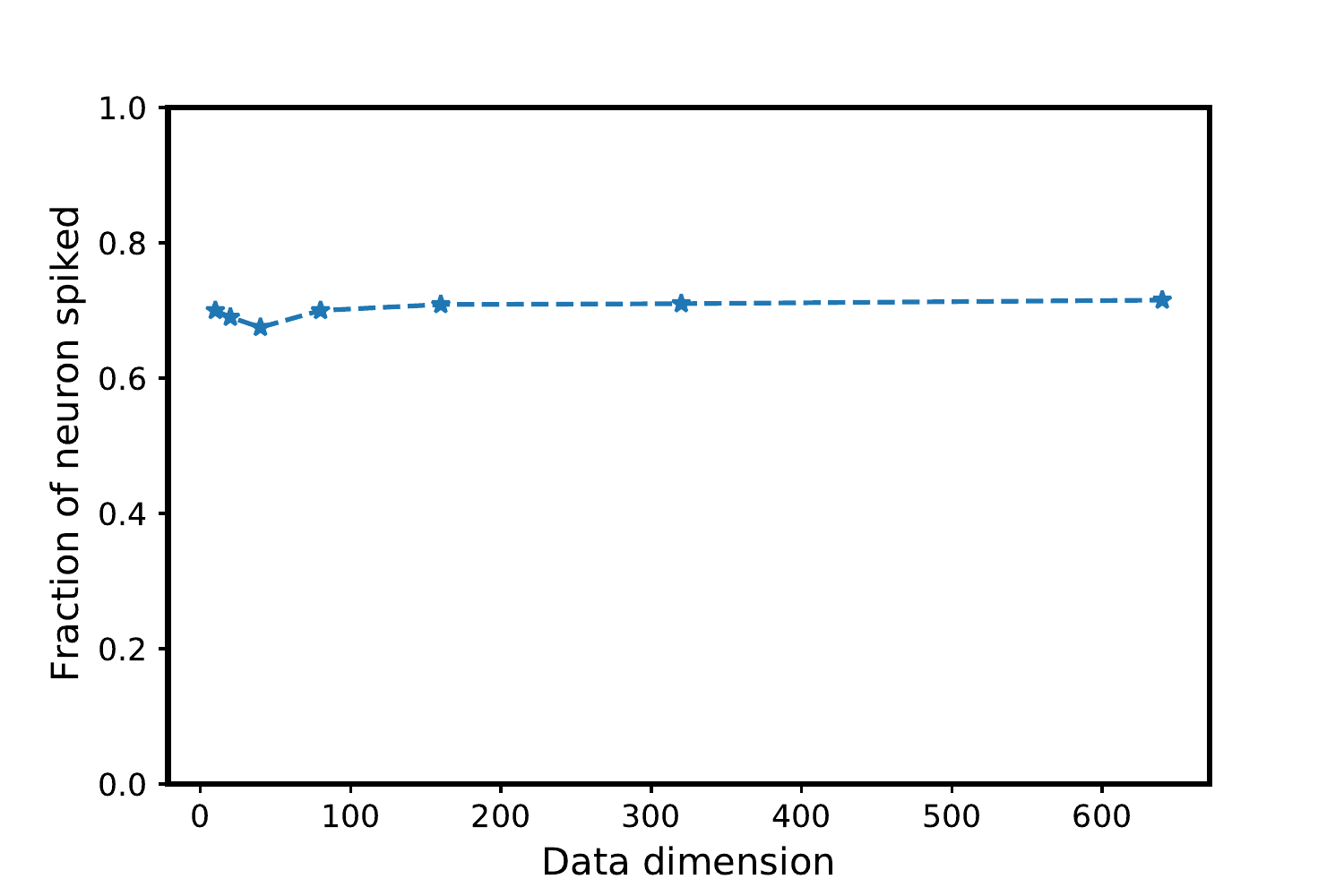}}
  \hspace{0.01in}
  \subfloat[]{\includegraphics[width=0.32\linewidth]{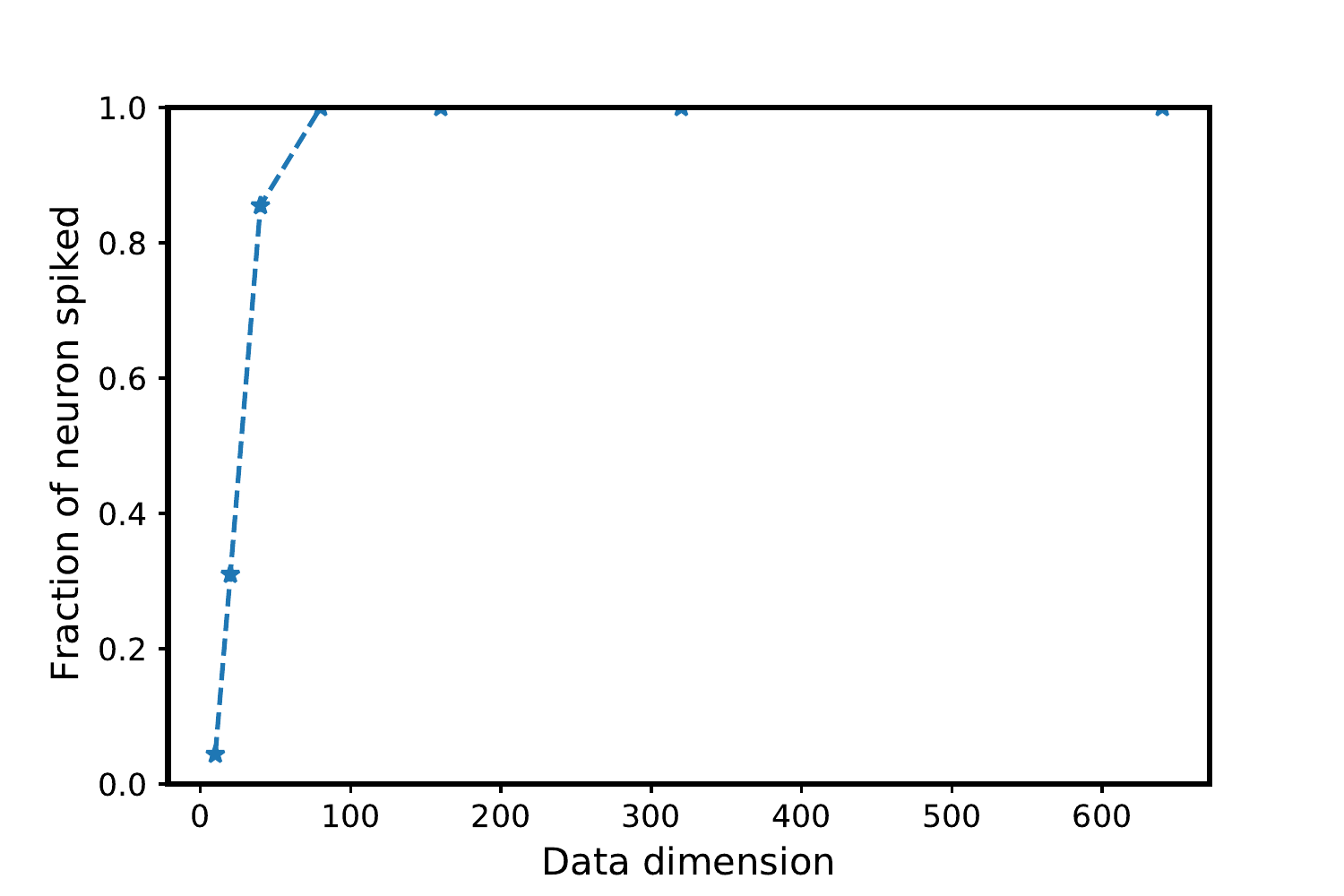}}
  \hspace{0.01in}
  \subfloat[]{\includegraphics[width=0.32\linewidth]{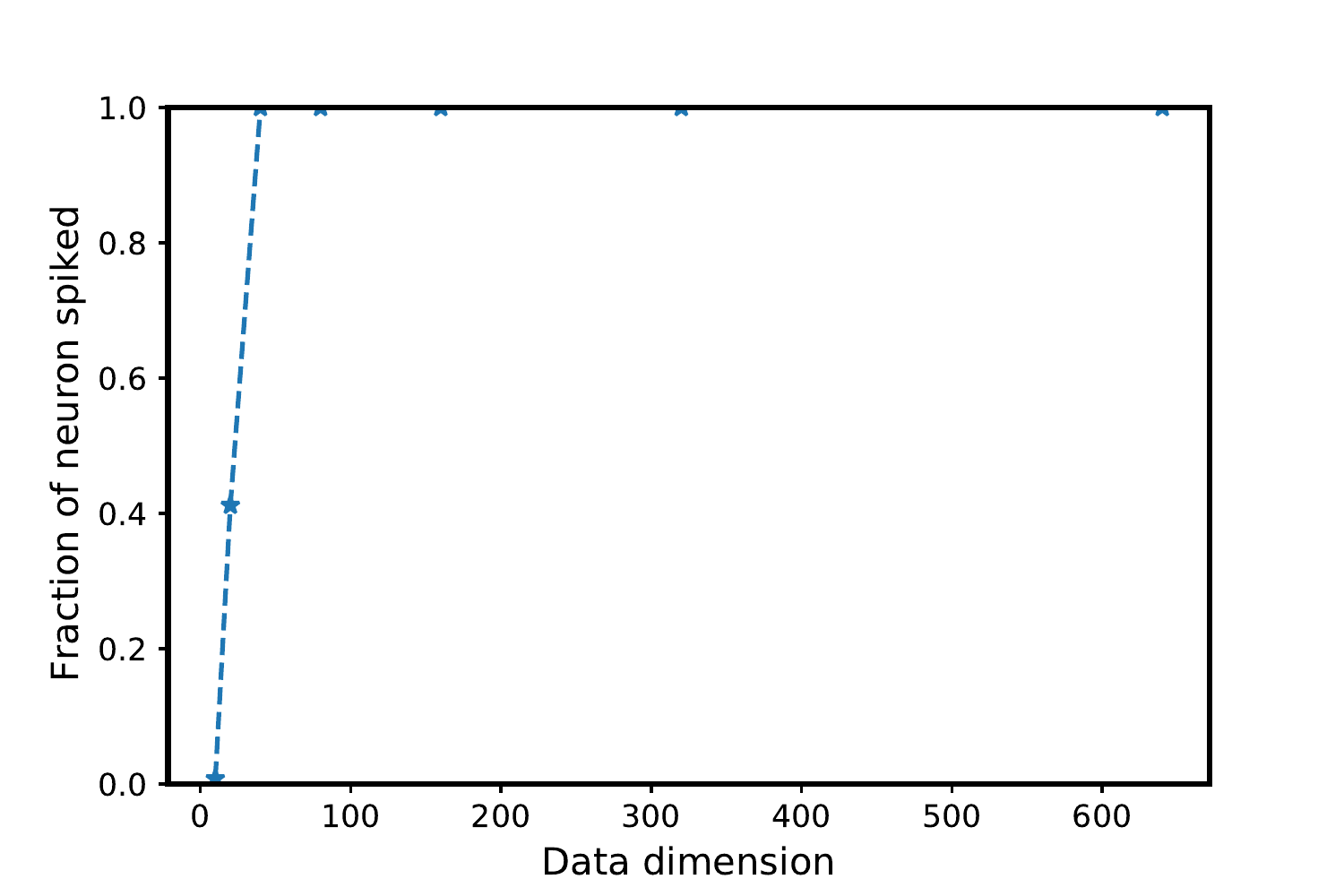}}
  \hspace{0.01in}
  \subfloat[]{\includegraphics[width=0.32\linewidth]{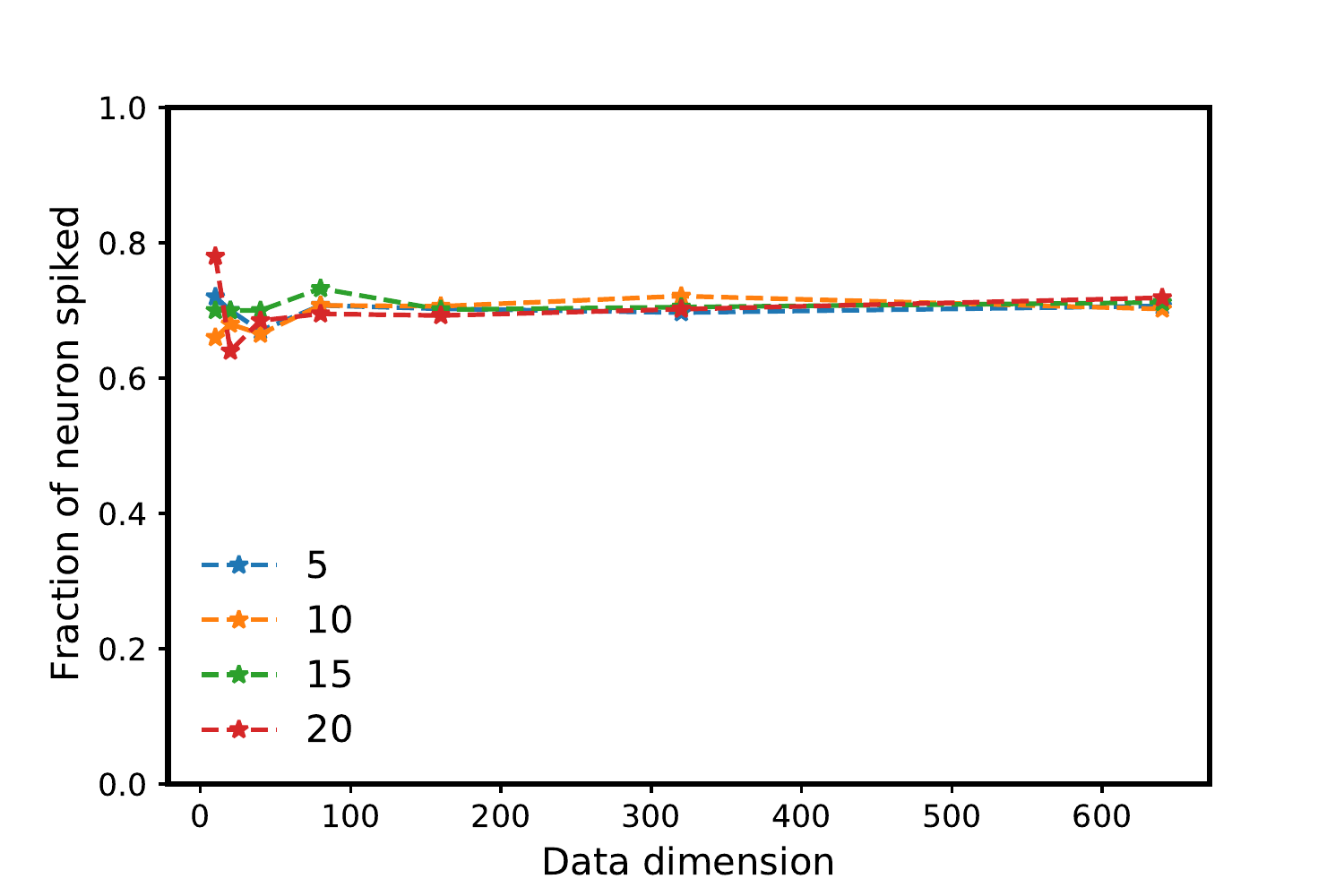}}
  \hspace{0.01in}
  \subfloat[]{\includegraphics[width=0.32\linewidth]{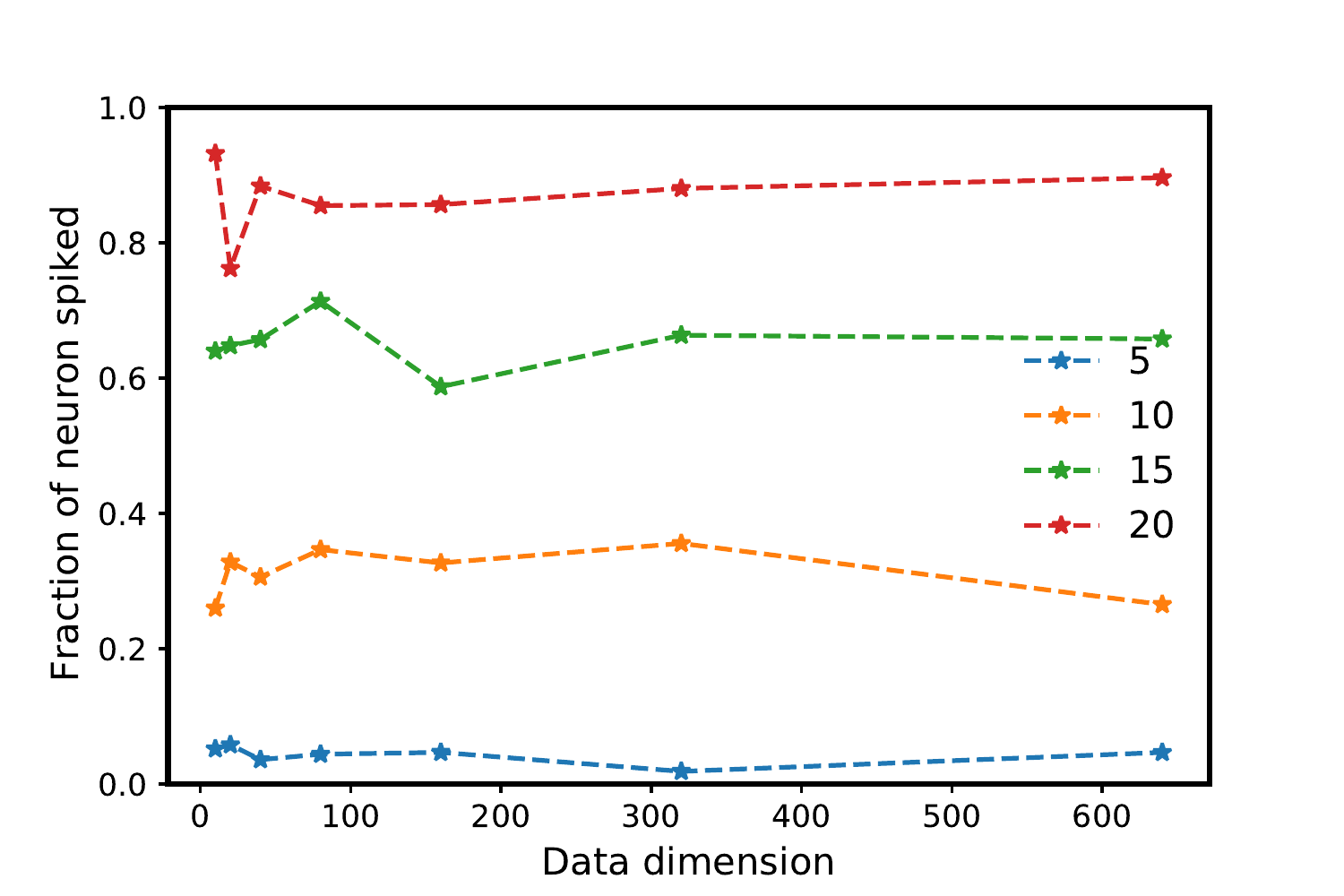}}
  \hspace{0.01in}
  \subfloat[]{\includegraphics[width=0.32\linewidth]{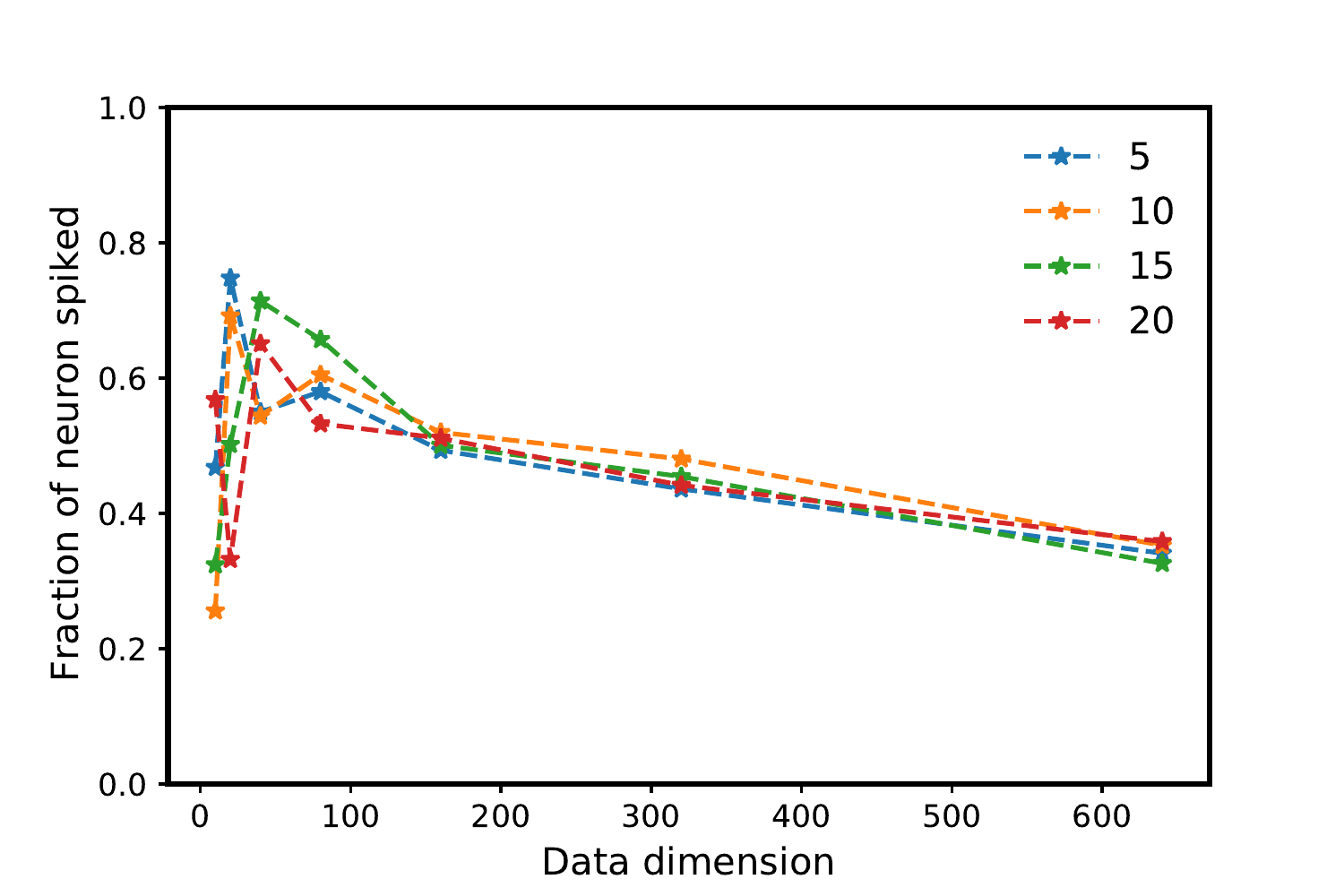}}
  \hspace{0.01in}
  \caption{Variation of MC ( left column ), non learning GC ( middle column ) and learning GC ( right column ) spike counts w.r.t dimension for various levels of convergence of MCs to GCS in a feedforward MC-GC network. a) Variation of fraction of MC spike count w.r.t variation in data dimension from $8$ to $640$ in multiples of $2$. b) Variation of fraction of non learning low threshold GC spike count w.r.t variation in data dimension from $8$ to $640$ in multiples of $2$. $50\%$ of total GCs were selected as non learning. c) Variation of fraction of learning high threshold GC spike count w.r.t variation in data dimension from $8$ to $640$ in multiples of $2$. $50\%$ of total GCs were selected as learning. d) Same as $a$ but after application of model scaling for MC spike count regularization. e, f) Same as $b, c$ after application of MC spike count regularization. g) Variation of fraction of MC spike counts w.r.t variation in data dimension from $8$ to $640$ in multiples of $2$ after application of MC spike count regularization, low threshold and high threshold GC spike count regularization. Color corresponds to different levels of convergence of MCs to low threshold GCs. h) Variation of low threshold non learning GC spike count w.r.t variation in data dimension after application of MC spike count regularization, low threshold and high threshold GC spike count regularization. i) Variation of high threshold learning GC spike count w.r.t variation in data dimension after application of MC spike count regularization, low threshold and high threshold GC spike count regularization.}  
  \label{mod_scaling}
\end{figure}

We here sought to regularize the mitral cells and the granule cells ( both the non learning low threshold and learning high threshold, see materials \& methods for details ) spike counts across dimensions. Accordingly, we used random synthetic odor samples drawn from a uniform distribution whose dimensions can be $10, 20, 40, 80, 160, 320, 640$. Of all the GC thresholds, lowest $50 \%$ were selected as non-learning and the remaining high threshold ones as learning. As the data dimension increases, the absolute spike counts of neurons are expected and also required to increase. But, for appropriate model functioning we expect the fraction of active neurons defined as $\frac{MC/GC \, spike \, count}{Total \, MC/GC \, spike \, count}$ to remain reasonably constant. 
\newline
Fig ~\ref{mod_scaling}shows that the fraction of active MCs decreases when the data dimension is varied from $10-640$ in multiples of $2$. Fig ~\ref{mod_scaling}b,c shows that the corresponding GC spike counts increase with dimension. This is undesirable. 
\newline
In order to regularize MC spike counts, we first regularize the MC for samples of a data dimension ( say $k$). Then after the application of unsupervised concentration tolerance, the data $X^d$ is set to $X^d \times \frac{d}{k}$. For this study $k=20$ and after this step, the fraction of MC spike counts are regularized  across dimensions, Fig ~\ref{mod_scaling}d. But the GC spike counts are not regularized, Fig~\ref{mod_scaling}e,f. 
\newline
To ensure sparsity of GC spiking, we introduced MC-GC convergence conservation policy for the non-learning low threshold GCs into the model. Accordingly, we kept the number of random MCs converging to a low threshold GC constant w.r.t dimension. As observed in Fig ~\ref{mod_scaling}h, when the number of MCs converging to a GC are $5, 10, 15, 20$, the fraction of active non-learning low threshold GCs are reasonably constant. And it is obvious, the fraction of active GCs increases with the convergence number of MCs. For regularizing the spike counts of high threshold GCs, the maximum spiking threshold ($v_{thmax}$) was set to $v_{thmax} \times \frac{d}{k}$. The GC spike counts for the high threshold group were reasonably regularized, Fig ~\ref{mod_scaling}i. 

\subsection{Attractor / oscillator in action}

After the studies on the network comprised of feedforward glomerular and external plexiform layer ( MC - GC projection), we next introduced the GC - MC inhibitory drive ( see materials \& methods ) to the network and created a spatio - temporal attractor / oscillator which can \textit{learn in the wild}. 
With the synthetic odor method described in materials \& methods, we generated $5$ similar odors of dimension $20$ for sequential learning. For testing the attractor performance, we also generated $10$ noisy samples corresponding to each train odor (total $\, = \, 11 \times 5 \, = \, 55 $ samples). This attractor model had $20$ ET, PG, MC cells with heterogeneous duplication level of $5$ for MCs. In a naive network ( without any odor learning ), the number of non learning ( no neurogenesis ) low threshold GCs was $75$ per column (total $\, = 75 \times 20$) and the number of learning ( with neurogenesis ) high threshold GCs was $50$ per column (total $\, = 50 \times 20$). Hence, the model had $125$ different heterogeneous thresholds of GCs ( same values for all columns ).  

\begin{figure}
  \centering
  \includegraphics[width=0.85\linewidth]{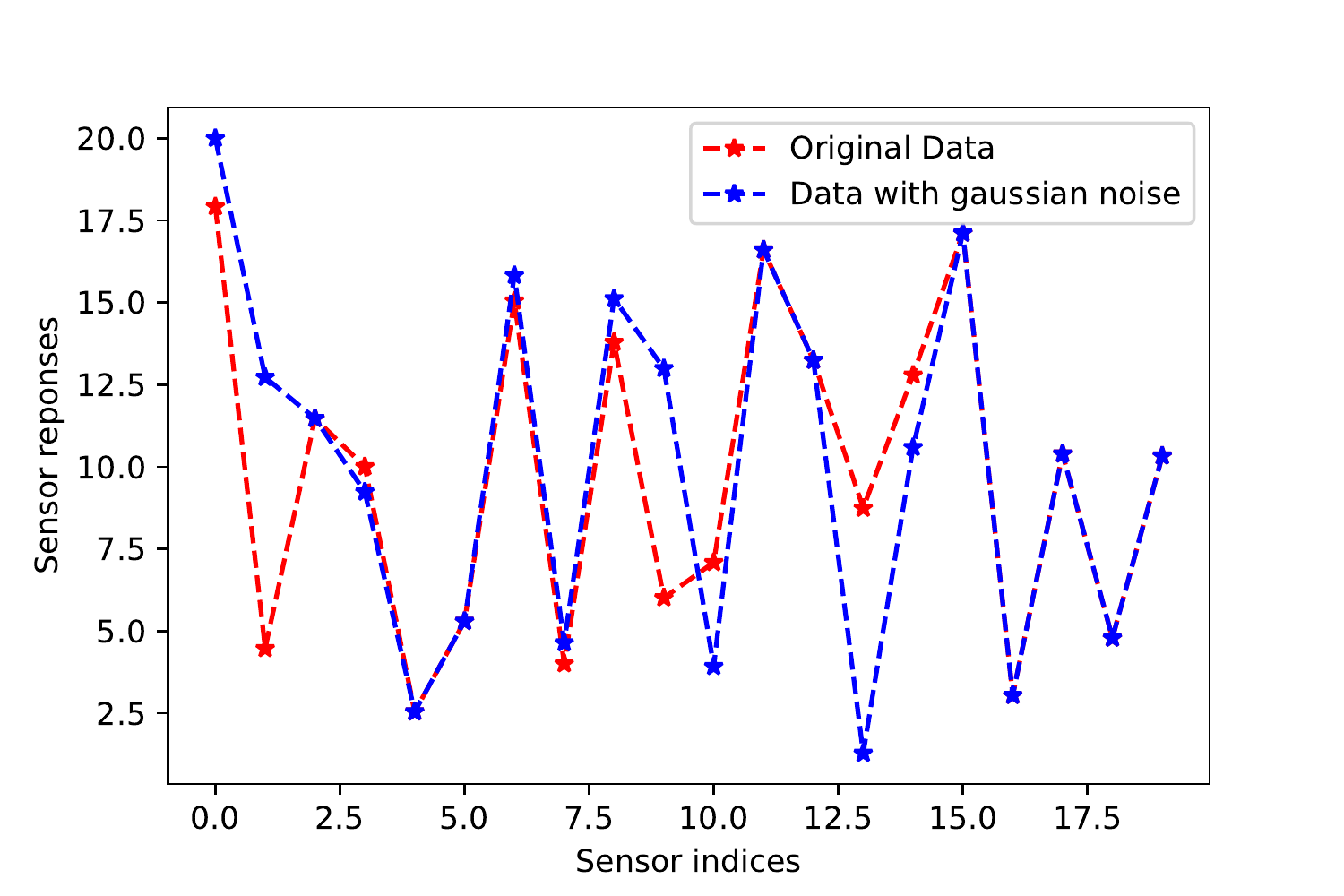}
  \caption{Sample synthetic odor of dimension $20$ (in red) and its corresponding noisy version. $Noise \, type \, = \, gaussian; mean \, = \, 0.; standard \, deviation \, = \, 6.;occlusion \, level \, = \, 50\%;$}  
  \label{gauss_noise}
\end{figure}

\begin{figure}
  \centering
  \subfloat[$1^{st}$ $\gamma$ cycle]{\includegraphics[width=0.48\linewidth]{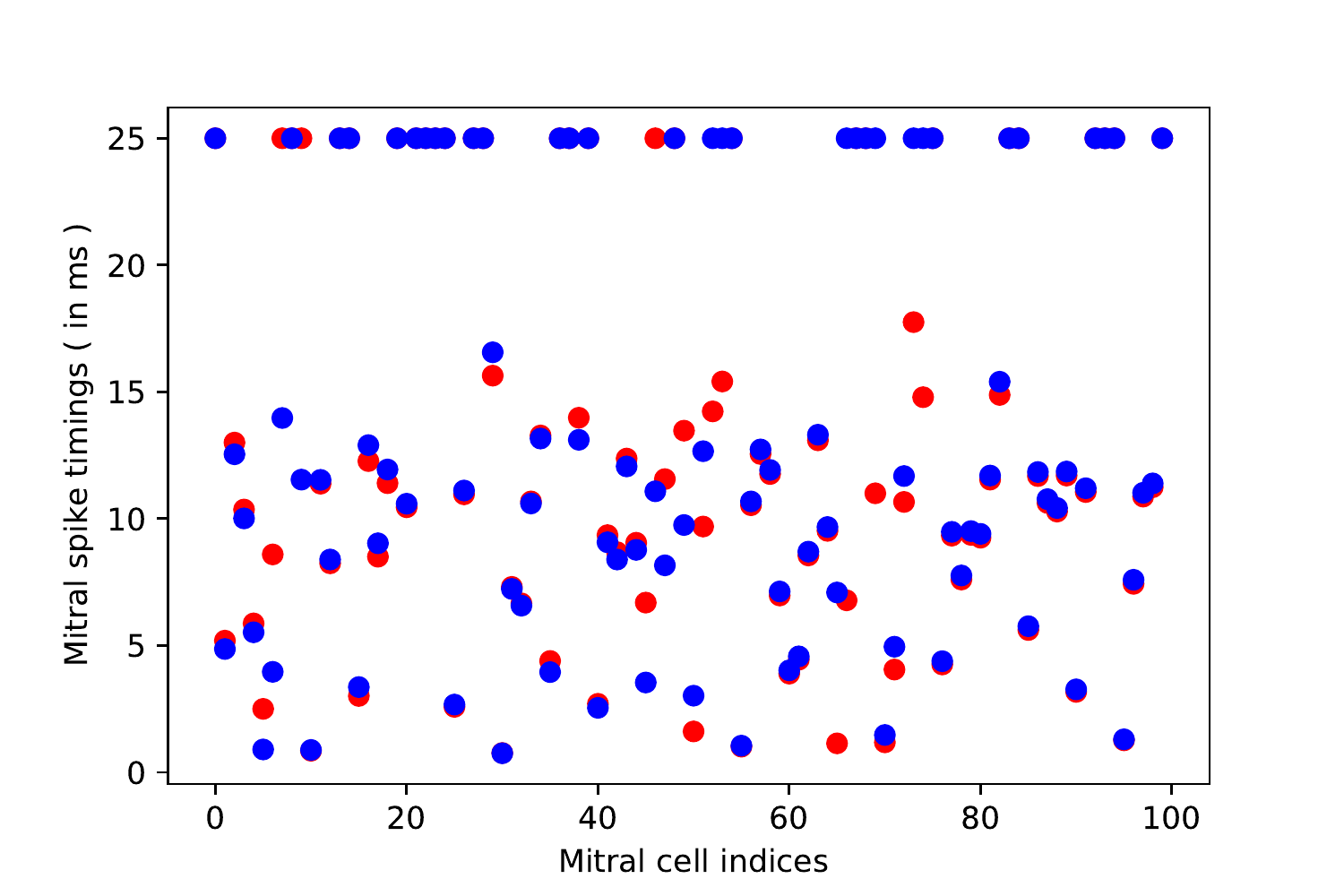}}
    \hspace{0.001in}
\subfloat[$2^{nd}$ $\gamma$ cycle]{\includegraphics[width=0.48\linewidth]{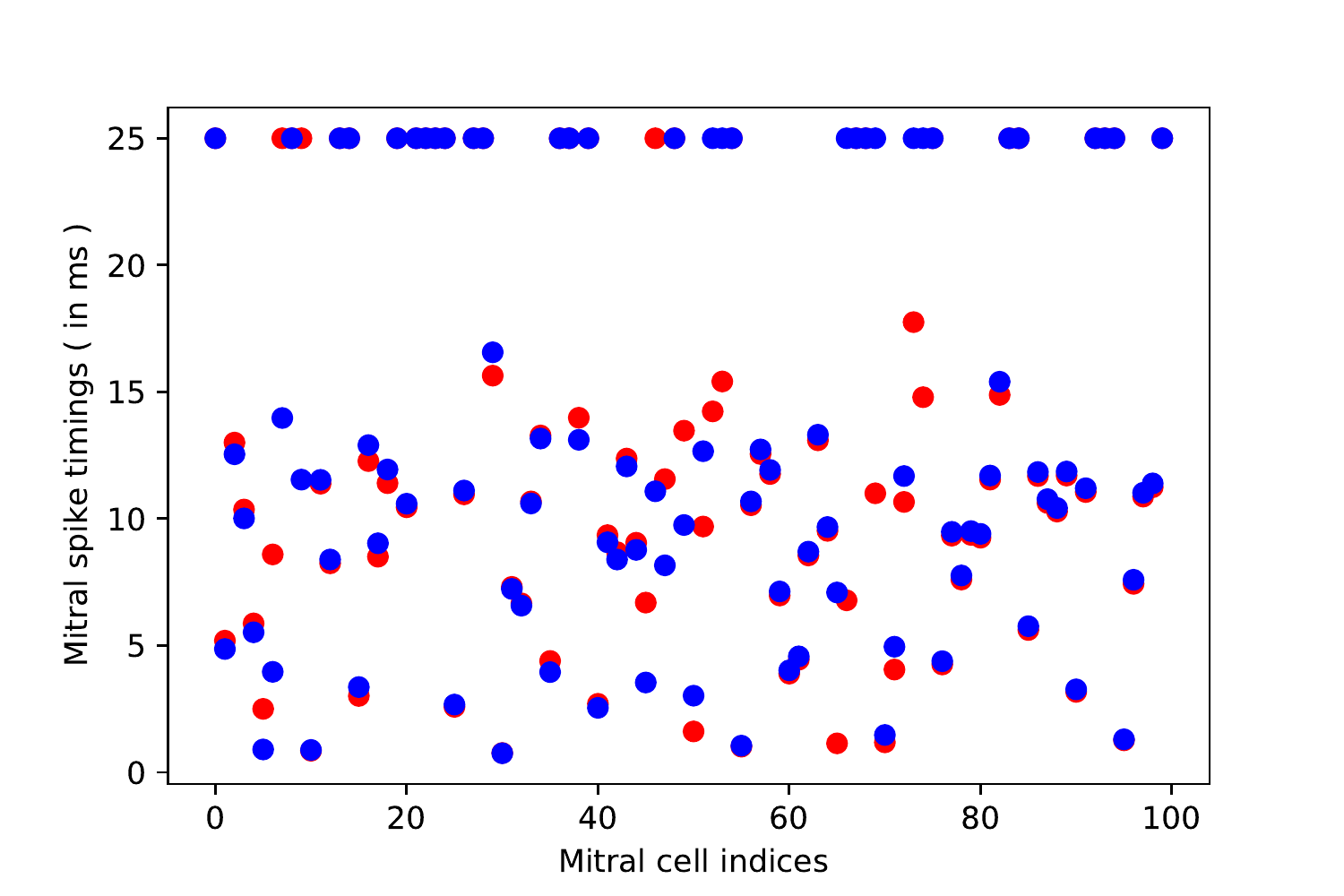}}
\hspace{0.001in}
\subfloat[$3^{rd}$ $\gamma$ cycle]{\includegraphics[width=0.48\linewidth]{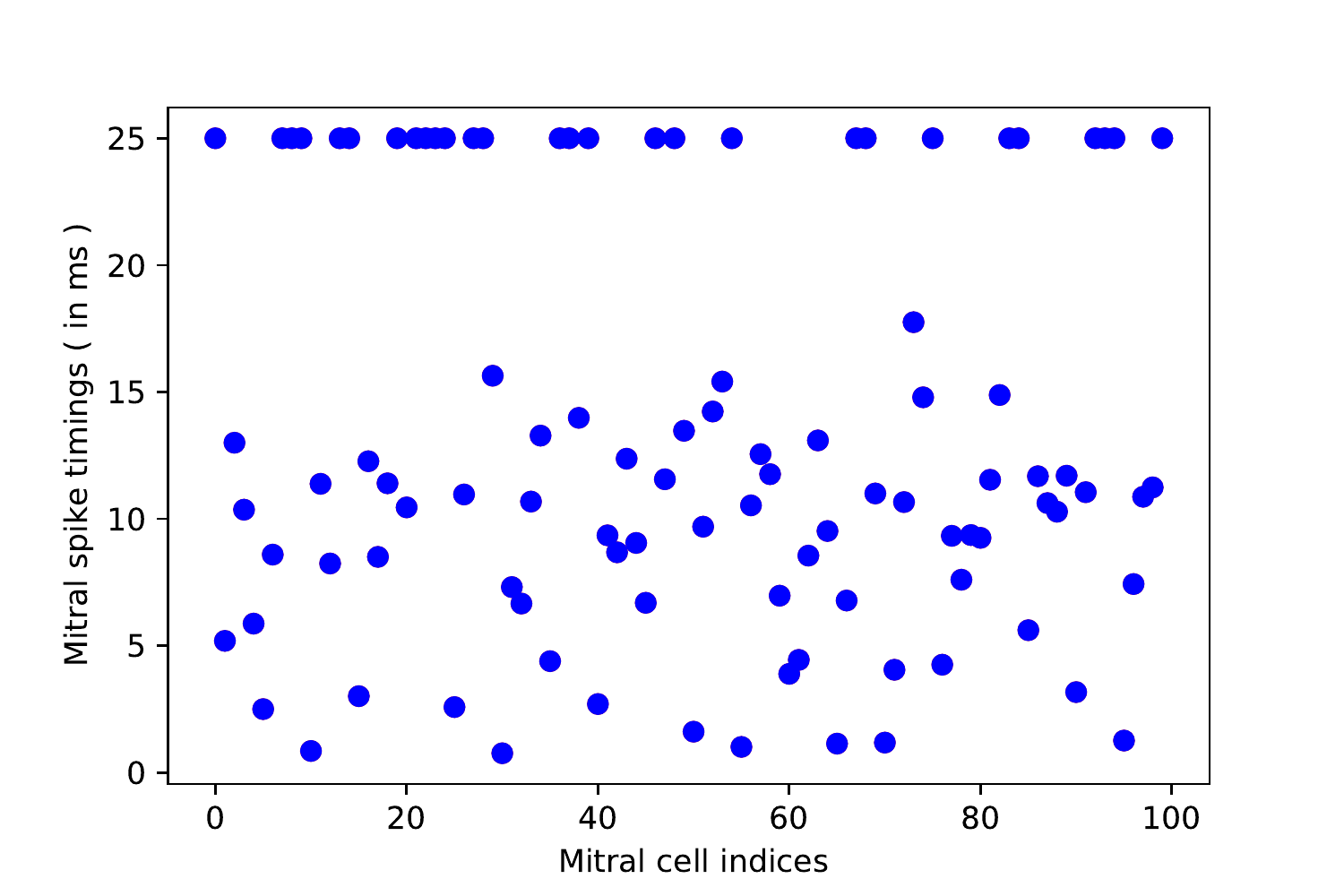}}
\hspace{0.001in}
\subfloat[$4^{th}$ $\gamma$ cycle]{\includegraphics[width=0.48\linewidth]{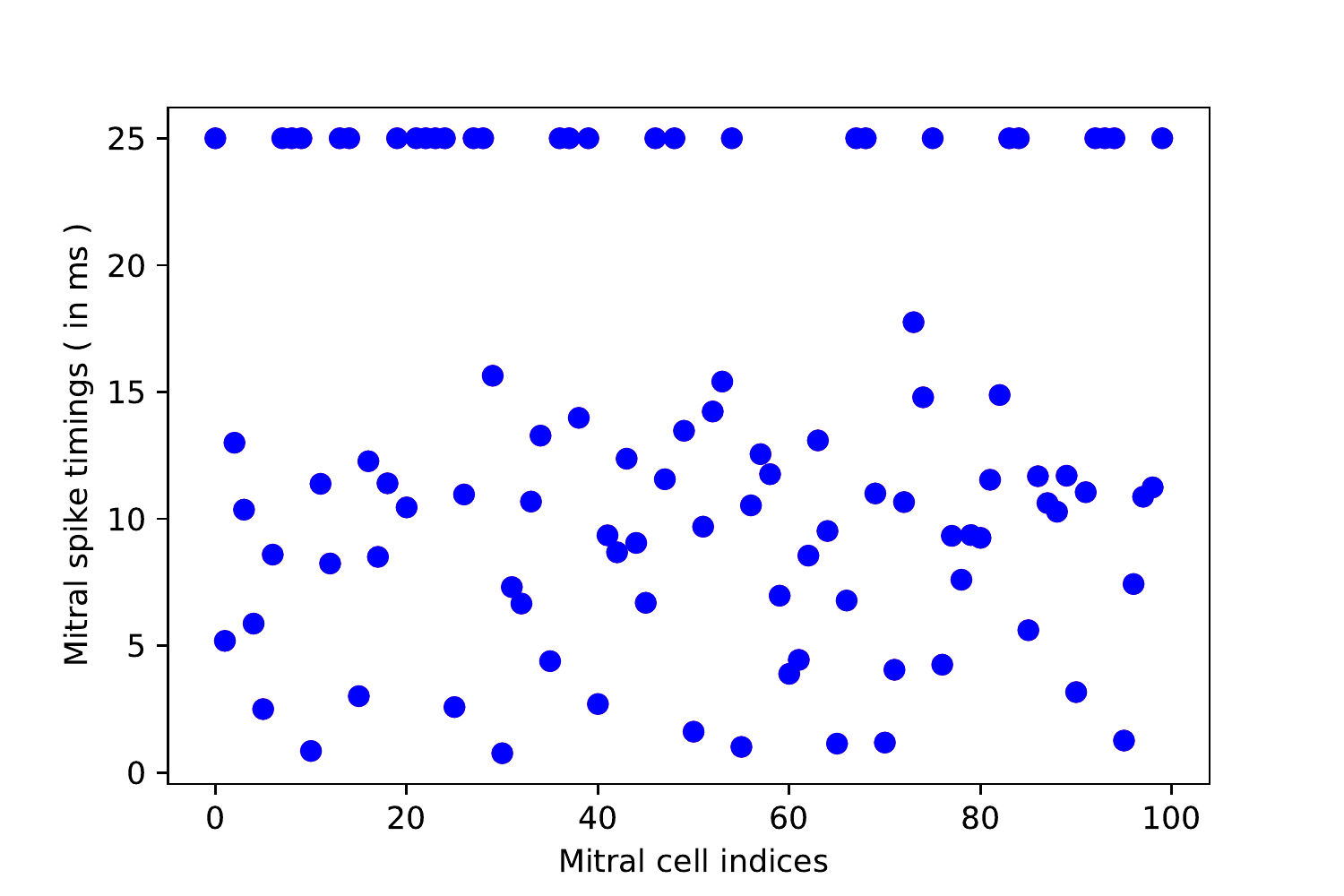}}

\caption{Spike time raster of mitral cells (MCs) for clean and noisy odor sample described in Fig ~\ref{gauss_noise}. Only the first four out of eight $\gamma$ cycles are shown here. }  
  \label{raster}

\end{figure}

Fig ~\ref{gauss_noise} depicts a sample train odor and its noisy version ($Noise \, type \, = \, gaussian; mean \, = \, 0.; standard \, deviation \, = \, 6.;occlusion \, level \, = \, 50\%;$). Fig ~\ref{raster}shows the MCsoma spike timings in the $1_{st}$ $\gamma$ cycle. Due to noise, the spike timings for train and test ( noisy version of train sample ) are different - red and blue dots don't overlap. inhibitory drive is applied from the $3^{rd} cycle$ and as a result the MCsoma spike timings for the noisy test sample is shaped by the differentiated GCs as per the train pattern. Hence, the train and test spike timings overlap perfectly - the attractor / oscillator converges to a solution. 

\subsubsection{Performance w.r.t connection and threshold heterogeneity, classifier confidence} 

\begin{figure}
  \centering
  \subfloat[MC - GC connection heterogeneity]{\includegraphics[width=0.49\linewidth]{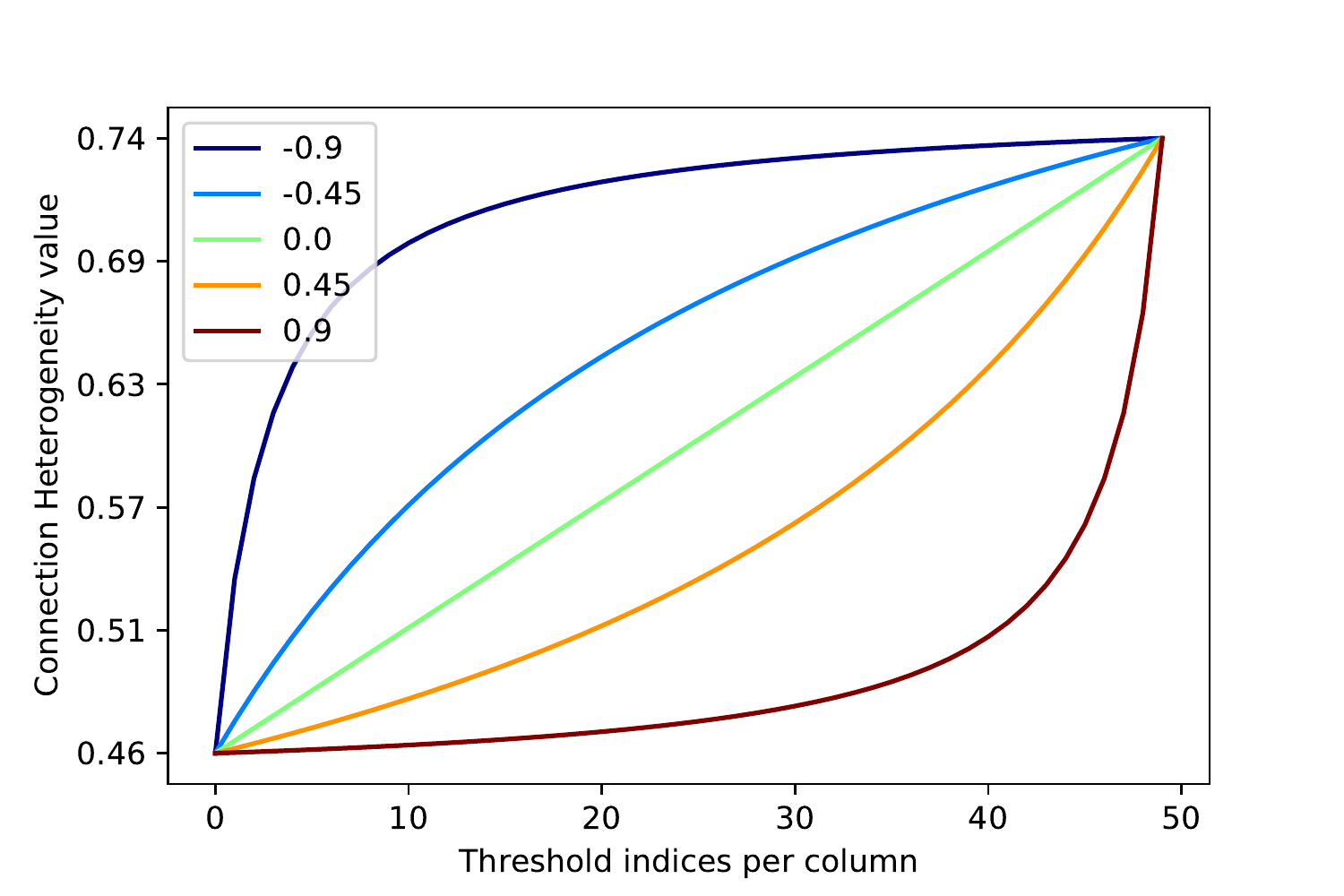}}
    \hspace{0.001in}
\subfloat[GC threshold heterogeneity]{\includegraphics[width=0.49\linewidth]{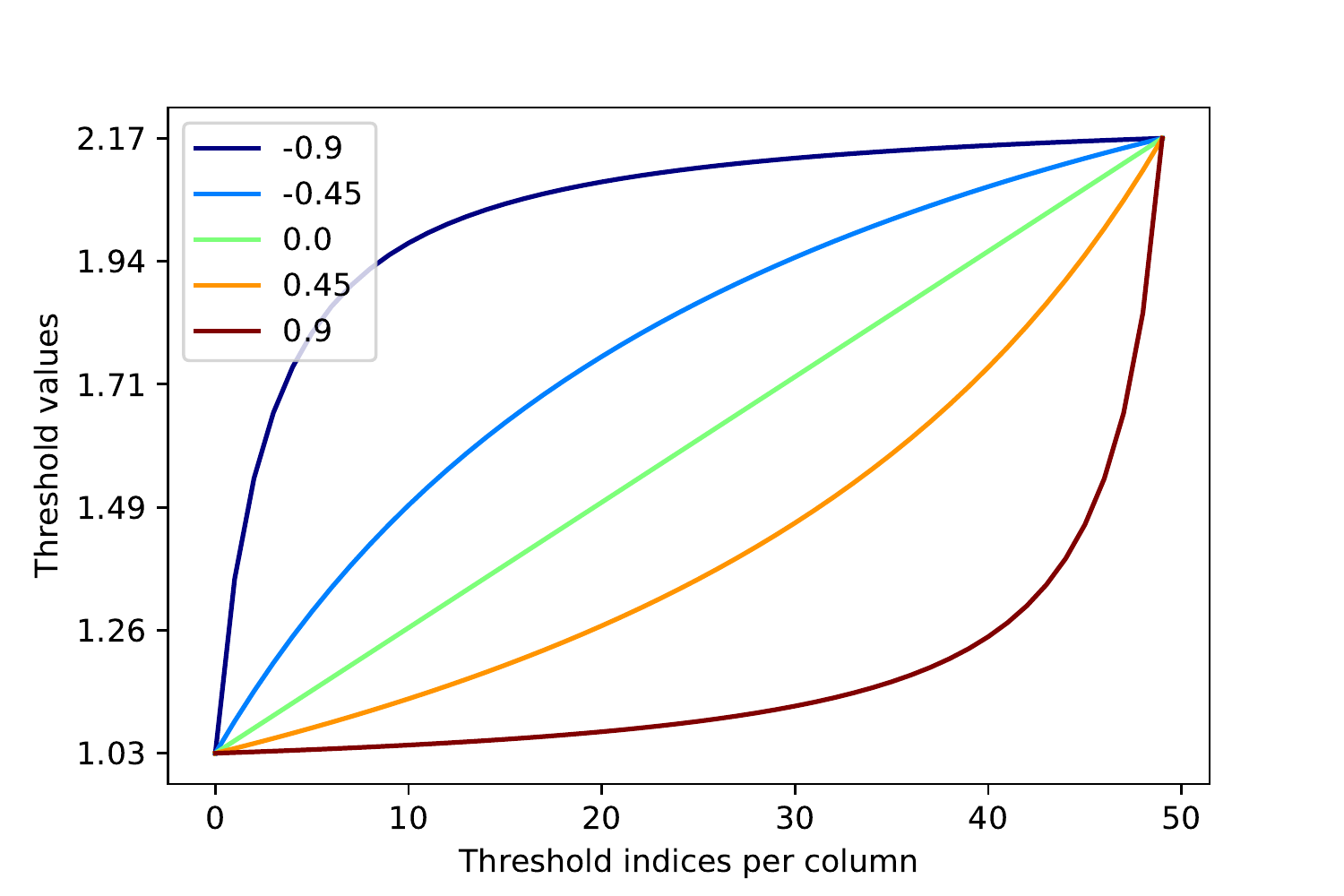}}

\caption{MC - GC connection heterogeneity and GC threshold heterogeneity profiles for ($k_{cp}, k_{vth}$ values of $-0.9, \, -0.45, \, 0., \, 0.45, \, 0.9$. Labels in figures indicate corresponding $k_{cp}, k_{vth}$ values. }  
  \label{k_description}

\end{figure}

\begin{figure}
  \centering
  \subfloat[Classifier confidence $=0.25$]{\includegraphics[width=0.48\linewidth]{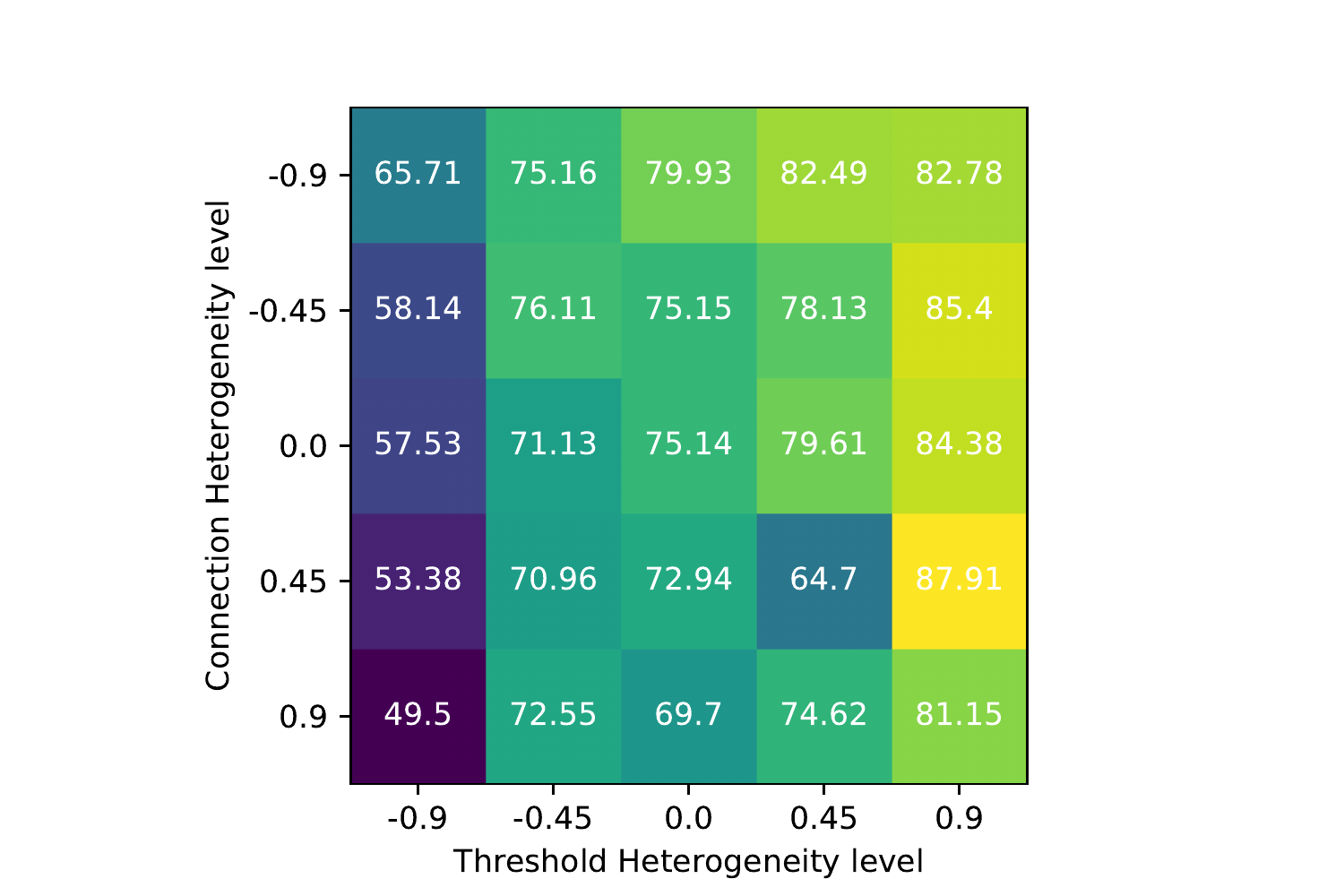}}
    \hspace{0.001in}
\subfloat[Classifier confidence $=0.5$]{\includegraphics[width=0.48\linewidth]{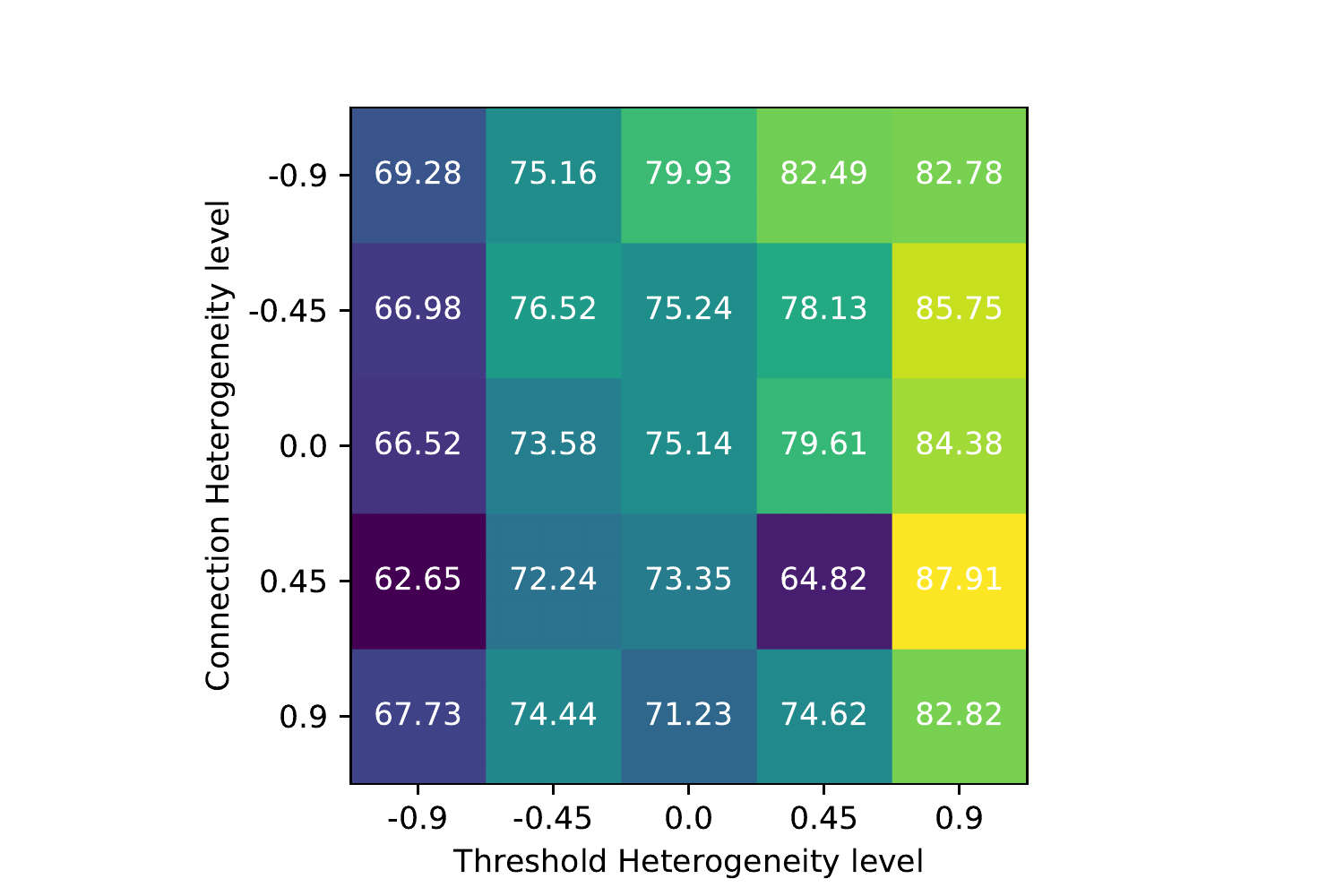}}
\hspace{0.001in}
\subfloat[Classifier confidence $=0.75$]{\includegraphics[width=0.48\linewidth]{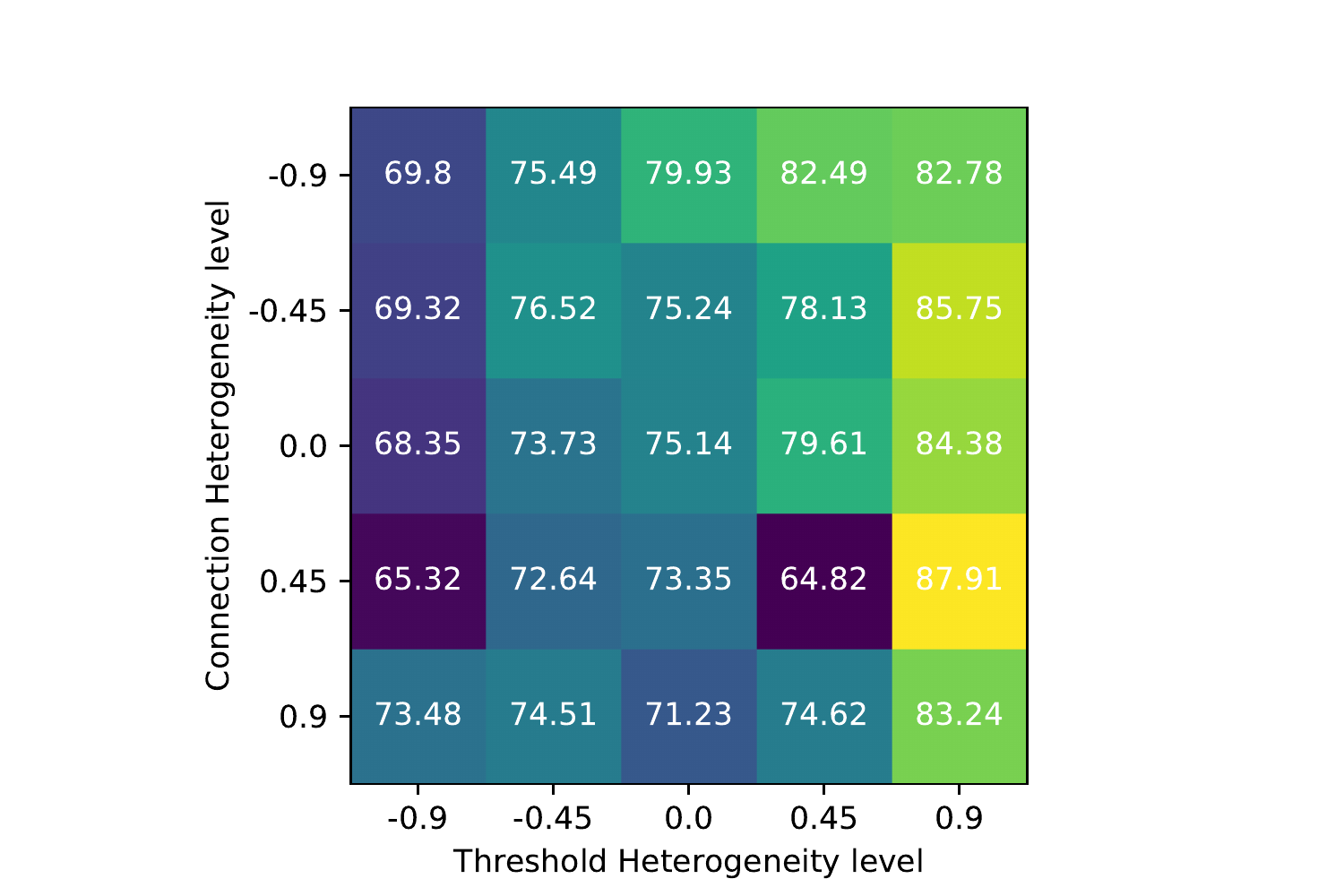}}

\caption{Classification accuracies ( average of 5 runs ) w.r.t $k_{cp}$ and $k_{vth}$ variations. The train set consisted of $4$ sequentially similar and $1$ non overlapping odors of dimension 100. The test set consists of $11 \times 5 \, = \, 55$ noisy samples. $Noise \, type \, = \, gaussian; mean \, = \, 0.; standard \, deviation \, = \, 6.;occlusion \, level \, = \, 50\%$.}
  \label{k_accuracy}

\end{figure}

\begin{figure}
  \centering
\includegraphics[width=0.8\linewidth]{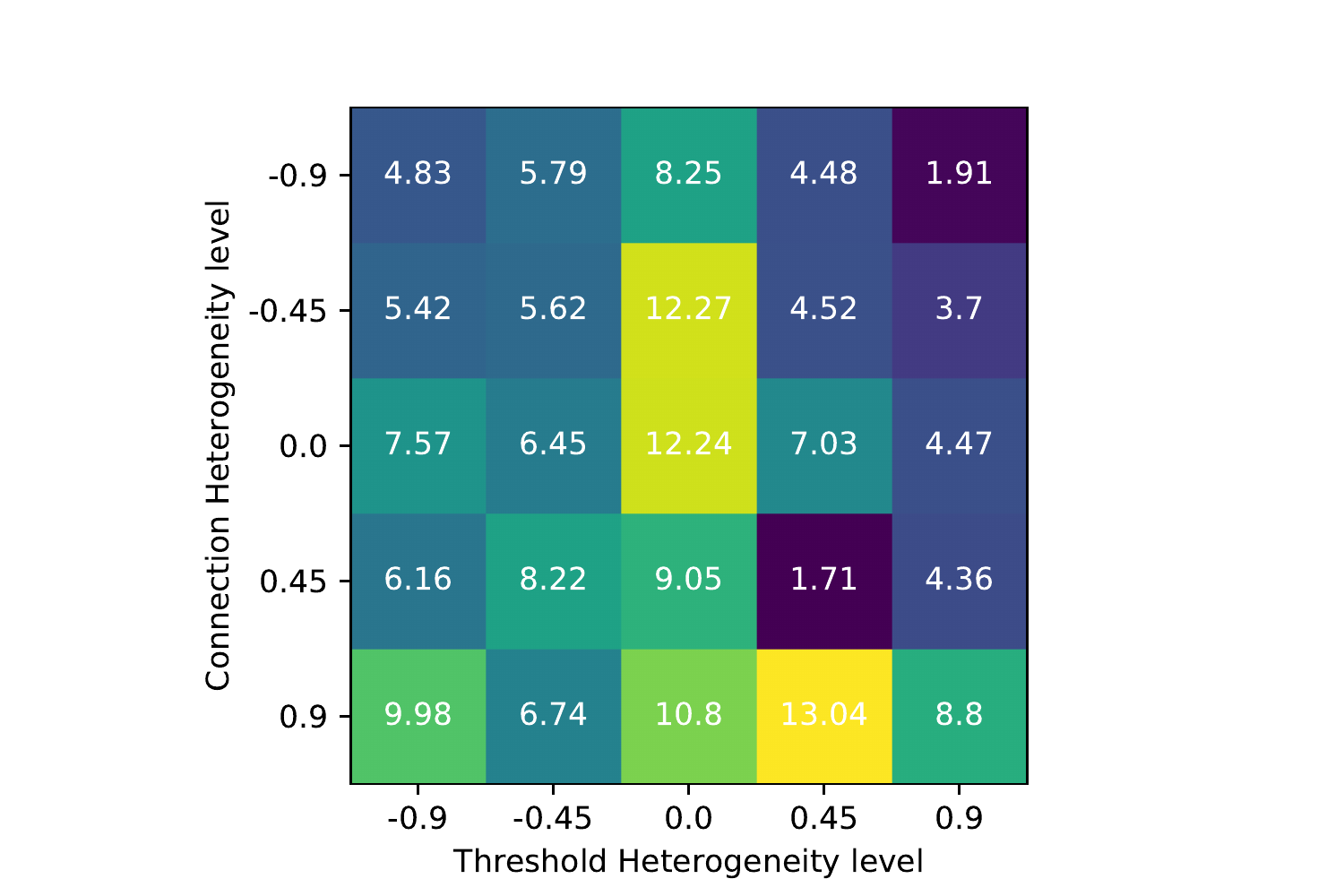}

\caption{Standard deviation of classification accuracies (for $5$ runs) described in Fig ~\ref{k_accuracy}b for classifier confidence $\,=0.5$.}  
  \label{accu_std}

\end{figure}

Above we observed that GC threshold heterogeneity and MC-GC connection probability heterogeneity are useful for network performance. We next sought a relationship between connection probability heterogeneity levels and threshold heterogeneity levels and their respective distributions. Accordingly, in order to compute the distribution of connection probabilities / thresholds,  we used the normalized tunable sigmoidal function~\cite{Normaliz95:online}:
\begin{equation}
    Connection \, probability / threshold \, values = \frac{x-kx}{k-2k\abs{x}+1} 
\end{equation}
where for connection probability, $k$ is $k_{cp}$ and for threshold $k$ is $k_{vth}$. Fig ~\ref{k_description}describes the distribution of connection probabilities / convergence ratios within $0.4-0.8$ for different values of $k_{cp}$. Lower the value of $k_{cp}$, higher is the shift of connection probabilities towards the maximum ($0.8$) - a large fraction of the GCs receive higher initial connection probabilities / convergence ratios, $k_{cp}=0$ refer to uniform connection probability distribution. Positive value of $k_{cp}$ implies shift of connection probabilities / convergence towards the minimum ($0.4$) - a large fraction of the GCs receive lower connection probabilities. Similarly, Fig ~\ref{k_description}b describe the threshold distributions within $0.8-2.4 \, mV$.

To predict the class of test sample, we compared the MCsoma spike timings of the test sample with all the train samples using Jaccard distance. The minimum distance train sample is selected as the class. In addition, the network also has a tunable \textit{classifier confidence}. If the distance is greater than \textit{classifier confidence}, the class of the test sample is set to \textit{None \, of \, the \, above}. Lower the  \textit{classifier \, confidence}, stricter is the requirement for similarity of train/test samples for classification. Fig ~\ref{k_accuracy} shows the average classification performances for $k_{cp}, k_{vth}$ values of $-0.9, -0.45, 0., 0.45, 0.9$. Comparing Fig ~\ref{k_accuracy}a, b, c we arrive at the conclusion that the attractor performance is not heavily influenced by classifier confidence. Which implies, the attractor is highly efficient in converging to one of the previously trained MCsoma spike timing patterns. Hence, for all subsequent studies, we set the \textit{classifier confidence} to $0.5$.

Fig ~\ref{k_accuracy} depicts the variation of classifier prediction performance for a test set comprised of $11 \times 5 \, = \, 55$ noisy samples with gaussian noise of mean $\, = 0$, standard deviation $\, = \, 6.$, occlusion level $\, = \, 50\%$. Overall, the performance is observed to be better in the above diagonal regions of the color map. Based upon the average highest classification performance, we set $k_{cp}=0.45$ and $k_{vth}=0.9$ for all subsequent studies.

Fig ~\ref{accu_std} depicts the standard deviation of the classifier performance across $5$ runs. For the parameter combination of our concern, $k_{cp}=0.45$ and $k_{vth}=0.9$, the standard deviation of accuracy is very low ($=4.36$). This establishes the robustness of accuracy with respect to variation in network configuration. 

\subsubsection{Classifier performance w.r.t number of learning high threshold GCs per column in a naive network}
\begin{figure}
  \centering
  \includegraphics[width=0.85\linewidth]{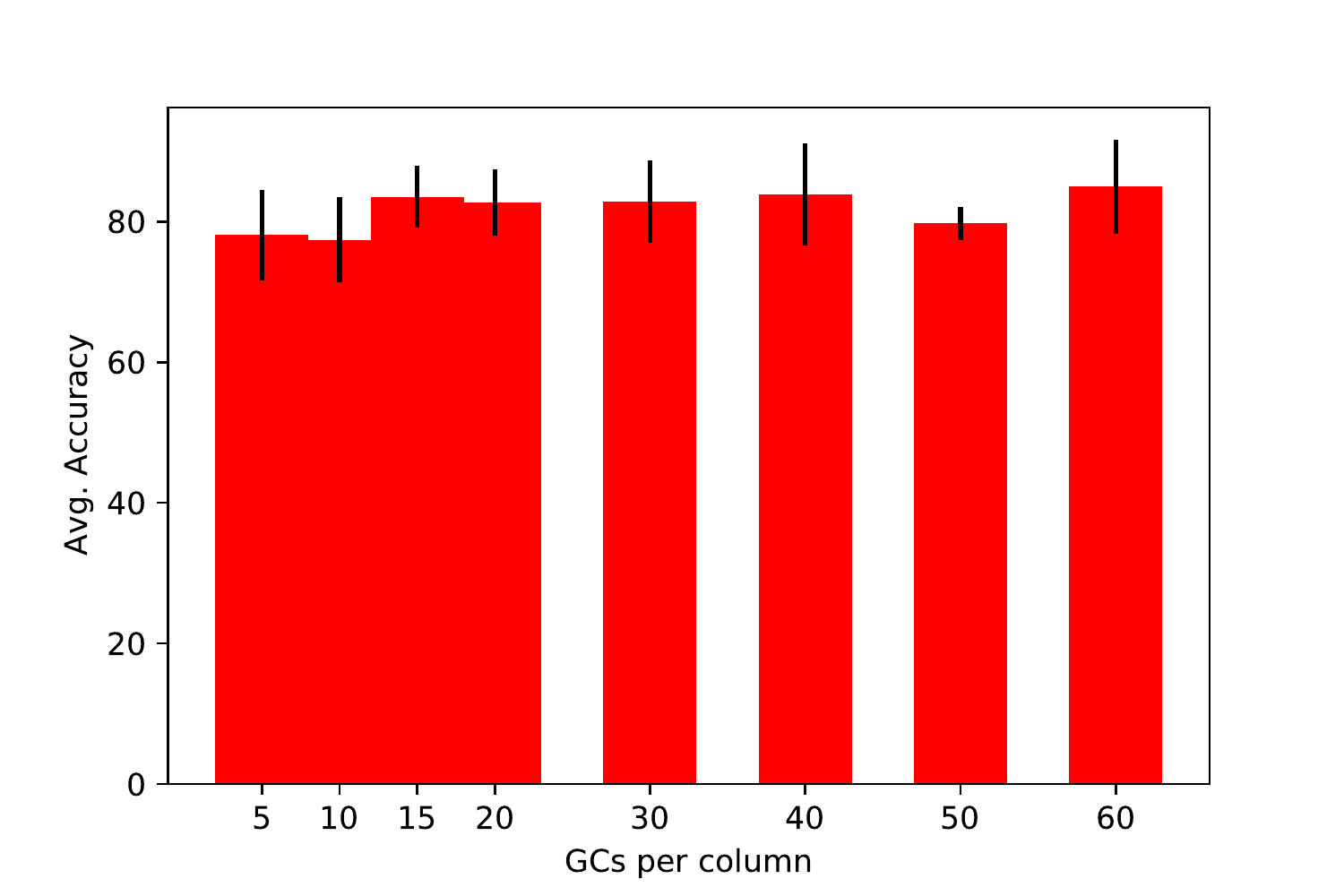}
  \caption{Classification accuracies ( average of 5 runs ) w.r.t variations in the number of high threshold learning GCs per column in a naive network for $k_{cp}=0.45$, $k_{vth}=0.9$. The train set consisted of $4$ sequentially similar and $1$ non overlapping odors of dimension 100. The test set consists of $11 \times 5 \, = \, 55$ noisy samples. $Noise \, type \, = \, gaussian; mean \, = \, 0.; standard \, deviation \, = \, 6.;occlusion \, level \, = \, 50\%$}  
  \label{gcs_col}
\end{figure}

We next studied the dependence of the network's performance on the number of learning high threshold GCs per column in a naive network using the same network parameters and synthetic data set as earlier (The test set is composed of $11 \times 5 \, = \, 55$ noisy samples. $Noise \, type \, = \, gaussian; mean \, = \, 0.; standard \, deviation \, = \, 6.;occlusion \, level \, = \, 50\%$). For this type, level of noise, the average prediction accuracy is high for larger number of GCs ( for 5 GCs, accuracy: $78.07 \pm 6.4$; 10 GCs, accuracy:$77.41 \pm 6.$; for 15 GCs, accuracy: $83.51 \pm 4.37$) but this effect plateaus after $20$ GCs per column ($82.66 \pm 4.72$). 

\subsubsection{Ablation study}
\begin{figure}
  \centering
  \includegraphics[width=0.85\linewidth]{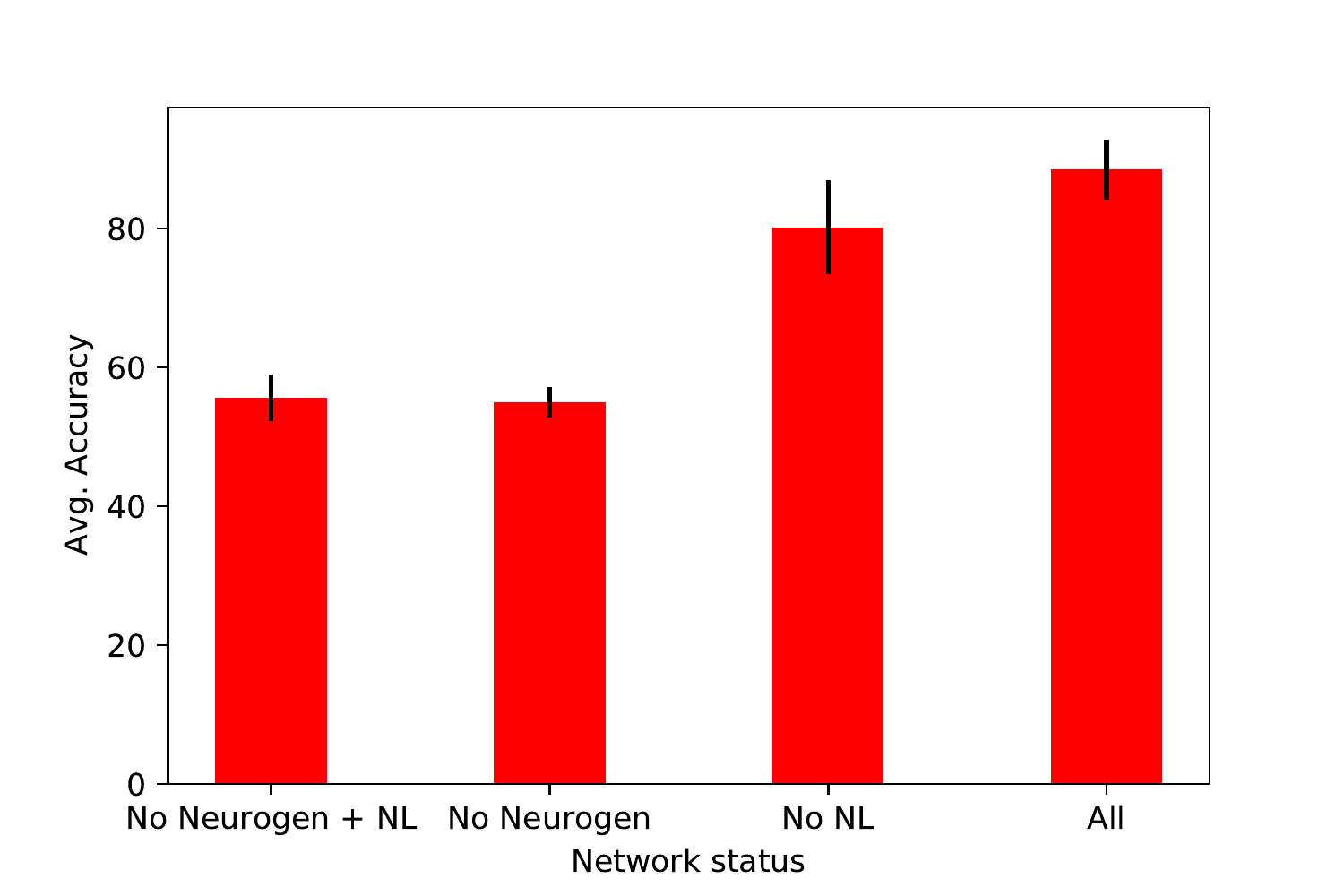}
  \caption{Classification accuracies ( average of 5 runs ) w.r.t variations in the network configuration. $No \, Neurogen \, + \, NL$ implies absence of neurogenesis and non learning low threshold GCs; $No \, Neurogen$ implies no neurogenesis but with non learning low threshold GCs; $No \, NL$ implies no learning low threshold GCs but with neurogenesis; $All$ implies complete network ( with neurogenesis and with non learning low threshold GCs). 
  The train set consisted of $4$ sequentially similar and $1$ non overlapping odors of dimension 100. The test set consists of $11 \times 5 \, = \, 55$ noisy samples. $Noise \, type \, = \, gaussian; mean \, = \, 0.; standard \, deviation \, = \, 6.;occlusion \, level \, = \, 50\%$}  
  \label{ablation}
\end{figure}

Earlier, we observed that in a feed forward network of glomerular layer and MC-GC projection, neurogenesis is needed to maintain GC spikes during sequential learning. Keeping the network parameters and the train/test data unchanged, we removed neurogenesis and non-learning low threshold GCs, the prediction accuracy was $55.63 \pm 3.37$. We then added non learning low threshold GCs, the prediction accuracy was almost the same as before $54.99 \pm 2.20$. Next, we added neurogenesis but removed non learning low threshold GCs, the prediction accuracy was significantly better $80.19 \pm 6.79$. Lastly, we added non-learning low threshold GCs, the prediction accuracy was the best of all with $88.5 \pm 4.3$, Fig ~\ref{ablation}. 

\subsubsection{Network performance on impulse noise}
\begin{figure}
  \centering
  \subfloat[Without non learning GCs]{\includegraphics[width=0.49\linewidth]{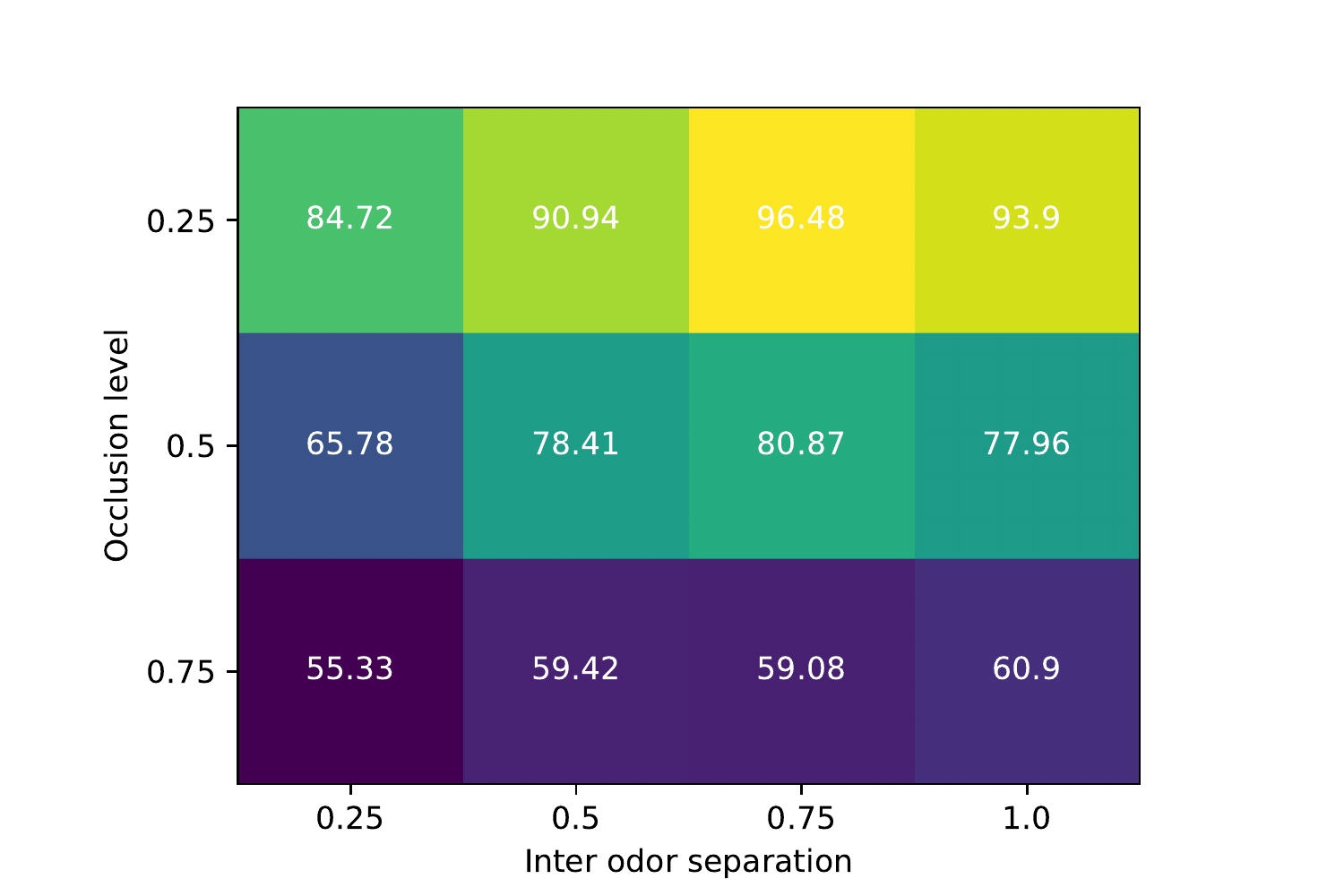}}
    \hspace{0.001in}
\subfloat[With non learning GCs]{\includegraphics[width=0.49\linewidth]{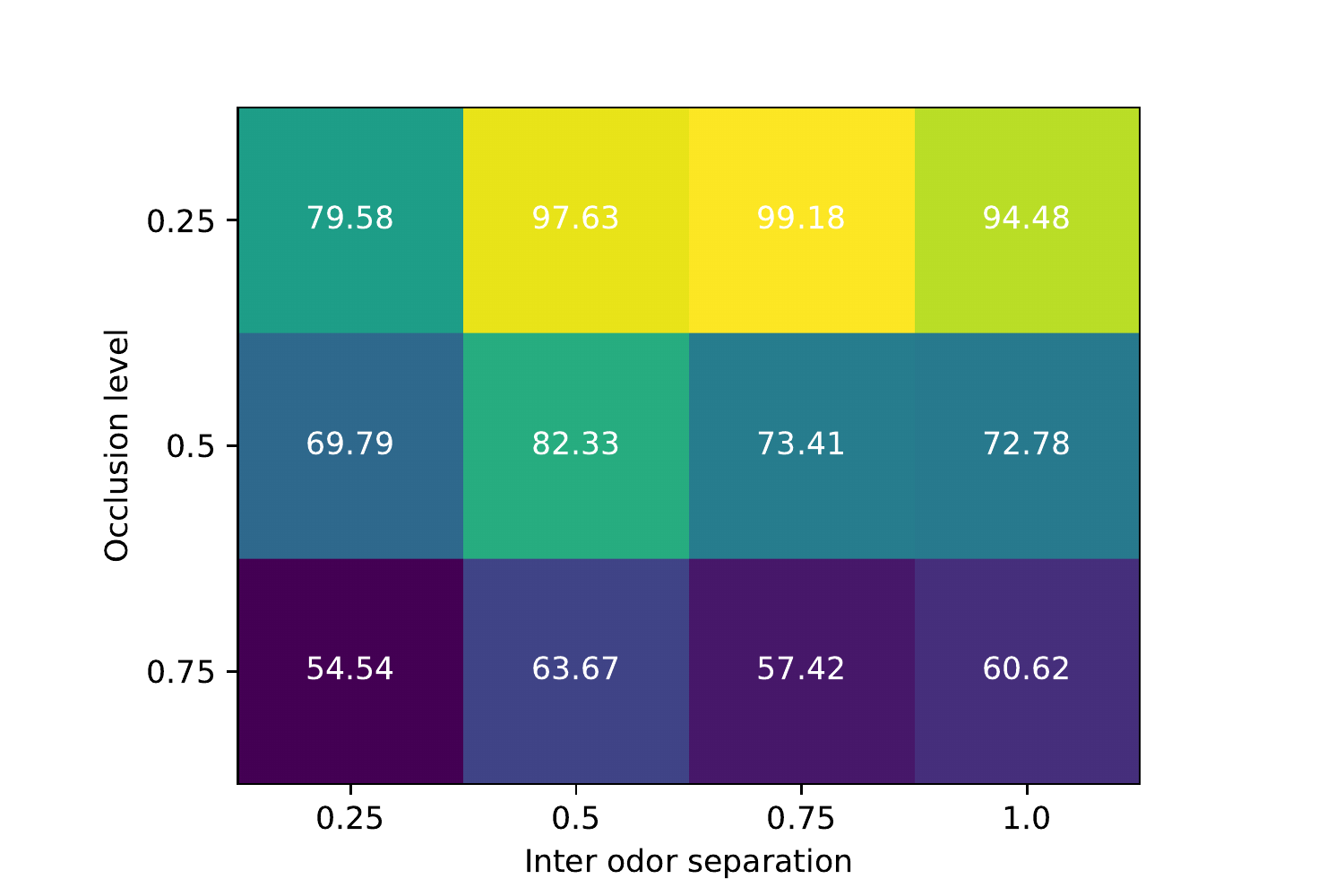}}

\caption{Classifier performance ( average of 5 runs) on impulse noise at occlusion levels of $0.25, 0.5, \& 0.75$ and inter-odor distances of $0.25, 0.5, 0.75, \& 1.$ }  
  \label{impulse_perf}
\end{figure}

We introduced impulse noise in the data by setting a percentage of the sensor responses equal to values ( between $0-20$) drawn from a uniform random distribution. These altered sensor responses are meant to have no relation with their original sensor responses. 

We generated $12$ datasets of $5$ sequentially similar odors each with $10$ noisy test samples ( total $\,=55$ samples). Each dataset had an inter-odor distance of $0.25, 0.5, 0.75$ or $1$ and an occlusion level of $25\%, 50\%$ or $75\%$ ( $ 4 \times 3 \, = 12$ samples ). 
Keeping the network parameters unaltered, we trained and tested the network performance on all $12$ datasets.  

Comparison of Fig ~\ref{impulse_perf}a and Fig ~\ref{impulse_perf}b indicates that for impulse noise, there is no clear benefit of using non-learning low threshold GCs as the prediction accuracies are not always better. For impulse noise, this is not very surprising. \\
In Fig ~\ref{impulse_perf}b, for inter odor distance of $0.25$, as the occlusion level is increased from $0.25$ to $0.75$, the prediction accuracy drops down to $54.54\%$ from $79.58\%$. This is because a higher level of sensor contamination fails to sufficiently activate GCs. Whereas, in Fig ~\ref{impulse_perf}b, for occlusion level of $0.25$, compared to inter odor distance of $0.25$, the classification performance is higher for higher inter odor distances - $97.63\%, 99.18\%, 94.48\%$ for inter odor distances of $0.5, 0.75, \, \& \, 1.$ respectively.

\subsubsection{Gaussian noise}

\begin{figure}
  \centering
\subfloat[$iod \, = \, 0.25$]{\includegraphics[width=0.48\linewidth]{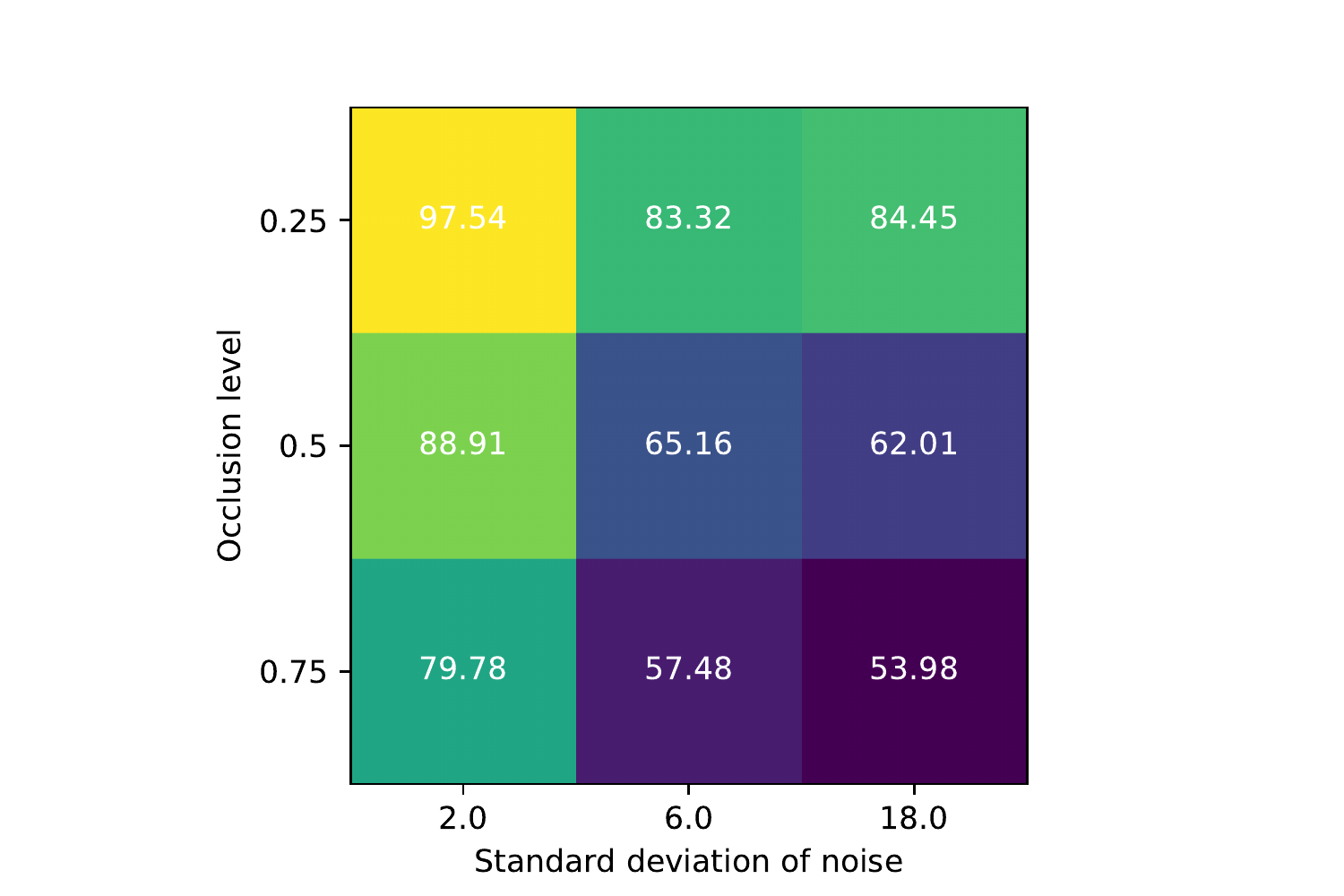}}
\hspace{0.001in}
\subfloat[$iod \, = \, 0.5$]{\includegraphics[width=0.48\linewidth]{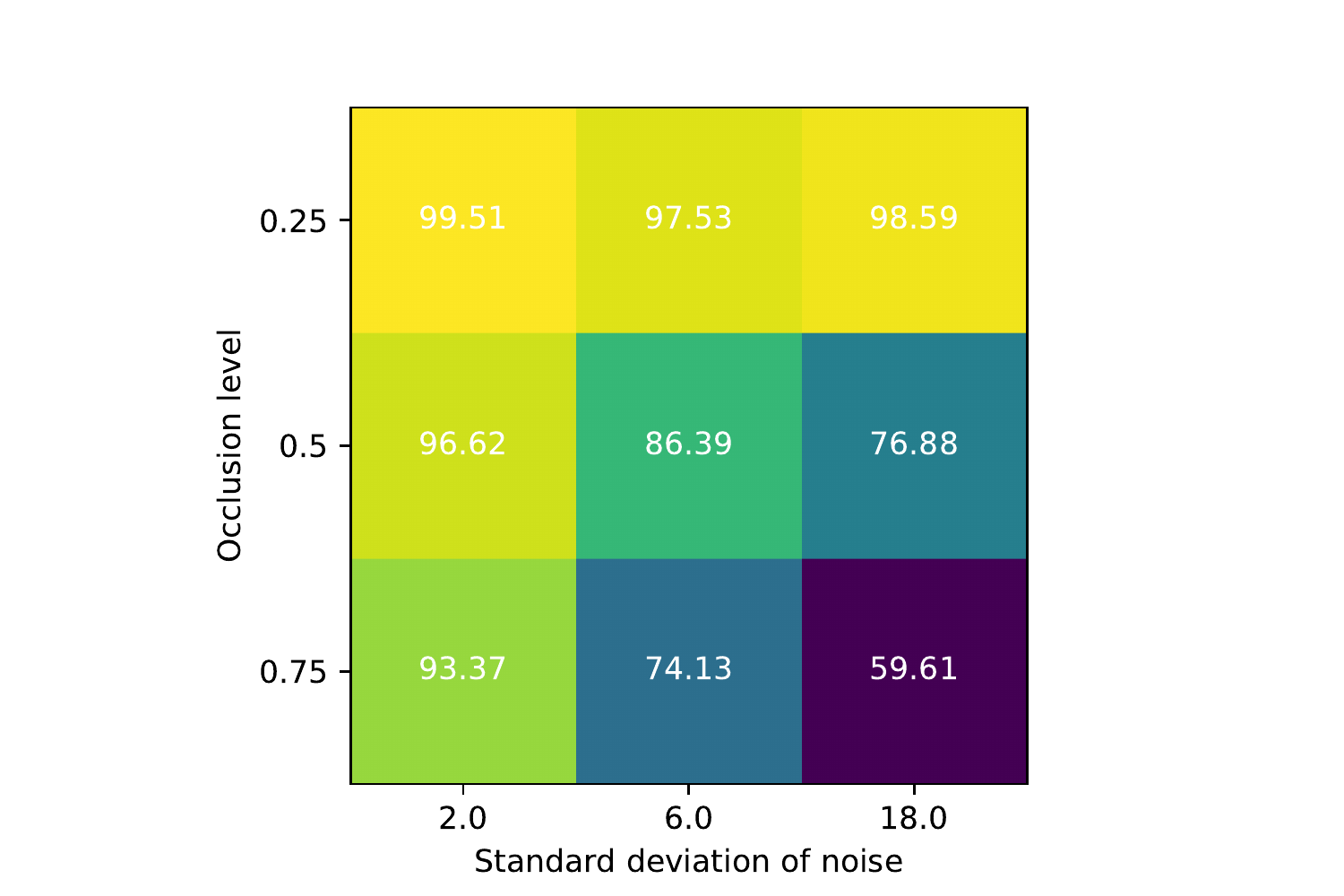}}
\hspace{0.001in}
\subfloat[$iod \, = \, 0.75$]{\includegraphics[width=0.48\linewidth]{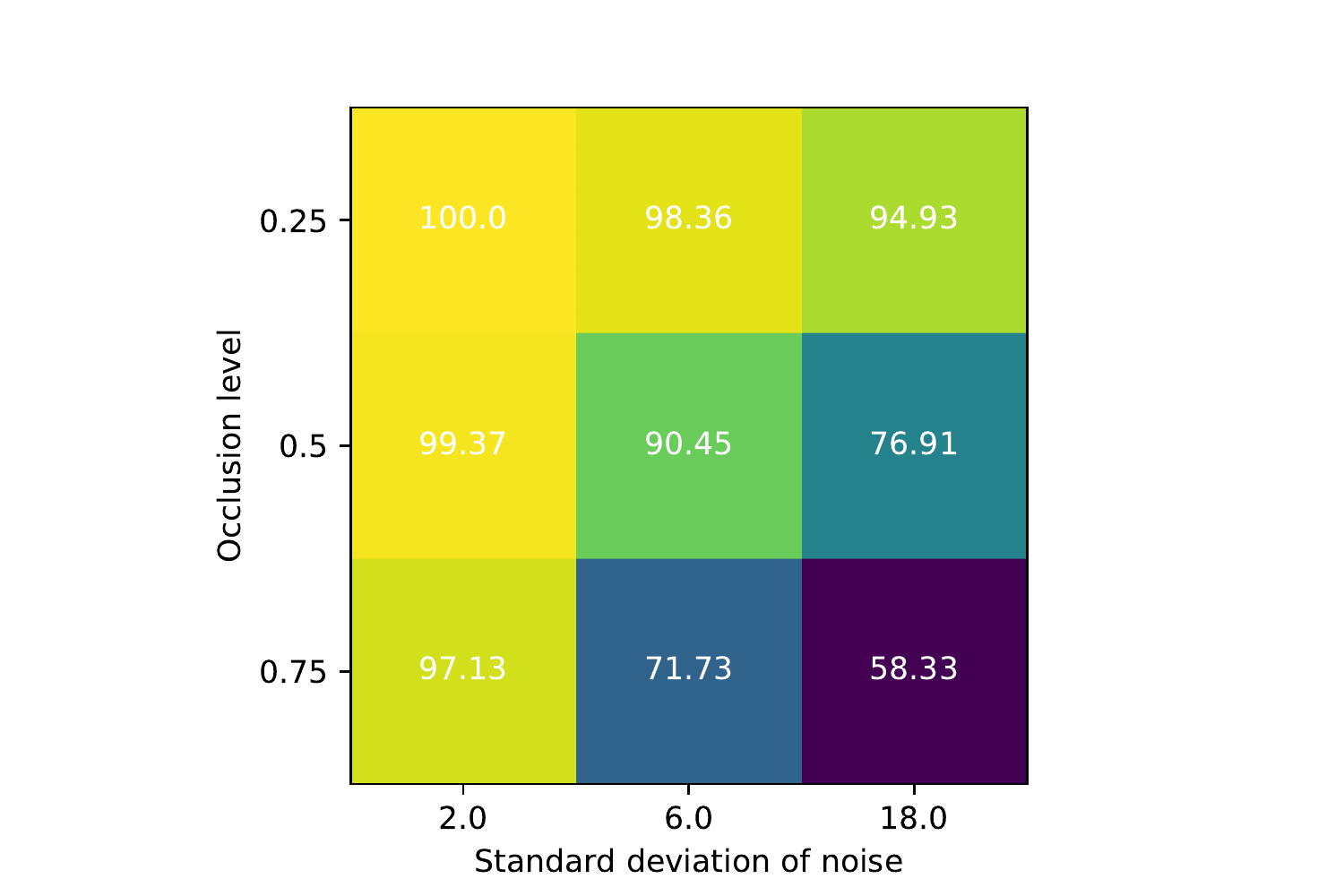}}
\hspace{0.001in}
\subfloat[$iod \, = \, 1.$]{\includegraphics[width=0.48\linewidth]{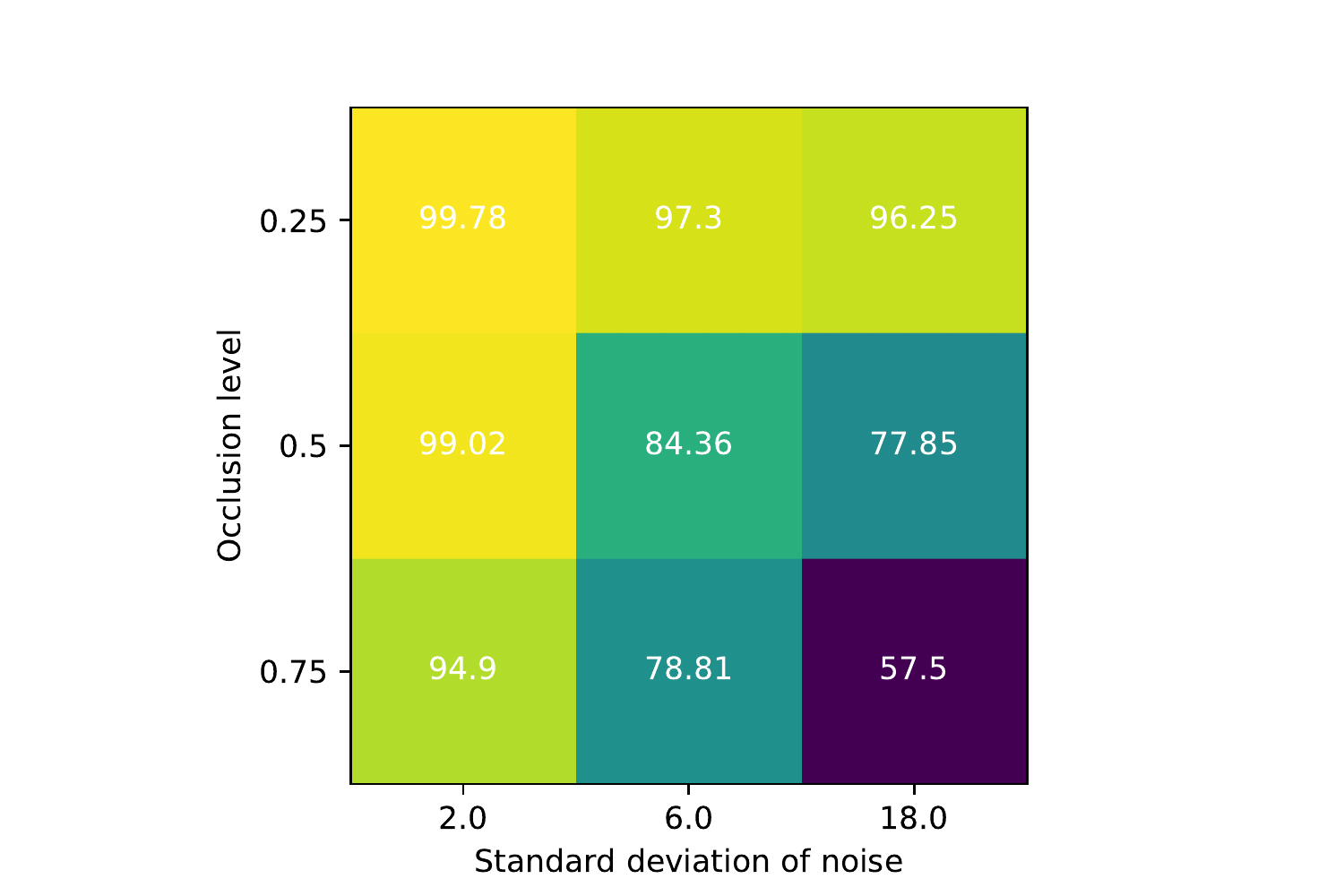}}

\caption{Average ( of five runs ) classification accuracies of the attractor when gaussian of standard deviations $2., 6., 18.$ and occlusion levels of $0.25, 0.5, 0.75$ were introduced into test data ( total number of test samples $\, = 11 \times 5 \, = \, 55$). Fig a, b, c \& d depict classification accuracies for inter odor distances of odors of $0.25, 0.5, 0.75, \& 1.$ respectively.}  
  \label{accu_gauss_noise}

\end{figure}

Using the same network as above and without any parameter change, we next trained/tested its performance with gaussian noise. Fig ~\ref{accu_gauss_noise}a shows the average classification accuracy percentage of the network when trained/tested with sequentially similar odors with inter distance of $0.25$, occlusion levels of $0.25, 0.5, 0.75$ and standard deviation of $2., 6., 18.$. The classification accuracy was highest ($=97.54$) for low standard deviation ($=0.25$), low occlusion level ($=0.25$) and was lowest ($=53.98$) for high standard deviation ($=18.$) and high occlusion levels ($=0.75$). When the inter odor difference was raised, the network became more tolerant to noise - the performance improved for high levels of occlusion and high levels of standard deviation. For example, for occlusion level of $0.25$ and standard deviation of $6.$, the classification accuracy for inter odor distance of $0.5$ was higher ($=97.53$) compared to an inter odor distance of $0.25$ ($=83.32$). 

\begin{figure}
  \centering
  \subfloat[$iod \, = \, 0.25, occlusion \, = \, 0.25, std \, = \, 2.$]{\includegraphics[width=0.49\linewidth]{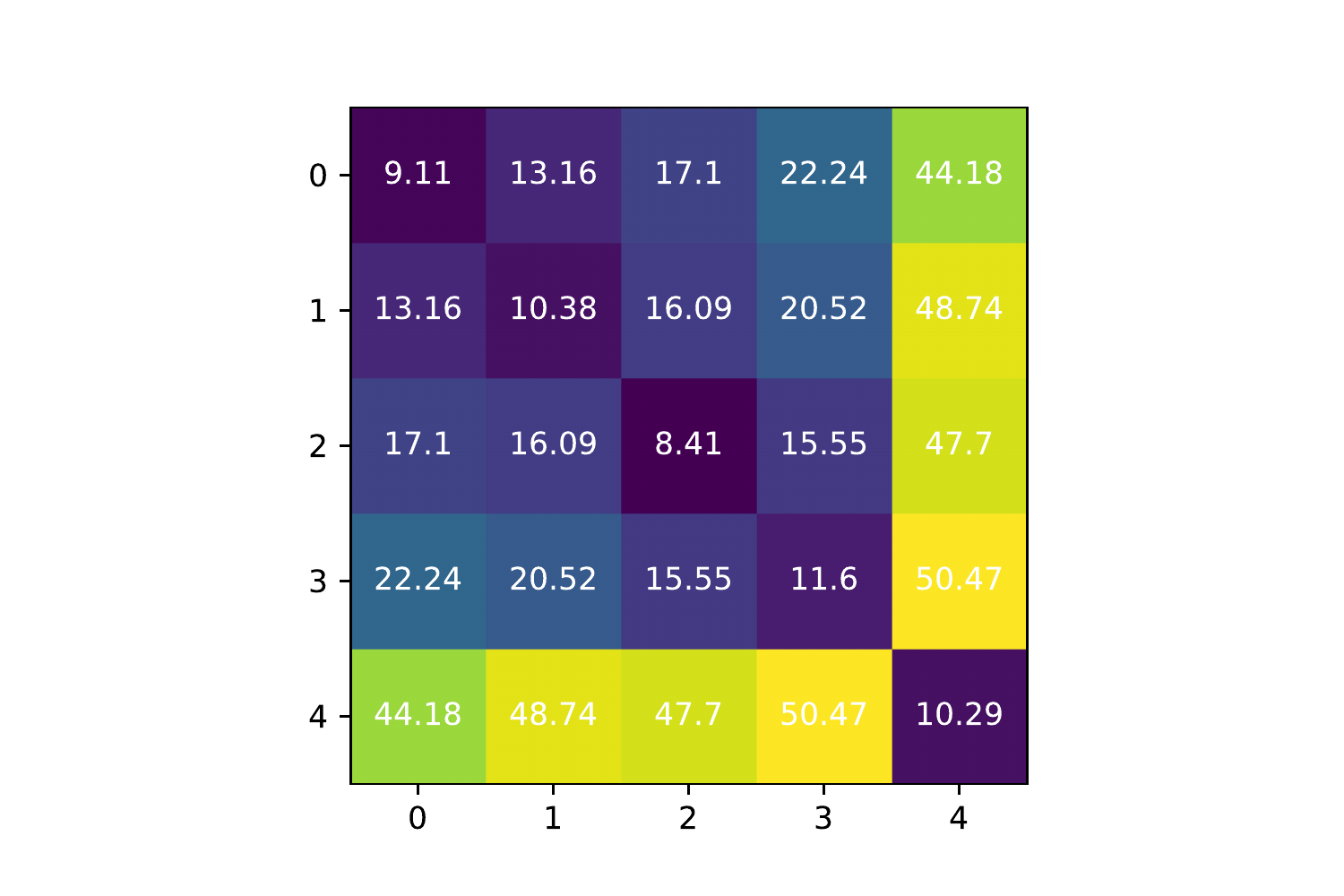}}
    \hspace{0.001in}
\subfloat[$iod \, = \, 1.0, occlusion \, = \, 0.75, std \, = \, 18.$]{\includegraphics[width=0.49\linewidth]{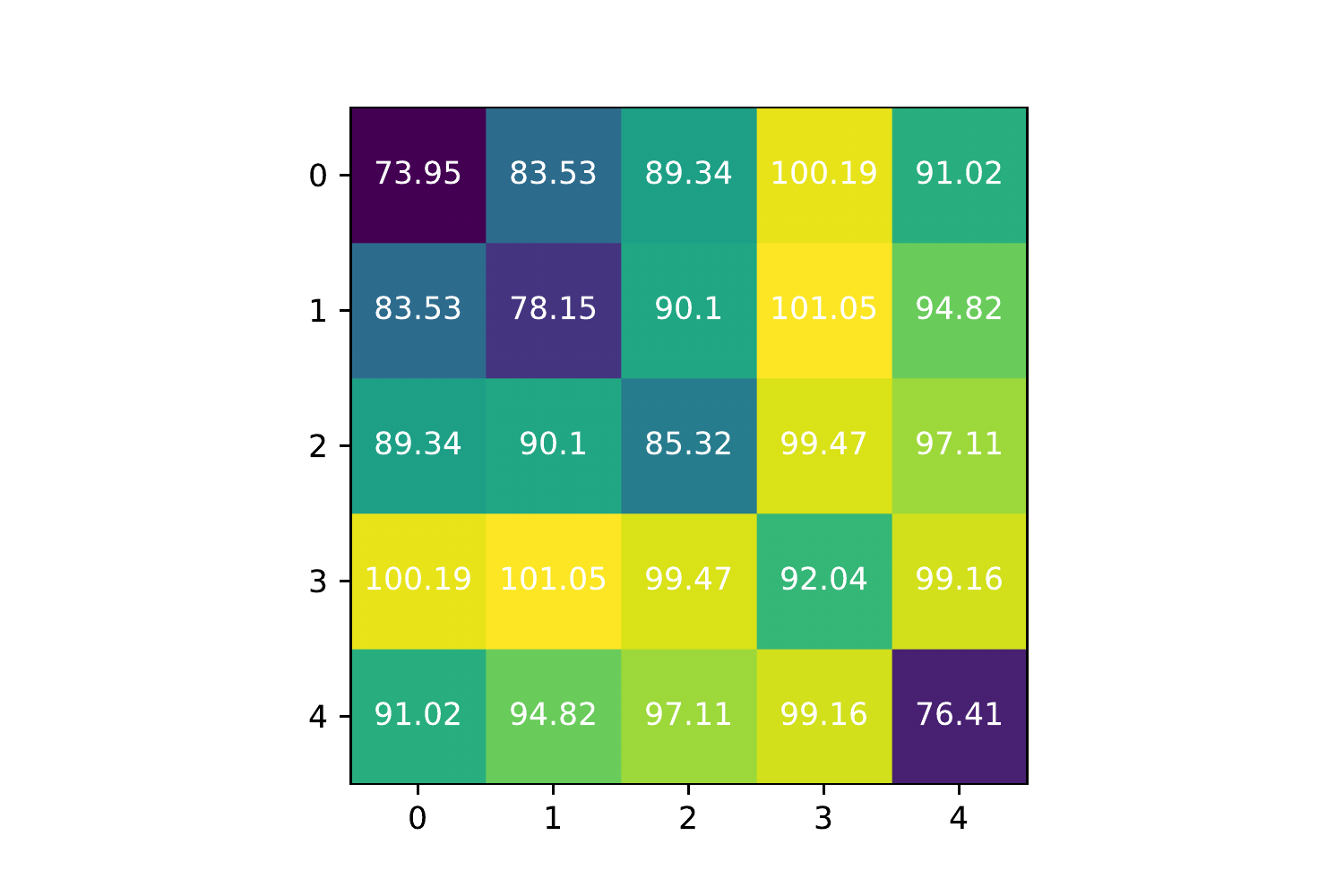}}

\caption{Inter and intra cluster euclidean distances between $5$ sequentially similar odors ( clusters ) of dimension $\, = 20$ with gaussian noise in the test samples ( total $\, = 11 \times 5 \, = \, 55$). a) Inter and Intra cluster distance for low level noise, $iod \, = \, 0.25, occlusion \, = \, 0.25, std \, = \, 2.$ b) Inter, intra cluster distance for high level of noise, $iod \, = \, 1.0, occlusion \, = \, 0.75, std \, = \, 18.$}  
  \label{cluster_dist_gauss}

\end{figure}

We next sought to understand and predict from data the drop in Sapinet's performance. Fig ~\ref{cluster_dist_gauss} a,b plots the inter and intra cluster euclidean distances between the $5$ sequentially similar odor clusters with gaussian noise. Fig ~\ref{cluster_dist_gauss} a as an input to the network predicted with an average accuracy of $97.54 \%$, whereas Fig ~\ref{cluster_dist_gauss} b as an input predicted with an accuracy of only $57.5 \%$. It is observed that, when the inter /intra distances are low and intra is less than inter, the network's performance is high, Fig ~\ref{cluster_dist_gauss} a. 

\subsection{Illustration of learning in the wild -- based on one-shot learning and real world datasets}
\subsubsection{Data regularization on real datasets}

\begin{figure}
  \centering
\subfloat[Raw data ( all sorted)]{\includegraphics[width=0.32\linewidth]{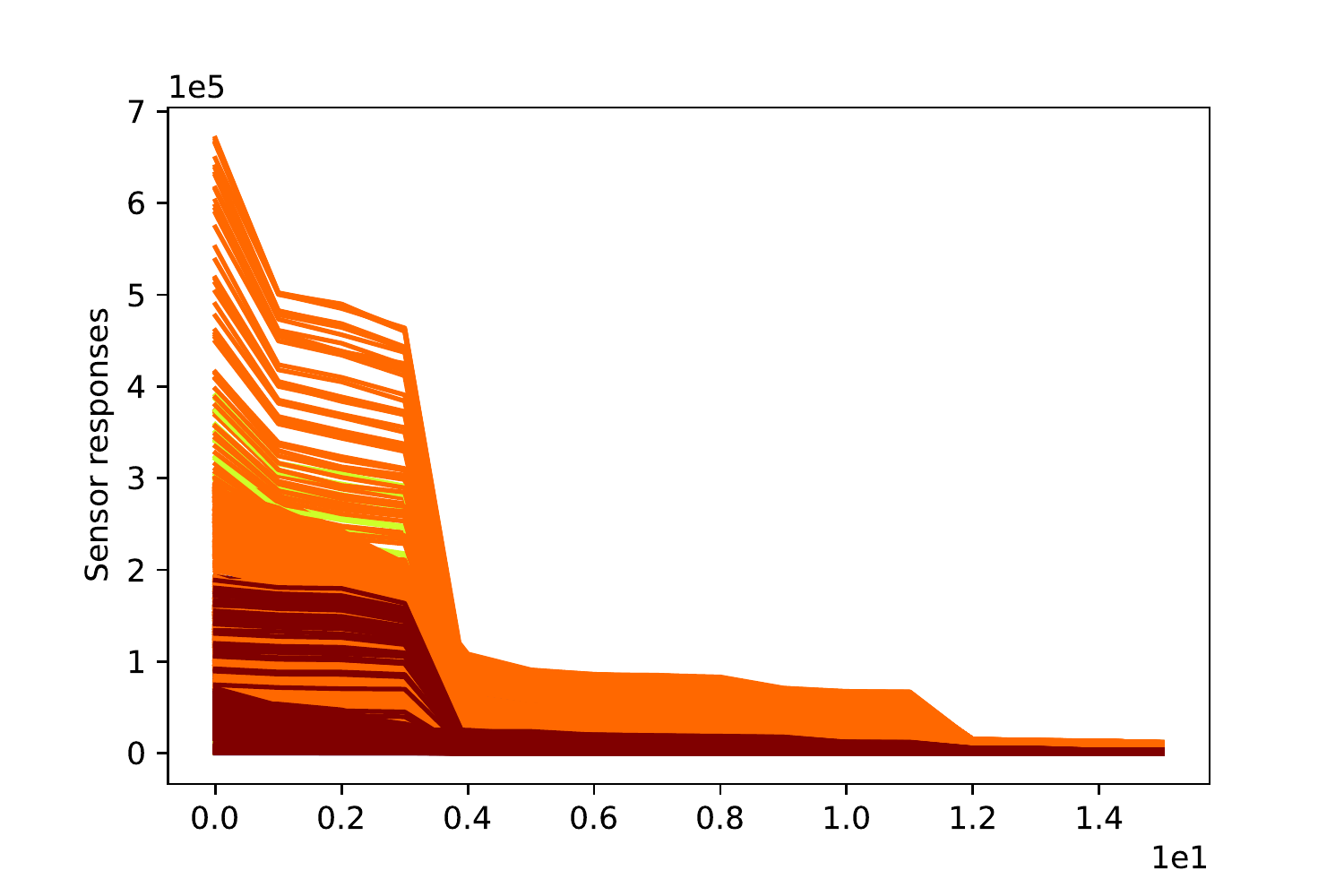}}
\hspace{0.001in}
\subfloat[MC spike count distribution]{\includegraphics[width=0.32\linewidth]{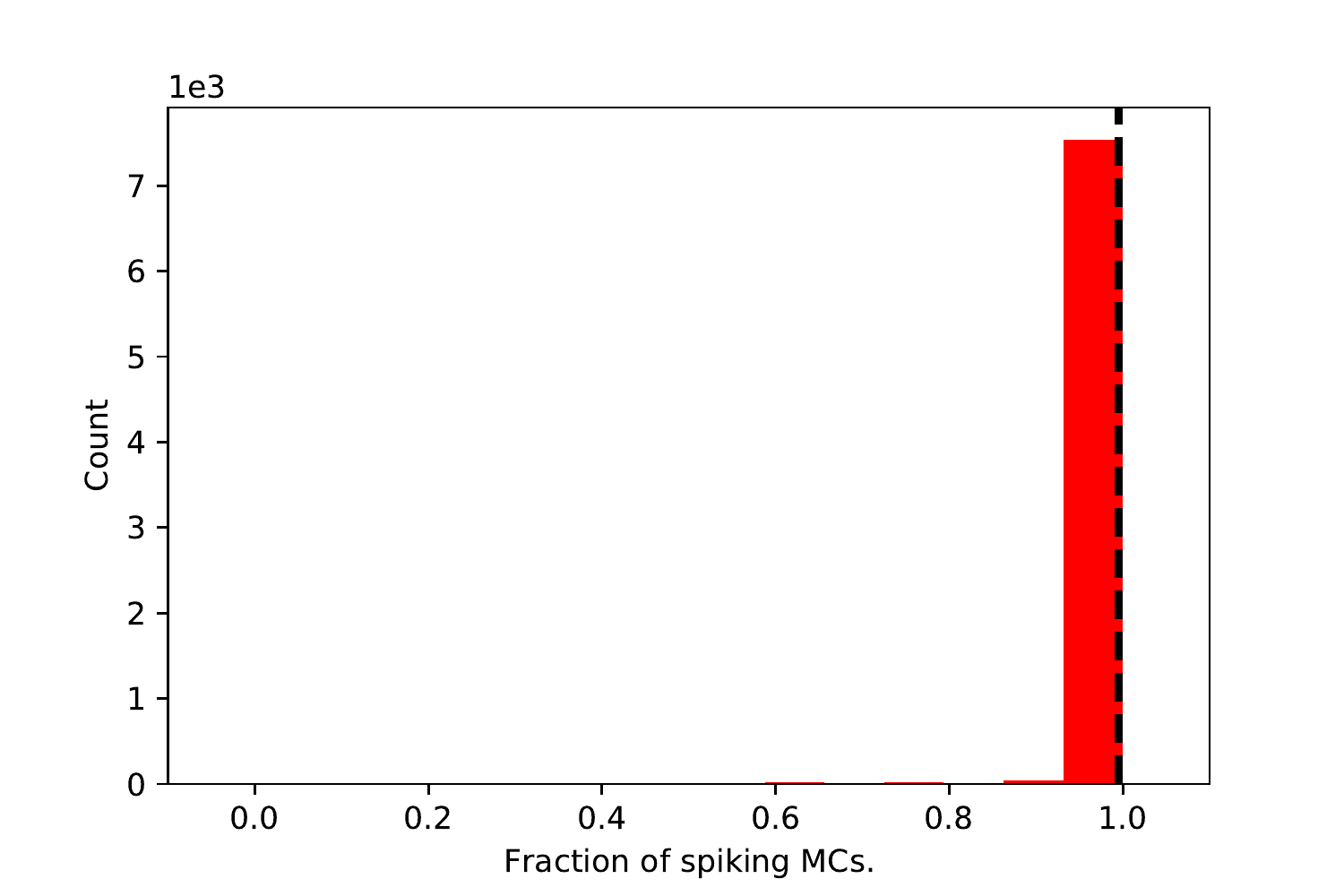}}
\hspace{0.001in}
\subfloat[GC spike count distribution]{\includegraphics[width=0.32\linewidth]{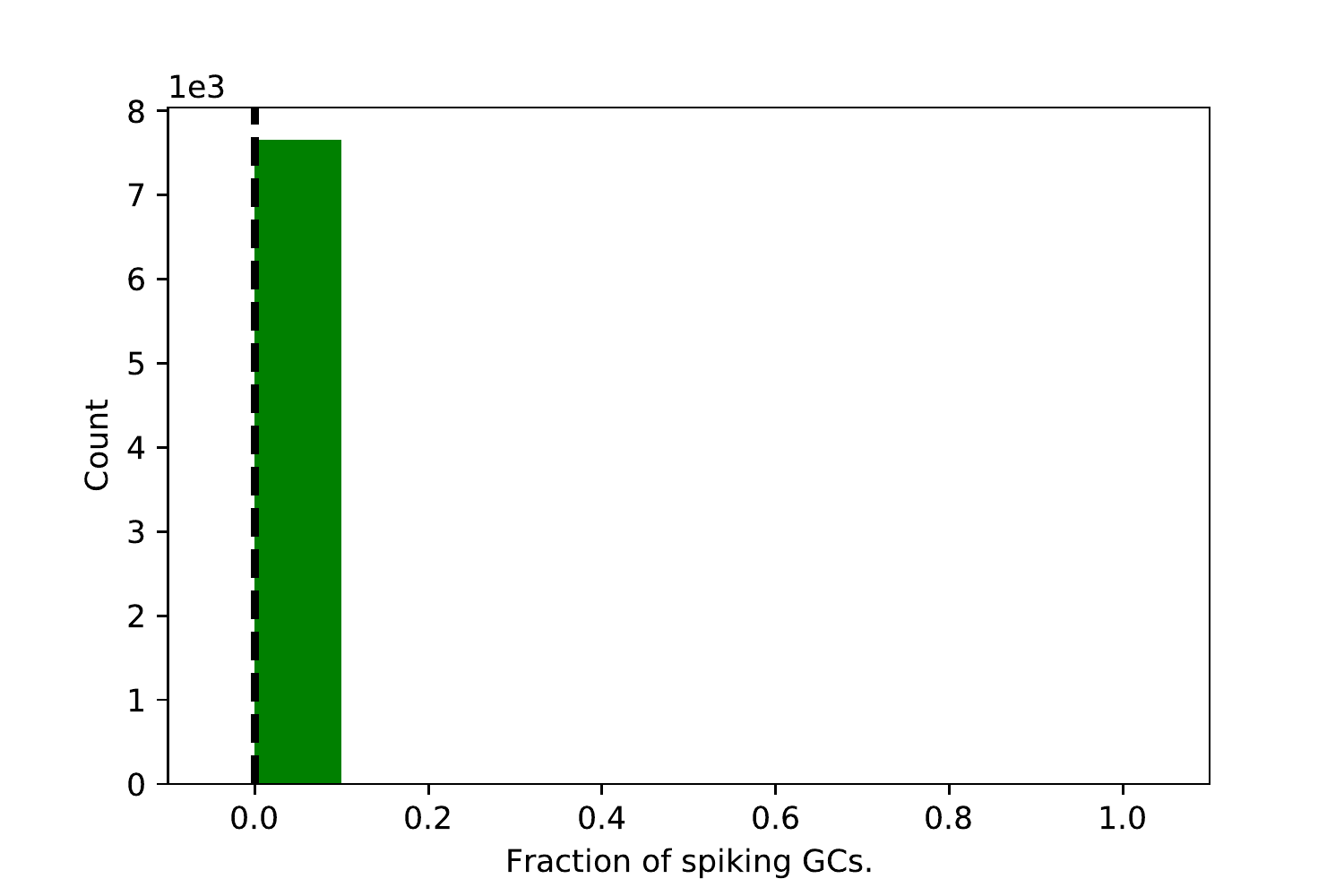}}
\hspace{0.001in}

\subfloat[Scaled data ( all sorted)]{\includegraphics[width=0.32\linewidth]{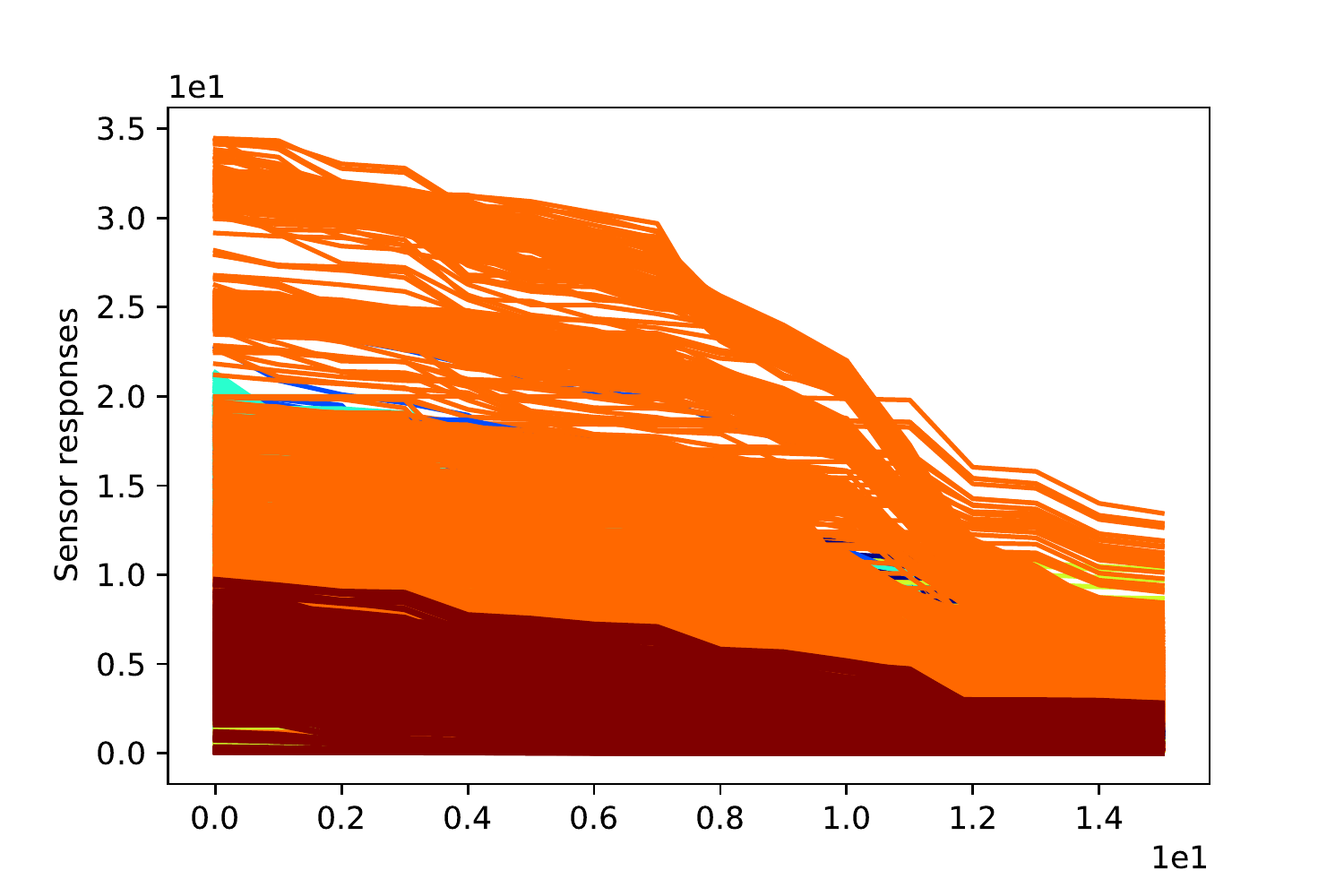}}
\hspace{0.001in}
\subfloat[MC spike count distribution]{\includegraphics[width=0.32\linewidth]{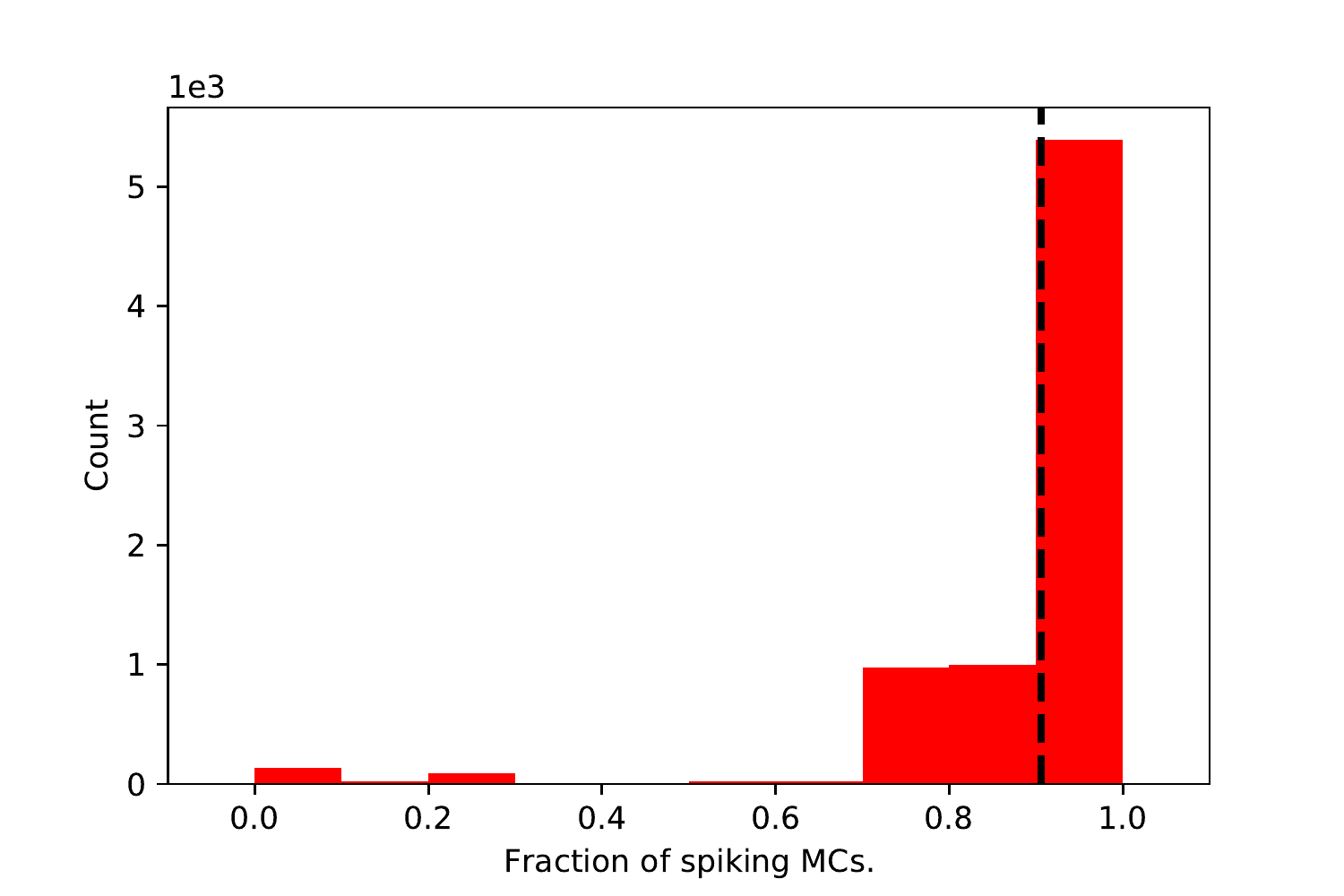}}
\hspace{0.001in}
\subfloat[GC spike count distribution]{\includegraphics[width=0.32\linewidth]{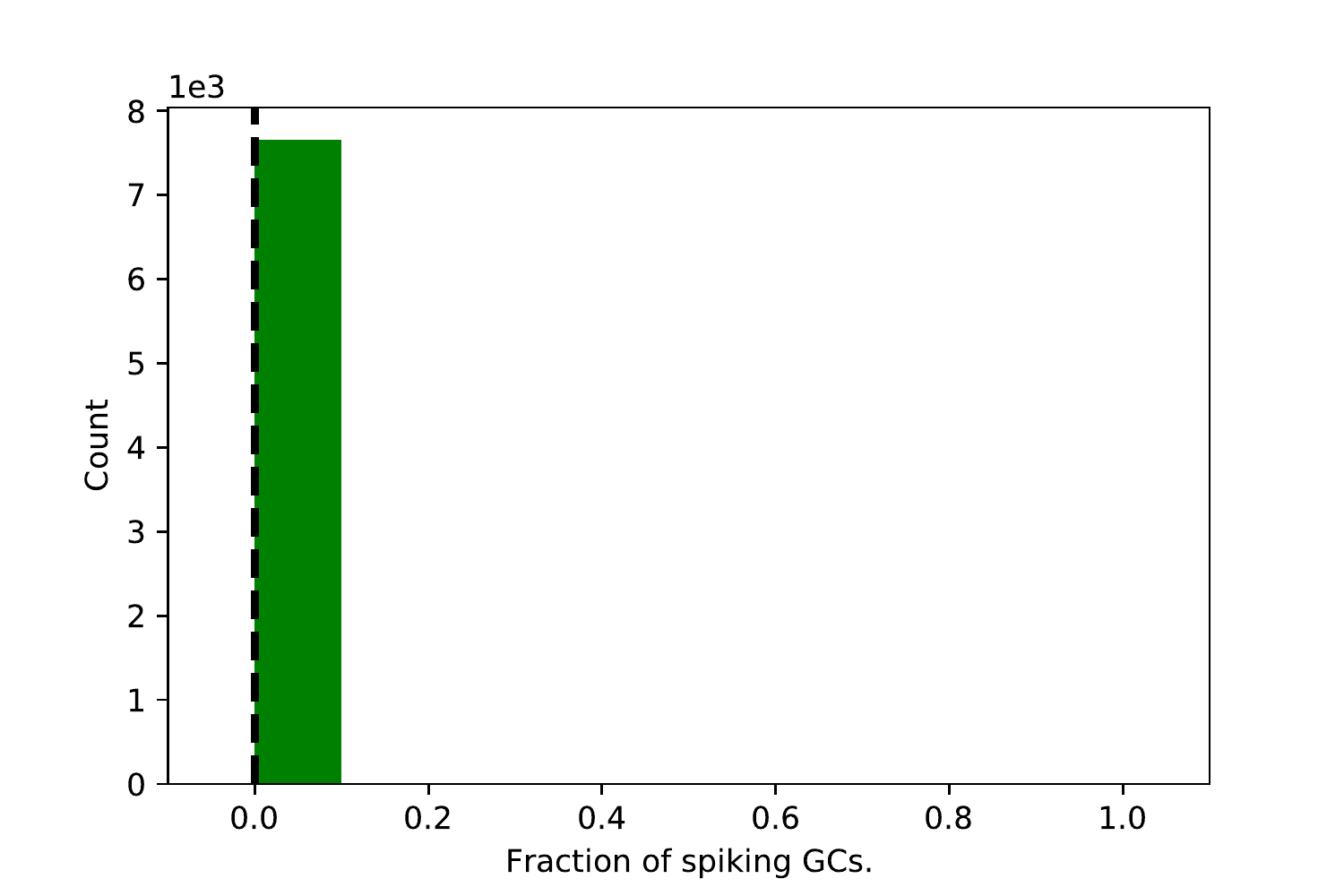}}
\hspace{0.001in}

\subfloat[Regularized \& model scaled data (all sorted)]{\includegraphics[width=0.32\linewidth]{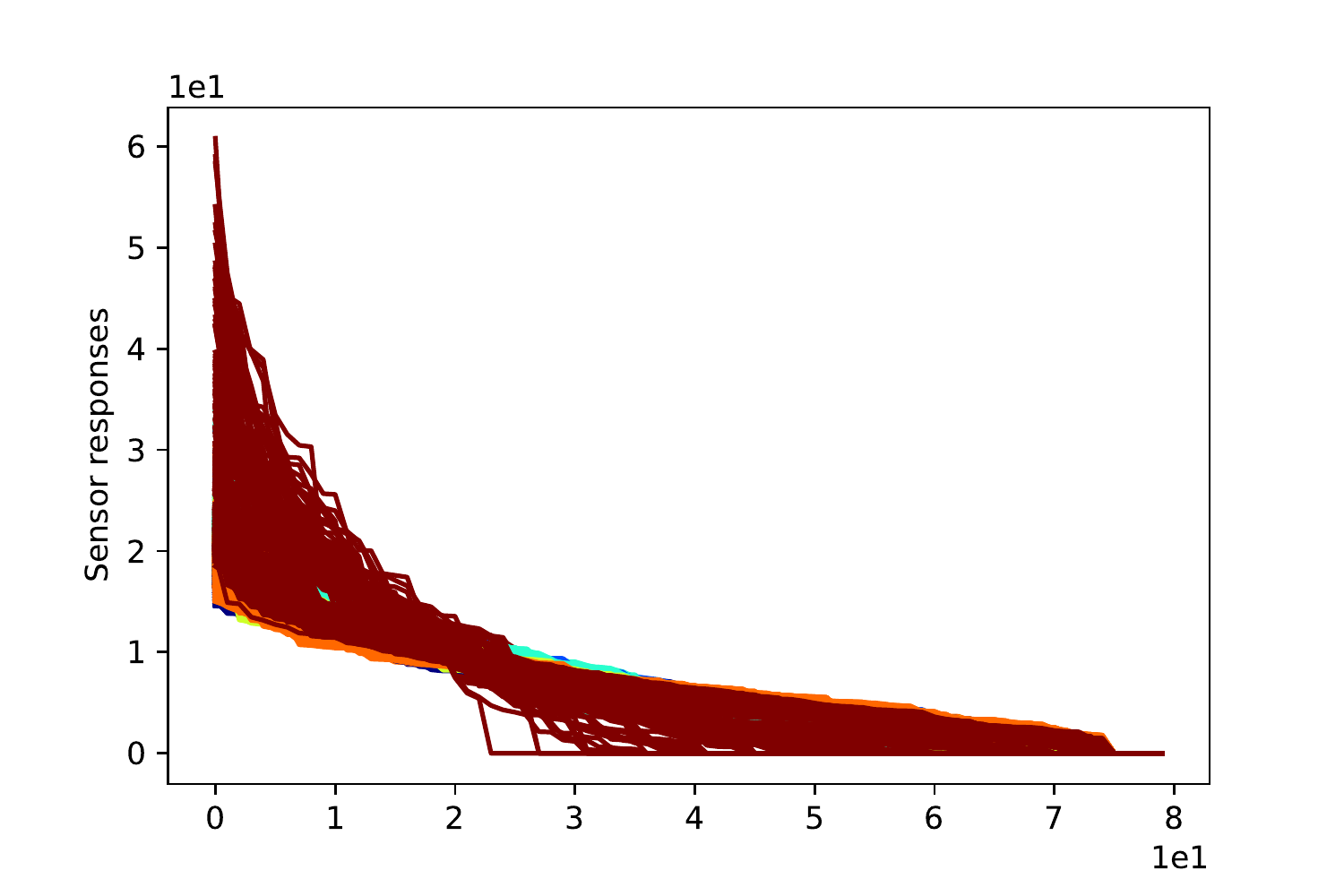}}
\hspace{0.001in}
\subfloat[MC spike count distribution]{\includegraphics[width=0.32\linewidth]{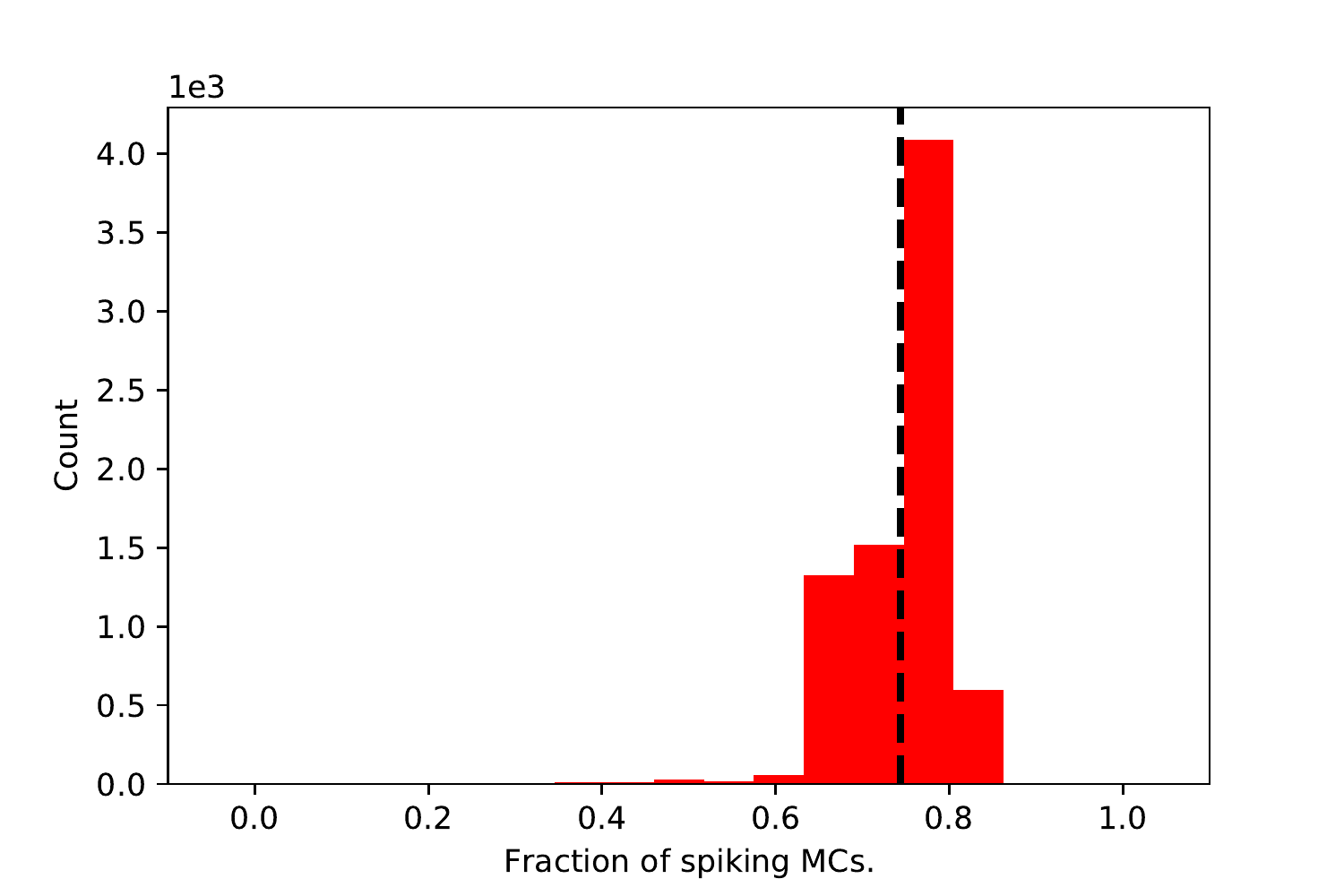}}
\hspace{0.001in}
\subfloat[GC spike count distribution]{\includegraphics[width=0.32\linewidth]{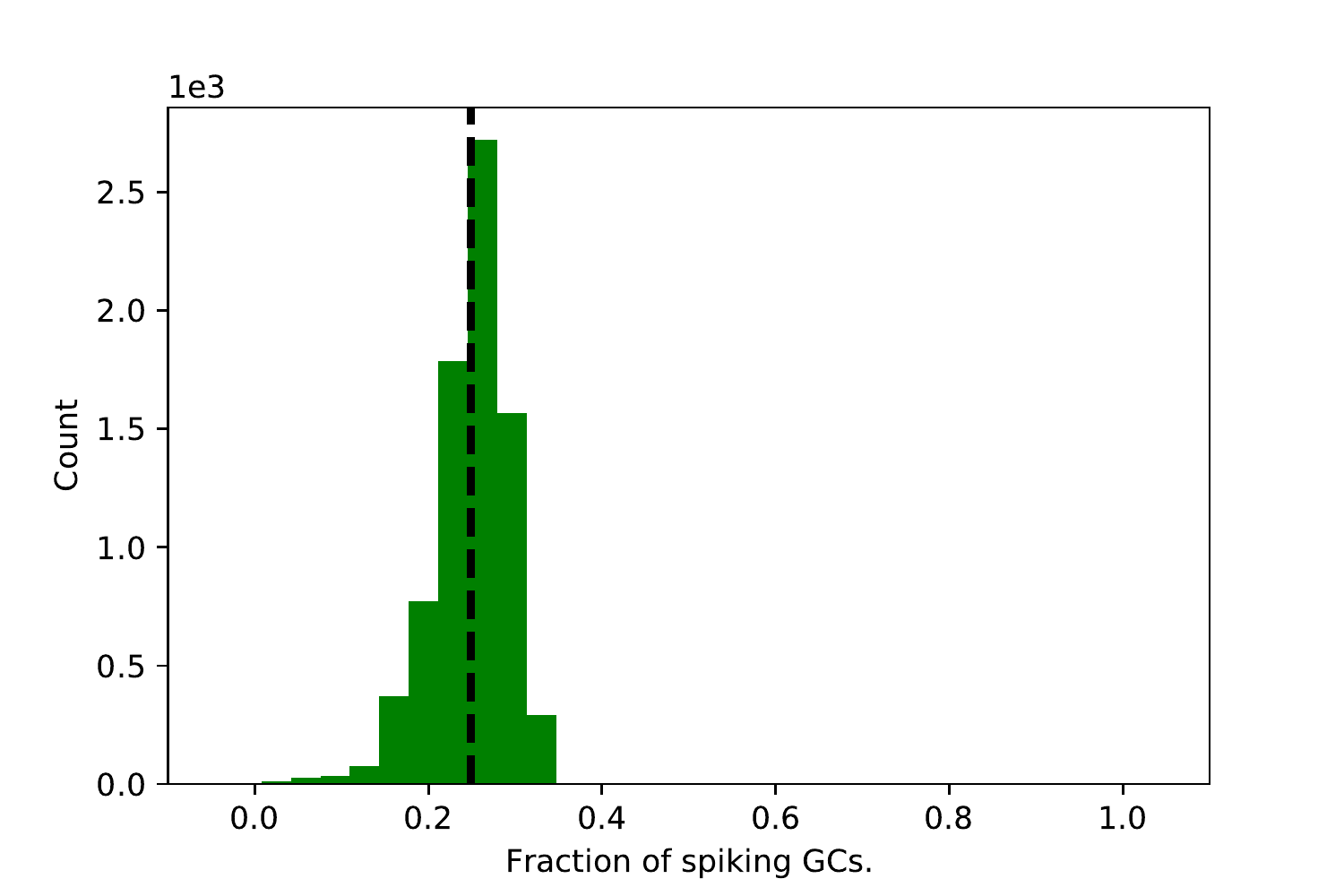}}
\hspace{0.001in}

\caption{Distributions of MC and GC spike counts for batch $1, 7 \, \& \, 10$ of UCSD gas sensor drift dataset ~\cite{vergara_chemical_2012} at different stages of data regularization. a) Raw MOS sensor responses ( sorted by amplitudes ) to $6$ gas types at different levels of concentration from batch $1, 7, 10$ of the UCSD gas sensor drift dataset ~\cite{vergara_chemical_2012}. b) MC spike count ( out of $16$) when raw data is input to the network. The average spike count ( marked in black) is $1.0$. c)  GC spike count ( out of $16 \times 25 \, = \, 400$) when raw data is input to the network. The average spike count ( marked in black) is $0.$. d) Same as (a) but after data scaling. e) Same as (b) after data scaling. Average MC spike count is $0.91$. f) Same as (c) after scaling. Average GC spike count is $0.$. g) Sensor data after application of data regularization ( including heterogeneous duplication) \& model scaling. h) Same as (b,e) after application of data regularization \& model scaling. Average MC spike count is $0.74$. i) Same as (c, f) after application of data regularization \& model scaling. Average GC spike count is $0.25$.}  
  \label{drift_reg}

\end{figure}

\begin{figure}
  \centering
\subfloat[Raw data ( all sorted)]{\includegraphics[width=0.32\linewidth]{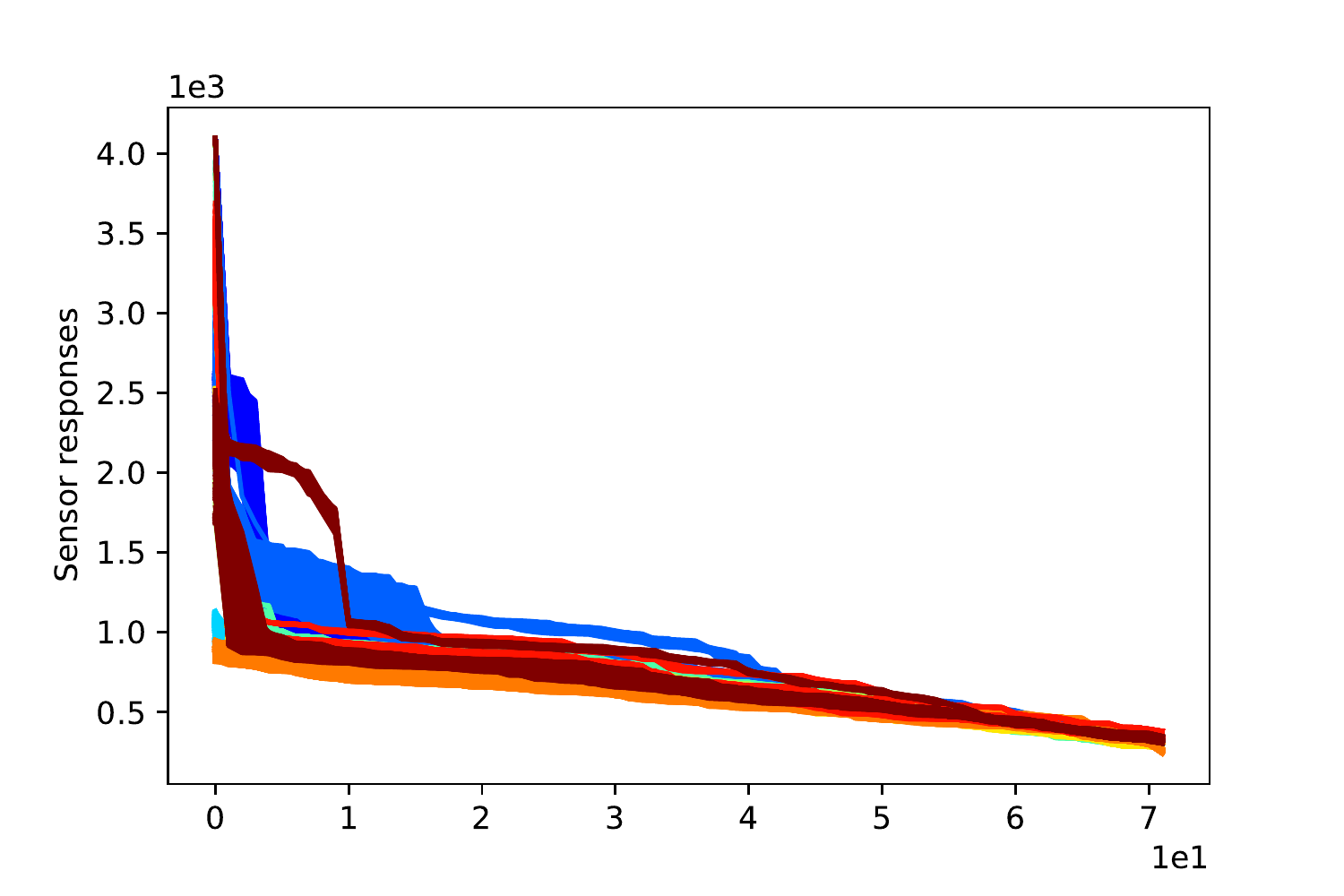}}
\hspace{0.001in}
\subfloat[MC spike count distribution]{\includegraphics[width=0.32\linewidth]{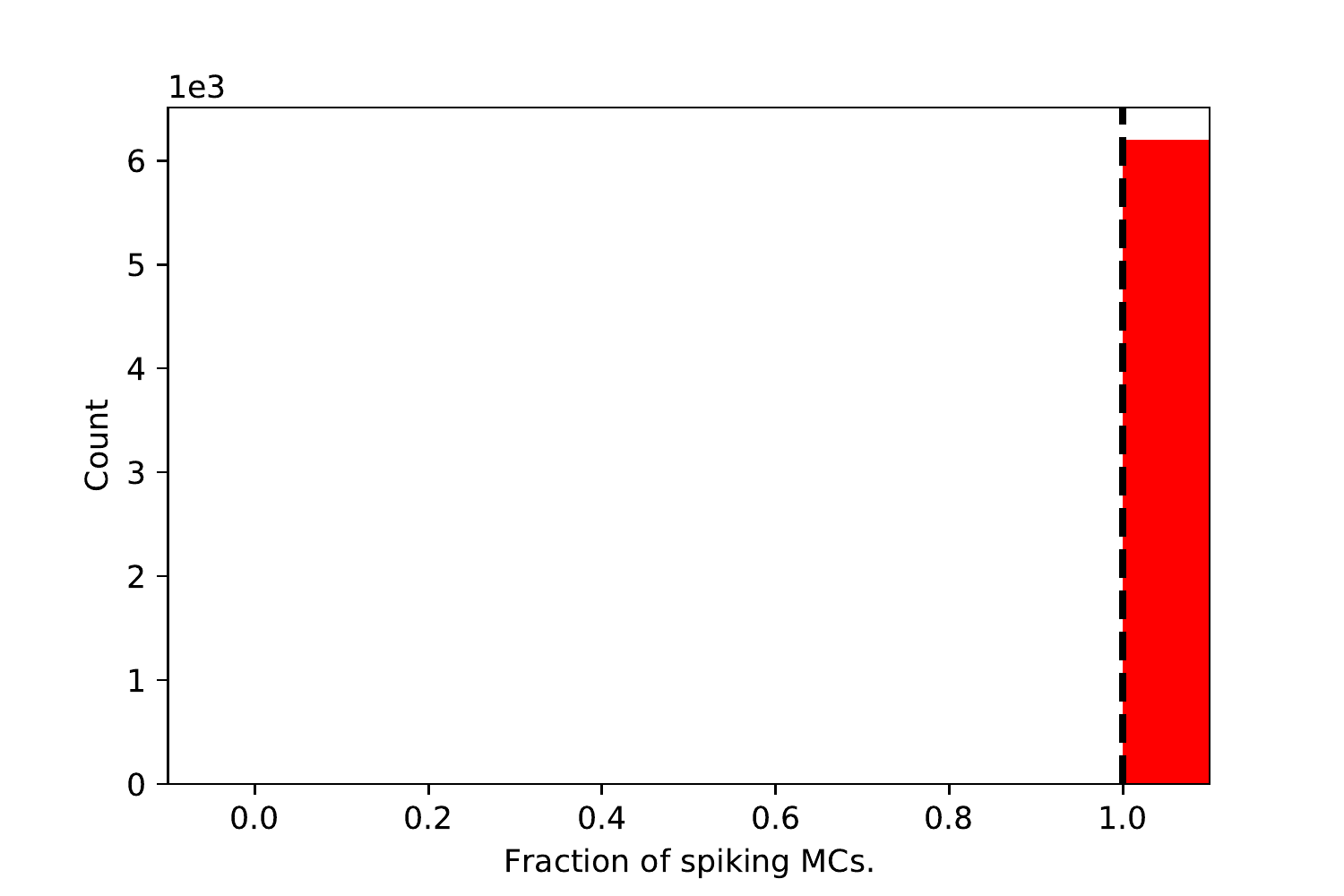}}
\hspace{0.001in}
\subfloat[GC spike count distribution]{\includegraphics[width=0.32\linewidth]{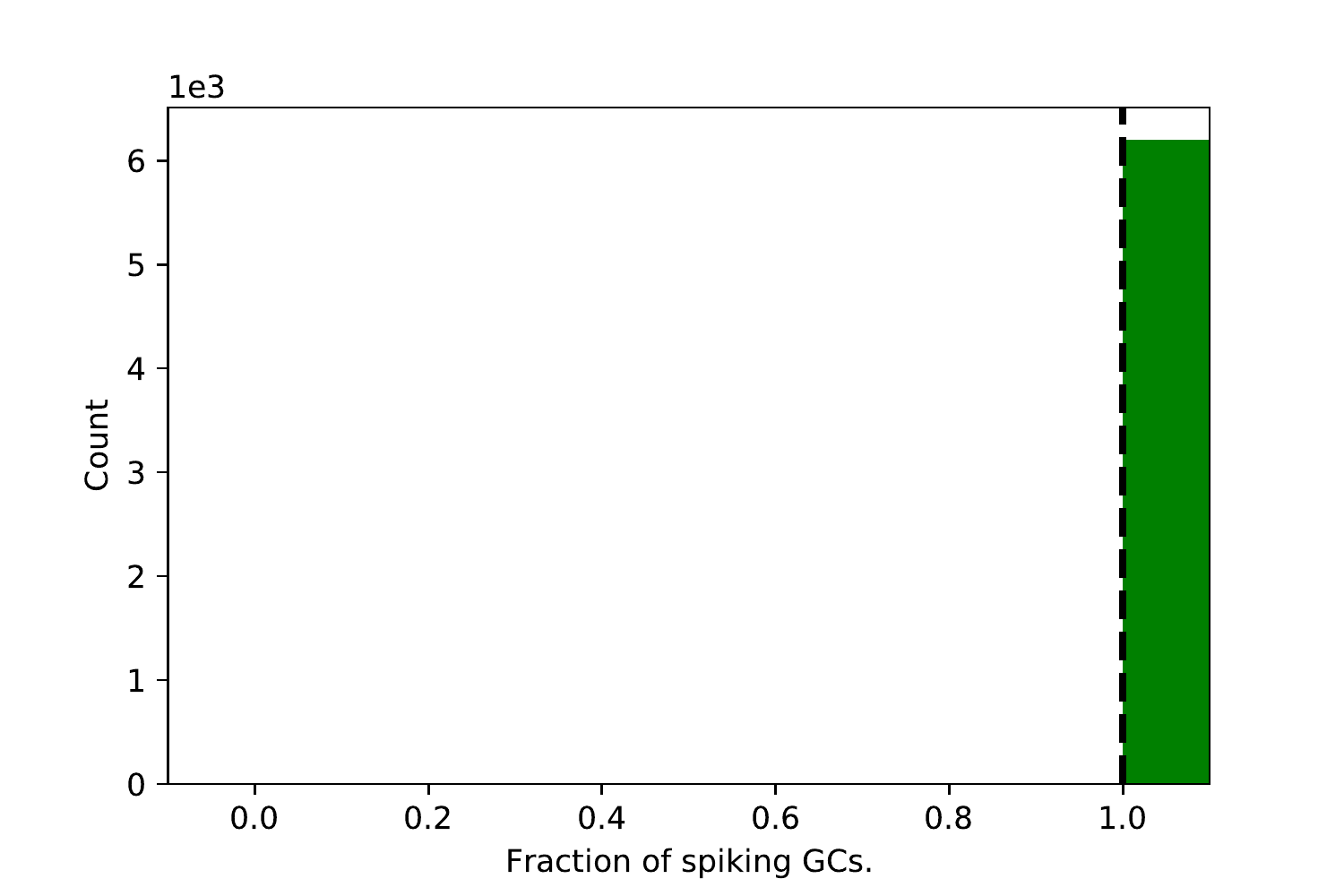}}
\hspace{0.001in}

\subfloat[Scaled data ( all sorted)]{\includegraphics[width=0.32\linewidth]{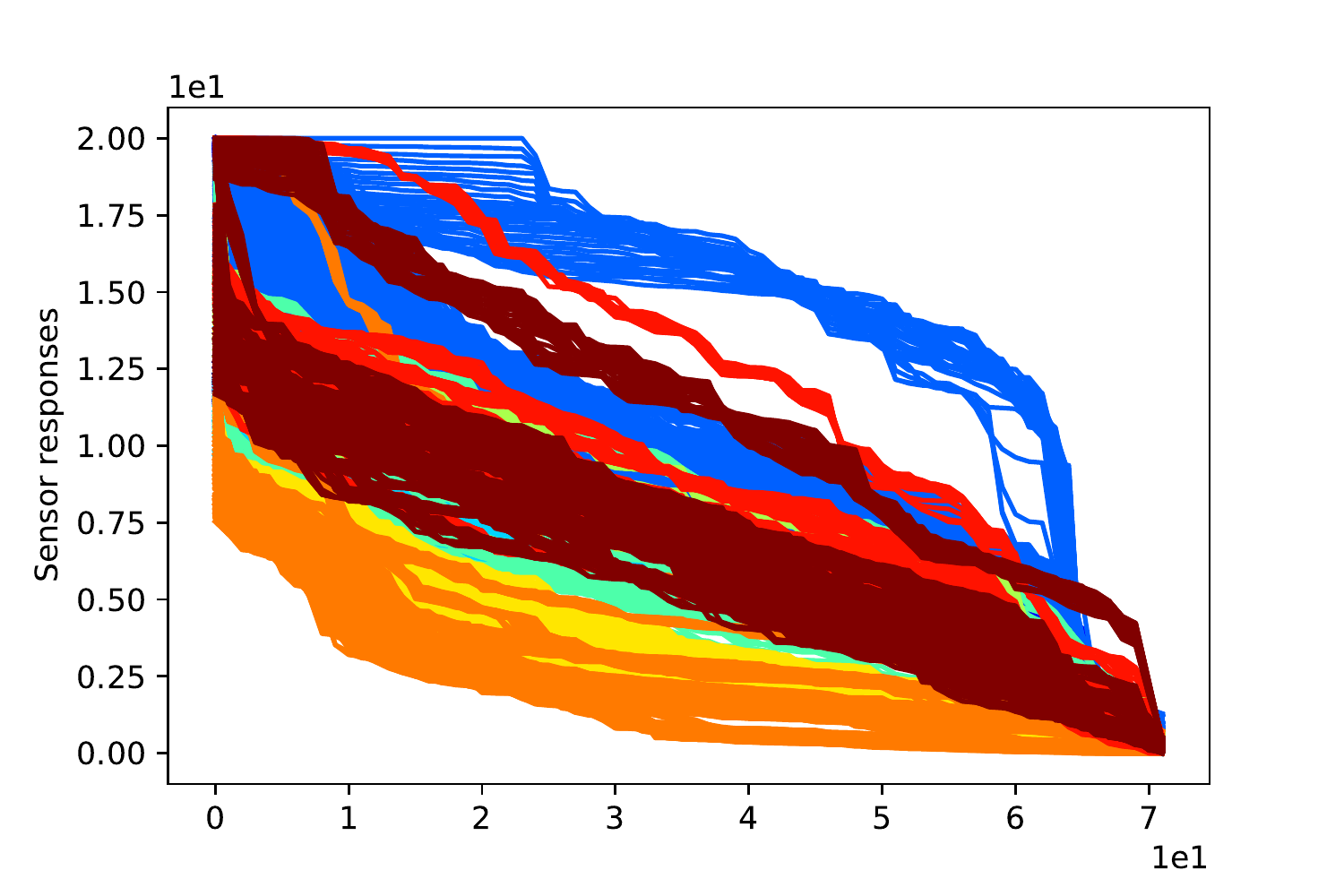}}
\hspace{0.001in}
\subfloat[MC spike count distribution]{\includegraphics[width=0.32\linewidth]{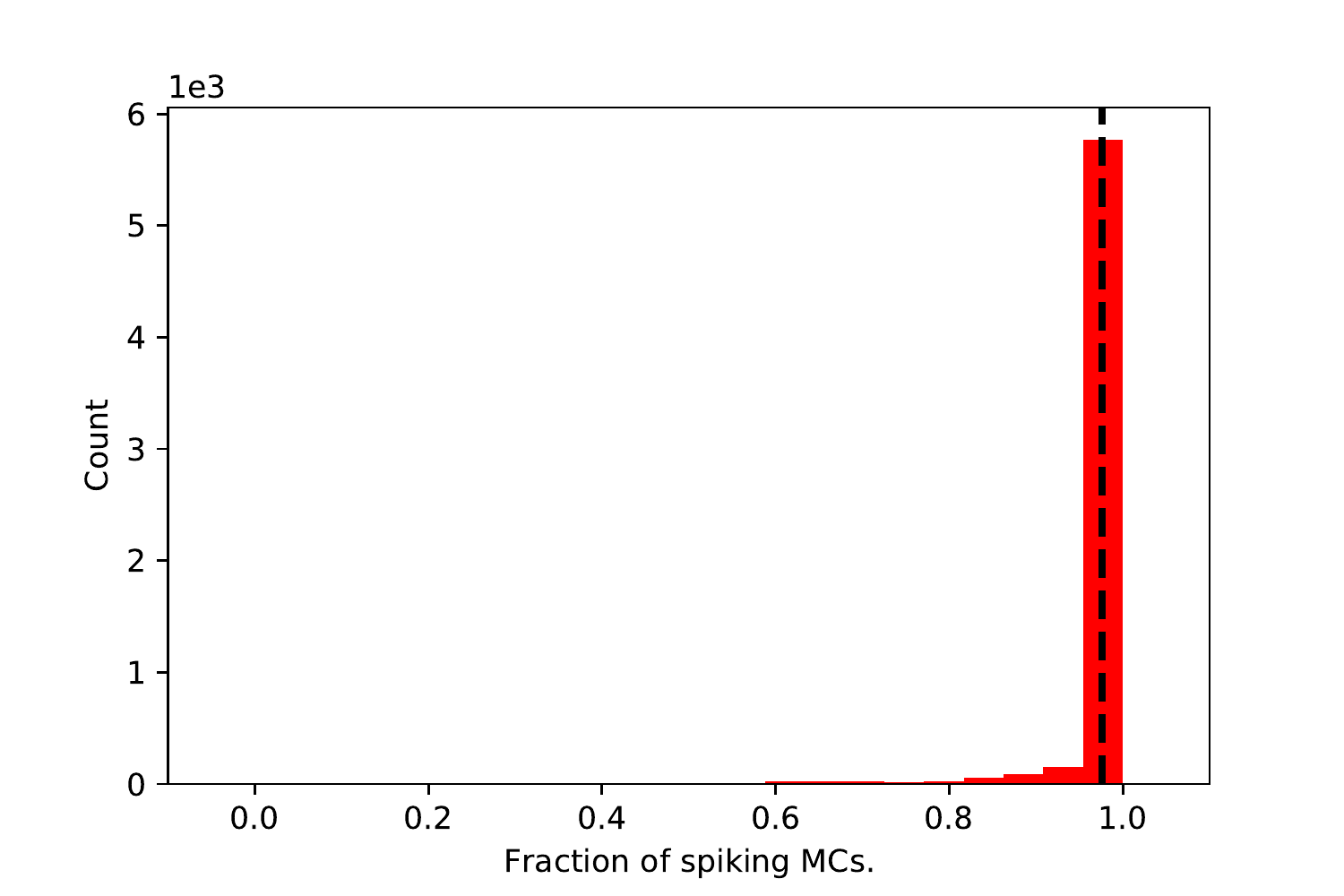}}
\hspace{0.001in}
\subfloat[GC spike count distribution]{\includegraphics[width=0.32\linewidth]{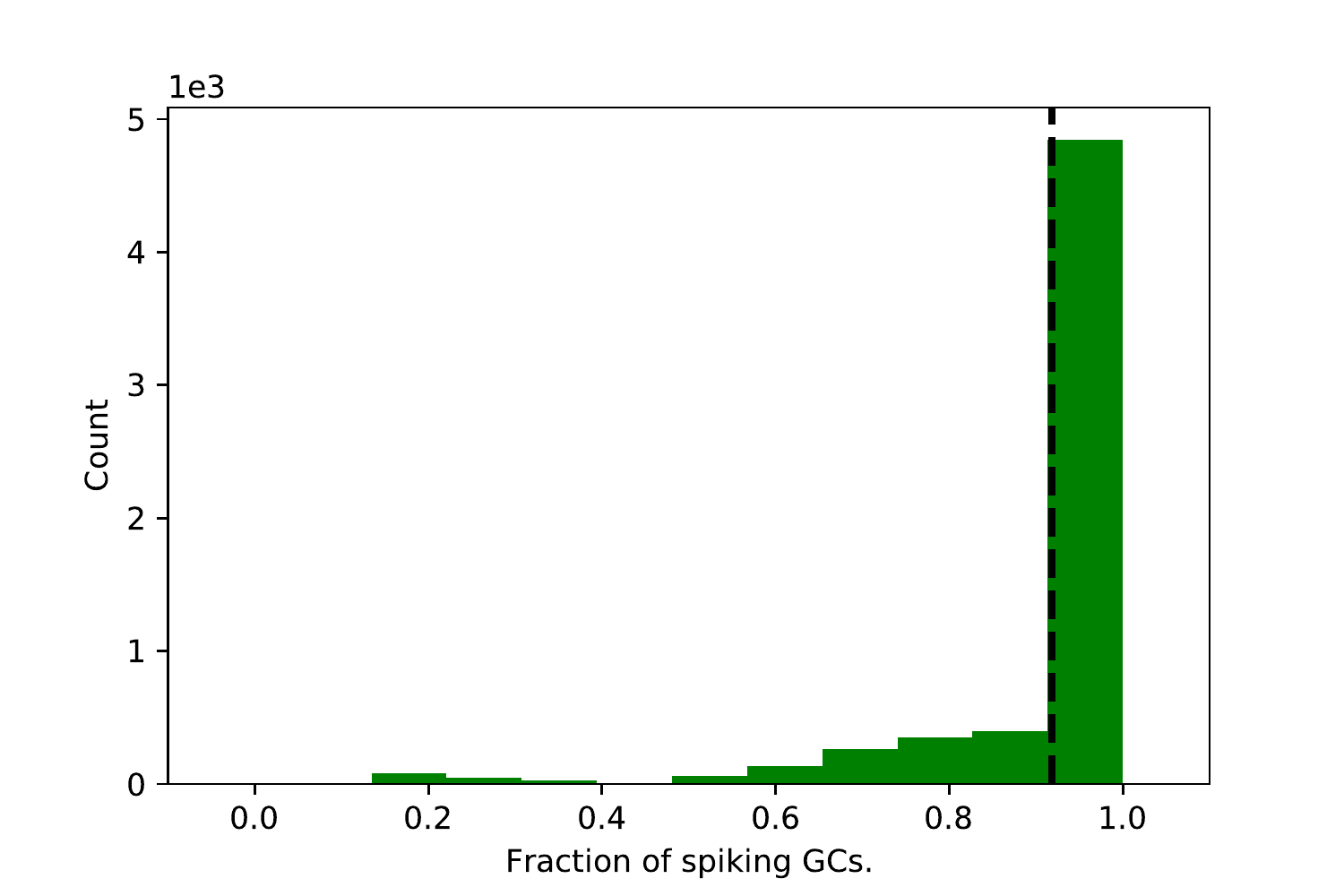}}
\hspace{0.001in}

\subfloat[Regularized \& model scaled data (all sorted)]{\includegraphics[width=0.32\linewidth]{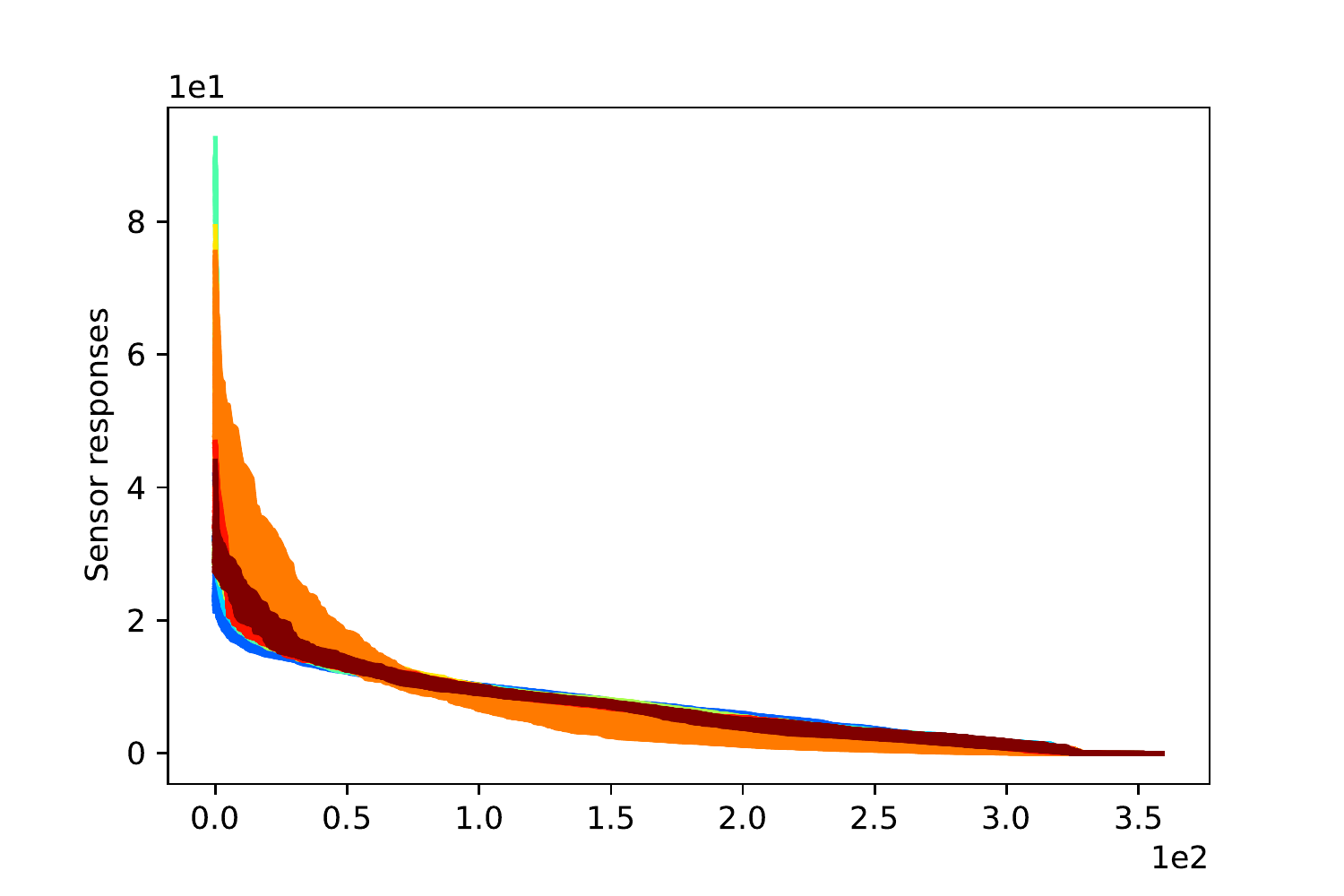}}
\hspace{0.001in}
\subfloat[MC spike count distribution]{\includegraphics[width=0.32\linewidth]{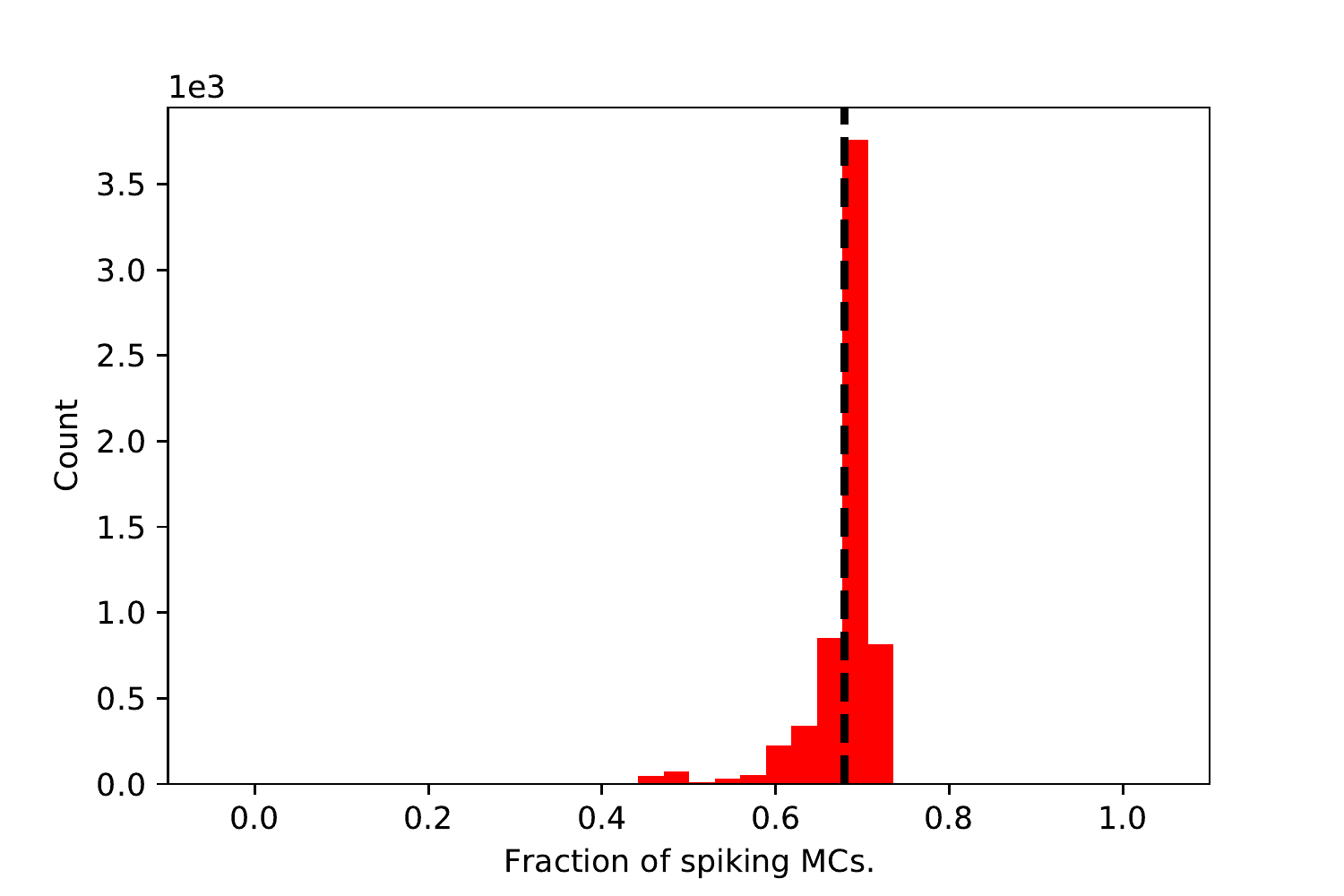}}
\hspace{0.001in}
\subfloat[GC spike count distribution]{\includegraphics[width=0.32\linewidth]{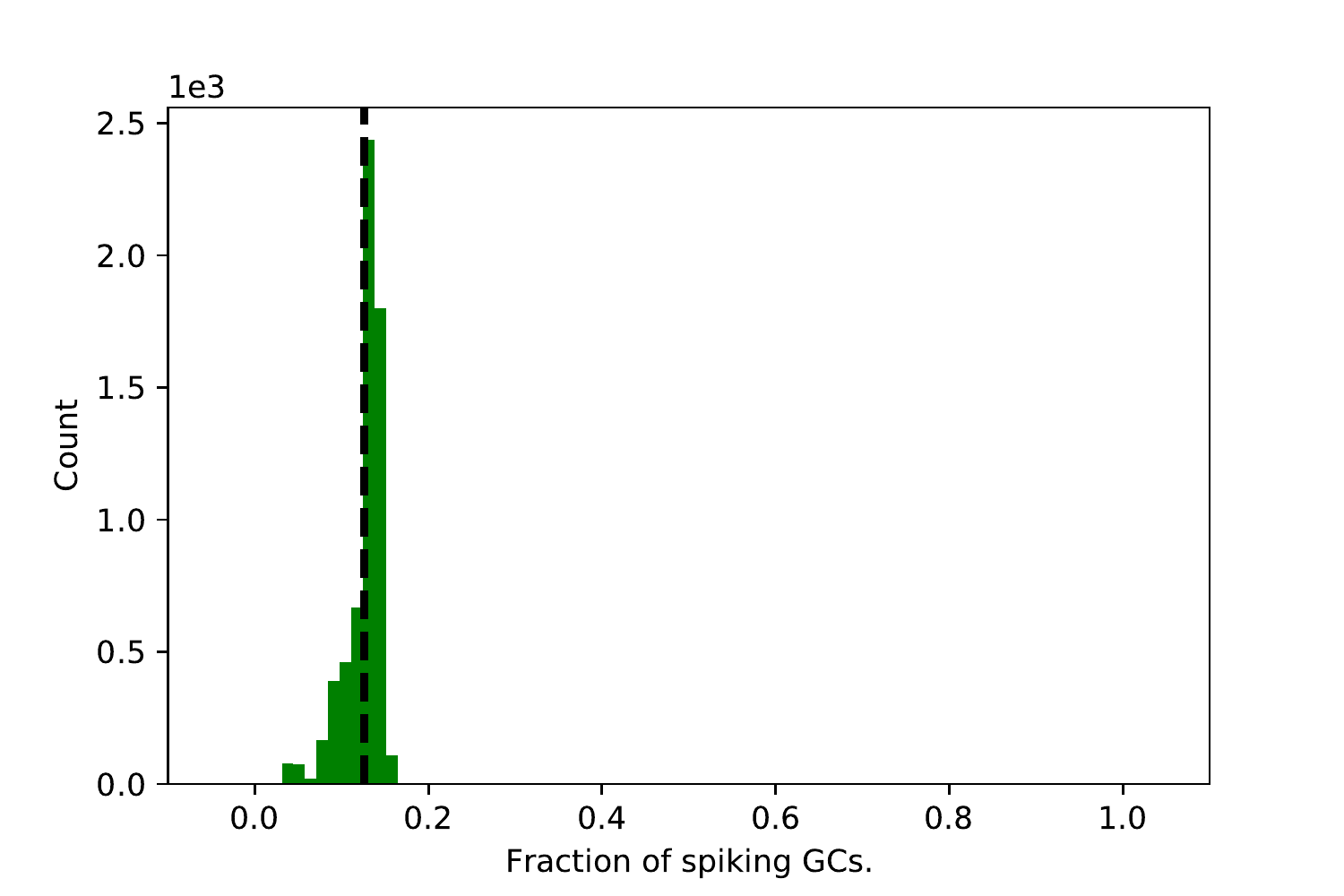}}
\hspace{0.001in}

\caption{Distributions of MC and GC spike counts at different stages of data regularization of "Gas sensor arrays in open sampling
settings" dataset ~\cite{VERGARA2013462}. a) Raw MOS sensor responses ( sorted by amplitudes ) to $10$ gas types. b) MC spike count ( out of $72$) when raw data is input to the network. The average spike count ( marked in black) is $1.0$. c)  GC spike count ( out of $72 \times 25 \, = \, 1800$) when raw data is input to the network. The average spike count ( marked in black) is $1$. d) Same as (a) but after data scaling. e) Same as (b) after data scaling. Average MC spike count is $0.98$. f) Same as (c) after scaling. Average GC spike count is $0.92$. g) Sensor data after application of data regularization ( including heterogeneous duplication) \& model scaling. h) Same as (b,e) after application of data regularization \& model scaling. Average MC spike count is $0.68$. i) Same as (c, f) after application of data regularization \& model scaling. Average GC spike count is $0.13$.}  
  \label{wt_reg}

\end{figure}

\textit{Learning in the wild} is concerned with the ability of training / testing the network with intrinsically high dimensional data of unpredictable distribution and of various dimensions ( consistent for a particular task though ). As described earlier, data regularization and model scaling can together regularize MC/GC spike counts across data from various distributions and dimensions. We here describe the effectiveness of this technique on real world datasets. \\

\paragraph{UCSD gas sensor array drift dataset ~\cite{UCIMachineLearningRepositoryGasSen,vergara_chemical_2012}}

Fig ~\ref{drift_reg}a plots the raw sensor responses from batch $1$ ( $1^{st}$ month ), batch $7$ ( $21^{st}$ month ), and batch $10$ ( $36^{th}$ month ) of the UCSD gas sensor drift dataset all sorted by their response amplitudes. 
Due to drift, the sensor responses to the gasses degrade with time and consequently the data statistics alter. With raw data as input, the average fraction of MC spike count was $1.$ and the GC spike average spike count was $0.$, Fig ~\ref{drift_reg}b,c. This is not optimal for good model performance. Fig ~\ref{drift_reg}d shows the corresponding scaled data. The mean MC and GC spike counts are $0.91$ and $0.$ respectively. Fig ~\ref{drift_reg}g shows the sensor responses after application of data regularization and model scaling.  This step ensures an optimal MC and GC spike count fraction (mean spike fraction for MC: $0.74$ and for GC: $0.25$). The goodness of preprocessing ($g_{p}$) of MC is $0.86$ and the for GC is $0.72$, Table ~\ref{tab:mc_gp}, ~\ref{tab:gc_gp}.

\paragraph{Gas sensor arrays in open sampling settings dataset ~\cite{UCIMachineLearningRepositoryGassen_wind,VERGARA2013462}}

Fig ~\ref{wt_reg}shows the raw sensor responses comprising plume dynamics for $10$ gases (see materials \& methods for details). The MC and GC average spike counts are $1.$, Fig ~\ref{wt_reg}b, c.Fig ~\ref{wt_reg}d shows the data after scaling. Fig ~\ref{wt_reg}e,f shows the MC and the GC average spike counts ($0.98$, $0.92$) - which are high. After application of data regularization and model scaling, the data becomes regularized Fig ~\ref{wt_reg}g. The MC and GC average spike counts are within acceptable ranges, $0.68$ and $0.13$ respectively. The goodness of preprocessing ($g_{p}$) of MC is $0.92$ and the for GC is $0.76$, Table ~\ref{tab:mc_gp}, ~\ref{tab:gc_gp}.

\subsubsection{Classification performance on the UCSD gas sensor drift dataset while learning in the wild}

\begin{table}[h!]
\centering
\resizebox{\textwidth}{!}{
 \begin{tabular}{||c c c c c c c c c c c ||} 
 \hline
  & Batch 1 & Batch 2 & Batch 3 & Batch 4 & Batch 5 & Batch 6 & Batch 7 & Batch 8 & Batch 9 & Batch 10\\ [0.5ex] 
 \hline\hline
 Months & 1-2 & 3-10 & 11-13 & 14-15 & 16 & 17-20 & 21 & 22-23 & 24-30 & 36\\ 
 \#Samples & 445 & 1,244 & 1,586 & 161 & 197 & 2,300 & 3,613 & 294 & 470 & 3,600\\
 \hline\hline
 
 \end{tabular}}
  \caption{ Properties of the UCSD gas sensor drift dataset.\\
  Months denote the age of the sensor array during the sampling of the corresponding dataset. \#Samples denote the number of samples provided by the dataset in that particular batch. \\
 }
\label{table:2}
\end{table}

\begin{table}[h!]
\centering
\resizebox{\textwidth}{!}{
 \begin{tabular}{||c c c c c c c c c c c c||} 
 \hline
   & Batch 1 & Batch 2 & Batch 3 & Batch 4 & Batch 5 & Batch 6 & Batch 7 & Batch 8 & Batch 9 & Batch 10 & Average\\ [0.5ex] 
 \hline
 1 shot & 93.84 & 82.53 & 94.36 & 71.84 & 91.04 & 78.38 & 79.54 &  86.43 & 91.93 & 82.54 & 85.24\\ 
 \hline
 \end{tabular}}
  \caption{Mean classification accuracy across all test odorants on the UCSD drift data set by Sapinet using jaccard distance.}

\label{table:4}
\end{table}

\begin{figure}
  \centering
  \includegraphics[width=0.8\linewidth]{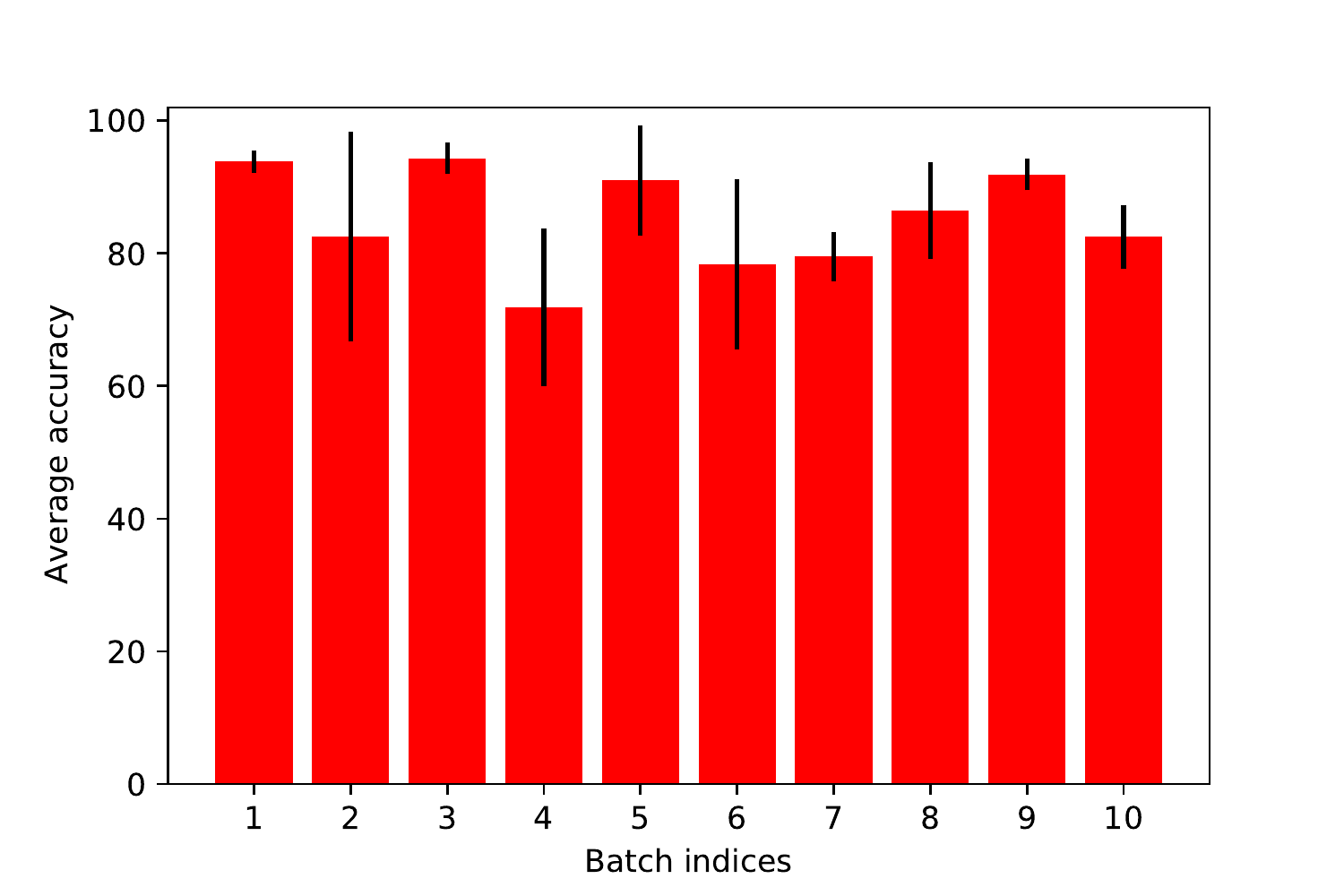}
  \caption{Mean classification ( of 5 runs) accuracy  on 1 shot  online learning of gas sensor responses across all batches of the UCSD drift dataset. The train samples for 1 shot learning are drawn randomly in each run and hence are different across runs.}  
  \label{drift_ol_acc}
\end{figure}

\begin{figure}
  \centering
\subfloat[Ethanol]{\includegraphics[width=0.48\linewidth]{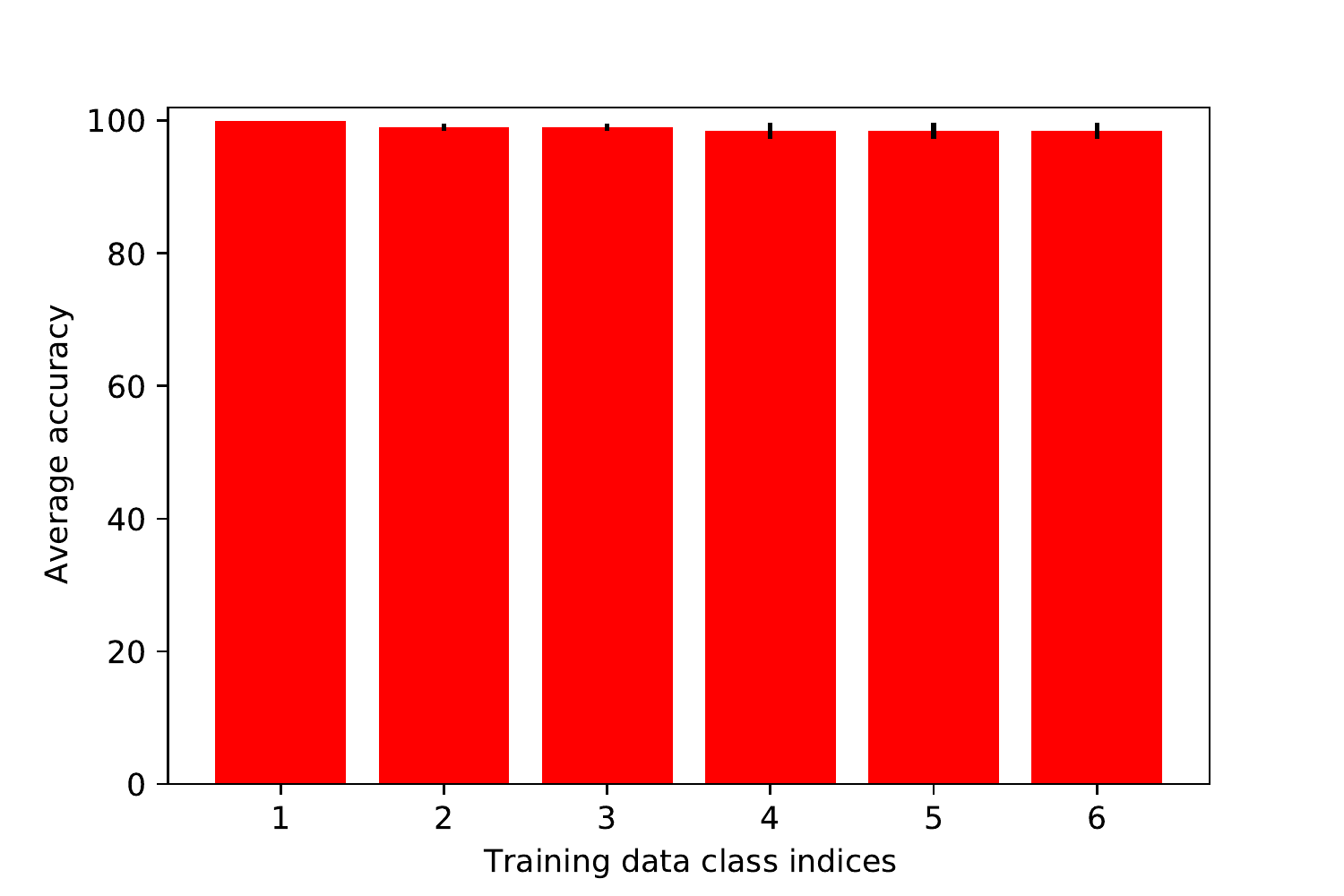}}
\hspace{0.001in}
\subfloat[Ethylene]{\includegraphics[width=0.48\linewidth]{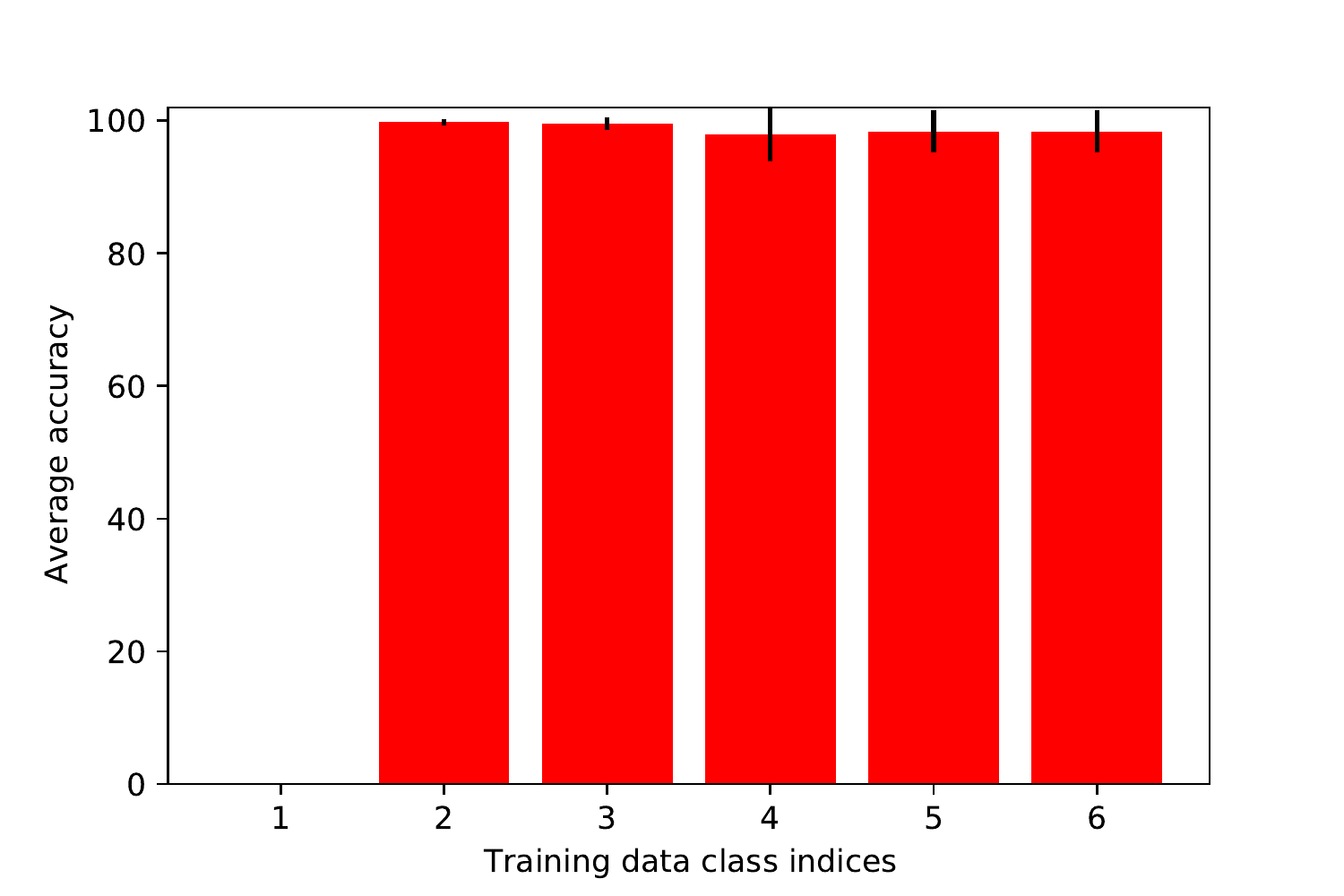}}
\hspace{0.001in}
\subfloat[Ammonia]{\includegraphics[width=0.48\linewidth]{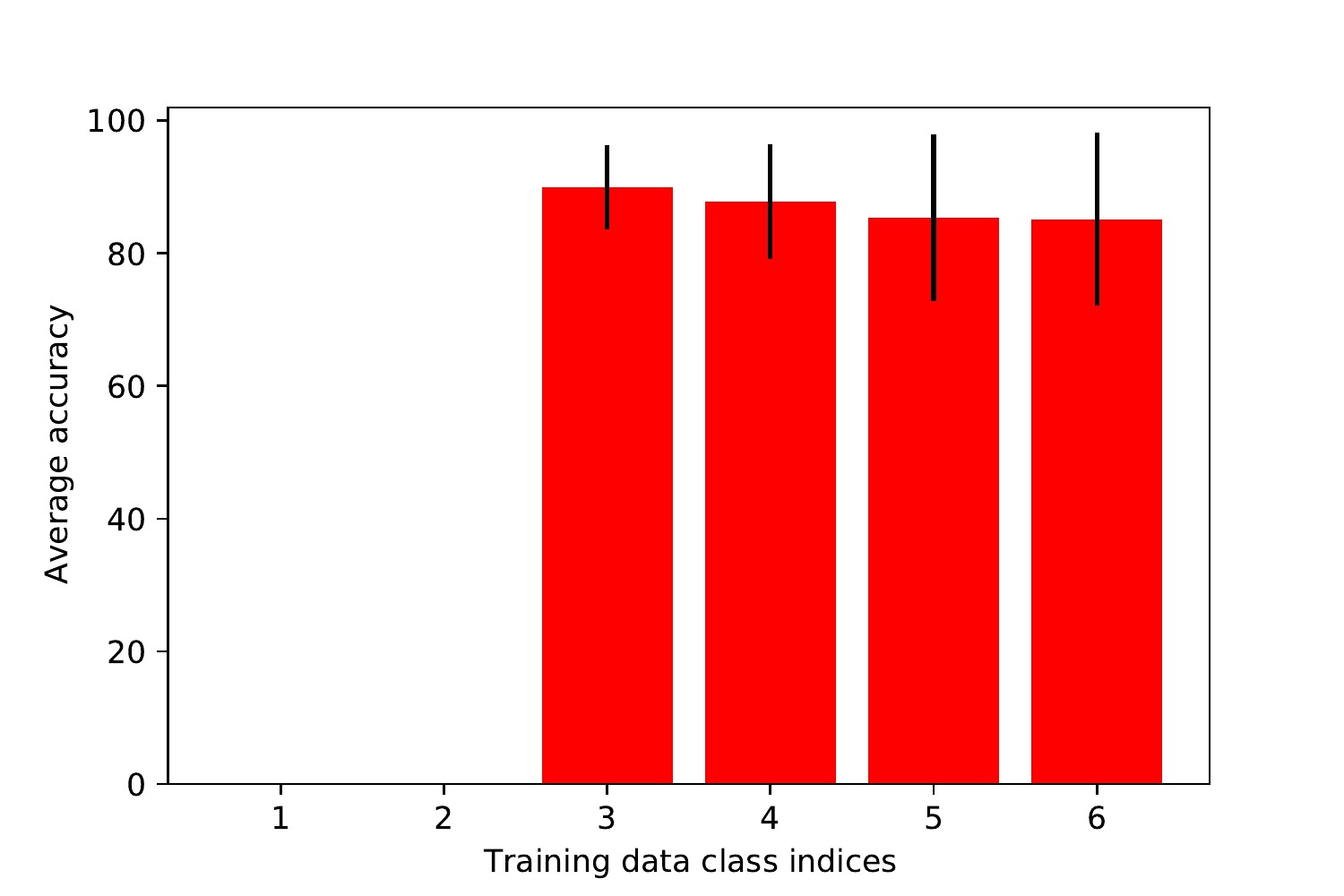}}
\hspace{0.001in}
\subfloat[Acetaldehyde]{\includegraphics[width=0.48\linewidth]{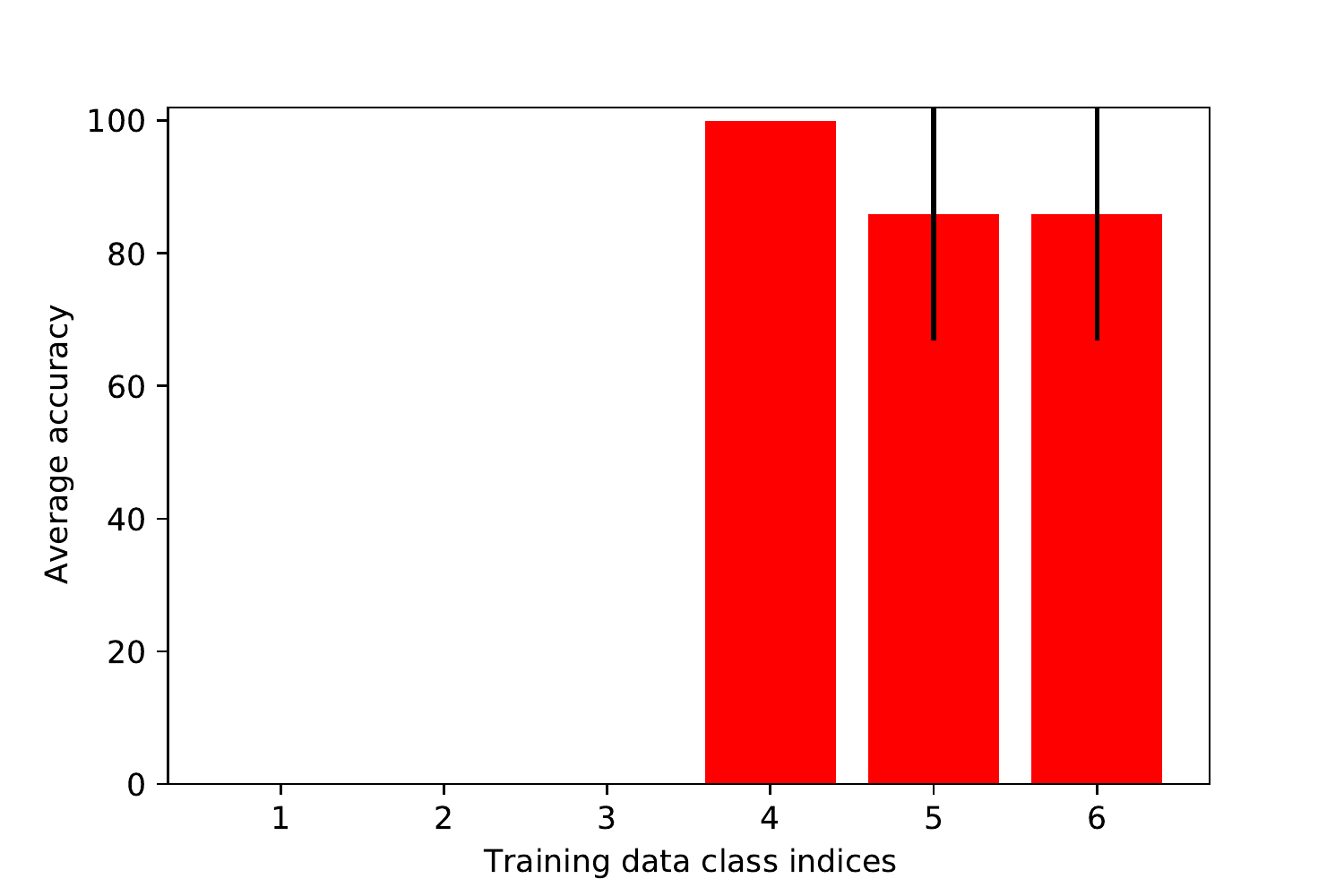}}
\hspace{0.001in}
\subfloat[Acetone]{\includegraphics[width=0.48\linewidth]{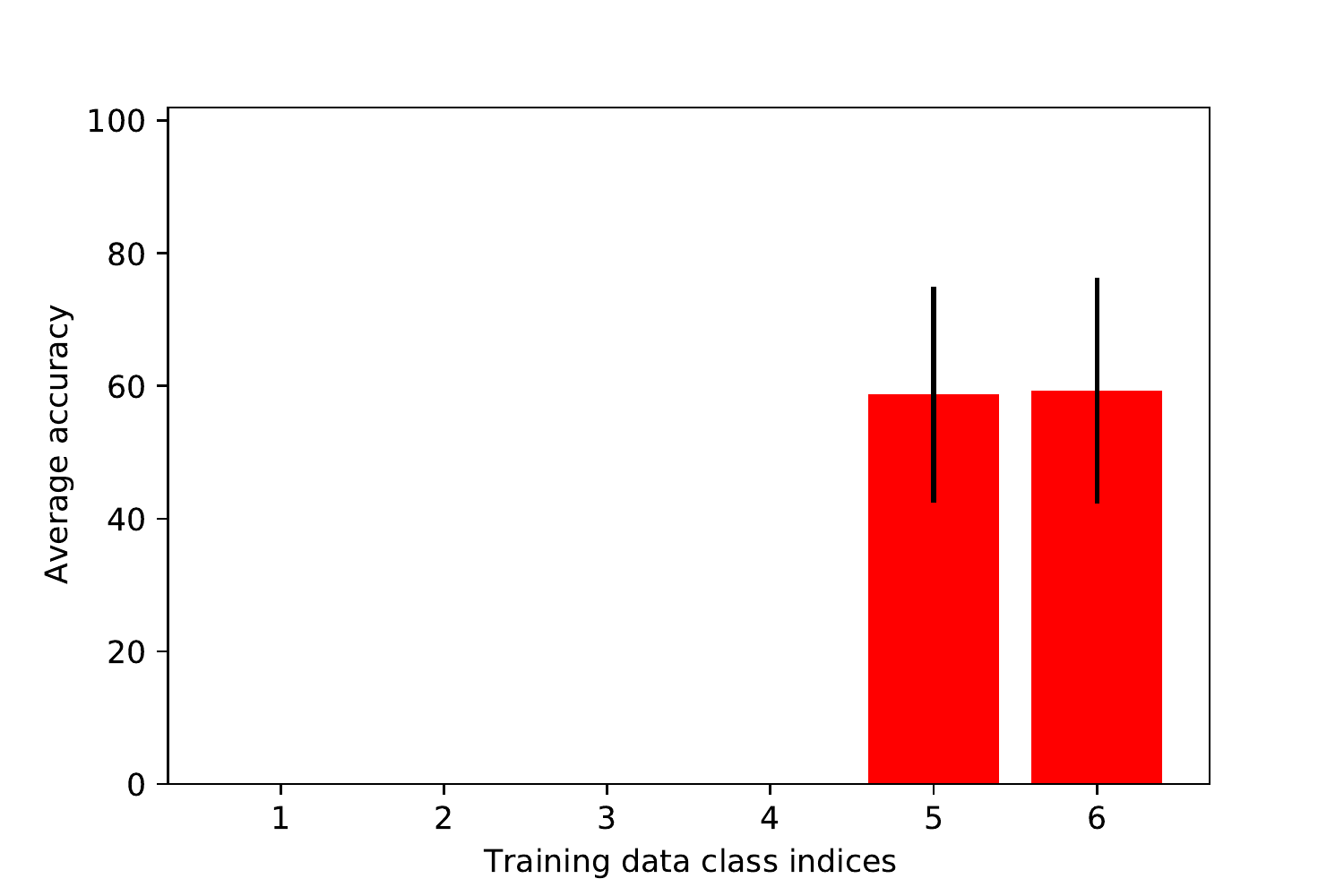}}
\hspace{0.001in}
\subfloat[Toluene]{\includegraphics[width=0.48\linewidth]{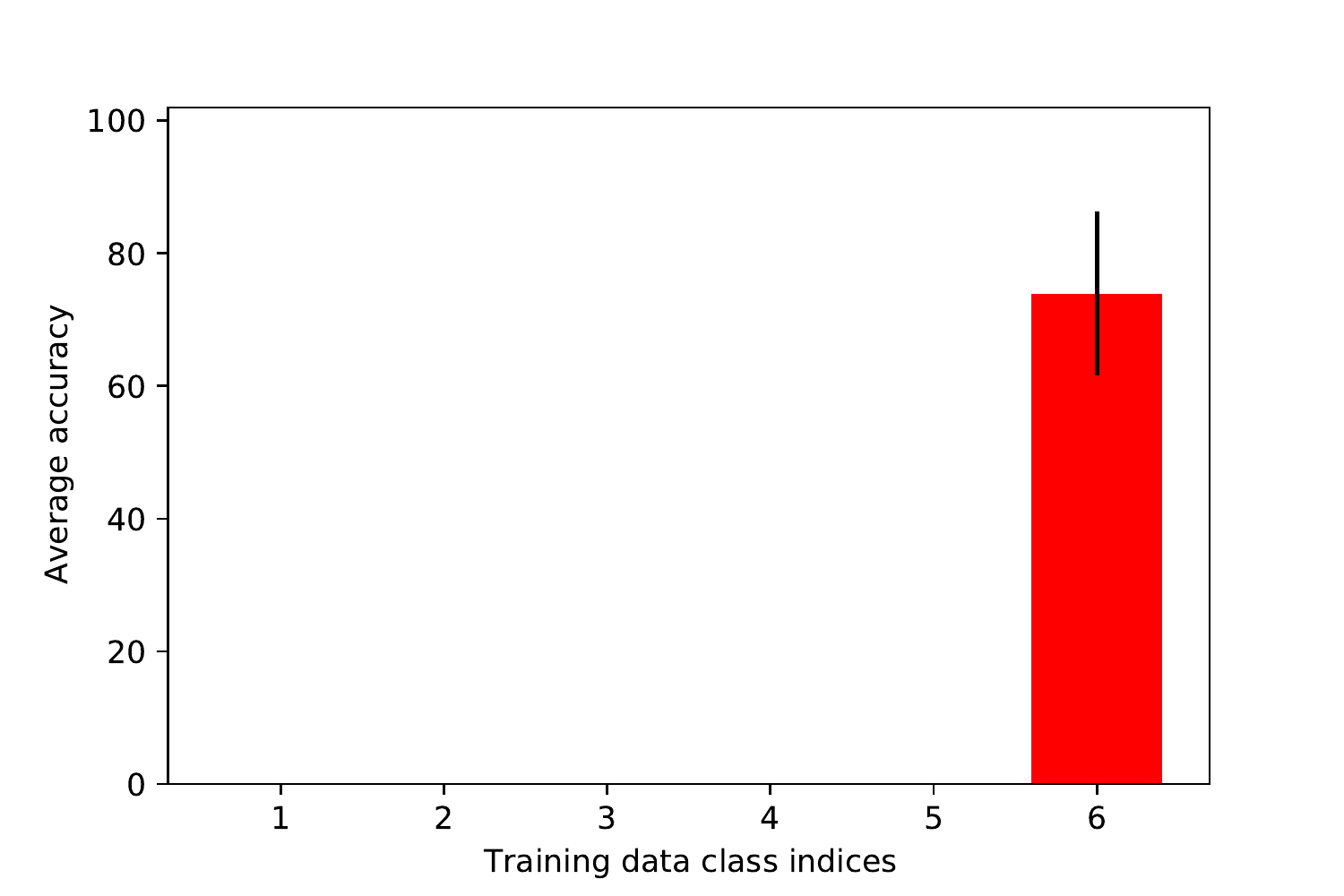}}
\hspace{0.001in}
\caption{Mean ( of five runs ) classification accuracies of all $6$ gas types of batch 1 of the UCSD gas sensor drift dataset during online learning. The accuracies are shown separately for each of the gases. The network was first trained and tested with ethanol (a) followed by ethylene(b), ammonia (c), acetaldehyde (d), acetone (e) and toluene (f).}  
  \label{accu_ol_drift}

\end{figure}

\begin{figure}
  \centering
  \includegraphics[width=0.8\linewidth]{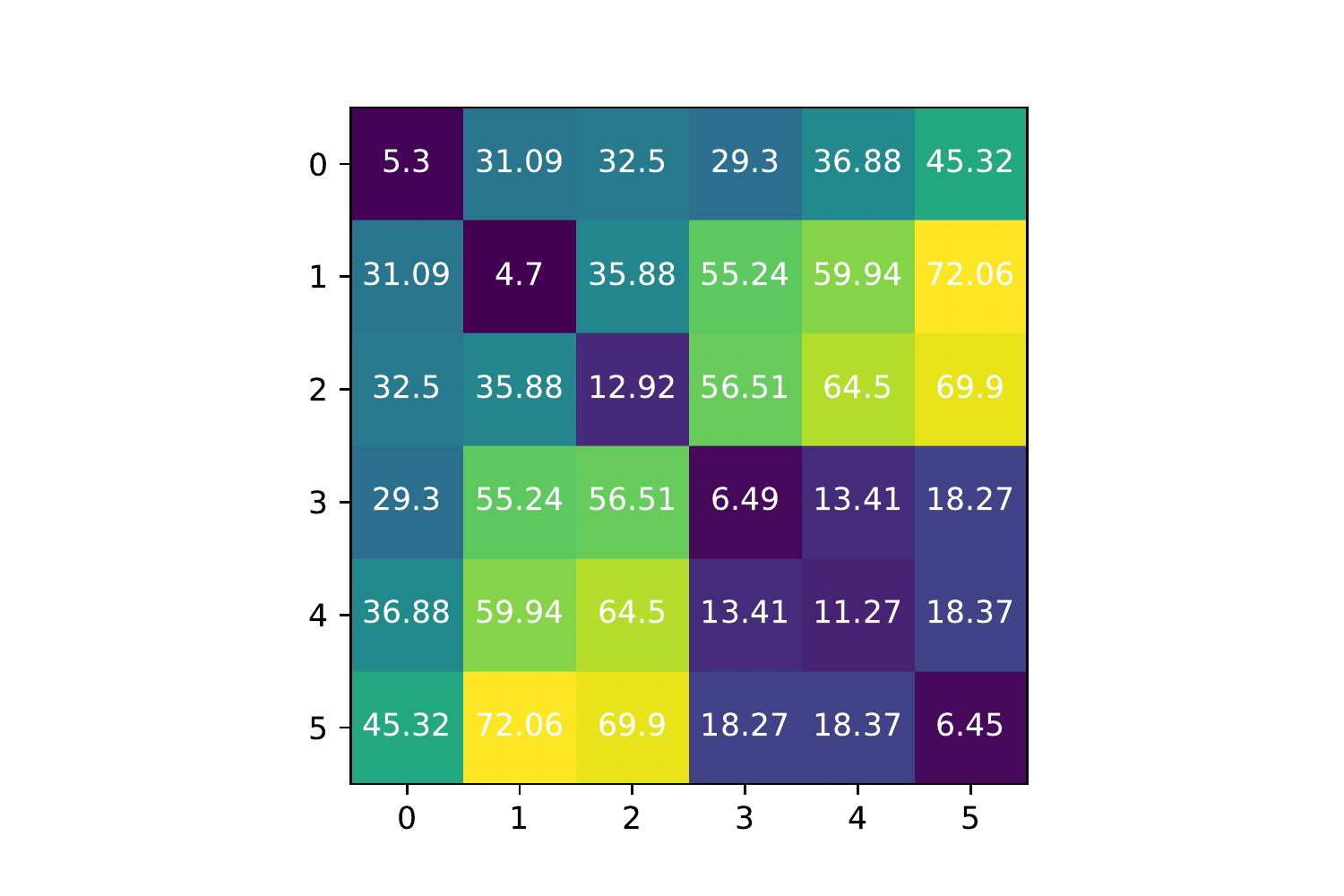}
  \caption{Inter and intra euclidean cluster distance between the $6$ gas types of the UCSD gas sensor drift dataset (batch 1). Label 0: Ethanol; Label 1: Ethylene; Label 2: Ammonia; Label 3: Acetaldehyde; Label 4: Acetone; Label 5: Toluene}  
  \label{drift_euc_b1}
\end{figure}

\begin{table}[h!]
\centering
\resizebox{\textwidth}{!}{
 \begin{tabular}{||c c c c c c c ||} 
 \hline
   & Group 1 & Group 2 & Group 3 & Group 4 & Group 5 & Group 6\\ [0.5ex] 
 \hline
 Ethanol & 100 & 99. & 99. & 98.52 & 98.52 & 98.52\\ 
 \hline
 Ethylene & 0. & 99.77 & 99.55 & 97.95 & 98.41 & 98.41\\
 \hline
 Ammonia & 0. & 0. &  90.  &  87.84 &  85.40 &  85.13\\
 \hline
 Acetaldehyde & 0. & 0. & 0. & 100 & 85.93 &  85.93\\
 \hline
 Acetone & 0. & 0.& 0. & 0. & 58.73 & 59.36\\
 \hline
 Toluene & 0. & 0. & 0. & 0 & 0. & 73.94\\
 \hline 
 \end{tabular}}
  \caption{Mean classification accuracy of Sapinet on batch 1 of UCSD gas sensor drift dataset during online learning. The accuracies are shown w.r.t the $6$ gas types. }

\label{table:5}
\end{table}

\begin{figure}
  \centering
\subfloat[Batch 1]{\includegraphics[width=0.45\linewidth]{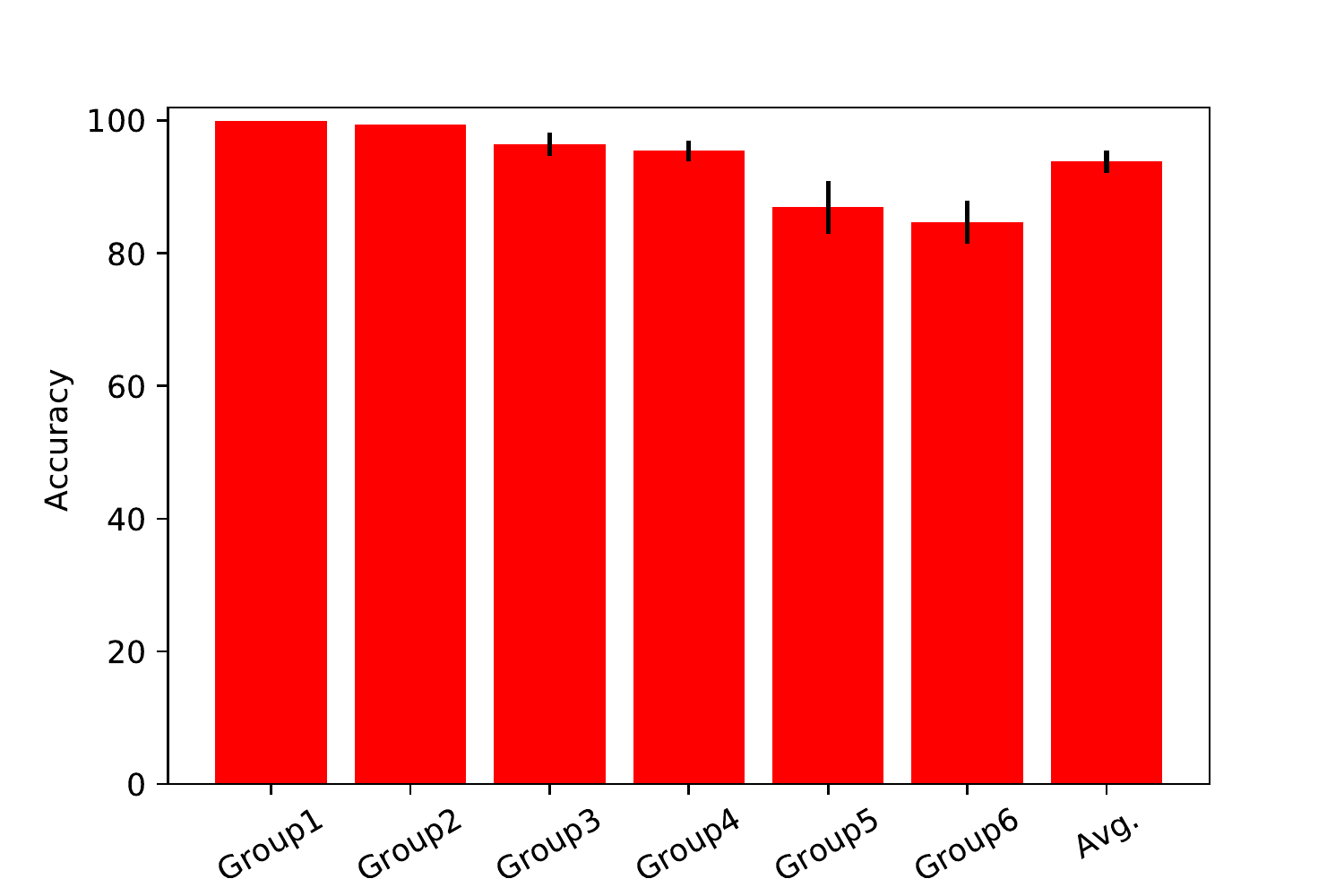}}
\hspace{0.001in}
\subfloat[Batch 10]{\includegraphics[width=0.45\linewidth]{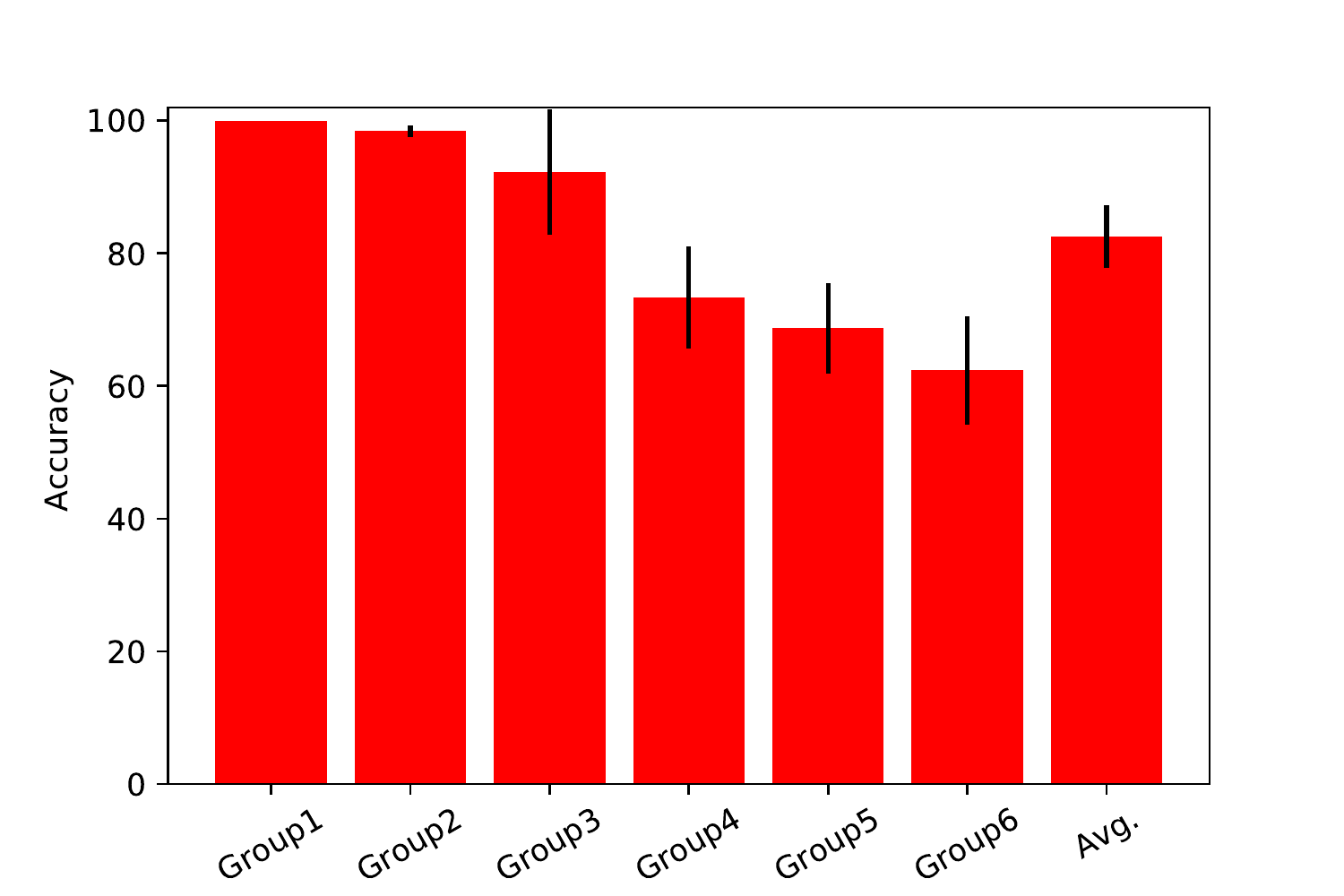}}
\hspace{0.001in}
\caption{Average ( of five runs ) classification accuracies at various stages of online learning. "Group" indicates a 1 shot learning phase. a) Average accuracy during online learning of batch 1 (data from the $1^{st}$ month. b) Average accuracy during online learning of batch 10 (data from the $36^{th}$ month). }  
  \label{ol_all_b110}

\end{figure}

After data regularization and model scaling studies, we tested the multilayer attractor, Sapinet's data classification and denoising performance on all batches of the UCSD gas sensor drift dataset. Table ~\ref{table:4} describes in detail the properties of the UCSD drift dataset. For batch 1, the average online learning ( of five runs ) was $93.84 \pm 1.64$ on a 1 shot 1 epoch sequential online learning task, Fig ~\ref{drift_ol_acc}, Table ~\ref{table:4}. Since the classification was implemented by comparing train / test MCsoma exact spike timing patterns using jaccard distance, this also indicates that Sapinet performed a robust denoising. 

Due to sensor drift, the sensor response to the same gases changes with time. In our previous study on this dataset ~\cite{BorthakurCleland2019, Borthakur2019B} using a feedforward MC - GC network, we proposed the use of rapid online learning as a solution for mitigating sensor drift. We followed the same procedure. Sapinet performed well on all batches. Even on batches with huge drift such as batches 7 - 10, the model achieved accuracies of $79.54 \pm 3.73, 86.42 \pm 7.29, 91.93 \pm 2.41, 82.54 \pm 4.79$ respectively after resetting the network at the start of each batch and training on 1 shot 1 epoch of gas samples. 

We next analyzed the robustness to catastrophic forgetting of Sapinet. Accordingly, we observed the prediction accuracy of the gas samples after learning of subsequent gas samples in an online manner. Fig ~\ref{accu_ol_drift} and Table ~\ref{table:5} describe the online learning performance of all the gas types ( ethanol, ethylene, ammonia, acetaldehyde, acetone, and toluene ) from batch 1. Ethanol was first used for training Sapinet. After learning of all six gases, the prediction accuracy of ethanol remained almost unchanged ( dropped from $100 \pm 0.$ to $98.52 \pm 1.21$ ).   Similarly, for ethylene: $99.77 \pm 0.46$ to $98.41 \pm 3.18$; Ammonia: $90 \pm 0. $ to $85.13 \pm 13.02$; Acetaldehyde: $100. \pm 0.$ to $85.93 \pm 19.098$; Acetone: $58.73 \pm 0.$ to $59.36 \pm 17.02$. Overall, the network was robust enough to perform online learning ( doesn't suffer from catastrophic forgetting ). 

We then analyzed the performance of acetone as the accuracy was only $58.73 \pm 0.$. Fig ~\ref{drift_euc_b1} shows that for acetone, intra cluster distance was $11.27$ which is comparable to its inter cluster distance with Acetaldehyde ($13.41$ ) and also Toluene ($18.37$). Due to this high degree of overlap, the prediction accuracy of Acetone was low. 

Fig ~\ref{ol_all_b110} a, b shows the average prediction accuracies of all gas types during online learning of batch 1 ( with no drift ) and batch 10 ( maximum drift ). 

\subsubsection{Classification performance on the gas sensor arrays in open sampling settings dataset while learning in the wild}

\begin{figure}
  \centering
  \includegraphics[width=0.8\linewidth]{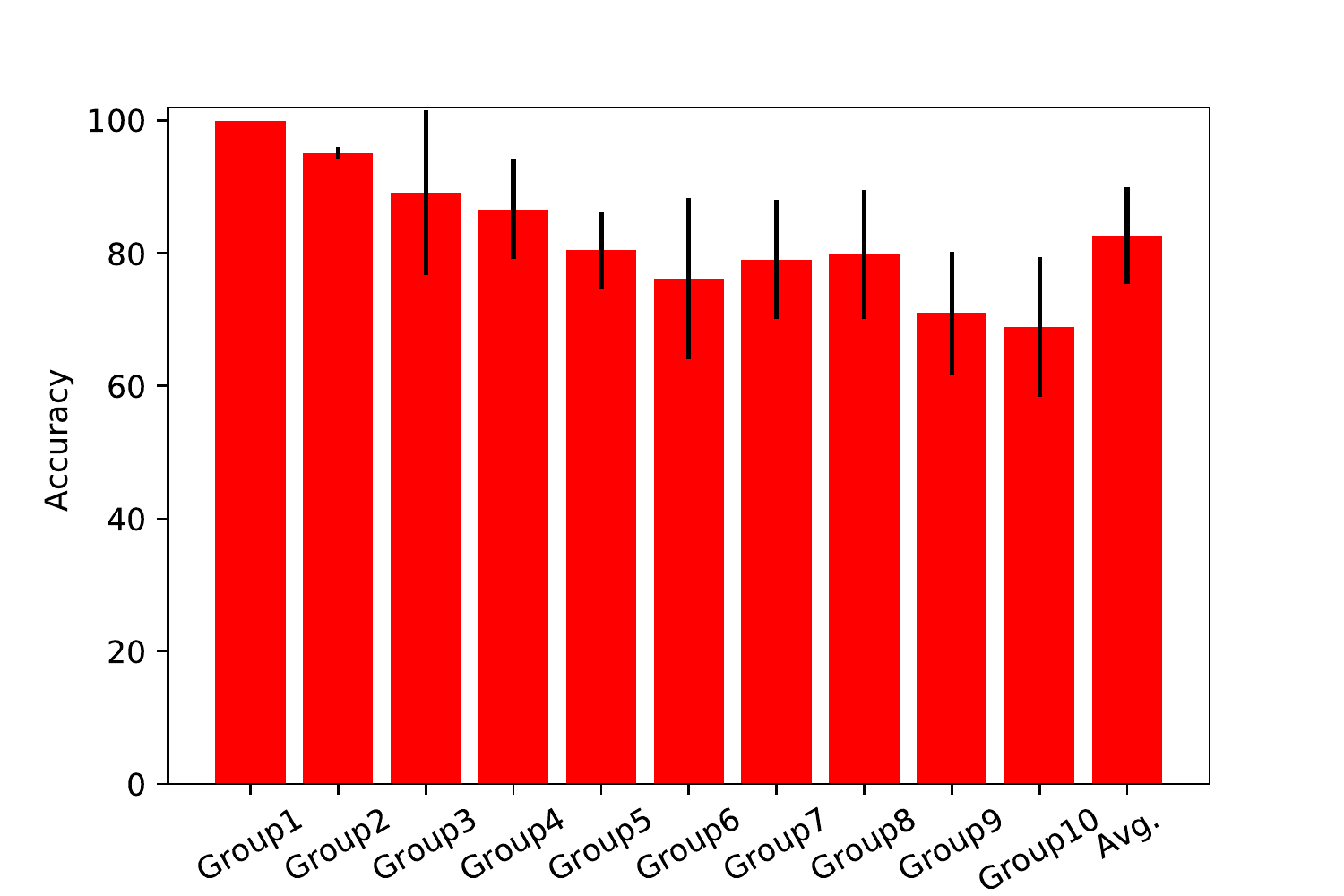}
  \caption{Average ( of five runs ) classification accuracies at various stages of online learning of the $10$ gas sensor response of "Gas sensor arrays in open sampling settings" dataset ~\cite{UCIMachineLearningRepositoryGassen_wind,VERGARA2013462}. "Group" indicates a 1 shot learning phase.}  
  \label{wt_acc}
\end{figure}

\begin{table}[h!]
\centering
\resizebox{\textwidth}{!}{
 \begin{tabular}{||c c c c c c c c c c c c||} 
 \hline
   & Group 1 & Group 2 & Group 3 & Group 4 & Group 5 & Group 6 & Group 7 & Group 8 & Group 9 & Group 10 & Average\\ [0.5ex] 
 \hline
 1 shot & 100. & 95.14  & 89.12 & 86.65  & 80.46 & 76.24 & 79.07 & 79.88 & 71.04 & 68.91 & 82.65\\ 
 \hline
 \end{tabular}}
  \caption{Average ( of five runs ) classification accuracies during online learning of the $10$ gas types of the $10$ gas sensor response of "Gas sensor arrays in open sampling settings" dataset ~\cite{UCIMachineLearningRepositoryGassen_wind,VERGARA2013462}. "Group" indicates a 1 shot learning phase.}

\label{table:6}
\end{table}

We used the "gas sensor arrays in open sampling settings' ' dataset for testing the classification and denoising performance of Sapinet after data regularization and model scaling - without any further fine tuning on a test set with plume dynamics ( see materials \& methods for details ). In order to reduce the model size, we selected only the middle bank of MOS sensors ( $24$ dimension data ). All of the studies above used $75$ non learning and $50$ learning GCs per column. Fig ~\ref{gcs_col} showed that beyond $~20$ GCs per column, the classification performance doesn't change significantly. So, in order to reduce computation time, we used only $20$ high threshold learning GCs per column. Fig ~\ref{wt_acc} and Table ~\ref{table:6} shows the average ( of 5 runs ) performance during online learning. We observed that Sapinet's average prediction accuracy was $82.65 \pm 7.3$.

\section{Discussion}


We present a mammalian olfactory bulb inspired algorithm for \textit{learning in the wild}. The unique characteristics of the algorithm makes it amenable for optimal implementation in neuromorphic chips such as Loihi~\cite{loihi}. In this work, we make a systematic study on methods to build spike timing code, model components contributing towards data regularization, model scaling, data classification and data denoising. \\

For assessing the efficacy of data regularization and model scaling, we used $40$ synthetic samples of $100$ dimension drawn from various distributions ( see results for description ), batch $1, 7, \,  \& \, 10$ of the UCSD gas sensor drift dataset of dimension $16$, and samples from odor plume of the wind tunnel data set of dimension $72$. Specifically, we observed that concentration tolerance, heterogeneity ( in the form of spiking threshold, synaptic weights ) and sparse random projections are effective for regularizing data. This also indicates a possible importance of heterogeneous spiking activities of sister mitral cells ~\cite{Dhawale2010} in data regularization. In ANNs, techniques such as batch normalization ~\cite{ioffe_batch_2015} are most commonly used. But unlike batch normalization, the data regularization techniques discussed here doesn't require knowledge about the data statistics ( mean, standard deviation ) and is suitable for sequential online learning - where there is uncertainty regarding the future data to be encountered.  Next, we developed techniques for tuning connectivity, thresholds according to data dimension as part of model scaling. This allowed us to train/test the algorithm without any other parameter change on synthetic ( gaussian, impulse noises ), drift and wind tunnel datasets. \\

Like the previous implementation by Imam \& Cleland~\cite{imam_rapid_2019} of data classification and denoising, the current implementation has similarities to a Hopfield autoassociative network ~\cite{Hopfield1982}. This current version, Sapinet includes a glomerular layer ( mostly for data regularization ) a modified model of the external plexiform layer. The data classification and denoising implementation includes a novel neurogenesis method which can aid in mapping stimuli similarity unlike the previous implementation by Imam \& Cleland ~\cite{imam_rapid_2019}. Moreover, the granule cell to mitral cell synapse is modified to a inhibitory drive version for robust performance in  various scenarios. The granule cell layer in Sapinet is comprised of a heterogeneous population of neurons - varying spiking thresholds with a percentage of them ( low threshold group ) not having neurogenesis. \\

To ensure the range of success and limitations of our current version of Sapinet, we train/tested the network on multiple datasets  - with different dimensions and challenges such as noise type, concentration, sensor drift, and plume dynamics. Using a sequentially similar synthetic dataset, we assessed Sapinet's performance on gaussian and impulse noise. It was observed that Sapinet's performance degraded when the occlusion level and/ or standard deviation was high and/or the inter odor distance was low ( highly overlapping train data  - similarity for train odors were high).  We experimentally observed that all of these model failure scenarios can be predicted by computing the intra-inter euclidean distances of data clusters. If the intra - inter cluster distances are equivalent, the model's performance degrades. On the other hand, if the intra cluster distance is less than inter cluster distance, the model performance is very good. This is a limitation of Sapinet. These findings were later validated on real world gas sensor datasets, UCSD gas sensor drift and wind tunnel datasets - performance on the highly overlapping gas sensor data was poor which also necessitates the need for better gas sensors. Future work will include methods like discrimination learning for improving performance on highly overlapping datasets. \\

This work also has implications for machine learning, neuroscience and neuromorphic computing in general ~\cite{ClelandBorthakur2020}. Enabling continual learning in ANN ( without catastrophic forgetting) is an active area of research  ~\cite{paris}. Hayes et al. ~\cite{hayes2021replay} proposed the replay during sleep as a mechanism for mitigating catastrophic forgetting. Velez et al ~\cite{velez_diffusion-based_2017} proposed the use of neuromodulation for continual online learning without catastrophic forgetting. Our work indicate that for \textit{ learning in the wild} which is a superset of online learning, neurogenesis is an essential mechanisms for robust performance. For SNN based neuromorphic chips, spike rate coding has not been observed to be optimal in terms of energy consumption and latency ~\cite{imam_rapid_2019, Davies2019}. Whereas, spike timing based coding techniques introduced in this work, Sapinet can significantly improve neuromorphic performance. Neuromorphic chips such as Loihi ~\cite{Davies2019} have been found to be useful for non batch ( or For batch size = 1) train/test scenarios ~\cite{davies2021}.  The sequential learning paradigm and the corresponding good accuracy on test set proposed in this work is a step towards increasing the range of applications of neuromorphic systems. Neuromorphic computing system are non von neumann architectures with computing units ( neurons ) and memory ( synapses ) co-localized.  Computing architectures such as Loihi ~\cite{Davies2019, davies2021} support programmable on-chip weight update algorithms. Use of such local weight update algorithms can significantly reduce training time and energy of neural networks ~\cite{Roy2019}. The proposed local excitatory and inhibitory drive algorithms will reduce training time of networks if implemented on chips. \\

\section{Materials \& Methods}

\subsection{Datasets}

\subsubsection{Synthetic odors}
Odor representations are intrinsically high dimensional data ~\cite{cleland_construction_2014}. Proximity in the data space corresponds to odor similarity.  Owing to the intrinsic variability of odor sources, odor source representations constitute manifolds (\lq{clouds}\rq) within this high-dimensional space. We developed techniques for generating sequentially similar odors of desired dimension for developing and tuning the network as described in Borthakur \& Cleland ~\cite{borthakur_neuromorphic_2017}. 

To generate sequentially similar odor series, we added a small random variation to each of the odor-receptor binding affinities; specifically, $odor \:B = odor \:A + X_{1}; odor \:C = odor \:B + X_{2}$, etc., where $X_{i}$ is a vector whose elements are independently drawn from a skewed normal distribution. Because random walks in high-dimensional spaces reliably move away from the point of origin (P\'{o}lya's theorem), this procedure generated odor series ordered in terms of decreasing similarity; non-overlapping control odors also were generated by randomly shuffling the response values w.r.t the receptor indices. The similarity between odors is tunable by adjusting the \textit{inter odor} distance between odors - a fixed parameter for a set of sequentially similar odors. 

\paragraph{Gaussian noise}
For sequential training, we generated $5$ sequentially similar odors. For testing, in addition to the train samples, we generated $10$ test samples by adding gaussian noise to each of the raw data samples. For that, we used gaussian distributions of $mean \, = \, 0.$ and variable standard deviations ( $2., 6., 18.$ ). The number of noise values added depends upon the occlusion levels ( $0.25, 0.5, 0.75$ ). For example, $0.25$ occlusion level implies $25\%$ of the total number of receptors /sensors were added gaussian noise and hence equivalent number of values from a gaussian distribution were generated. 

\paragraph{Impulse noise}
Impulse noise was generated by replacing a percentage of the total receptors / sensors ($p$) raw responses by values drawn from a uniform distribution in the range $\mathit{0 \mhyphen 20}$ ( see results regarding the selection of this range). The percentage ($p$) is determined by the occlusion level ( $0.25, 0.5, \, 0.75 \, $ ).
 
\subsubsection{UCSD Gas Sensor Array Drift Dataset at Different Concentrations Data Set}
We tested our algorithm on the publicly available UCSD gas sensor drift dataset ~\cite{UCIMachineLearningRepositoryGasSen,vergara_chemical_2012, rodriguez-lujan_calibration_2014}, slightly reorganized to better demonstrate online learning as in our previous work ~\cite{BorthakurCleland2019, Borthakur2019B}. The original dataset contains $13,910$ measurements from an array of $16$ MOS chemosensors exposed to six gas-phase odors spanning a wide range of concentrations ( $\mathit{10 \mhyphen  1000 \, ppmv}$ ) and distributed across $10$ batches that were sampled over a period of $3$ years to emphasize the challenge of sensor drift over time. Owing to drift, the sensors' output statistics change drastically over the course of the $10$ batches. For the online learning scenario, we sorted each batch of data according to the odor trained, but did not organize the data according to concentration. Hence, each training set comprised $1$ odor stimuli of the same type but at randomly selected concentrations. Test sets always included all samples from previously trained odors, again at randomly selected concentrations. For sensor scaling, we used $100\%$ of the Batch 1 data set in order to assess the maximum sensor response before drift. The six odors in the dataset are, in the order of training used herein: ammonia, acetaldehyde, acetone, ethylene, ethanol, and toluene. Batches $\mathit{2 \mhyphen 5}$ included only five different odor stimuli, omitting toluene.

\subsubsection{UCSD Gas Sensor Arrays in Open Sampling Settings}
We also used the " Gas sensor arrays in open sampling setting " dataset ~\cite{UCIMachineLearningRepositoryGassen_wind, VERGARA2013462} for testing data regularization, model scaling and classification, denoising capacity of the attractor. The dataset was created using $72$ MOS chemo sensors distributed across a wind tunnel. For this study, we used recordings from the mid point of the tunnel ( "L4" ), wind speed of $0.21 \, mS^{-1}$ and heater voltage of $500 V$. The size of the tunnel was $1.2 \, m \,$ wide $\, \times 0.4 \,  m \,$ tall $\times 2.5 \, m \,$ long and sensors were deployed in $9$ modules with each module containing $8$ MOS chemo sensors. $10$ different gases - acetone, acetaldehyde, ammonia, butanol, ethylene, methane, methanol, carbon monoxide, benzene and toluene were presented to the gas sensors. Each gas was presented $\mathit{10 \mhyphen 20}$ times for $180 \, seconds$. We trained our network with near peak sensor response ( sensor response at timepoint $90 \, s$  was selected for training set ). The test set comprised of plume variance of data sampled between timepoints $\mathit{30 \mhyphen 180 \, s}$. 

\subsection{Data sampling \& network modeling}

In order to sample an odor, we introduce a slow ( $~5 \, Hz$ ) sampling cycle in the model - which is similar to theta band oscillation in the mammalian olfactory ( can also be referred to as a "sniff" cycle). During each such cycle, a single steady odor is presented to the network as current input across multiple $\gamma$ oscillation cycles ($25 Hz$). Hence, a "sniff" cycle consists of eight $\gamma$ oscillation cycles in this present study.    

The bulb network is composed of mitral cells (MC), external tufted cells (ET) and inhibitory periglomerular cells (PG), and granule cells (GCs). \\
External tufted (ET) and Periglomerular (PG) cells were modeled as non-spiking neurons.

As in previous work ~\cite{imam_rapid_2019}, the mitral cell had two compartments -- apical dendrite (ApiMC) and soma (MCsoma). 

The ET, PG, and ApiMC together form the glomerular layer; and the MCsoma and GC recurrent excitatory-inhibitory network form the external plexiform layer. The primary role of the glomerular layer is data regularization and that of the external plexiform layer is signal identification and denoising. 

We modeled the apical dendrites of mitral cells as leaky integrate and fire equation
\begin{equation}\label{eq:leakint1}
  \tau_{\rm m} \frac{{\rm d}v(t)}{{\rm d}t} =  - v(t) + I(t)/G
\end{equation}

where $v$ is the membrane potential, $G$ is the membrane conductance, $I$ is the input current, $\tau_{m}$ is the membrane time constant, and $v_{th}$ is the Spiking threshold.

In eq.\ref{eq:leakint1}, the apical dendrite spike when the voltage $v$ exceeds the threshold $v_{th}$ and , $v$ is set to $0.$. 

The sensor data is first fed to the ET and the PG cells simultaneously. 
The apical dendrite receives input from the external tufted cells (ET). In the ApiMC, the data is converted into a $\gamma$ spike precedence code. Accordingly, ApiMC spike in order of magnitude -- higher input makes ApiMC spike earlier in a $\gamma$ cycle. 

A spike in ApiMC initiates spontaneous spike in mitral soma (MCsoma) when MCsoma is not influenced by the granule cells (GCs). Post spike, ApiMC is set to refractory mode for the remaining duration of the $\gamma$ cycle. 

\begin{figure}
  \centering
  \includegraphics[width=0.85\linewidth]{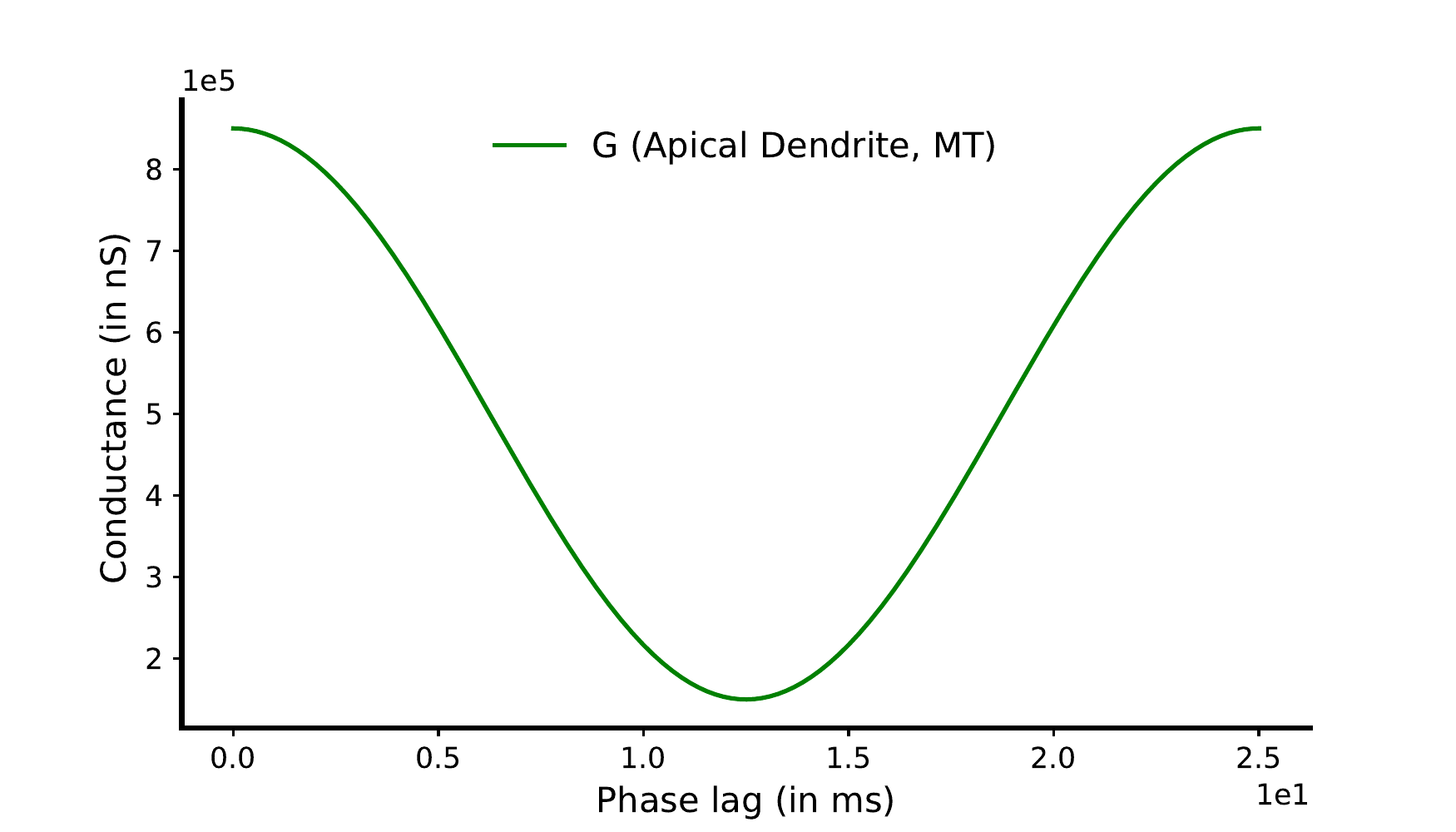}

  \caption{Conductance $G$ varies sinusoidally ($\gamma$ oscillation) for apical dendrite of mitral cells.}  
  \label{gamma_conduct}
\end{figure}

We implemented $\gamma$ oscillation of $ 40 \, Hz $ as a sinusoidal variation of the mitral cell apical dendrite conductance ($G$) as shown in Fig. \ref{gamma_conduct}. 

The mitral cell soma was modeled as an integrator
 \begin{equation}\label{eq:leakint2}
  \frac{{\rm d}v(t)}{{\rm d}t} =  I(t)/c_{mem}
\end{equation}

In integration and fire neuron eq.\ref{eq:leakint2}, if $v >= v_{th}$, set $v=0$, $c_{mem}$ is the membrane capacitance. Both ApiMC and MCsoma spike only once per $\gamma$ cycle and the activities are reset every $\gamma$ cycle. MCsoma is propagated to the GCs and is also used for classifying odors. The current $I$ of MCsoma is tuned by ApiMC and the inhibitory drive (see details below ) applied by the granule cells (GCs). In the $1_{st}$ and the $2_{nd}$  $\gamma$ oscillation cycle, the MCsoma spike follows the ApiMC spike timing - GCs don't apply any inhibitory drive on MCsoma as the network is expected to be unstable.  After the $2_{nd}$ $\gamma$ cycle, MCsoma spike timing is driven by ApiMC and inhibitory drive. Across multiple $\gamma$ cycles, the MCsoma spike timing is shaped towards a previously trained odor MCsoma spike timing pattern. 
Hence, the separation of ApiMC and MCsoma is important to segregate input dependent spike of ApiMC and learning dependent spike of MCsoma. 

The GCs were also modeled as single compartment leaky integrate and fire neurons similar to ApiMCs, eq. ~\ref{eq:leakint1}. Unlike ApiMC, the conductance ( $G$ ) is constant with respect to time. In general, multiple MCsoma spikes are required to make a GC spike. Like the MC compartments, GCs also spike once per cycle and the activities are reset at the end of every $\gamma$ cycle. The total excitatory current to GCs was modeled as

\begin{equation}
    I = g_{w}(E_{n}-v)
\end{equation}

where $E_{n}$ was the Nernst potential of the excitatory current ( $+70mv$ ), $v$ was the GC membrane potential, and $g_{w}=\sum_{i=1}^{n}w_{i}g_{max}\frac{\tau_{1}\tau_{2}}{\tau_{1}-\tau_{2}}(e^\frac{-(t-t_{i})}{\tau_{1}} - e^\frac{-(t-t_{i})}{\tau_{2}})$ describes the open probability of the AMPA-like synaptic conductances. Here, $t_{i}$ denotes presynaptic spike timing, $w_{i}$ denotes the synaptic weight, and $g_{max}$ is a scaling factor ~\cite{borthakur_neuromorphic_2017, BorthakurCleland2019, Borthakur2019B}.

\subsection{Neurogenesis}

Differentiation of interneurons ( GCs ) during learning depletes available GCs (computing resources) for future odor learning. To avoid reduction of network performance, we introduced new neurons ( GCs ) into the network. The rate of introduction of new neurons is slower than the synaptic plasticity rate. In the previous implementation by Imam \& Cleland ~\cite{imam_rapid_2019}, a fixed number of GCs ( $5$ in the model ) were introduced into the network before every odor learning. The model size grew at the same rate after new odor learning. This approach has scalability issues and is not optimal for similarity mapping. We here modify this method and introduce a \textit{use dependent column filler} technique. As per this method, the number of neurons ( GCs ) introduced into the network is equal to the number of active previously non-differentiated GCs during the last odor learning. All GCs can spike during training \& inference. The initial number of GCs per column is high ( e.g. $50$ for UCSD drift dataset, synthetic dataset ) for the studies here and the subsequent addition of GCs per learning cannot exceed this value. The rate of GC layer size may not be constant across learning. The new neurons receive random connection from MCs. The spiking threshold, MC-GC initial convergence ratio and the column location is determined by the differentiated GC being replaced. This process of adding neurons ( GCs ) is analogous to neurogenesis in the granule cell layer of the rodent olfactory bulb which as been experimentally found to be important for odor coding ~\cite{sultan, kermen, lledo, herman, Moreno17980, Shani-Narkiss2020, murthy}.  

\subsection{Excitatory plasticity}

The Mitral soma ( MCsoma ) to granule cell ( GC ) excitatory synapse implemented a local pre to post neuron spike timing dependent plasticity rule described below. The network was initialized with a non zero synaptic weight $w$ ($=18$) in this study. Post learning, granule cells develop sparse responses to odors due to formation of higher order receptive fields (HORF) after repetitive presentation of odor. In our present implementation, MCsoma - GC synaptic weight values get locked to the first odor learned. This locking is essential for effective reconstruction of odors during inference / testing. The synaptic values during learning are capped to a range of $0 - w_{max}$, where $w_{max}$ is a value between $w$ and $1.5 \times w$, randomly drawn from a uniform distribution. 

\paragraph{Spike timing-dependent plasticity rule:}
We used a modified spike timing-dependent plasticity rule (STDP~\cite{song_competitive_2000, dan_spike_2004}) to regulate MCsoma-GC synaptic weight modification.  Briefly, synaptic weight changes were initiated by GC spikes and depended exponentially upon the spike timing difference between the postsynaptic GC spike and the presynaptic MCsoma spike.  When MCsoma spikes preceded the GC spike within the same gamma cycle, $w$ was increased; when MCsoma spikes followed GC spikes, or when a GC spike occurred without a presynaptic MCsoma spike, $w$ was decremented.  Synaptic weights were limited by a maximum weight $w_{max}$. The pairing of STDP with MCsoma spike precedence coding discretized by the gamma clock generated a time-governed \textit{k-winners take all} rule, in which \textit{k} depended substantially on the GC spike threshold $v_{th}$ and on the maximum excitatory synaptic weight $w_{max}$. \\

\begin{algorithm}
Array w[0...\textit{m}, 0...\textit{n}] = 0  \tcp{\textit{w} is the synaptic weight matrix of \textit{m} MCsomas and \textit{n} GCs}
 Let $v_{1}, v_{2}, ...,v_{m}$ be the spike times of MCsomas\\
 Let $u_{1}, u_{2}, ...,u_{n}$ be the spike times of GCs\\
 \For{i=1,2,...,m}{
  \For{j=1,2,...,n}{
  if{$v_{i}$ $\leq$ $u_{j}$}\\
  {set $w[i][j]$ = min($w[i][j] + a_{p}e^{\frac{v_{i}-u_{j}}{tau_{p}}},	w_{max}$) \tcp{$a_{p}$ is the learning rate, $tau_{p}$ is the time constant of synaptic weight increase, $w_{max}$ is the maximum permitted synaptic weight.}}
  else\\
  {if {$v_{i}$ is infinity}\\{
  set $w[i][j]$ = max(0, $w[i][j] - w_{scale}$) \tcp{$w_{scale}$ is a parameter that heavily penalizes  synapses in which the presynaptic MCsomas do not spike.}}
  
  else\\
  {set $w[i][j]$ = max($w[i][j] - a_{n}e^{-\frac{v_{i}-u_{j}}{tau_{n}})}$)\tcp{$a_{n}$ is the learning rate, $tau_{n}$ is the time constant of synaptic weight decrease}}
 
  }}}
\Return w
 \caption{Spike timing dependent plasticity (STDP) at the end of $\gamma$ cycle}
\end{algorithm}

The parameters $a_{p}, a_{m}, tau_{p}, tau_{m},$ and $w_{scale}$ were tuned using a synthetic dataset~\cite{borthakur_neuromorphic_2017, BorthakurCleland2019}, whereas the maximum synaptic weight $w_{max}$ was tuned based on training and validation set performance on the synthetic dataset.  

\subsection{Inhibitory drive}

The MCs can excite almost any GCs. But for GCs, there exists a biologically appropriate columnar organization ~\cite{bi}. Like the previous work ~\cite{imam_rapid_2019}, we here retain the goal of odor-specific spike timing learning but only in the high threshold learning GC. But GCs now learn spike timings of all sister MCsomas. In addition, we modify the GC action method on MCsomas. Also, GCs of multiple spiking thresholds are present in a column. 

\subsubsection{Low threshold non learning GCs}
The low threshold GCs are non-learning GCs in the model. Low threshold GCs require less MCs activities to spike. Hence, they are responsive to more odors and are less selective. Absence of learning ensures explosion of the number of newborn low threshold neurons ( due to neurogenesis ) in the model. But they are hypothesized as essential for denoising very highly occluded signals.  During learning, spiking low threshold GCs are grouped with the co-columnar high threshold learning GCs - this grouping is used for application of inhibitory drive. 

\subsubsection{High threshold learning GC to MC weight update at the end of $\gamma$ cycle}
GCs attain non zero weights after learning only for their synapses to the co-columnar sister mitral cells ($5$ sister mitral cells in our implementation ).  Let $t_{mc}^1, t_{mc}^2, t_{mc}^3, t_{mc}^4, t_{mc}^5$ be the  spike timings of $5$ co-columnar sister mitral cell somas. Let $GC_{1}, GC_{2}, ..., GC_{n}$ be $n$ high threshold learning granule cells belonging to the same column that spiked. Then during learning, 
The synaptic weights from a spiked GC to a co-columnar MC is set equal to the corresponding MCsoma spike timing. For example, if $w_{1}^{1, ..., 5}$ are the weights of $GC_{1}$ to its sister mitral cells, after learning, $w_{1}^{1, ..., 5}$ are set equal to $t_{mc}^{1, ..., 5}$. 

\subsubsection{Action of GC on MCsoma during training/ testing }

Unlike the previous implementation ~\cite{imam_rapid_2019}, GC action on MC is the same both during testing and inference / testing. We define this GC action on MC as inhibitory drive. Due to application of inhibitory drive, MCsoma spike timing is determined by the selected GC-MC weight ($w_{gcmc}$) -- if ApiMC spike earlier than $w_{gcmc}$, MCsoma spike is delayed whereas MCsoma spike excited with a strong current input if ApiMC spike later than $w_{gcmc}$. $w_{gcmc}$ is selected as the maximum number of nonzero similar weight value in a column for a sister mitral cells. \\
\subsection{What is learning in the wild ?}
\textit{Learning in the wild} defines an aspirational set of capacities for artificial neural networks that reflect the performance of biological systems operating in natural environments (discussed in detail in Borthakur \& Cleland ~\cite{Borthakur2019B}). Some key requisites of \textit{learning in the wild} algorithm, Sapinet discussed here are - 

\begin{itemize}
\item It must support online learning (no catastrophic forgetting, ideally without storage of  trained data).
\item It must exhibit rapid one or few-shot learning of novel stimuli.
\item It must be robust to wild, poorly-matched inputs without resorting to hyperparameter re-tuning.
\item It must provide a technique for model scaling - tuning model with respect to data dimension. 
\item It must be robust to environmental and stimulus variance, including unpredictable stimulus intensities (e.g., odor concentrations), other forms of stimulus heterogeneity, and the effects of environmental temperature and humidity.
\item It must exhibit concentration tolerance where appropriate, and also provide an estimate of concentration.
\item It must adapt to sensor drift owing to time and/or contamination [5].
\item It must provide none of the above options during classification (classifier confidence) [5].
\item It must be able to identify the signatures of known inputs despite substantial interference from background stimuli (whether previously or simultaneously delivered).
\end{itemize}

The mammalian primary olfactory receptor responses are altered by factors such as wind, presence of other odors in the background, etc. The mammalian olfactory system is capable of detecting odors despite these interferences. We here sought to copy these characteristics in our artificial neural network implementation. \\

For assessing the efficacy of data regularization and model scaling, we used $40$ synthetic samples of $100$ dimension drawn from various distributions ( see results for description ), batch $1, 7, \& 10$ of the UCSD gas sensor drift dataset of dimension $16$, and samples from odor plume of the wind tunnel data set of dimension $72$. \\

For understanding the attractor's performance in a $1$ shot learning task, we used the synthetic similar odors ($4$ sequentially similar and $1$ non overlapping of dimension $20$) of varying inter odor distances ($0.25, 0.5, 0.75 \& 1.$) on a sequential learning task. During testing, we introduced gaussian noise of different levels of  standard deviations: $2., 6., 18.$ and occlusion levels: $0.25, 0.5, \& 0.75$ respectively ( total number of datasets $\, =\, 4 \times 3 \times 3 \, = \, 36$). Using the same network and synthetic train odors, for testing we next introduced impulse noises of occlusion levels: $0.25, 0.5, 0.75$, total number of datasets: $\, = \, 4 \times 3 \, = \, 12$. After testing and tuning the network with synthetic odor and noise, we next tested the network's performance on real world datasets. We tested our algorithm on the publicly available UCSD gas sensor drift dataset ~\cite{vergara_chemical_2012, rodriguez-lujan_calibration_2014} for \textit{learning in the wild} as described eariler. The original dataset contains $13,910$ measurements from an array of $16$ polymer chemosensors exposed to six gas-phase odorants spanning a wide range of concentrations ( $10 – 1000 \, ppmv$ ) and distributed across $10$ batches that were sampled over a period of $3$ years to emphasize the challenge of sensor drift over time. In order to account for change in dimension from the synthetic dataset, we used data regularization and model scaling techniques. In order to ensure generality of our algorithm, we next tested our network's performance after data regularization and model scaling on the plume dynamics of the wind tunnel dataset ~\cite{VERGARA2013462}. We trained the network on a sequential learning task using $10$ different gas sensor responses of the data drawn from around $90 s$. For testing, we extracted $30$ samples ( per gas sensor exposure to an an odor ) per plume by sampling at an interval of $5 \, s$ between $30 - 180 s$. 

The classification method is similar to Imam \& Cleland ~\cite{imam_rapid_2019}. Briefly, MCsoma spikes of the last $\gamma$ oscillation cycle are recorded at the end of training. During inference / testing, similarities of MCsoma spike timing pattern of all $\gamma$ cycles of the unknown are compared with all the previously trained MCsoma spike timing patterns. If the similarity is less than classifier confidence, then the unknown odor class is set to \textit{None of the above}. For similarities less than classifier confidence, the greatest similarity class is selected. In our implementation we report similarities as $1 \, - \, Jaccard \, distance \, between \, samples $.

\section{Acknowledgements}
I am most grateful to my advisor, Prof. Thomas A. Cleland, for giving me
the freedom to design this exciting project by merging ideas from electrical engineering, computer science, neuroscience \& psychology. I am also very thankful to all of my committee members: Prof. Alyosha Molnar, Prof. David Field, Prof. David Smith, and Prof. Thorsten Joachims.

\section{Funding details}
This work was funded in part by a Cornell Sage fellowship, an Intel Neuromorphic Research Community project grant, and a Cornell Center for Technology Licensing IGNITE Research Acceleration grant. 

\bibliographystyle{unsrt}  
\bibliography{references}  

\end{document}